\documentclass{article}

    \PassOptionsToPackage{numbers, compress}{natbib}

\usepackage{algorithm}
\usepackage{algorithmic}
\usepackage{wrapfig}

\usepackage[final]{neurips_2023}
\usepackage[utf8]{inputenc} 
\usepackage[T1]{fontenc}    
\usepackage{hyperref}       
\usepackage{url}            
\usepackage{booktabs}       
\usepackage{amsfonts}       
\usepackage{nicefrac}       
\usepackage{microtype}      
\usepackage{xcolor}         

\usepackage{amsmath, amssymb}
\usepackage{bm}
\usepackage{subfig}
\usepackage{paralist}

\usepackage{mathtools}
\mathtoolsset{showonlyrefs}

\usepackage{multirow, makecell}

\newcommand{\Transpose}{^{\mathsf{T}}}

\newcommand{\SolutionSymb}{u}
\newcommand{\Solution}{{\SolutionSymb}}

\newcommand{\HiddenStateSymb}{h}
\newcommand{\HiddenState}{\bm{\HiddenStateSymb}}

\newcommand{\HiddenDim}[1]{n_{#1}}

\newcommand{\NNWeightSymb}{W}
\newcommand{\NNWeight}{\NNWeightSymb}

\newcommand{\NNBiasSymb}{b}
\newcommand{\NNBias}{\bm{\NNBiasSymb}}

\newcommand{\NonlinearActivation}{\sigma}

\newcommand{\LeftSingularVectorSymb}{U}
\newcommand{\LeftSingularVector}{\LeftSingularVectorSymb}

\newcommand{\RightSingularVectorSymb}{V}
\newcommand{\RightSingularVector}{\RightSingularVectorSymb}

\newcommand{\SingularValueSymb}{\Sigma}
\newcommand{\SingularValue}{\SingularValueSymb}

\newcommand{\SingularValueElemSymb}{s}
\newcommand{\SingularValueElem}{\bm{\SingularValueElemSymb}}

\newcommand{\LRPINN}{{LR-PINN}}
\newcommand{\LRPINNs}{{LR-PINNs}}
\newcommand{\LRPINNplain}{{LR-PINN}}
\newcommand{\HyperLRPINN}{{Hyper-LR-PINN}}
\newcommand{\NaiveLRPINN}{{Na\"ive-LR-PINN}}

\newcommand{\HLRPINN}{{Hyper-LR-PINN}}

\title{Hypernetwork-based Meta-Learning\\for Low-Rank Physics-Informed Neural Networks}

\author{%
  Woojin Cho\footnote[2]{} \quad \; Kookjin Lee\footnote[3]{}\, \thanks{Co-corresponding authors} \quad \; Donsub Rim\footnote[4]{} \quad \; Noseong Park\footnote[2]{}\, \footnote[1]{} \\
  \footnote[2]{} \, Yonsei University \\
  \footnote[3]{} \, Arizona State University \\
  \footnote[4]{} \, Washington University in St. Louis \\
  \texttt{snowmoon@yonsei.ac.kr, kookjin.lee@asu.edu,} \\
  \texttt{rim@wustl.edu, noseong@yonsei.ac.kr} \\
}

\begin{document}

\maketitle

\begin{abstract}
In various engineering and applied science applications, repetitive numerical simulations of partial differential equations (PDEs) for varying input parameters are often required (e.g., aircraft shape optimization over many design parameters) and solvers are required to perform rapid execution. In this study, we suggest a path that potentially opens up a possibility for physics-informed neural networks (PINNs), emerging deep-learning-based solvers, to be considered as one such solver. Although PINNs have pioneered a proper integration of deep-learning and scientific computing, they require repetitive time-consuming training of neural networks, which is not suitable for \textit{many-query} scenarios. To address this issue, we propose lightweight low-rank PINNs containing only hundreds of model parameters and an associated hypernetwork-based meta-learning algorithm, which allow efficient solution approximations for varying PDE input parameters. Moreover, we show that the proposed method is effective in overcoming a challenging issue, known as ``failure modes’’ of PINNs.
\end{abstract}

\section{Introduction}
Physics-informed neural networks (PINNs) \cite{raissi2019physics} are a particular class of coordinate-based multi-layer perceptrons (MLPs), also known as implicit neural representations (INRs), to numerically approximate solutions of partial differential equations (PDEs). That is, PINNs are taking spatiotemporal coordinates $(\pmb{x},t)$ as an input and predict PDE solutions evaluated at the coordinates $u_{\Theta}(\pmb{x},t)$ and are trained by minimizing (implicit) PDE residual loss and data matching loss at initial and boundary conditions. PINNs have been successfully applied to many different important applications in computational science and engineering domain: computational fluid dynamics \cite{mao2020physics,yang2019predictive}, cardiac electrophysiology simulation \cite{sahli2020physics}, material science \cite{zhang2022analyses}, and photonics \cite{ma2021deep}, to name a few. 

PINNs are, however, sharing the same weakness with coordinate-based MLPs (or INRs), which hinders the application of PINNs/INRs to more diverse applications; for a new data instance (e.g., a new PDE for PINNs or a new image for INRs), training a new neural network (typically from scratch) is required. Thus, using PINNs to solve PDEs (particularly, in parameterized PDE settings) is usually computationally demanding, and this burden precludes the application of PINNs to important scenarios that involve \textit{many queries} in nature as these scenarios require the parameterized PDE models to be simulated thousands of times (e.g., design optimization, uncertainty propagation), i.e., requiring PDE solutions $u(\pmb x,t;\pmb \mu)$ at many PDE parameter settings $\{\pmb \mu^{(i)}\}_{i=1}^{N_{\mu}}$ with very large $N_\mu$.

To mitigate the above described issue, we propose i) a low-rank structured neural network architecture for PINNs, denoted as low-rank PINNs (LR-PINNs), ii) an efficient rank-revealing training algorithm, which adaptively adjust ranks of LR-PINNs for varying PDE inputs, and iii) a two-phase procedure (offline training/online testing) for handling  \textit{many-query} scenarios. This study is inspired by the observations from the studies of numerical PDE solvers \cite{kressner2011low,grasedyck2013literature,lee2017preconditioned,bachmayr2018parametric,lee2020model} stating that numerical solutions of parameteric PDEs can be often approximated in a low-rank matrix or tensor format with reduced computational/memory requirements. In particular, the proposed approach adopts the computational formalism used in reduced-order modeling (ROM) \cite{holmes2012turbulence, benner2015survey}, one of the most dominant approaches in solving parameteric PDEs, which we will further elaborate in Section \ref{sec:hlrpinn}.  

In essence, \LRPINNs{} represent the weight of some internal layers as a low-rank matrix format. Specifically, we employ a singular-value-decomposition (SVD)-like matrix decomposition, i.e., a linear combination of rank-1 matrices: the weight of the $l$-th layer is $W^{l} = \sum_{i=1}^r s_i^l \pmb{u}_i^l \pmb{v}_i^l{}\Transpose$ with the rank $r = \min( n_{l}, n_{l+1})$, where $W^{l} \in \mathbb{R}^{n_{l+1} \times n_{l}}$, $\pmb{u}_i^l \in \mathbb{R}^{n_{l+1}}$,  $\pmb{v}_i^l \in \mathbb{R}^{n_{l}}$, and $s^l_i\in\mathbb{R}$. 
The ranks of the internal layers, however, typically are not known a priori. To address this issue, we devise a novel hypernetwork-based neural network architecture, where the rank-structure depending on the PDE parameters $\pmb \mu$ is learned via training. In short, the proposed architecture consists of i) a \textit{lightweight} hypernetwork module and ii) a low-rank solution network module; the hypernetwork takes in the PDE parameters and produces the coefficients of the rank-1 series expansion (i.e., $\pmb{s}(\pmb{\mu}) = f^{\text{hyper}}(\pmb{\mu})$). The low-rank solution network module i) takes in the spatiotemporal coordinates $(\pmb{x},t)$, ii) takes the forward pass through the linear layers with the weights $W^{l}(\pmb{\mu}) = \sum_{i=1}^r s_i^l(\pmb{\mu}) \pmb{u}_i^l \pmb{v}_i^l{}\Transpose$, of which $s_i(\pmb{\mu})$ comes from the hypernetwork, and iii) produces the prediction $u_{\Theta}(\pmb{x},t;\pmb{\mu})$. Then, the training is performed via minimizing the PINN loss, i.e., a part of the PINN loss is the PDE residual loss, $\| \mathcal R(u_{\Theta}(x,t;\pmb{\mu});\pmb{\mu})\|_2$, where $\mathcal R(\cdot,\cdot;\pmb \mu)$ denotes the parameterized PDE residual operator. 


We show the efficacy and the efficiency of our proposed method for 
solving some of the most fundamental parameterized PDEs, called \emph{convection-diffusion-reaction equations}, and \emph{Helmholtz equations}. Our contributions include:
\begin{compactenum}
    \item We employ a low-rank neural network for PINNs after identifying three research challenges to address.
    \item We develop a hypernetwork-based framework for solving parameterized PDEs, which computes solutions in a rank-adaptive way for varying PDE parameters. 
    \item We demonstrate that the proposed method resolves the ``\textit{failure modes}'' of PINNs.  
    \item We also demonstrate that our method outperforms baselines in terms of accuracy and speed.
\end{compactenum}

\section{Na\"ive low-rank PINNs} \label{sec:naive}
Let us being by formally defining \LRPINNs{} and attempt to answer relevant research questions. \LRPINNs{} are a class of PINNs that has hidden fully-connected layers (\textsc{FC}) represented as a low-rank weight matrix. We denote this intermediate layer as \textsc{lr-FC}: the $l$-th hidden layer is defined such that 
\begin{equation}\label{eq:factored_fc}
    \HiddenState^{l+1} = \textsc{lr-FC}^{l}(\HiddenState^{l}) 
    \quad \Leftrightarrow \quad \HiddenState^{l+1} = \LeftSingularVector_r^l(\SingularValue_r^l (\RightSingularVector_r^l{}\Transpose \HiddenState^l)) + \NNBias^l, 
\end{equation}
where $\LeftSingularVector_r^l \in \mathbb{R}^{\HiddenDim{l+1} \times r}$ and $\RightSingularVector_r^l \in \mathbb{R}^{\HiddenDim{l} \times r}$ denote full column-rank matrices (i.e., rank $r \ll n_l, n_{l+1}$) containing a set of orthogonal basis vectors, and $\SingularValue_r^l \in \mathbb{R}^{r \times r}$ is a diagonal matrix, $\SingularValue_r^l = \text{diag}\left(\SingularValueElem_r^l\right)$ with $\SingularValueElem_r^l\in\mathbb{R}^{r}$. 

\paragraph{Memory efficiency:} 
\LRPINN{}s with a rank of $r$ and $L$ hidden layers require $O((2n_l+1)rL)$ memory as opposed to $O(n_l^2 L)$ required by regular PINNs. 

\paragraph{Computational efficiency:} 
The forward/backward pass of \LRPINN{}s can be computed efficiently by utilizing a factored representation of the weights. To simplify the presentation, we describe only the  forward pass computation; 
the forward pass is equivalent to perform three small matrix-vector products (MVPs) in sequence as indicated by the parentheses in Eq.~\eqref{eq:factored_fc}. 

\paragraph{Challenges:} Representing the weights of hidden layers itself is straightforward and indeed has been studied actively in many different fields of deep learning, e.g., NLP \cite{chen2021drone,hu2021lora}. However, those approaches typically assume that there exist pre-trained models and approximate the model weights by running the truncated SVD algorithm. Our approach is different from these approaches in that we attempt to reveal the ranks of internal layers as the training proceeds, which brings  unique challenges. These challenges can be summarized with some research questions, which include: \textbf{C1}) ``should we make all parameters learnable (i.e., ($U^l, V^l, \Sigma^l)$)?'',  \textbf{C2}) ``how can we determine the ranks of each layer separately, and also adaptively for varying $\pmb{\mu}$?'', and \textbf{C3}) ``can we utilize a low-rank structure to avoid expensive and repetitive training of PINNs for every single new $\pmb{\mu}$ instances?''. In the following, we address these questions by proposing a novel neural network architecture.  

\begin{figure*} [t]
\centering
\includegraphics[width=1.0\columnwidth]{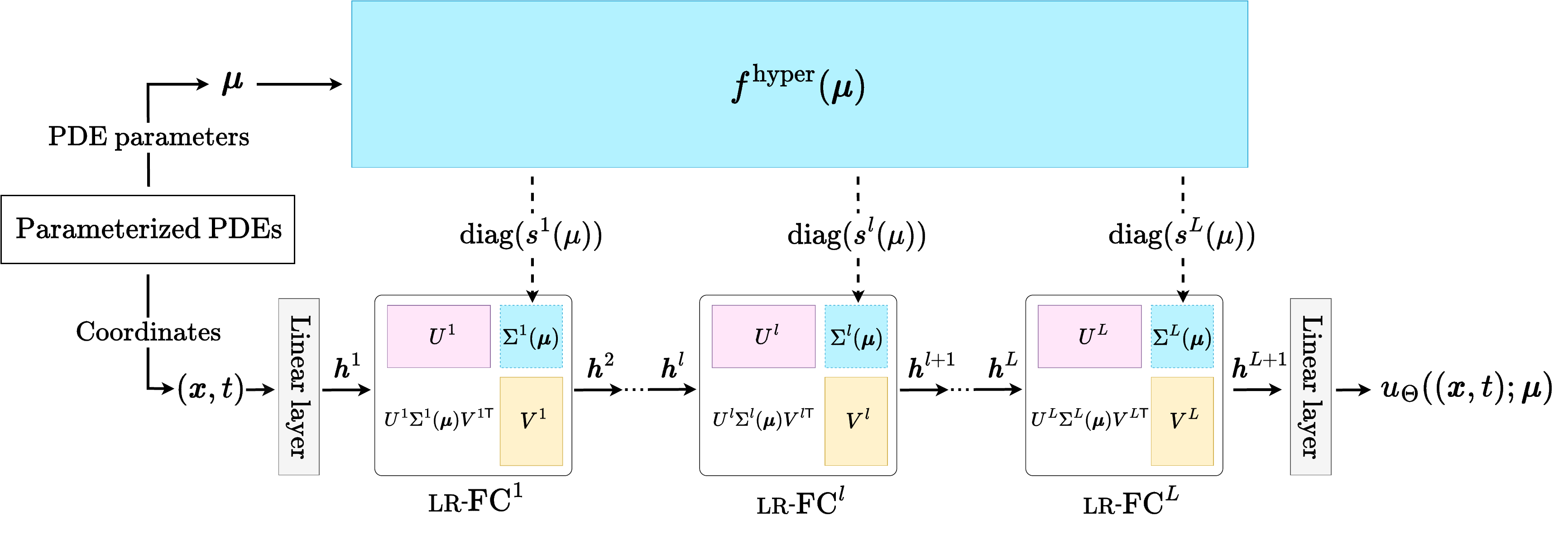}\vspace{-2mm}
\caption{The architecture of \HLRPINN{} consisting of i) the hypernetwork generating model parameters (i.e., diagonal elements) of \LRPINNplain{} and ii) \LRPINNplain{} approximating solutions.}\label{fig:archi}
\end{figure*}

\section{Hyper-LR-PINNs: hypernetwork-based meta-learning low-rank PINNs}\label{sec:hlrpinn}
We propose a novel neural network architecture based on a hypernetwork and an associated training algorithm, which address the predescribed challenges. To distinguish them, we hereinafter call \LRPINNplain{} with (resp. w/o) our proposed hypernetwork as ``\HyperLRPINN{}'' (resp. ``\NaiveLRPINN{}'').

\paragraph{Design goals:} 
Here we attempt to resolve all challenges by setting up several design goals that are inspired from our domain knowledge in the field (and also from some preliminary results that can be found in Appendix~\ref{a:prelim_exp}):
\begin{compactenum}
\item build a \textit{single set} of basis vectors $U^l$ and $V^l$, preferably as \textit{orthogonal} as possible, that perform well over a range of PDE parameters, 
\item build LR-PINNs with an \textit{adaptive and layer-wise rank structure} that depends on PDE parameters (e.g., a higher rank for a higher convective PDE) and,
\item make \textit{only the diagonal elements}, denoted $\SingularValueElem_r^l$ in Eq.~\eqref{eq:factored_fc}, \textit{learnable} to achieve high efficiency once a proper set of basis vectors and the rank structure are identified.
\end{compactenum}

Our design largely follows the principles of ROMs, where the expensive computation is offloaded to an \textit{offline} phase to build a cheap surrogate model that can perform an efficient computation in an \textit{online} phase for a test set. To make an analogy, consider parameterized dynamical systems (which may arise in semi-discretization of time-dependent PDEs):  $\frac{\mathrm d \pmb u(t;\pmb{\mu})}{\mathrm d t} = \pmb f(\pmb u(t;\pmb{\mu}); \pmb{\mu})$, where $\pmb u \in \mathbb R^{N}$. In the offline phase, the method seeks a low-dimensional linear trial basis where the reduced representation of the solution lie on, which is achieved by performing high-fidelity simulations on a set of PDE parameter instances $\{\pmb \mu^{(i)}\}_{i=1}^{n_{\text{train}}}$ and constructing a trial linear subspace $\Psi := \text{range}(\Psi_p)$ with $\Psi_p = [\psi_1,\cdots,\psi_p] \in \mathbb{R}^{N\times p}$ from the solution snapshots collected from the high-fidelity simulations. In the online phase, the solutions at a set of test PDE parameter instances $\{\pmb \mu^{(i)}\}_{i=1}^{n_{\text{test}}}$ are approximated as $\pmb u(t, \pmb \mu) \approx \Psi_p \pmb c(t, \pmb \mu)$ with $\pmb c \in \mathbb{R}^{p}$, and a low-dimensional surrogate problem is derived as $\frac{\mathrm d \pmb c(t;\pmb{\mu})}{\mathrm d t} = \Psi_p\Transpose \pmb f (\Psi_p \pmb c(t;\pmb{\mu}); \pmb{\mu}) =\hat{\pmb f}(\pmb c(t;\pmb{\mu}); \pmb{\mu}) \in \mathbb{R}^{p}$, which can be rapidly solved, while not losing too much accuracy. See Appendix~\ref{a:rom} for an illustrative explanation and details of ROMs.

Taking a cue from the ROM principles, we design our model to operate on a common set of basis vectors $\{U^l_r\}$ and $\{V^l_r\}$, which are obtained during the offline phase (analogous to $\Psi_p$ in ROMs) and update only the diagonal elements $\{\pmb s_r^l\}$ during the online phase (analogous to $\pmb c$ in ROMs). Now, we elaborate our network design and the associated two-phase algorithm. For the connection to the ROM, which explains details on the context/query sets, can be found in Appendix~\ref{a:rom}. 



\subsection{Hypernetwork-based neural network architecture}\label{sec:hypernet}
The proposed framework has two computational paths: a path for the hypernetwork and a path for \LRPINNplain{} (Figure~\ref{fig:archi}). The hypernetwork path reads the PDE parameter $\pmb{\mu}$ and outputs the diagonal elements $\{\pmb{s}^l\}_{l=1}^L$ of \textsc{lr-FC}s. The \LRPINNplain{} path reads 
$(\pmb{x}, t)$ and the output of the hypernetwork, and outputs the approximated solution 
$u_{\Theta}$ at  $(\pmb{x}, t;\pmb{\mu})$, which can be written as follows:
\begin{equation}
u_{\Theta}((\pmb{x},t); \pmb{\mu}) = u_{\Theta}((\pmb{x},t); f^{\text{hyper}}(\pmb{\mu})),
\end{equation}
where $f^{\text{hyper}}(\pmb{\mu})$ denotes the hypernetwork such that $\{\pmb{s}^l(\pmb{\mu})\}_{l=1}^L = f^{\text{hyper}}(\pmb{\mu})$. We denote $\pmb{s}$ as a function of $\pmb{\mu}$ to make it explicit that it is dependent on $\pmb{\mu}$. 
The internals of \LRPINNplain{} can be described as with $\pmb{h}^{0} = [\pmb{x}, t]\Transpose$
\begin{equation}
    \begin{split}
        \pmb{h}^1 &= \sigma(W^0 \pmb{h}^0 + \pmb{b}^0),\\
        \pmb{h}^{l+1} &= \sigma(U^l ( \Sigma^l(\pmb{\mu}) (V^l{}\Transpose \pmb{h}^l)) + \pmb{b}^l), l=1,\ldots,L,\\
        u_{\Theta}((\pmb{x},t);\pmb{\mu}) &= \sigma(W^{L+1} \pmb{h}^{L+1} + \pmb{b}^{L+1}), 
    \end{split}
\end{equation} where $\Sigma^l(\pmb{\mu}) = \text{diag}({\pmb{s}^l(\pmb{\mu})})$. The hypernetwork can be described as the following initial embedding layer, where $\pmb{e}^{0} = \pmb{\mu}$, followed by an output layer:
\begin{equation}
    \begin{split}
    \pmb{e}^{m} = \sigma( W^{\text{emb},m} \pmb{e}^{m-1} + \pmb{b}^{\text{emb},m}),\;\;\; m=1,\ldots,M,\\
    \pmb{s}^l(\pmb{\mu}) = \text{ReLU} ( W^{\text{hyper},l} \pmb{e}^{M} + \pmb{b}^{\text{hyper},l}),\;\;\; l=1,\ldots, L,
    \end{split}
\end{equation} where ReLU is employed to automatically truncate the negative values so that the adaptive rank structure for varying PDE parameters can be revealed (i.e., the number of non-zeros (NNZs)\footnote{The NNZs is not equal to the rank as the learned basis is only soft-constrained to be orthogonal (Eq.~\eqref{eq:ortho_const}). Nevertheless, to avoid  notation overloading, we denote NNZs as the rank in the following.} in $\pmb{s}^l(\pmb{\mu})$ varies depending on $\pmb{\mu}$).

\subsection{Two-phase training algorithm}\label{sec:train_alg}
Along with the framework, we present the proposed two-phase training algorithm. Phase 1 is for learning the common set of basis vectors and the hypernetwork and Phase 2 is for fine-tuning the network for a specific set of test PDE parameters. 
Table~\ref{tab:param_summary} shows the sets of model parameters that are being trained in each phase. (See Appendix~\ref{a:algorithm} for the formal algorithm.)

In Phase 1, we train the hypernetwork and the \LRPINNplain{} jointly on a set of collocation points that are collected for varying PDE parameters. 
Through the computational procedure described in Section~\ref{sec:hypernet}, the approximated solutions at the collocation points are produced $u_{\Theta}((\pmb{x}_j,t_j); \pmb{\mu}^{(i)})$. Then, as in regular PINNs, the PDE residual loss and the data matching loss can be computed. The small difference is that the PDE operator, $\mathcal F$,  for the residual loss is also parameterized such that $\mathcal F(u_{\Theta}((\pmb{x}_j,t_j); \pmb{\mu}^{(i)}); \pmb{\mu}^{(i)})$.
As we wish to obtain basis vectors that are close to orthogonal, we add the following orthogonality constraint based on the Frobenius norm to the PINN loss \cite{yang2020learning}:
\begin{equation}\label{eq:ortho_const}
w_1 \| U^l{}\Transpose U^l - I \|_F^2  + w_2 \| V^l{}\Transpose V^l - I \|_F^2,
\end{equation}
where $w_1$ and $w_2$ are penalty weights. (See Appendix~\ref{a:ablation}.)

In Phase 2, we continue training \LRPINNplain{} for approximating the solutions of a target PDE parameter configuration. We i) fix the weights, the biases, and the set of basis vectors of \textsc{lr-FC} obtained from  Phase 1 , ii) convert the diagonal elements to a set of learnable parameters after initializing them with the values from the hypernetwork, and iii) detach the hypernetwork. Thus, only the trainable parameters from this point are the set of diagonal elements, first and last linear layers. The hypernetwork-initialized diagonal elements serve as a good starting point in (stochastic) gradient update optimizers (i.e., requires less number of epochs). Moreover, significant computational savings can be achieved in the gradient update steps as only the diagonal elements are updated (i.e., $\pmb s_{l+1} \leftarrow \pmb s_l + \eta \nabla_{\pmb s} L$ instead of $\Theta_{l+1} \leftarrow \Theta_l + \eta \nabla_{\Theta} L$). 
\begin{table}[t]
\centering
\small
\caption{Learnable parameters in each phase}\label{tab:param_summary}
\begin{tabular}{l|l}
\toprule
\multirow{2}{*}{Phase 1} & $\{(U^l, V^l, \pmb{b}^{l})\}_{l=1}^{L}$, $W^0, W^{L+1}, \pmb{b}^0, \pmb{b}^{L+1}$ \hfill (\textsc{LR-PINN}),\\
& $\{( W^{\text{emb},m},  \pmb{b}^{\text{emb},m})\}_{m=1}^{M}$,  $\{(W^{\text{hyper},l}, \pmb{b}^{\text{hyper},l})\}_{l=1}^{L} $\hfill \text{ (hypernetwork)},\\
\midrule
Phase 2 & $\{\pmb{s}^l\}_{l=1}^L$, $W^0, W^{L+1}, \pmb{b}^0, \pmb{b}^{L+1}$ \hfill (\textsc{LR-PINN})\\
\bottomrule
\end{tabular}
\end{table}

\section{Experiments}
We demonstrate that our proposed method significantly outperforms baselines on the 1-dimensional/2-dimensional PDE benchmarks that are known to be very challenging for PINNs to learn \cite{ mcclenny2020self,krishnapriyan2021characterizing}. We report the average accuracy and refer readers to Appendix~\ref{a:std} for the std. dev. of accuracy after 3 runs.

\paragraph{Baselines for comparison:}
Along with the vanilla PINN~\cite{raissi2019physics}, our baselines include several variants. 
PINN-R~\cite{kim2021dpm} denotes a model that adds skip-connections to PINN.
PINN-S2S~\cite{krishnapriyan2021characterizing} denotes a method that uniformly segments the temporal domain and proceeds by training each segment one by one in a temporal order. PINN-P denotes a method that directly extends PINNs to take $(\pmb{x},t,\pmb{\mu})$ as input and infer $u_{\theta}(\pmb{x},t,\pmb{\mu})$, i.e., $\pmb{\mu}$ is being treated as a coordinate in the parameter domain. 

We also apply various meta-learning algorithms to Na\"ive-LR-PINN: model-agnostic meta learning (MAML) \cite{finn2017model} and Reptile \cite{nichol2018first} --- recall that Na\"ive-LR-PINN means that LR-PINN without our proposed hypernetwork-based meta-learning and therefore, MAML and Reptile on top of Na\"ive-LR-PINN can be compared to Hyper-LR-PINN. In the parameterized PDE setting, we can define a task, $\tau^{(i)}$, as a specific setting of the PDE parameters, $\pmb{\mu}^{(i)}$. 
Both MAML and Reptile seek an initial weights of a PINN, which can serve as a good starting point for gradient-based optimizers when a solution of a new unseen PDE parameter setting is sought. See Appendix~\ref{a:maml_reptile} for details. For reproducibility, we refer readers to Appendix~\ref{a:rep}, including hyperparameters and software/hardware environments.



\paragraph{Evaluation metrics:} Given the $i$-th PDE parameter instance $\pmb{\mu}^{(i)}$, the ground-truth solution evaluated at the set of test collocation points can be defined collectively as  
$
\pmb{u}^{(i)} = [u(x_1, t_1; \pmb{\mu}^{(i)}), \ldots, u(x_N, t_N; \pmb{\mu}^{(i)})]\Transpose$
and likewise for PINNs as $\pmb{u}_{\theta}^{(i)}$. Then the absolute error and the relative error can be defined as $\frac{1}{N}\| \pmb{u}^{(i)} - \pmb{u}_{\theta}^{(i)} \|_1$ and $\|\pmb{u}^{(i)} - \pmb{u}_{\theta}^{(i)} \|_2 / \| \pmb{u}^{(i)} \|_2$, respectively. In Appendix~\ref{a:fail_result}, we measure the performance on more metric: max error and explained variance score.

\begin{figure}[t]
\vspace{-1.0em}
\subfloat[Exact solution]{\includegraphics[width=0.32\columnwidth]{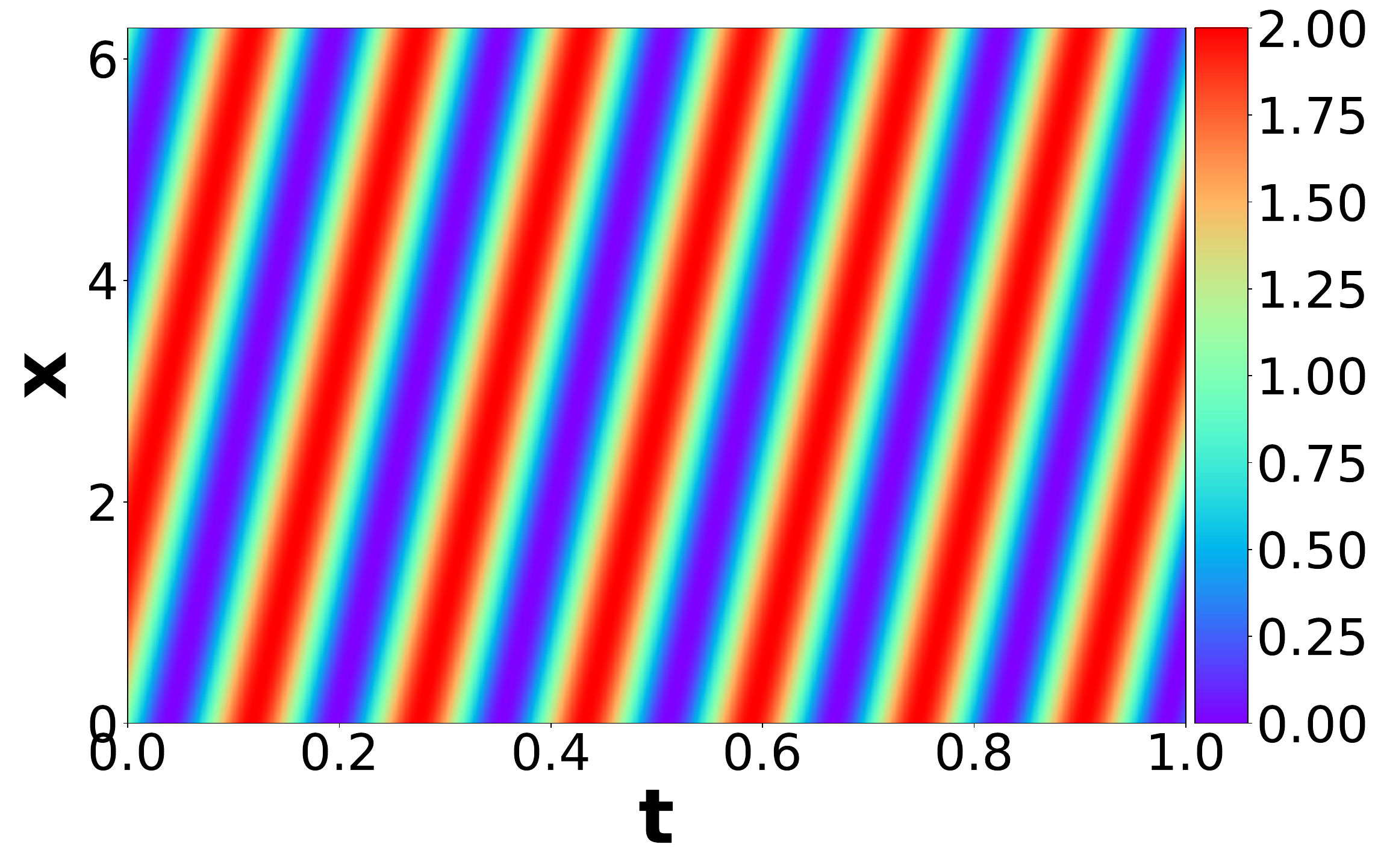}}\hfill
\subfloat[PINN]{\includegraphics[width=0.32\columnwidth]{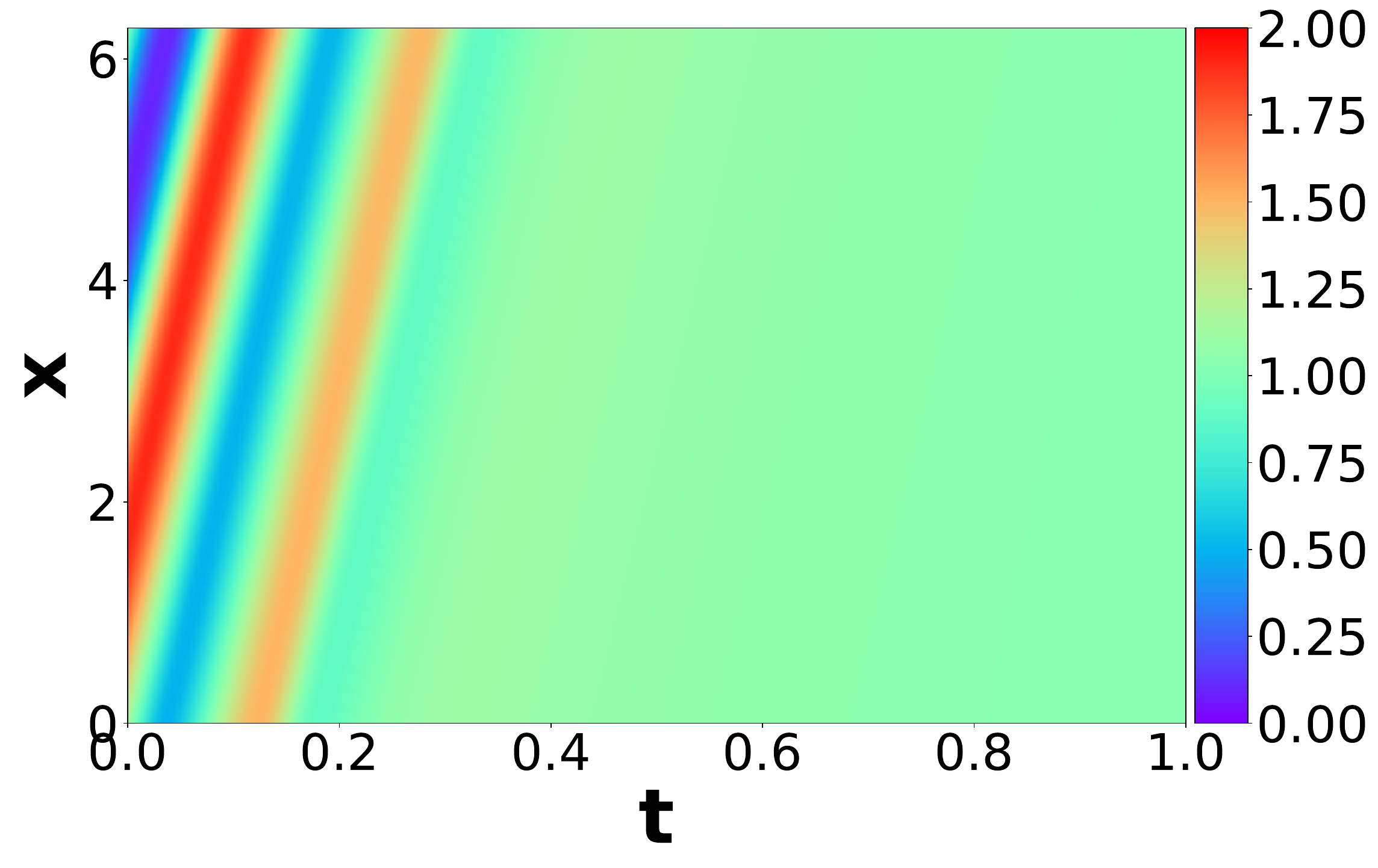}}\hfill
\subfloat[Ours]{\includegraphics[width=0.32\columnwidth]{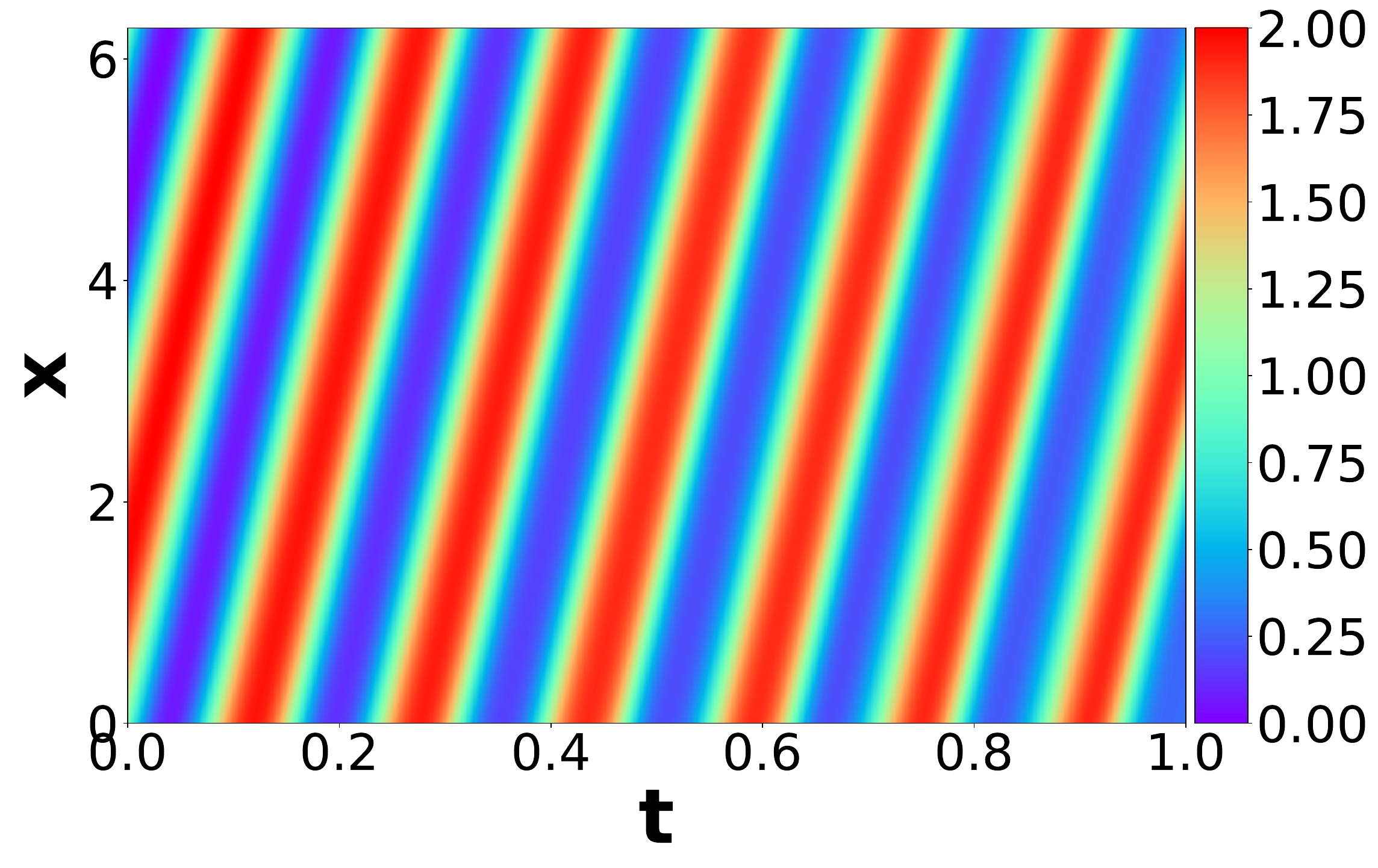}}\\
\caption{[Convection equation] Solution snapshots for $\beta=40$}\label{fig:conv_40_result}
\vspace{-.5em}
\end{figure}

\subsection{Benchmark parameterized PDEs}
\paragraph{1D PDEs:} For our first benchmark parameterized PDEs, we consider the following  parameterized convection-diffusion-reaction (CDR) equation:
\begin{equation}\label{eq:cdr}
    u_t + \beta u_x - \nu u_{xx} - \rho u (1-u) = 0, \quad x \in \Omega, \; t \in [0,T], 
\end{equation} 
where the equation describes how the state variable $u$ evolves under convective (the second term), diffusive (the third term), and reactive (the fourth term) phenomena.\footnote{Note that we consider the Fisher's form $\rho{u(1-u)}$ as the reaction term following \cite{krishnapriyan2021characterizing}.} 
The triplet, $\pmb{\mu} = (\beta, \nu, \rho)$, defines the characteristics of the CDR equations: how strong convective/diffusive/reactive the equation is, respectively. 
A particular choice of $\pmb{\mu}$ leads to a realization of a specific type of CDR processes.

There is a set of $\pmb{\mu}$ which makes 
training PINNs very challenging, known as ``failure modes'' 
\cite{krishnapriyan2021characterizing}: i) convection equations 
with high convective terms ($\beta\geq30$) and ii) reaction(-diffusion) 
equations with high reactive terms ($\rho \geq 5$). We demonstrate that our method does not require specific-PDE-dedicated algorithms (e.g., \cite{krishnapriyan2021characterizing}) while producing comparable/better accuracy in low rank.

\paragraph{2D PDEs:} 
As the second set of benchmarks, we consider the 2-dim parameterized Helmholtz equation: 
\begin{equation}
    u_{xx} + u_{yy} + k^2 u - q(x,y;a_1,a_2) = 0,
\end{equation}
where $q(x, y; a_1, a_2)$ denotes a specific parameterized forcing term and the solution $u$ can be calculated analytically (See Appendix~\ref{a:helmholtz_appendix}). As observed in the failure modes of convection equations (high convective terms), certain choices of the parameters in the forcing term, i.e., $a_1, a_2$, make the training of PINNs challenging as the solutions become highly oscillatory.


\begin{figure}[t]
    \vspace{-1.0em}
    \centering
    \subfloat[Train loss]
    {\includegraphics[width=0.33\columnwidth]
    {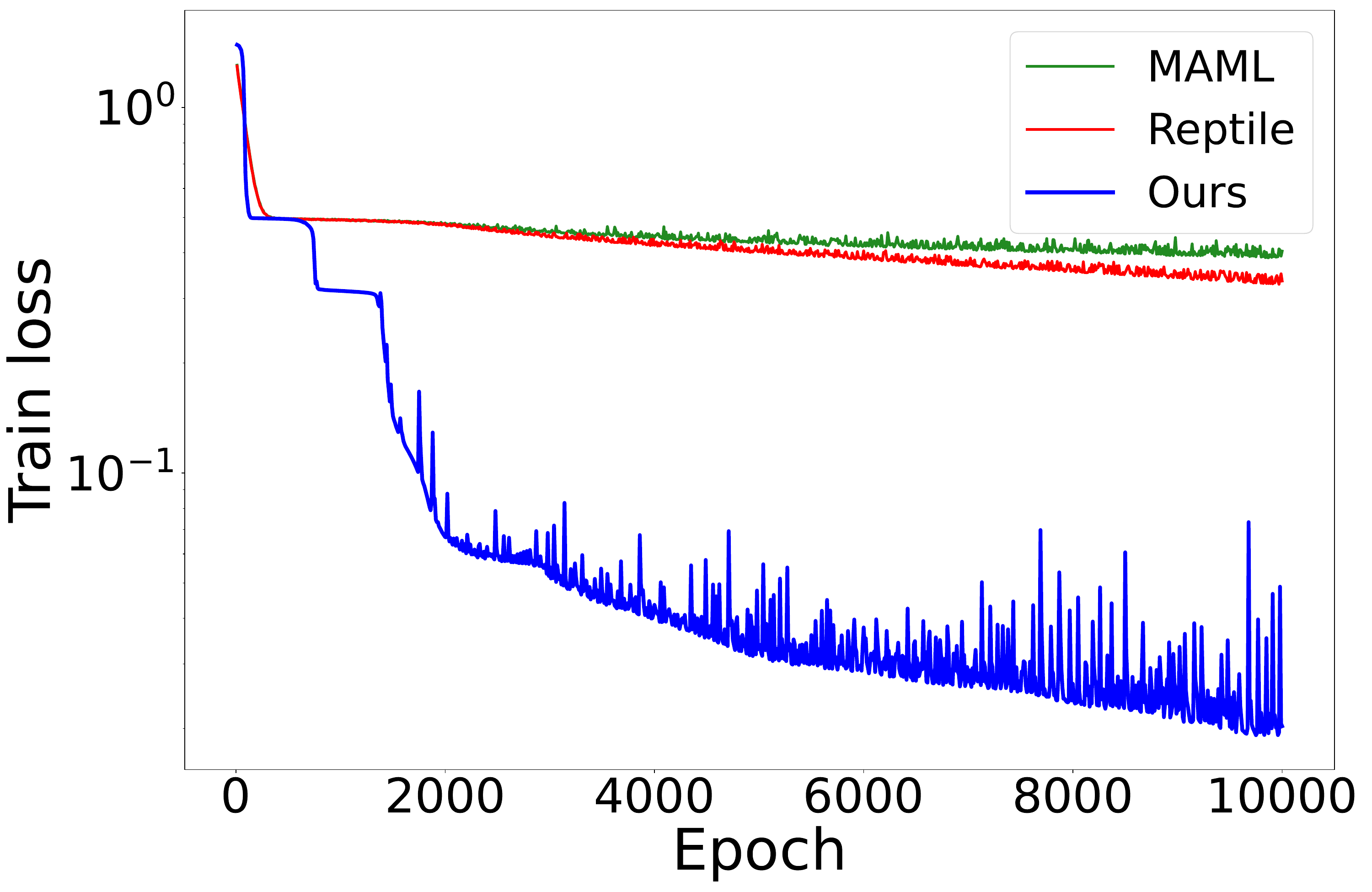}}\hfill
    \subfloat[The absolute error]
    {\includegraphics[width=0.33\columnwidth]
    {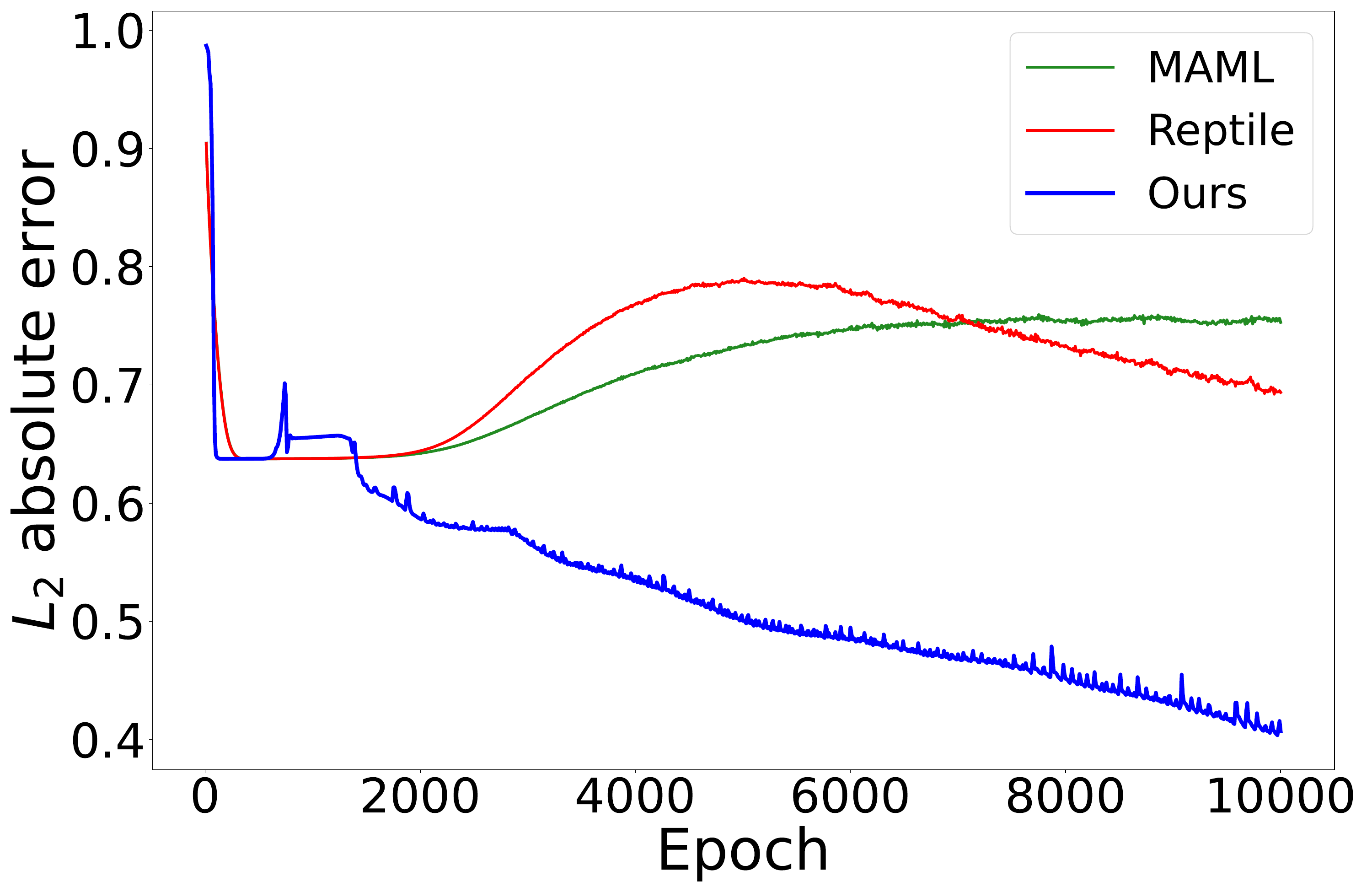}}\hfill
    \subfloat[The relative error]
    {\includegraphics[width=0.33\columnwidth]
    {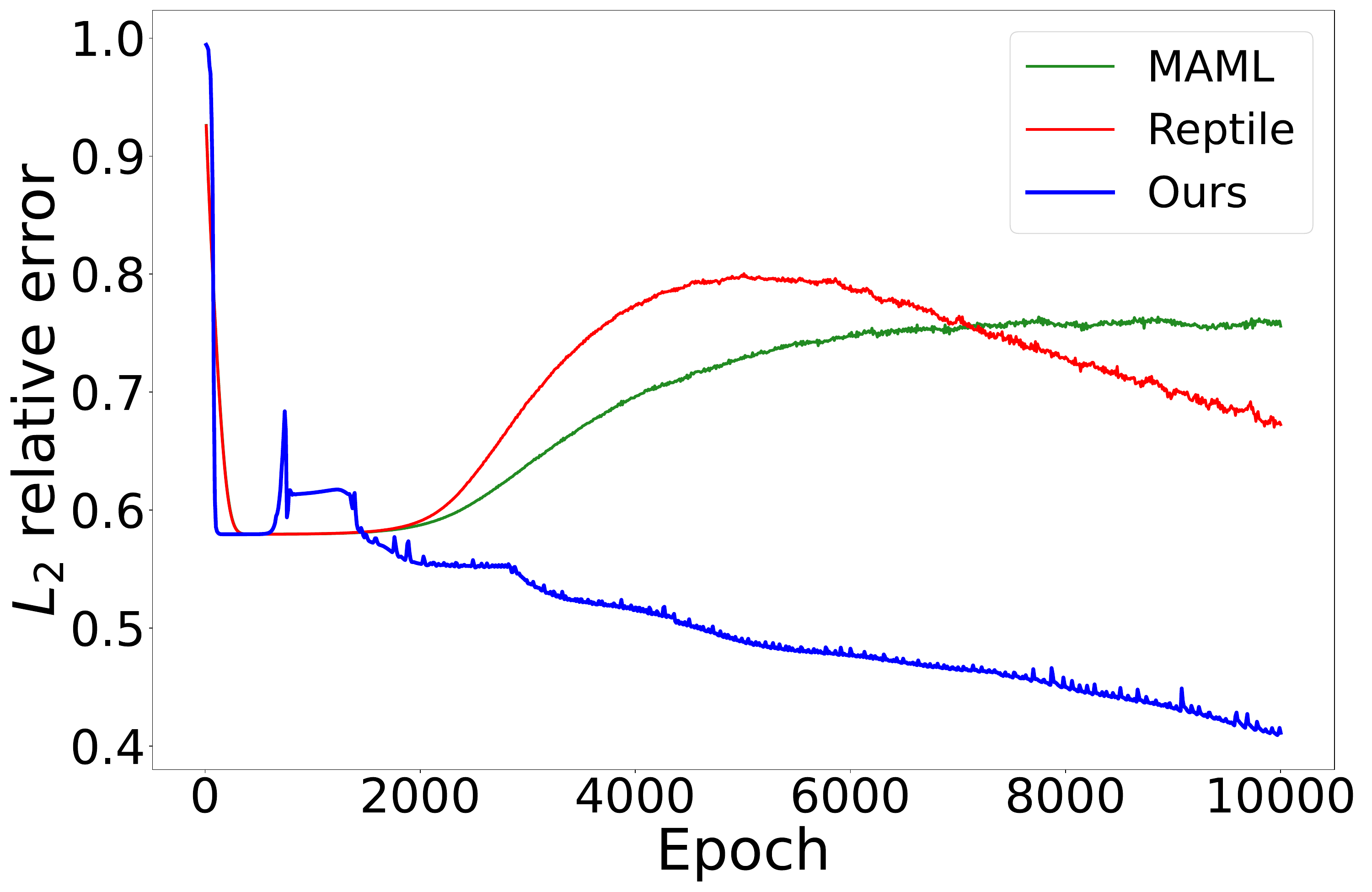}} \\
    \caption{[Convection equation] Per epoch averaged train loss and test errors for $\beta\in [30,40]$ for Phase 1 (meta-learning phase). We refer readers to Appendix \ref{a:loss_curve_appendix} for Phase 2 loss curve.}
    \label{fig:train_test_beta_30}
    \vspace{-1.0em}
\end{figure}
\begin{table}[]
\renewcommand{\arraystretch}{0.85}
\centering
    \caption{[Convection equation] The absolute and relative errors of the solutions of convection equations with $\beta = \{30,35,40\}$. We apply the curriculum learning proposed in~\cite{krishnapriyan2021characterizing}, MAML, and Reptile to Na\"ive-LR-PINN. Therefore, we focus on comparing various Na\"ive-LR-PINN-based enhancements and our Hyper-LR-PINN. See Appendix~\ref{a:fail_result} for other omitted tables.}\label{tab:result_conv_err}
\resizebox{\textwidth}{!}{
\begin{tabular}{lccccccccccccc}
\specialrule{1pt}{2pt}{2pt}
\multirow{5}{*}{$\pmb{\beta}$} & \multirow{5}{*}{\textbf{Rank}} & \multicolumn{2}{c}{\textbf{{[}w/o{]} Pre-training}} & \multicolumn{10}{c}{\textbf{{[}w{]} Pre-training}} \\ \cmidrule(lr){3-4} \cmidrule(lr){5-14}
 &  & \multicolumn{2}{c}{\textbf{Na\"ive-LR-PINN}} & \multicolumn{2}{c}{\begin{tabular}[c]{@{}c@{}}\textbf{Curriculum}\\ \textbf{learning}\end{tabular}} & \multicolumn{2}{c}{\textbf{MAML}} & \multicolumn{2}{c}{\textbf{Reptile}} & \multicolumn{2}{c}{\begin{tabular}[c]{@{}c@{}}\textbf{Hyper-LR-PINN}\\ \textbf{(Full rank)}\end{tabular}} & \multicolumn{2}{c}{\begin{tabular}[c]{@{}c@{}}\textbf{Hyper-LR-PINN}\\ \textbf{(Adaptive rank)}\end{tabular}} \\ \cmidrule(lr){3-14}
 &  & \multicolumn{1}{c}{Abs. err.} & \multicolumn{1}{c}{Rel. err.} & \multicolumn{1}{c}{Abs. err.} & \multicolumn{1}{c}{Rel. err.} & \multicolumn{1}{c}{Abs. err.} & \multicolumn{1}{c}{Rel. err.} & \multicolumn{1}{c}{Abs. err.} & \multicolumn{1}{c}{Rel. err.} & Abs. err. & Rel. err. & Abs. err. & Rel. err. \\
\specialrule{1pt}{2pt}{2pt}
\multirow{7}{*}{\textbf{30}} & 10 & 0.5617 & 0.5344 & 0.4117 & 0.4098 & 0.6757 & 0.6294 & 0.5893 & 0.5551 & \multirow{7}{*}{0.0360} & \multirow{7}{*}{0.0379} & \multirow{7}{*}{0.0375} & \multirow{7}{*}{0.0389} \\ \cmidrule(lr){2-10}
 & 20 & 0.5501 & 0.5253 & 0.4023 & 0.4005 & 0.6836 & 0.6452 & 0.6144 & 0.5779 &  &  &  &  \\ \cmidrule(lr){2-10}
 & 30 & 0.5327 & 0.5126 & 0.4233 & 0.4204 & 0.5781 & 0.5451 & 0.6048 & 0.5704 &  &  &  &  \\ \cmidrule(lr){2-10}
 & 40 & 0.5257 & 0.5076 & 0.3746 & 0.3744 & 0.5848 & 0.5515 & 0.5757 & 0.5442 &  &  &  &  \\ \cmidrule(lr){2-10}
 & 50 & 0.5327 & 0.5126 & 0.4152 & 0.4127 & 0.5898 & 0.5562 & 0.5817 & 0.5496 &  &  &  &  \\
\specialrule{1pt}{2pt}{2pt}
\multirow{7}{*}{\textbf{35}} & 10 & 0.5663 & 0.5357 & 0.5825 & 0.5465 & 0.6663 & 0.6213 & 0.5786 & 0.5446 & \multirow{7}{*}{0.0428} & \multirow{7}{*}{0.0443} & \multirow{7}{*}{0.0448} & \multirow{7}{*}{0.0461} \\ \cmidrule(lr){2-10}
 & 20 & 0.5675 & 0.5369 & 0.6120 & 0.5673 & 0.6814 & 0.6433 & 0.5971 & 0.5606 &  &  &  &  \\ \cmidrule(lr){2-10}
 & 30 & 0.6081 & 0.5670 & 0.5864 & 0.5503 & 0.5819 & 0.5466 & 0.5866 & 0.5506 &  &  &  &  \\ \cmidrule(lr){2-10}
 & 40 & 0.5477 & 0.5227 & 0.5954 & 0.5548 & 0.5809 & 0.5462 & 0.5773 & 0.5435 &  &  &  &  \\ \cmidrule(lr){2-10}
 & 50 & 0.5449 & 0.5208 & 0.6010 & 0.5619 & 0.5870 & 0.5514 & 0.5731 & 0.5404 &  &  &  &  \\
\specialrule{1pt}{2pt}{2pt}
\multirow{7}{*}{\textbf{40}} & 10 & 0.5974 & 0.5632 & 0.5978 & 0.5611 & 0.6789 & 0.6446 & 0.5992 & 0.5632 & \multirow{7}{*}{0.0603} & \multirow{7}{*}{0.0655} & \multirow{7}{*}{0.0656} & \multirow{7}{*}{0.0722} \\ \cmidrule(lr){2-10}
 & 20 & 0.5890 & 0.5563 & 0.6274 & 0.5820 & 0.7008 & 0.6801 & 0.6189 & 0.5853 &  &  &  &  \\ \cmidrule(lr){2-10}
 & 30 & 0.6142 & 0.5724 & 0.6011 & 0.5652 & 0.6072 & 0.5700 & 0.6126 & 0.5810 &  &  &  &  \\ \cmidrule(lr){2-10}
 & 40 & 0.5560 & 0.5293 & 0.6126 & 0.5715 & 0.6149 & 0.5832 & 0.6004 & 0.5638 &  &  &  &  \\ \cmidrule(lr){2-10}
 & 50 & 0.6161 & 0.5855 & 0.6130 & 0.5757 & 0.6146 & 0.5799 & 0.6007 & 0.5645 &  &  &  &  \\
\specialrule{1pt}{2pt}{2pt}
\end{tabular}}
\end{table}

\subsection{Experimental results}
\subsubsection{Performance on the failure mode of the CDR equation}
\paragraph{Solution accuracy:} As studied in prior work, vanilla PINNs fail to approximate solutions exhibiting either highly oscillatory (due to high convection) or sharp transient (due to high reaction) behaviors. Here, we present the results of 
convection equations and leave the results on reaction(-diffusion) equations in Appendix~\ref{a:fail_result}. 
For convection equations\footnote{Initial condition: $1+\sin(x)$ and boundary condition: periodic}, failures typically occur with high $\beta$, e.g., $\beta \geq 30$. 
Table~\ref{tab:result_conv_err} reports the results of all considered models trained on $\beta \in [30,40]$\footnote{Following \cite{krishnapriyan2021characterizing}, we consider only the coefficients in the natural numbers (e.g., $\beta \in \mathbb{N}_{+}$) and [30,40] indicates the set $\{30,31,\ldots,40\}$.} and essentially shows that \HLRPINN{} (Figure~\ref{fig:conv_40_result}(c)) is the only low-rank method that can resolve the failure-mode and there is only marginal decreases in accuracy compared to \HLRPINN{} in full rank. For reaction(-diffusion) equations, we observe similar results, \HLRPINN{} outperforms the best baseline by more than an order of magnitude (See Appendix~\ref{a:fail_result}). Moreover, 
we report that \HLRPINN{} outperforms in most cases to the baselines that operate in ``full-rank'': meta-learning methods (MAML, Reptile),  PINN-S2S, and PINN-P (Appendix~\ref{a:general_result}).


\paragraph{Loss curves:} Figure~\ref{fig:train_test_beta_30} depicts two curves of train loss and test errors as the meta-learning algorithms proceed (MAML, Reptile, and ours). As opposed to the optimization-based meta-learning algorithms, which tend to learn meta-initial weights that perform well ``on average'' over randomly sampled training tasks, our hypernetwork-based method minimizes the loss for each individual task simultaneously. Loss curves for Phase 2 are reported in Appendix~\ref{a:loss_curve_appendix}, which essentially show that the baseline meta-learners do not provide good initializations.

\begin{table}[t]
    \centering
    \caption{Comparisons of model size}\label{tab:rank_comp}
    \begin{tabular}{lcccccccc}
    \Xhline{2\arrayrulewidth}
    \textbf{Model} & \multicolumn{5}{c}{\textbf{Na\"ive-LR-PINN}} & \textbf{Ours} & \textbf{PINN}\\
    \Xhline{2\arrayrulewidth}
    \textbf{Rank} & 10 & 20 & 30 & 40 & 50 & Adaptive & -\\
    \Xhline{2\arrayrulewidth}
    \textbf{\# Parameters} & 381 & 411 & 441 & 471 & 501 & $\sim$351 & 10,401\\
    \Xhline{2\arrayrulewidth}
    \end{tabular}
\end{table}

\begin{figure}[t]
    \centering
    \subfloat[$\beta\in\text{[30,40]}$]
    {\includegraphics[width=0.33\columnwidth]{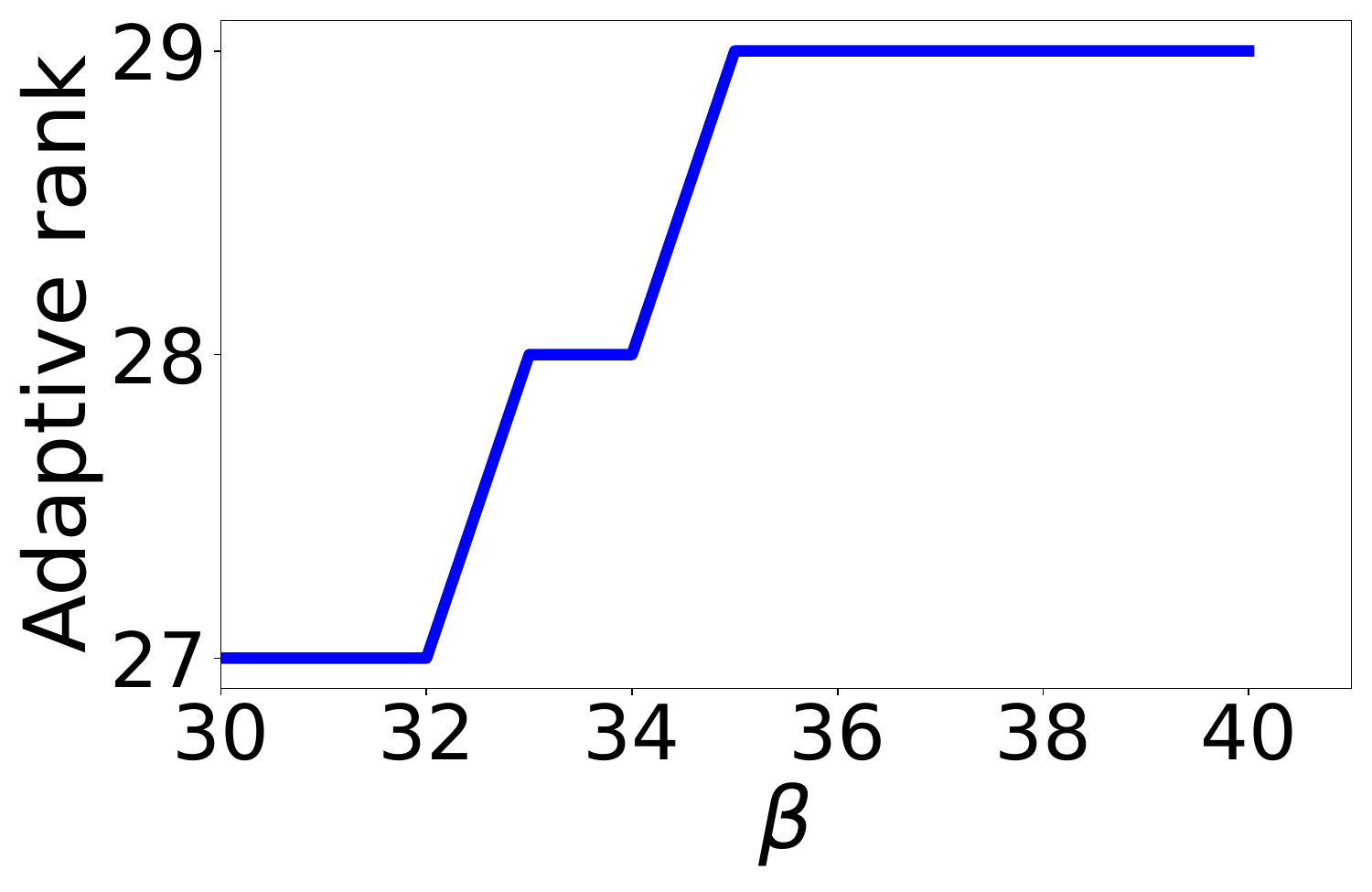}}\hfill
    \subfloat[$\beta\in\text{[1,20]}$]
    {\includegraphics[width=0.33\columnwidth]
    {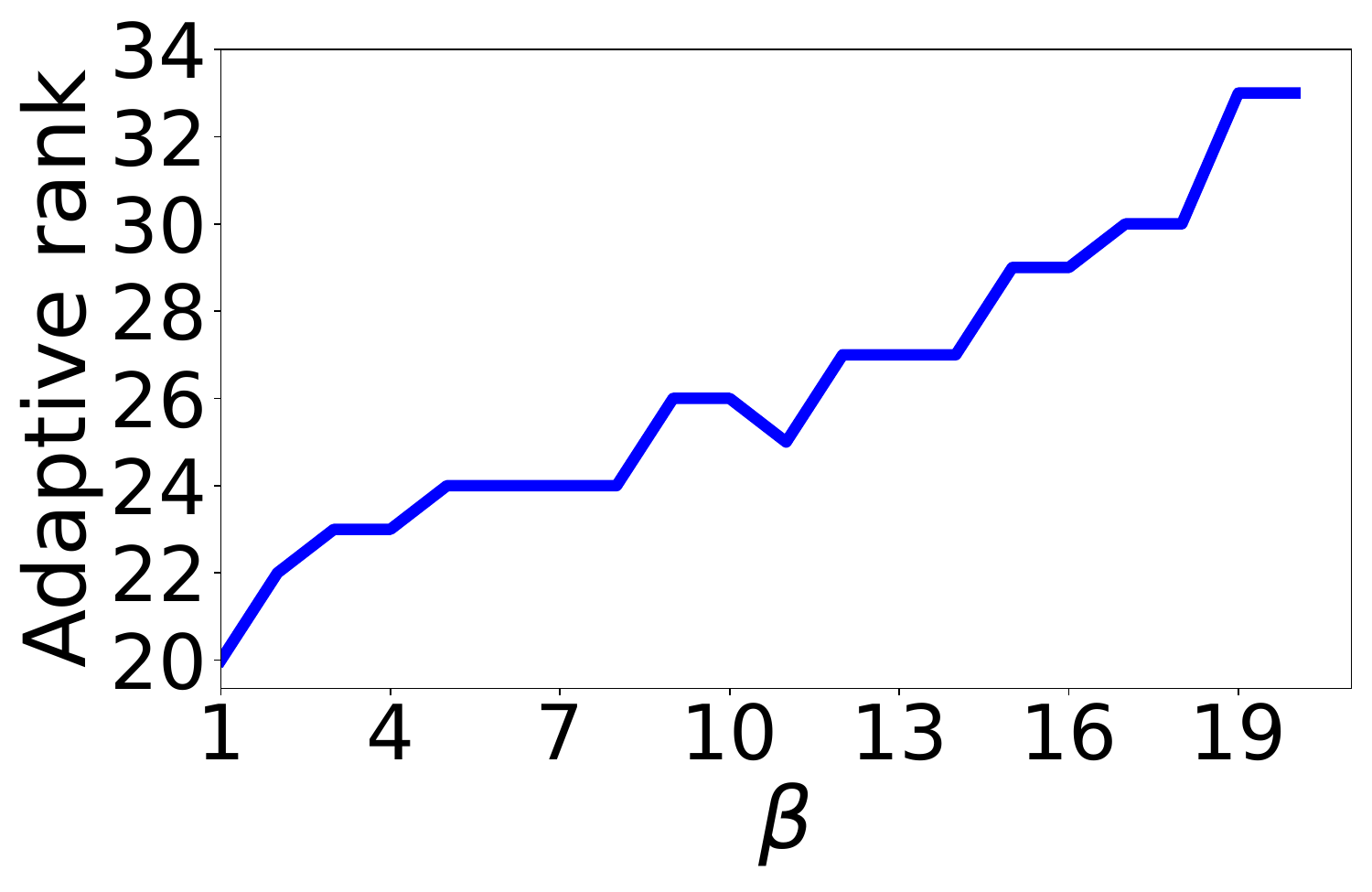}}\hfill
    \subfloat[$\beta\in\text{[1,20]}$]
    {\includegraphics[width=0.33\columnwidth]  {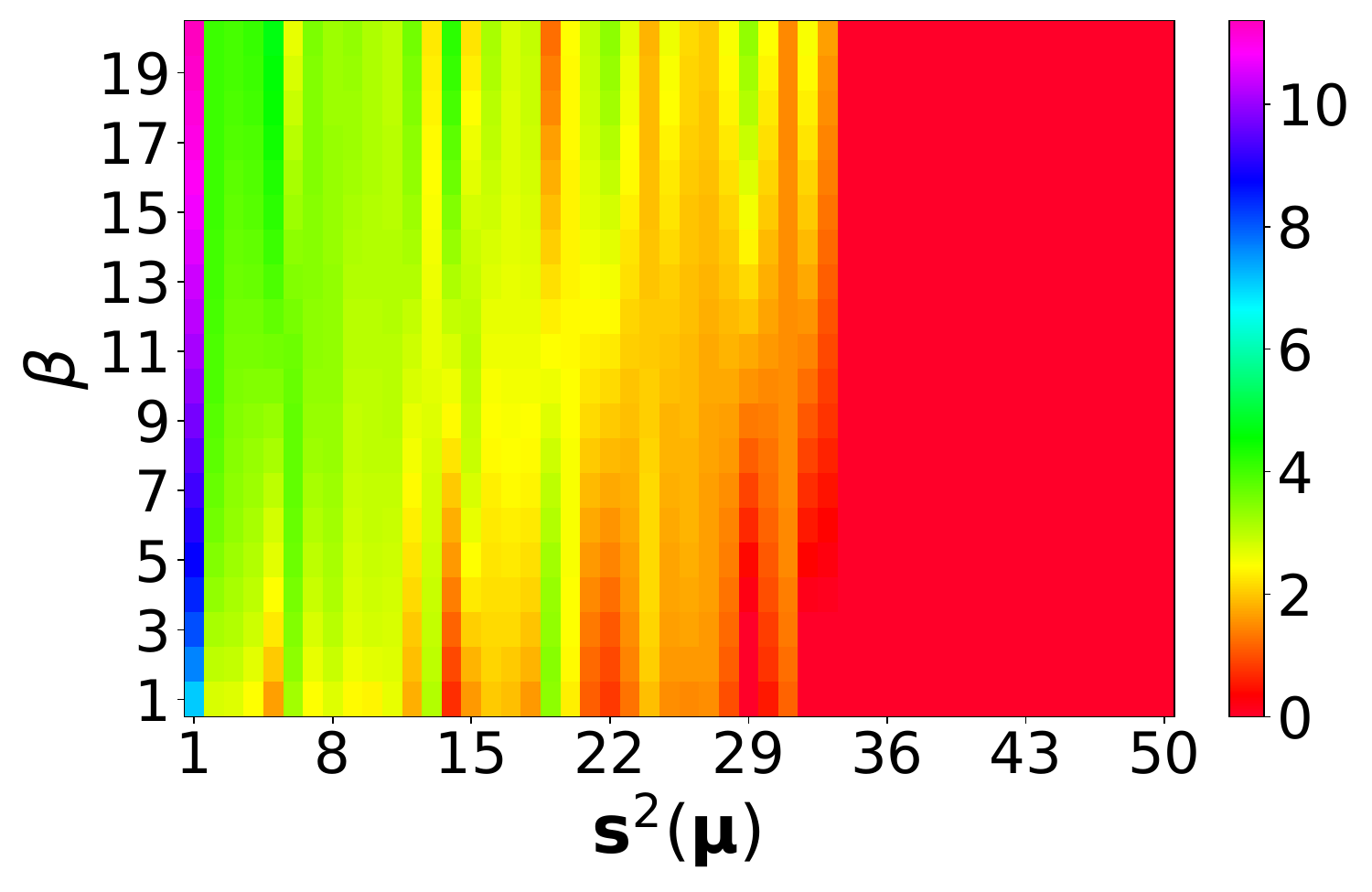}}
\\
    \caption{Adaptive rank on convection equation (the left and the middle panels).
    The magnitude of the learned diagonal elements $\pmb{s}^{2}$ of the second hidden layer for varying $\beta \in [1,20]$ (the right panel).}
    \label{fig:adap_rank_conv}    
\end{figure}


\begin{figure}[t]
\subfloat[The absolute error]{\includegraphics[width=0.32\columnwidth]{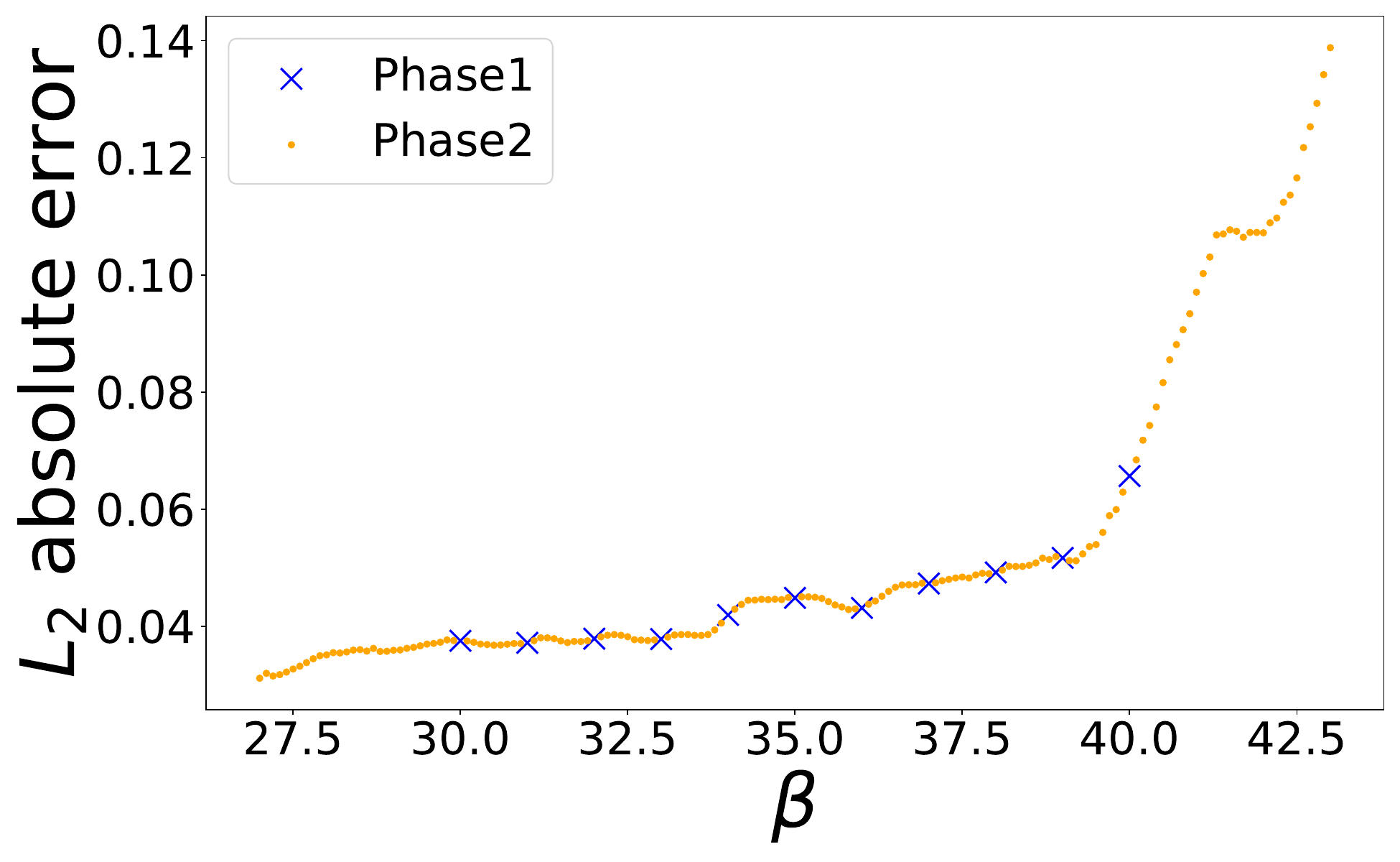}}\hfill
\subfloat[The relative error]{\includegraphics[width=0.32\columnwidth]{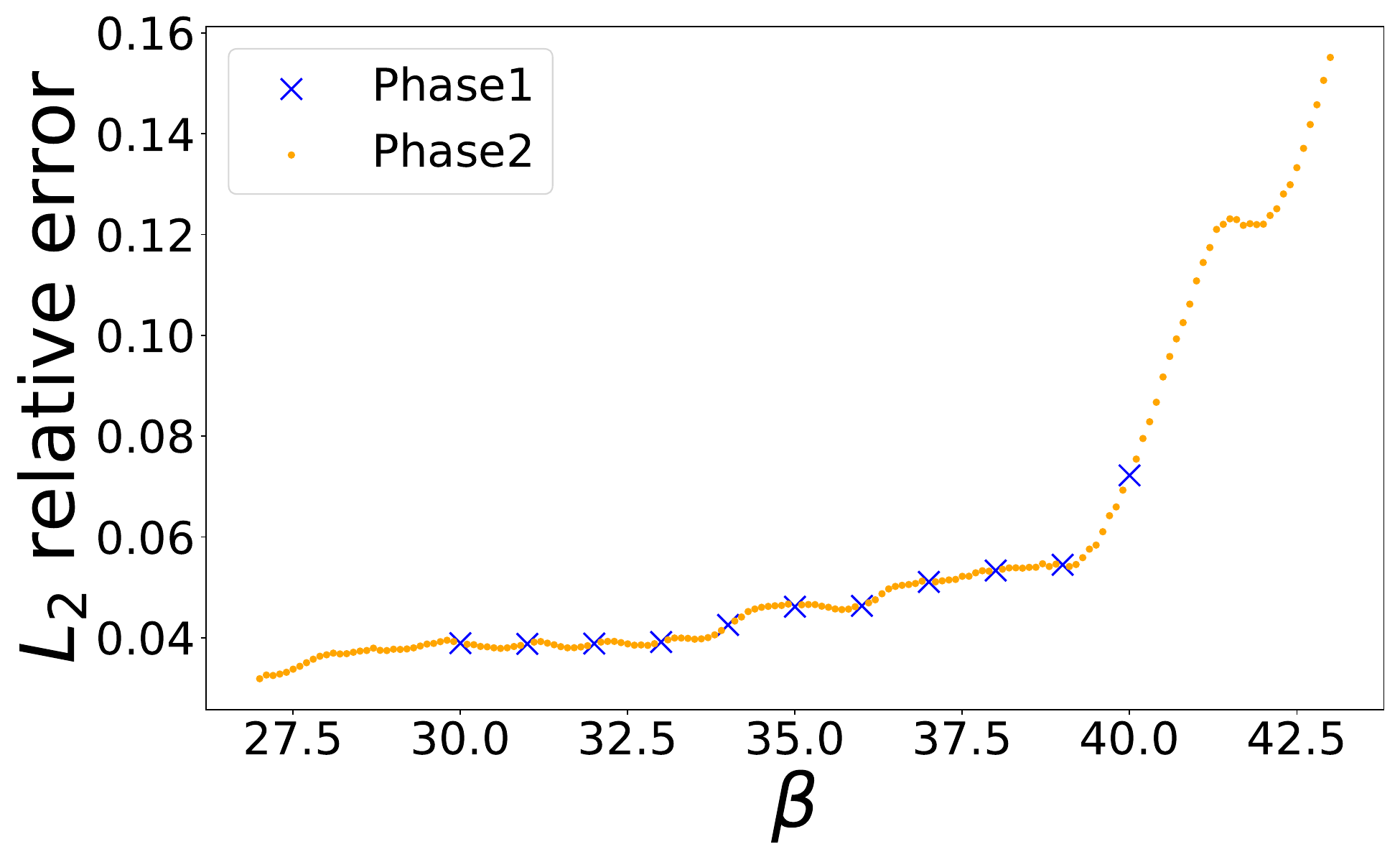}}\hfill
\subfloat[Model size]{\includegraphics[width=0.32\columnwidth]{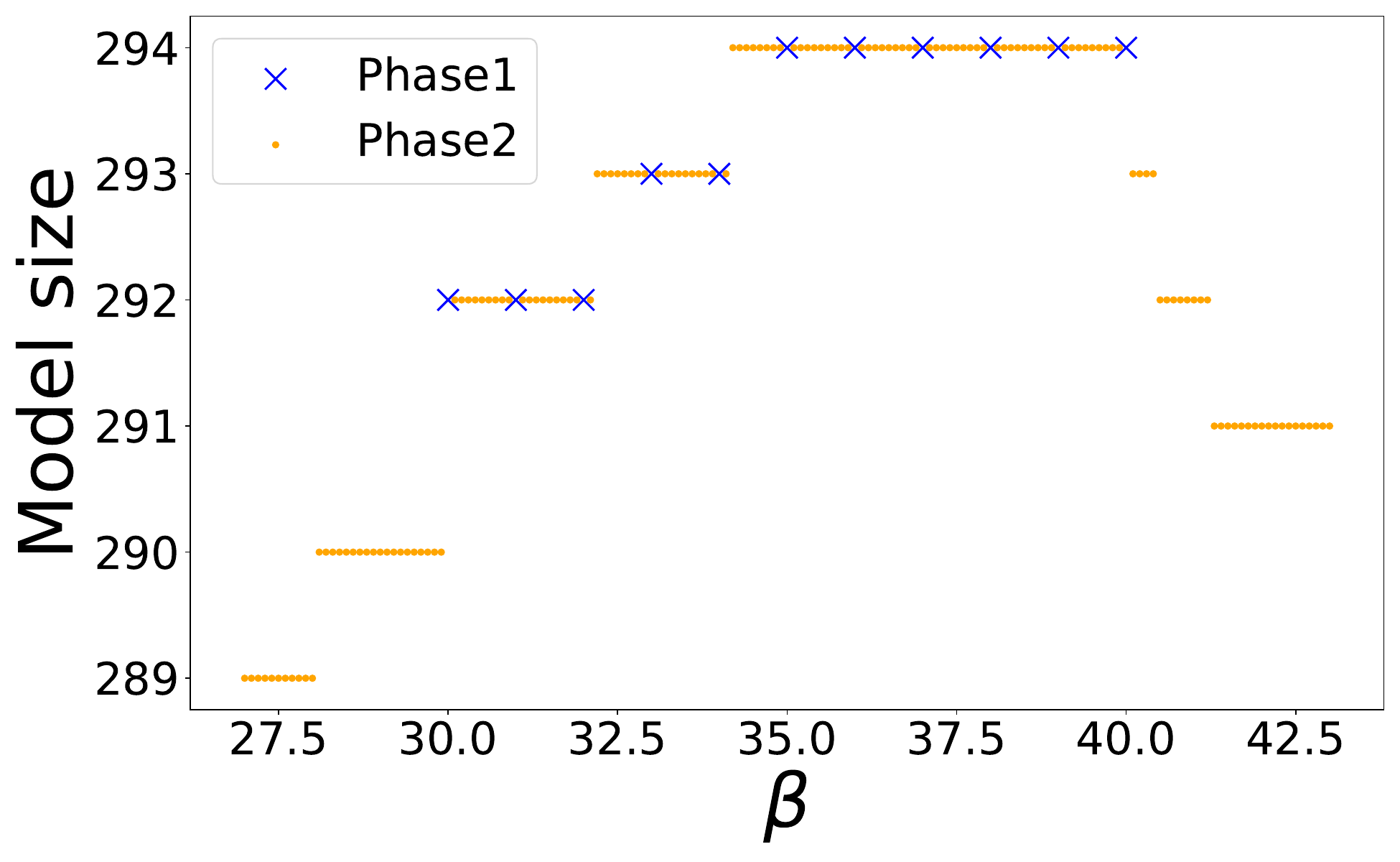}}\\
\caption{Multi-query scenario: the absolute/relative errors of the predicted solutions at unseen test parameters (left and middle), and the number of trainable model parameters (right).}\label{fig:interpolation_task}
\end{figure}

\paragraph{Rank structure:} Table~\ref{tab:rank_comp} compares the number of trainable model parameters for \NaiveLRPINN, PINN, and our method. In our model, each hidden layer has a different rank structure, leading to 351 trainable parameters, which is about ($\times$30) smaller than that of the vanilla PINN, Cf. merely decreasing the model size of PINNs lead to poor performance (Appendix~\ref{sec:other_exps}). Figure~\ref{fig:adap_rank_conv}(a) shows how the model adaptively learns the rank structure for varying values of $\beta$. We observe that, with the low-rank format, Hyper-LR-PINN provides more than ($\times 4$) speed up 
 in training time compared to the vanilla PINN. More comparisons on this aspect will be presented in the general CDR settings.

\paragraph{Ablation on fixed or learnable basis vectors:} We compare the performance of the cases, where the basis vectors $\{U_r^l, V_r^l\}$ are trainable. We observe that the fixed basis vectors lead to an order of magnitude more accurate predictions as well as faster convergence (See Appendix~\ref{a:learn_fixed_basis} for plots). 

\paragraph{Ablation on orthogonality constraint:} We compare the performances of the cases, where the proposed model is trained without the orthogonality constraint, Eq.~\eqref{eq:ortho_const}. Experimental results show that the orthogonality constraint is essential in achieve good performance (See Appendix~\ref{a:ablation}). 

\paragraph{Many-query settings (inter/extra-polation in the PDE parameter domain):} Next, we test our algorithm on the \textit{many-query} scenario, 
where we train the proposed framework with the 11 training PDE parameters $\beta\in [30,40]$ (Phase 1), freeze the hypernetwork and the learned basis, and retrain only the diagonal elements for 150 unseen test PDE parameters (Phase 2). Figure~\ref{fig:interpolation_task} depicts the absolute and relative errors of the solutions, which are as accurate as the ones on the training PDE parameters, in particular, for $\beta < 30$ (i.e., extrapolation). Although the model produces an increasing error for  $\beta>40$, the relative error is still around $\sim$10\%, which cannot be achievable by vanilla PINNs. The right chart of Figure~\ref{fig:interpolation_task} shows the number of trainable model parameters, which is determined by the rank (i.e., the output of the hypernetwork). 
As PINNs fail to produce any reasonable approximation in this $\beta$ regime, we make comparisons in the general CDR settings below.

\subsubsection{Performance on the general case of the CDR equation}
\paragraph{Solution accuracy:} 
Next, we present the performance of the proposed method on solving the benchmark PDEs (not only in the failure mode but) in all available settings for $\beta, \nu, \rho$. Due to space reasons, we present only a subset of our results and leave the detailed results in Appendix~\ref{a:general_result}. Table~\ref{tab:l2_subset} shows that the proposed \HLRPINN{} mostly outperforms the baselines that operate in full rank. 

\paragraph{Some observations on learned low-rank structures:} 
Again, the proposed method learns rank structures that are adaptive to the PDE parameter (Figure~\ref{fig:adap_rank_conv}(b)). The learned rank structures vary more dynamically ($r$ from 20 to 34) as the PDE parameter range [1,20] is larger compared to [30,40]. 

Figure
~\ref{fig:adap_rank_conv}(c)
visualizes the magnitude of the learned diagonal elements $\pmb{s}^{2}$ in the second hidden layer. The horizontal axis indicates the index of the element in $\pmb{s}^2$, which are sorted in descending order for analysis, and the vertical axis indicates the $\beta$ value. The depicted result agrees with our intuitions in many aspects: i) the values decay fast (indicating there exist low-rank structures), ii) the value of each element $\pmb{s}_i^2$ either increases or decreases gradually as we vary $\beta$, and iii) for higher $\beta$, higher frequency basis vectors are required to capture more oscillatory behavior exhibited in the solutions, which leads to higher ranks. 
We  report the information for other layers and other problems in Appendix~\ref{a:diag_heat}. 

\begin{wrapfigure}{r}{0.25\textwidth}
\centering
    \vspace{-1.5em}
    \scriptsize
    \setlength{\tabcolsep}{5pt}
    \includegraphics[width=0.25\textwidth]{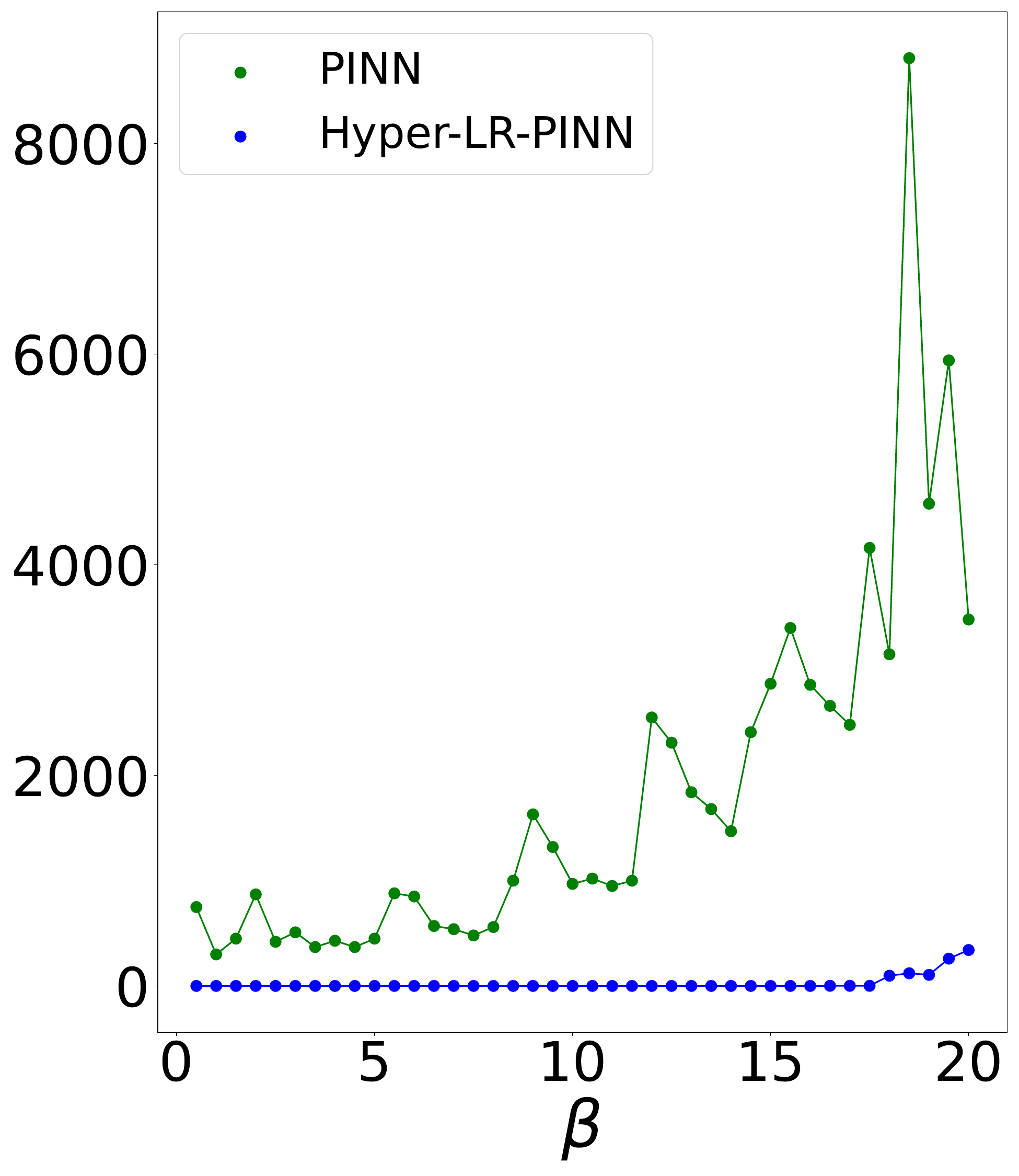}
    \vspace{-1em}
    \caption{Computational cost in epochs}
    \vspace{-4em}
    \label{fig:compare_20}
\end{wrapfigure}

\paragraph{Computational cost:} 
Here, we compare the computational cost of Hyper-LR-PINN and PINN in a \textit{many-query} scenario. We train Hyper-LR-PINN for $\beta \in [1,20]$ with interval 1 (Phase 1) and perform Phase 2 on $\beta \in [1,20]$ with interval 0.5 (including unseen test PDE parameters). We recorded the number of epochs required for PINN and Hyper-LR-PINN to reach an L2 absolute error of less than 0.05. Figure~\ref{fig:compare_20} reports the number of required epochs for each $\beta$, showing that the number of epochs required for Hyper-LR-PINN slowly increases while that for PINNs increases rapidly; for most cases, Hyper-LR-PINN reaches the target (i.e., 0.05) in only one epoch. We also emphasize that Hyper-LR-PINN in Phase 2 requires much less FLOPS in each epoch (See Section~\ref{sec:train_alg}).


\begin{table}[t]
\centering
\small
\vspace{-0.5em}
\renewcommand{\arraystretch}{0.5}
\caption{The relative errors of the solutions of parameterized PDEs with $\beta, \nu, \rho\in[1,5]$. The initial condition of each equation is Gaussian distribution $N(\pi, (\pi/2)^2)$.}\label{tab:l2_subset}
\begin{tabular}{lcccccc}

\specialrule{1pt}{2pt}{2pt}
 \multirow{3}{*}{\textbf{PDE type}} & \multicolumn{3}{c}{\textbf{No pre-training}} & \multicolumn{3}{c}{\textbf{Meta-learning}} \\ \cmidrule{2-4} \cmidrule{5-7}
 &  \textbf{PINN}  & \textbf{PINN-P} & \textbf{PINN-S2S} &  \textbf{MAML} & \textbf{Reptile} & \textbf{Ours} \\

\specialrule{1pt}{2pt}{2pt}

\textbf{Convection} & 0.0327  & 0.0217 & 0.2160 & 0.1036 & 0.0347 & 0.0085 \\
\cmidrule{1-7}
\textbf{Reaction} & 0.3907  & 0.2024 & 0.5907  &  0.0057 & 0.0064 & 0.0045 \\
\cmidrule{1-7}
\textbf{Conv-Diff-Reac} & 0.2210  & 0.2308 & 0.5983  &  0.0144 & 0.0701 & 0.0329 \\
\specialrule{1pt}{2pt}{2pt}
\end{tabular}
\end{table}

\subsubsection{Performance on the 2D Helmholtz equation}
Now we report experimental results on 2D Helmholtz equations for varying $a_1$ and $a_2$.  Under the same condition in Table~\ref{tab:result_conv_err}, we train Hyper-LR-PINN for $a\in [2, 3]$ with interval 0.1 $(a=a_1=a_2)$, and compare the performance with PINN. Surprisingly, Hyper-LR-PINN approximates the solution with high precision in only 10 epochs in Phase 2, while PINN struggles to find accurate solutions over 2,000 epochs (Figure~\ref{fig:helmholtz_main}). We show more visualizations and detailed results in Appendix~\ref{a:helmholtz_appendix}.

\begin{figure}[ht!]
\vspace{-1.0em}

\subfloat[PINN (Abs.err.=0.7403)]{\includegraphics[width=0.3\columnwidth]{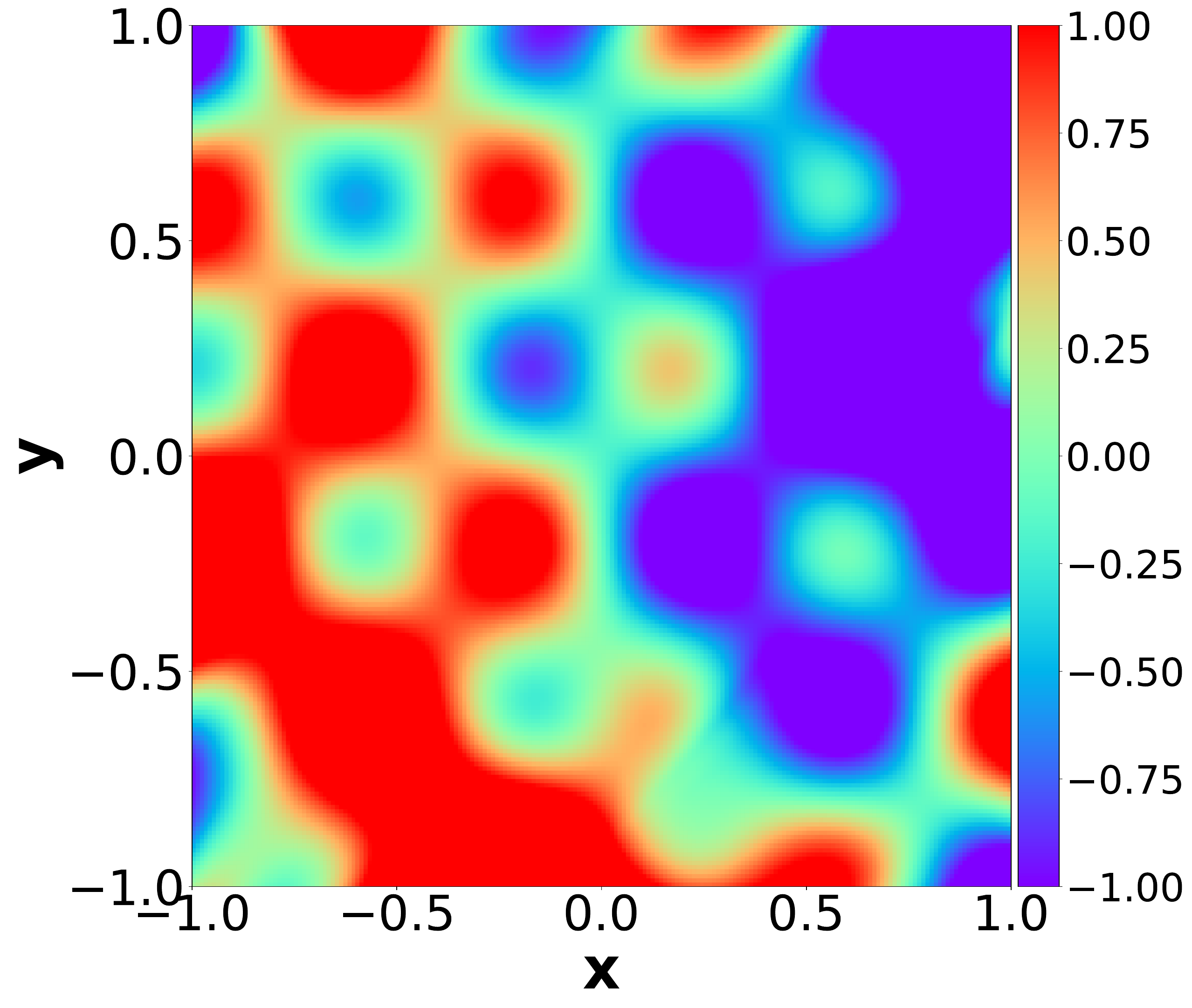}}\hfill
\subfloat[Ours (Abs.err.=0.0285)]{\includegraphics[width=0.3\columnwidth]{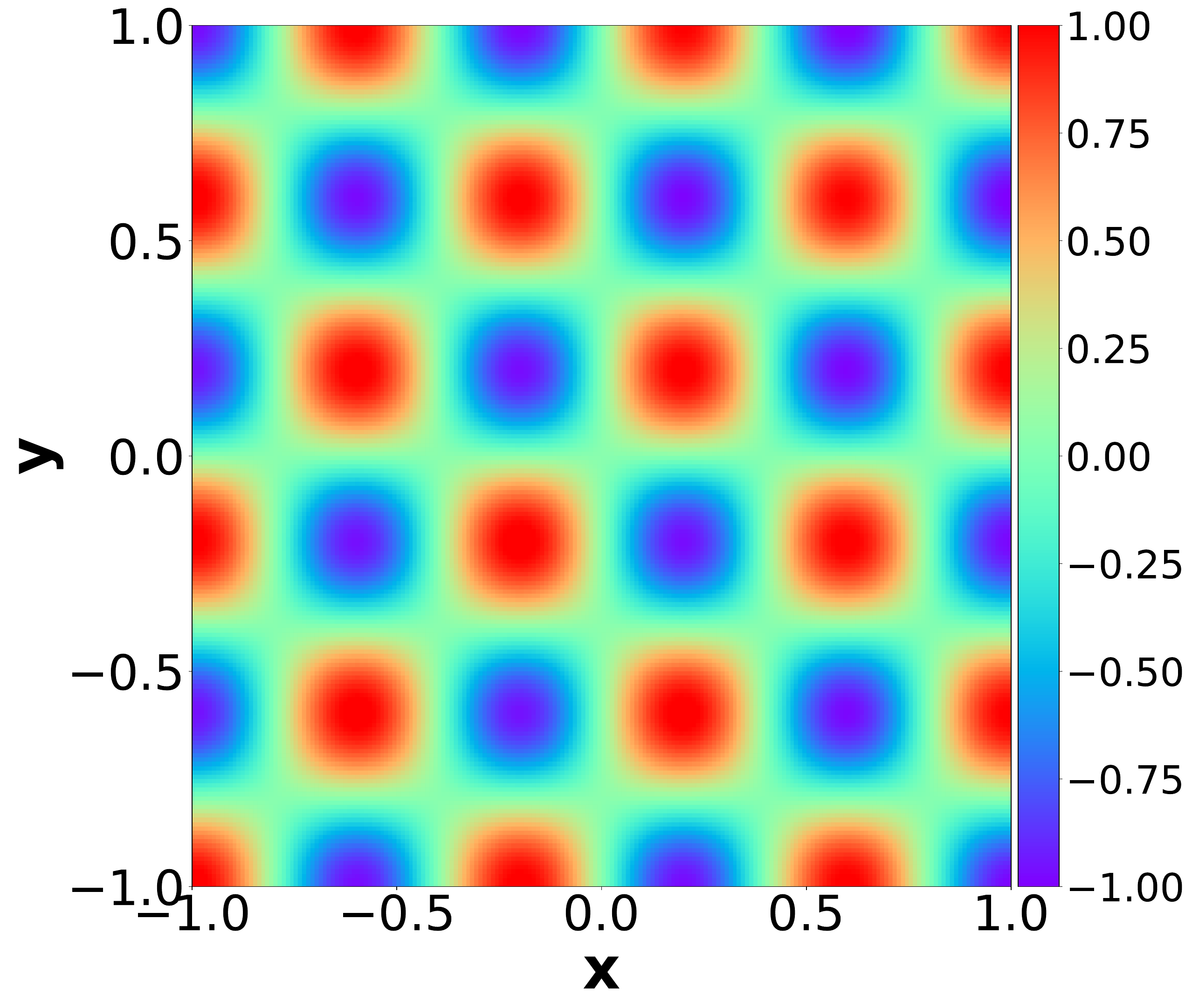}}\hfill
\subfloat[Exact solution]{\includegraphics[width=0.3\columnwidth]{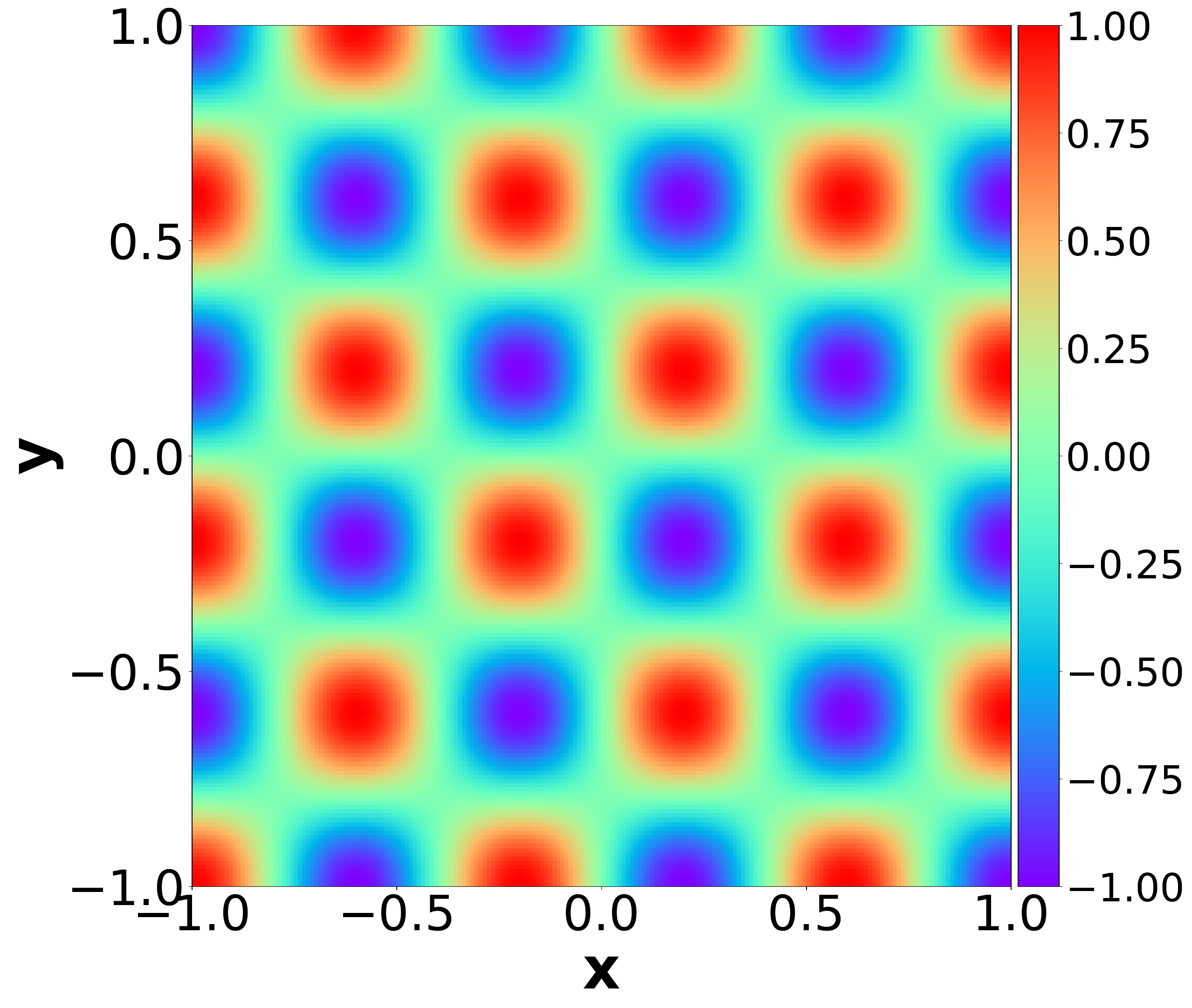}}\\

\caption{[2D-Helmholtz equation] Solution snapshots for $a=2.5$}\label{fig:helmholtz_main}
\vspace{-.5em}
\end{figure}

\section{Related Work}

\paragraph{Meta-learning of PINNs and INRs:} HyperPINNs \cite{de2021hyperpinn} share some commonalities with our proposed method in that they generate model parameters of PINNs via hypernetwork \cite{ha2016hypernetworks}. However, hyperPINNs can only generate full-rank weights and do not have capabilities to handle parameterized PDEs. Another relevant  work is optimization-based meta-learning algorithms; representatively MAML \cite{finn2017model} and Reptile \cite{nichol2018first}. 
In the area of INRs, meta-learning methods via MAML and Reptile for obtaining initial weights for INRs have been studied in \cite{tancik2021learned}.  In \cite{lee2021meta,schwarz2022meta}, meta-learning methods, which are based on MAML, have been further extended to obtain sparse representation of INRs.  

\paragraph{Low-rank formats in neural networks:} 
In natural language processing, the models being employed (e.g., \text{Bert}) typically have hundreds of millions of model parameters and, thus, making the computation efficient during the inference is one of the imminent issues. As a remedy, approximating layers in low-rank via truncated SVD has been studied \cite{chen2021drone,hu2021lora}. Modeling layers in low-rank in general has been studied for MLPs \cite{sainath2013low,zhang2014extracting,zhao2016low,khodak2020initialization} and convolutional neural network architectures \cite{jaderberg2014speeding}. In the context of PINNs or INRs, there is no low-rank format has been investigated. The work that is the closest to our work is SVD-PINNs \cite{gao2022svd}, which represents the hidden layers in the factored form as in Eq.~\eqref{eq:factored_fc}, but always in full-rank. 

\paragraph{PINNs and their variants:} 
There have been numerous sequels improving PINNs \cite{raissi2019physics} in many different aspects. One of the main issues is the multi-term objectives; in \cite{kim2021dpm}, a special gradient-based optimizer that balances multiple objective terms has been proposed, and in \cite{wang2021understanding}, a network architecture that enforce boundary conditions by design has been studied. Another main issue is that training PINNs often fails on certain classes of PDEs (e.g., fail to capture sharp transition in solutions over the spatial/temporal domains). This is due to the spectral bias \cite{rahaman2019spectral} and, as a remedy, in \cite{krishnapriyan2021characterizing}, training approaches that gradually increase the level of difficulties in training PINNs have been proposed. There are also other types of improvements including Bayesian-version of PINNs \cite{yang2021b} and PINNs that enforce conservation laws \cite{jagtap2020conservative}. All these methods, however, require training from scratch when the solution of a new PDE is sought.  

\section{Conclusion}
In this paper, we propose a low-rank formatted physics-informed neural networks (PINNs) and a hypernetwork-based meta-learning algorithm to solve parameterized partial differential equations (PDEs). Our two-phase method learns a common set of basis vectors and adaptive rank structure for varying PDE parameters in Phase 1 and approximate the solutions for unseen PDE parameters by updating only the coefficients of the learned basis vectors in Phase 2. From the extensive numerical experiments, we have demonstrated that the proposed method outperforms baselines  in terms of accuracy and computational/memory efficiency and does not suffer from the failure modes for various parameterized PDEs. 

\paragraph{Limitations:} Our Hyper-LR-PINN primarily concentrates on the cases where the PDE parameters are the coefficients of PDE terms. As a future study, we plan to extend our framework to more general settings, where the PDE parameters define initial/boundary conditions, or the shape of domains. See Appendix~\ref{a:lim} for more details.

\paragraph{Broader impacts:}

Hyper-LR-PINN can solve many equations even for PINN's failure modes with high precision.
However, despite its small errors, one should exercise caution when applying it to real-world critical applications.

\section*{Acknowledgments}
This work was supported by the Institute of Information \& Communications Technology Planning \& Evaluation (IITP) grant funded by the Korean government (MSIT) (No. 2020-0-01361, Artificial Intelligence Graduate School Program ay Yonsei University, 10\%), and (No.2022-0-00857, Development of AI/databased financial/economic digital twin platform, 90\%).





\bibliographystyle{unsrt}
\bibliography{main}

\begin{thebibliography}{10}

\bibitem{raissi2019physics}
Maziar Raissi, Paris Perdikaris, and George~E Karniadakis.
\newblock Physics-informed neural networks: {A} deep learning framework for solving forward and inverse problems involving nonlinear partial differential equations.
\newblock {\em Journal of Computational physics}, 378:686--707, 2019.

\bibitem{mao2020physics}
Zhiping Mao, Ameya~D Jagtap, and George~Em Karniadakis.
\newblock Physics-informed neural networks for high-speed flows.
\newblock {\em Computer Methods in Applied Mechanics and Engineering}, 360:112789, 2020.

\bibitem{yang2019predictive}
XIA Yang, Suhaib Zafar, J-X Wang, and Heng Xiao.
\newblock Predictive large-eddy-simulation wall modeling via physics-informed neural networks.
\newblock {\em Physical Review Fluids}, 4(3):034602, 2019.

\bibitem{sahli2020physics}
Francisco Sahli~Costabal, Yibo Yang, Paris Perdikaris, Daniel~E Hurtado, and Ellen Kuhl.
\newblock Physics-informed neural networks for cardiac activation mapping.
\newblock {\em Frontiers in Physics}, 8:42, 2020.

\bibitem{zhang2022analyses}
Enrui Zhang, Ming Dao, George~Em Karniadakis, and Subra Suresh.
\newblock Analyses of internal structures and defects in materials using physics-informed neural networks.
\newblock {\em Science advances}, 8(7):eabk0644, 2022.

\bibitem{ma2021deep}
Wei Ma, Zhaocheng Liu, Zhaxylyk~A Kudyshev, Alexandra Boltasseva, Wenshan Cai, and Yongmin Liu.
\newblock Deep learning for the design of photonic structures.
\newblock {\em Nature Photonics}, 15(2):77--90, 2021.

\bibitem{kressner2011low}
Daniel Kressner and Christine Tobler.
\newblock Low-rank tensor {K}rylov subspace methods for parametrized linear systems.
\newblock {\em SIAM Journal on Matrix Analysis and Applications}, 32(4):1288--1316, 2011.

\bibitem{grasedyck2013literature}
Lars Grasedyck, Daniel Kressner, and Christine Tobler.
\newblock A literature survey of low-rank tensor approximation techniques.
\newblock {\em GAMM-Mitteilungen}, 36(1):53--78, 2013.

\bibitem{lee2017preconditioned}
Kookjin Lee and Howard~C Elman.
\newblock A preconditioned low-rank projection method with a rank-reduction scheme for stochastic partial differential equations.
\newblock {\em SIAM Journal on Scientific Computing}, 39(5):S828--S850, 2017.

\bibitem{bachmayr2018parametric}
Markus Bachmayr, Albert Cohen, and Wolfgang Dahmen.
\newblock Parametric {PDEs: S}parse or low-rank approximations?
\newblock {\em IMA Journal of Numerical Analysis}, 38(4):1661--1708, 2018.

\bibitem{lee2020model}
Kookjin Lee and Kevin~T Carlberg.
\newblock Model reduction of dynamical systems on nonlinear manifolds using deep convolutional autoencoders.
\newblock {\em Journal of Computational Physics}, 404:108973, 2020.

\bibitem{holmes2012turbulence}
Philip Holmes, John~L Lumley, Gahl Berkooz, and Clarence~W Rowley.
\newblock {\em Turbulence, coherent structures, dynamical systems and symmetry}.
\newblock Cambridge university press, 2012.

\bibitem{benner2015survey}
Peter Benner, Serkan Gugercin, and Karen Willcox.
\newblock A survey of projection-based model reduction methods for parametric dynamical systems.
\newblock {\em SIAM review}, 57(4):483--531, 2015.

\bibitem{chen2021drone}
Patrick Chen, Hsiang-Fu Yu, Inderjit Dhillon, and Cho-Jui Hsieh.
\newblock {Drone: Data-aware low-rank compression for large NLP models}.
\newblock {\em Advances in neural information processing systems}, 34:29321--29334, 2021.

\bibitem{hu2021lora}
Edward~J Hu, Phillip Wallis, Zeyuan Allen-Zhu, Yuanzhi Li, Shean Wang, Lu~Wang, Weizhu Chen, et~al.
\newblock {LoRA: L}ow-rank adaptation of large language models.
\newblock In {\em International Conference on Learning Representations}, 2021.

\bibitem{yang2020learning}
Huanrui Yang, Minxue Tang, Wei Wen, Feng Yan, Daniel Hu, Ang Li, Hai Li, and Yiran Chen.
\newblock Learning low-rank deep neural networks via singular vector orthogonality regularization and singular value sparsification.
\newblock In {\em Proceedings of the IEEE/CVF conference on computer vision and pattern recognition workshops}, pages 678--679, 2020.

\bibitem{mcclenny2020self}
Levi McClenny and Ulisses Braga-Neto.
\newblock Self-adaptive physics-informed neural networks using a soft attention mechanism.
\newblock {\em arXiv preprint arXiv:2009.04544}, 2020.

\bibitem{krishnapriyan2021characterizing}
Aditi Krishnapriyan, Amir Gholami, Shandian Zhe, Robert Kirby, and Michael~W Mahoney.
\newblock Characterizing possible failure modes in physics-informed neural networks.
\newblock {\em Advances in Neural Information Processing Systems}, 34:26548--26560, 2021.

\bibitem{kim2021dpm}
Jungeun Kim, Kookjin Lee, Dongeun Lee, Sheo~Yon Jhin, and Noseong Park.
\newblock Dpm: A novel training method for physics-informed neural networks in extrapolation.
\newblock In {\em Proceedings of the AAAI Conference on Artificial Intelligence}, pages 8146--8154, 2021.

\bibitem{finn2017model}
Chelsea Finn, Pieter Abbeel, and Sergey Levine.
\newblock Model-agnostic meta-learning for fast adaptation of deep networks.
\newblock In {\em International conference on machine learning}, pages 1126--1135. PMLR, 2017.

\bibitem{nichol2018first}
Alex Nichol, Joshua Achiam, and John Schulman.
\newblock On first-order meta-learning algorithms.
\newblock {\em arXiv preprint arXiv:1803.02999}, 2018.

\bibitem{de2021hyperpinn}
Filipe de~Avila Belbute-Peres, Yi-fan Chen, and Fei Sha.
\newblock {HyperPINN: L}earning parameterized differential equations with physics-informed hypernetworks.
\newblock In {\em The Symbiosis of Deep Learning and Differential Equations}, 2021.

\bibitem{ha2016hypernetworks}
David Ha, Andrew Dai, and Quoc~V Le.
\newblock Hypernetworks.
\newblock {\em arXiv preprint arXiv:1609.09106}, 2016.

\bibitem{tancik2021learned}
Matthew Tancik, Ben Mildenhall, Terrance Wang, Divi Schmidt, Pratul~P Srinivasan, Jonathan~T Barron, and Ren Ng.
\newblock Learned initializations for optimizing coordinate-based neural representations.
\newblock In {\em Proceedings of the IEEE/CVF Conference on Computer Vision and Pattern Recognition}, pages 2846--2855, 2021.

\bibitem{lee2021meta}
Jaeho Lee, Jihoon Tack, Namhoon Lee, and Jinwoo Shin.
\newblock Meta-learning sparse implicit neural representations.
\newblock {\em Advances in Neural Information Processing Systems}, 34:11769--11780, 2021.

\bibitem{schwarz2022meta}
Jonathan~Richard Schwarz and Yee~Whye Teh.
\newblock Meta-learning sparse compression networks.
\newblock {\em arXiv preprint arXiv:2205.08957}, 2022.

\bibitem{sainath2013low}
Tara~N Sainath, Brian Kingsbury, Vikas Sindhwani, Ebru Arisoy, and Bhuvana Ramabhadran.
\newblock Low-rank matrix factorization for deep neural network training with high-dimensional output targets.
\newblock In {\em 2013 IEEE international conference on acoustics, speech and signal processing}, pages 6655--6659. IEEE, 2013.

\bibitem{zhang2014extracting}
Yu~Zhang, Ekapol Chuangsuwanich, and James Glass.
\newblock Extracting deep neural network bottleneck features using low-rank matrix factorization.
\newblock In {\em 2014 IEEE international conference on acoustics, speech and signal processing}, pages 185--189. IEEE, 2014.

\bibitem{zhao2016low}
Yong Zhao, Jinyu Li, and Yifan Gong.
\newblock Low-rank plus diagonal adaptation for deep neural networks.
\newblock In {\em 2016 IEEE International Conference on Acoustics, Speech and Signal Processing}, pages 5005--5009. IEEE, 2016.

\bibitem{khodak2020initialization}
Mikhail Khodak, Neil~A Tenenholtz, Lester Mackey, and Nicolo Fusi.
\newblock Initialization and regularization of factorized neural layers.
\newblock In {\em International Conference on Learning Representations}, 2020.

\bibitem{jaderberg2014speeding}
Max Jaderberg, Andrea Vedaldi, and Andrew Zisserman.
\newblock Speeding up convolutional neural networks with low rank expansions.
\newblock In {\em Proceedings of the British Machine Vision Conference. BMVA Press}, 2014.

\bibitem{gao2022svd}
Yihang Gao, Ka~Chun Cheung, and Michael~K Ng.
\newblock {SVD-PINNs: T}ransfer learning of physics-informed neural networks via singular value decomposition.
\newblock {\em arXiv preprint arXiv:2211.08760}, 2022.

\bibitem{wang2021understanding}
Sifan Wang, Yujun Teng, and Paris Perdikaris.
\newblock Understanding and mitigating gradient flow pathologies in physics-informed neural networks.
\newblock {\em SIAM Journal on Scientific Computing}, 43(5):A3055--A3081, 2021.

\bibitem{rahaman2019spectral}
Nasim Rahaman, Aristide Baratin, Devansh Arpit, Felix Draxler, Min Lin, Fred Hamprecht, Yoshua Bengio, and Aaron Courville.
\newblock On the spectral bias of neural networks.
\newblock In {\em International Conference on Machine Learning}, pages 5301--5310. PMLR, 2019.

\bibitem{yang2021b}
Liu Yang, Xuhui Meng, and George~Em Karniadakis.
\newblock {B-PINNs: B}ayesian physics-informed neural networks for forward and inverse {PDE} problems with noisy data.
\newblock {\em Journal of Computational Physics}, 425:109913, 2021.

\bibitem{jagtap2020conservative}
Ameya~D Jagtap, Ehsan Kharazmi, and George~Em Karniadakis.
\newblock Conservative physics-informed neural networks on discrete domains for conservation laws: Applications to forward and inverse problems.
\newblock {\em Computer Methods in Applied Mechanics and Engineering}, 365:113028, 2020.

\bibitem{golub2013matrix}
Gene~H Golub and Charles~F Van~Loan.
\newblock {\em Matrix computations}.
\newblock JHU press, 2013.

\bibitem{jain2013low}
Prateek Jain, Praneeth Netrapalli, and Sujay Sanghavi.
\newblock Low-rank matrix completion using alternating minimization.
\newblock In {\em Proceedings of the forty-fifth annual ACM symposium on Theory of computing}, pages 665--674, 2013.

\bibitem{doostan2009least}
Alireza Doostan and Gianluca Iaccarino.
\newblock A least-squares approximation of partial differential equations with high-dimensional random inputs.
\newblock {\em Journal of computational physics}, 228(12):4332--4345, 2009.

\bibitem{dolgov2014alternating}
Sergey~V Dolgov and Dmitry~V Savostyanov.
\newblock Alternating minimal energy methods for linear systems in higher dimensions.
\newblock {\em SIAM Journal on Scientific Computing}, 36(5):A2248--A2271, 2014.

\bibitem{lee2022enhanced}
Kookjin Lee, Howard~C Elman, Catherine~E Powell, and Dongeun Lee.
\newblock Enhanced alternating energy minimization methods for stochastic galerkin matrix equations.
\newblock {\em BIT Numerical Mathematics}, 62(3):965--994, 2022.

\bibitem{sb1}
K.~Dingle, C.~Q. Camargo, and A.~A. Louis.
\newblock Input-output maps are strongly biased towards simple outputs.
\newblock {\em Nature communications}, 2018.

\bibitem{10.5555/3495724.3497160}
Wei Hu, Lechao Xiao, Ben Adlam, and Jeffrey Pennington.
\newblock The surprising simplicity of the early-time learning dynamics of neural networks.
\newblock In {\em Proceedings of the 34th International Conference on Neural Information Processing Systems}, 2020.

\bibitem{huh2021lowranksimplicity}
Minyoung Huh, Hossein Mobahi, Richard Zhang, Pulkit Agrawal, and Phillip Isola.
\newblock The low-rank simplicity bias in deep networks.
\newblock {\em arXiv}, 2021.

\bibitem{schrauwen2007overview}
Benjamin Schrauwen, David Verstraeten, and Jan Van~Campenhout.
\newblock An overview of reservoir computing: theory, applications and implementations.
\newblock In {\em Proceedings of the 15th european symposium on artificial neural networks. p. 471-482 2007}, pages 471--482, 2007.

\bibitem{he2015delving}
Kaiming He, Xiangyu Zhang, Shaoqing Ren, and Jian Sun.
\newblock Delving deep into rectifiers: {S}urpassing human-level performance on {ImageNet} classification.
\newblock In {\em Proceedings of the IEEE international conference on computer vision}, pages 1026--1034, 2015.

\bibitem{saxe2013exact}
Andrew~M Saxe, James~L McClelland, and Surya Ganguli.
\newblock Exact solutions to the nonlinear dynamics of learning in deep linear neural networks.
\newblock {\em arXiv preprint arXiv:1312.6120}, 2013.

\end{thebibliography}

\clearpage
\appendix
\section{Limitations}\label{a:lim}
In this study, we mainly focus on the cases where PDE parameters are the coefficients of various PDE terms. In particular, we train the proposed model on a set of benchmark PDEs (e.g., convection equations or reaction equations or Helmholtz equations). As a future direction of this study, we list the following potential extensions:
\begin{compactenum}
    
    \item Extending our framework so that the set of basis vectors are parameter dependant to some extent, as we did to have adaptive rank (i.e, the rank structure is parameter dependant),

    \item (Related to the above extension) Extending our framework to be equipped with a specialized optimizer, which alternately update the basis vectors and diagonal elements, 
    
    \item Extending our framework to more general settings, attempting to learn the solutions of parameterized PDEs where the PDE parameters define initial/boundary conditions.
\end{compactenum}

Regarding the alternating solver: we expect that making the basis vector learnable in phase 2 would increase expressivity. However, in achieving low error, the difficulty is expected to come from the training algorithm, where the similar phenomena were observed in matrix decomposition methods. Consider finding a low-rank decomposition of matrix such that min  with a norm induced from an inner product. In such a problem, updating 
all together in an interactive solver typically introduces more complexity. To avoid such difficulty, special solvers have been developed such as direct linear algebraic decompositions \cite{golub2013matrix} which does not use a gradient-based optimization method, or notably, alternating minimization for matrix completion \cite{jain2013low}. In the context of low-rank approximation of solutions of PDEs, similar alternating approaches (e.g., alternating least-squares or alternating energy minimization) have been shown to be more effective \cite{doostan2009least,dolgov2014alternating,lee2022enhanced}.

\section{Remark on our low-rank approximation in general deep learning}
It had been reported that deep neural networks have low-rank biases for learned representations, which is the reason why over-parameterized neural networks do not always fail even when relatively small data is given~\cite{sb1,10.5555/3495724.3497160,huh2021lowranksimplicity}. One can consider that our proposed method imposes an effective learning bias, i.e., our adaptive low-rank approximation, on PINNs to make their learning process easier than before. Thus, we focus on the PINN's notorious failure modes and empirically prove the efficacy of our design.

\section{Reduced-Order Modeling (ROM)} \label{a:rom}
In the following, we summarize the offline and the online phases of traditional linear-subspace reduced-order modeling approaches. 
\paragraph{Offline phase} 
\begin{itemize}
    \item Perform high-fidelity simulations on a training PDE parameter instances $\{\pmb \mu^{(i)}_{\text{train}} \}_{i=1}^{n_{\text{train}}}$, i.e., 
    \begin{equation}
        \frac{\mathrm d \pmb u}{\mathrm d t} = \pmb f ( \pmb u; \pmb \mu^{(i)}_{\text{train}})
    \end{equation}
    \item Collect solution snapshots from the high-fidelity simulation
    \begin{equation}
        U = [\pmb u(t_1,\pmb \mu^{(1)}_{\text{train}}), \pmb u(t_2,\pmb \mu^{(1)}_{\text{train}}), \cdots, \pmb u(T,\pmb \mu^{(1)}_{\text{train}}), \pmb u(t_1,\pmb \mu^{(2)}_{\text{train}}), \cdots, \pmb u(T,\pmb \mu^{(n_{\text{train}})}_{\text{train}}) ]
    \end{equation}
    \item Compute SVD on $U$: $U = \Psi D \Phi\Transpose$
    \item Truncate the series $\Psi_p$
\end{itemize}
\paragraph{Online phase}
\begin{itemize}
    \item Perform inexpensive simulation on unseen test PDE parameter instances $\{\pmb \mu^{(i)}_{\text{test}} \}_{i=1}^{n_{\text{test}}}$, 
    \item Represent a solution as $\tilde u(t,\pmb \mu^{(i)}_{\text{test}}) = \Psi_p \pmb{c}(t, \pmb \mu^{(i)}_{\text{test}}) $
    \item Project the dynamical system into the low-dimensional space
    \begin{align*}
        \frac{\mathrm d (\Psi_p \pmb c)}{\mathrm d t} &= \pmb f (\Psi_p \pmb c; \pmb \mu^{(i)}_{\text{test}})\\
        \Leftrightarrow \frac{\mathrm d (\Psi_p\Transpose\Psi_p \pmb c)}{\mathrm d t} &= \Psi_p\Transpose \pmb f (\Psi_p \pmb c; \pmb \mu^{(i)}_{\text{test}}) \quad (\text{multiply } \Psi_p\Transpose \text{on both sides})\\
        \Leftrightarrow \frac{\mathrm d (\pmb c)}{\mathrm d t} &= \hat{\pmb f} (\Psi_p \pmb c; \pmb \mu^{(i)}_{\text{test}}) \quad (\Psi_p\Transpose\Psi_p = I \text{ and } \hat{\pmb f}  = \Psi_p\Transpose \pmb f \in \mathbb{R}^{p})\\
    \end{align*}
\end{itemize}

\begin{figure*} [t]
\centering
\includegraphics[width=1.0\columnwidth]{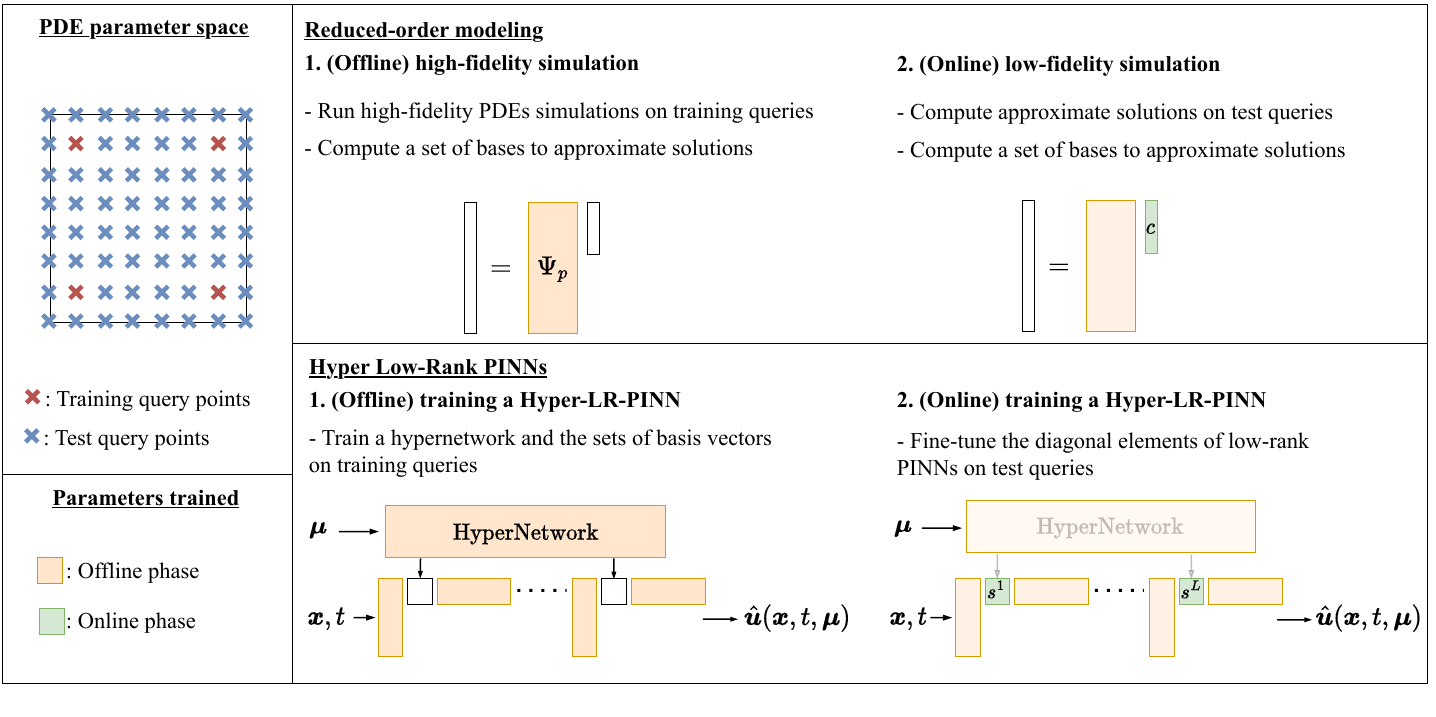}\vspace{-2mm}
\caption{Graphical comparisons between ROMs and Hyper-LR-PINNs.}\label{fig:analogy}
\end{figure*}

\paragraph{Analogy between ROMs and the proposed Hyper-LR-PINNs} Figure \ref{fig:analogy} draws an analogy between ROMs and Hyper-LR-PINNs. It is highlighted that the both approaches perform heavy-lifting in the offline phase to make the online phase less expensive so that solutions at the large number of test PDE parameter instances can be evaluated rapidly. 

\paragraph{Difference from reservoir computing}
We agree that there is a very high-level analogy that can be made to reservoir computing (RC) \cite{schrauwen2007overview}, since in the second phase of our two-phase training procedure only part of the parameters are trained. However, this is a feature shared with all models making use of encoding-decoding schemes. Moreover, an important distinguishing feature in RC is the recurrent neural network, whereas our neural network architecture is feedforward. The fact that we are specifically solving parametrized partial differential equations also makes our model more specialized. We believe these points makes our model much more similar to a ROM, where a dimension-reduced implicit representation is learned in the first phase. In the second phase, a family of functions is quickly approximated by training a few parameters.

\section{Preliminary experiments}\label{a:prelim_exp}
\subsection{Approximating trained PINNs' weights with low-rank matrices}

\begin{figure}[t!]
    \centering
     \includegraphics[width=.5\linewidth]{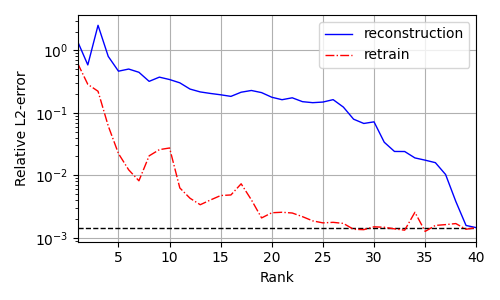} 
    \caption{[Viscous Burgers' equation] The relative L2 errors of the low-rank reconstructed models and the low-rank retrained models.}
    \label{fig:rel_l2_visc_burgers}
    \vspace{-1.em}
\end{figure}

To motivate our study, we perform some preliminary experiments on the canonical one-dimensional viscous Burgers' equation. We parameterize a PINN as a multi-layer perceptron (MLP) with 5 hidden layers and 40 neurons in each layer, followed by the \textsc{Tanh} nonlinearity; the $l$-th layer of the MLP can be represented as  $\HiddenState^{l+1} = \NonlinearActivation (\NNWeight^{l} \HiddenState^{l} + \NNBias^{l})$, where $\NNWeight$ and $\NNBias$ denote the weight and the bias, respectively, $\NonlinearActivation$ denotes the nonlinear activation, and $\HiddenState$ denotes the post activation. 
Training with L-BFGS results in  the solution accuracy measured in the relative L2 error (around .1\% error, the black dashed line in Figure~\ref{fig:rel_l2_visc_burgers}).  

After the training, we build a low-rank PINN in the following steps:  we i) take the weight matrices of hidden layers (except the input and the output layers), ii) take the SVDs, and iii) assemble weight matrices by retaining the $r$ largest singular values and corresponding singular vectors such that
\begin{equation}\label{eq:trunc}
    \NNWeight^l = \LeftSingularVector^l \SingularValue^l \RightSingularVector^l{}\Transpose, \quad \text{and }\quad \NNWeight_r^l = \LeftSingularVector_r^l \SingularValue_r^l \RightSingularVector_r^l{}\Transpose,
\end{equation}
which results in a low-rank PINN (\LRPINN{}) with a rank of $r$, which shares the same model parameters with the trained PINN, except the hidden layer weights obtained from the previous truncation step. 
We then vary the value of $r$ from 1 to 40 and evaluate the resulting \LRPINNs{} on the test set (``reconstruction'' in Figure~\ref{fig:rel_l2_visc_burgers}); 
the blue line depicts the error measured on the test set for \LRPINNs{} with varying $r$. Even truncating 3--4 smallest singular values seems to yield significant accuracy degradation by an order of magnitude. 

Next, we investigate what happens if we further train \LRPINN{} from the model parameters obtained from the truncation Eq.~\eqref{eq:trunc}. Instead of training all model parameters, $\LeftSingularVector,\SingularValue,\RightSingularVector$, we i) represent the hidden layer weights in a factored form $\NNWeight_r^l = \LeftSingularVector_r^l \SingularValue_r^l \RightSingularVector_r^l{}\Transpose$, ii) fix the basis matrices $(\LeftSingularVector_r^l, \RightSingularVector_r^l)$ and iii) make $\SingularValueElem_r$ only the trainable parameters, 
$\SingularValue_r^l = \text{diag}\left(\SingularValueElem_r^l\right) $. 
After training, we test the trained models on the same test data set. As shown by the red dashed line in Figure~\ref{fig:rel_l2_visc_burgers}, its accuracy becomes comparable to that of the full model when $r\geq20$, which shows that \emph{the weights of hidden layers 
have low-rank structures.}

\subsection{A study on the effect of rank and orthogonality of basis sets for varying PDE parameters} \label{sec:example_CD}
As an example parameterized PDE, we consider a one-dimensional convection-diffusion equations:
\begin{equation}
    \frac{\partial \Solution}{\partial t} + \beta\frac{\partial \Solution}{\partial x} - \nu\frac{\partial^2 \Solution}{\partial x^2}  = {0}, \quad x \in \Omega, \; t \in [0,T], 
    \label{eq:cd}
\end{equation}
where $\Solution(x,t;\pmb{\mu})$ denotes the solution satisfying the equation, with $\pmb{\mu} = (\beta, \nu)$ denote the convection and diffusion coefficients, respectively. 
The higher the ratio $\frac{\beta}{\nu}$ is, the more convection-dominated the problem is, which is considered to be a more challenging scenario for PINNs. 
In the following experiments, we set $\nu = 1$ fixed and vary $\beta$ from 2 to 40 to control the difficulty of 
training \LRPINNs{}. 
For training, 
we use a  curriculum-learning-type training approach proposed in \cite{krishnapriyan2021characterizing}; in a high $\beta$ regime (e.g., $\beta \geq 20)$, training PINNs directly on the target $\beta$ typically fails and the authors resolved this issue by curriculum learning which i) starts with a low $\beta$ value (e.g., $\beta=1$) and ii) over the course of training, gradually increases the value of $\beta$ until it reaches to the target $\beta$.

\begin{figure}[t!]
    \centering
    \subfloat[][Varying initializations]{
     \includegraphics[width=.47\columnwidth]{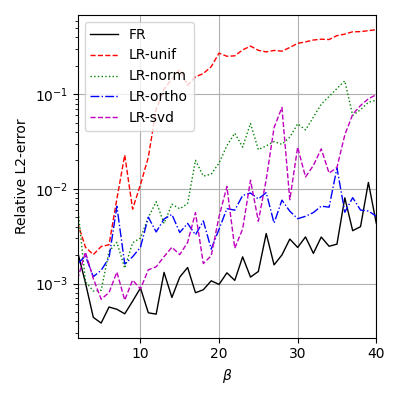} \label{fig:beta_l2_cd}} 
     \subfloat[][Varying ranks]{
     \includegraphics[width=.47\columnwidth]{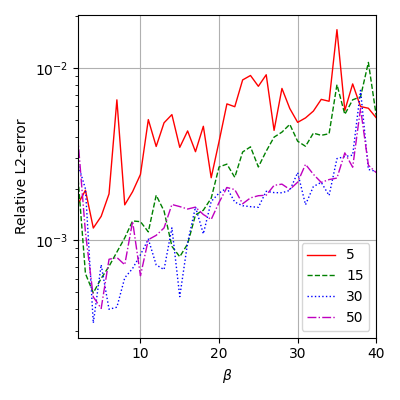} \label{fig:rank_l2_cd}}
    \caption{[Convection-diffusion equations] The relative L2 errors of full-rank PINNs (FR) and low-rank PINNs (LR): (a) \LRPINN{ } with varying weight initialization schemes including orthogonal, He-uniform, He-normal, and extracted basis from the pre-trained PINN, and (b) \LRPINN{} with  varying ranks ($r=5,15,30,50$) and the orthogonal initialization.}
    \label{fig:l2_cd}
    \vspace{-1.0em}
\end{figure}

\paragraph{Setting the basis vectors,   $\LeftSingularVector_r$ and $\RightSingularVector_r$:} We have tested four different schemes for setting the basis vectors: three schemes that are commonly used for initializing the weights of FC layers: \textsc{He-uniform}, \textsc{He-normal} \cite{he2015delving}, and \textsc{Orthogonal} \cite{saxe2013exact}, and one scheme that extracts the left and right singular vectors from a PINN trained for $\beta=1$ and use them for setting the basis vectors, which we denote by \textsc{SVD-init}.  

Figure~\ref{fig:l2_cd}(a) shows the errors of approximate solutions obtained by all trained full-rank PINNs (FR) and \LRPINNs{} with $r=5$ for varying $\beta$ values. Among the \LRPINNs{}, the ones initialized with \textsc{Orthogonal} (LR-ortho) and \textsc{SVD-init} (LR-svd) outperform the ones initialized with \textsc{He-uniform} (LR-unif) and \textsc{He-normal} (LR-norm). LR-ortho tends to perform better than LR-svd in very high $\beta$ regime ($\beta \geq 30$). Compared to LR-svd, LR-ortho has another advantage not requiring pretrainig of a PINN.
An apparent observation is that \emph{orthogonality in basis vectors helps achieving better accuracy} (Observation \#1).

\paragraph{The effect of rank:} Next, with LR-ortho, we vary the rank $r=\{5,15,30,50\}$. Figure~\ref{fig:l2_cd}(b) depicts the errors of \LRPINNs{} for varying $\beta$ and ranks. For lower $\beta$ ($\beta < 20$), \LRPINNs{} with $r=15,30,50$ produce comparable results. For higher $\beta$ values ($\beta \geq 20)$,
it appears that a higher rank is required to achieve a certain level of accuracy, presumably due to the difficulty of the numerical optimization problem. 
From this result, the second observation we make is that \emph{required ranks for varying $\pmb{\mu}$ to achieve a certain level of accuracy could vary} (Observation \#2) and, thus, a solution method that adaptively decides the ranks of each individual hidden layers is required.

\section{Proposed two-phase training algorithm}\label{a:algorithm}
Algorithm~\ref{alg:train} summarizes the two-phase algorithm we describe in Section~\ref{sec:train_alg}.

\begin{algorithm}[h]
   \caption{Two-phase training of the proposed method}
   \label{alg:train}
\begin{algorithmic}
   \STATE /* Phase 1 */
   \STATE {\bfseries Input:} 
   A set of training sampled points: $\{(\pmb{x},t,\pmb{\mu}^{(i)})\}_{i=1}^{n_{p_1}}$, where $n_{p_1}$ is number of PDE equations used in Phase 1.
   \STATE Initialize weights and biases $\{(U^l, V^l, \pmb{b}^{l})\}_{l=1}^{L}$, $W^0, W^{L+1}, b^0, b^{L+1}$ of  \textsc{LR-PINN}, and those $\{( W^{\text{emb},m},  \pmb{b}^{\text{emb},m})\}_{m=1}^{M}$, $\{(W^{\text{hyper},l}, \pmb{b}^{\text{hyper},l})\}_{l=1}^{L} $ of the hypernetwork.
   \FOR{$epoch=1$ to $ep_1$}
   \FOR{$i=1$ {\bfseries to} $n_{p_1}$}
       \STATE $\{\pmb{s}^l(\pmb{\mu}^{(i)})\}_{l=1}^L = f^{\text{hyper}}(\pmb{\mu}^{(i)})$
       \STATE Compute forward pass: $u_{\Theta}((\pmb{x},t); \{ \pmb{s}^l (\pmb{\mu}^{(i)}) \}_{l=1}^L)$
       \STATE Compute PINN loss and update model parameters via backpropagation
   \ENDFOR
   \ENDFOR
   \STATE
   \STATE /* Phase 2 */
   \STATE {\bfseries Input:} 
   A set of training sampled points: $\{(\pmb{x},t,\pmb{\mu}^{\text{target}})\}$\\
   \STATE Freeze $\{(U^l, V^l, \pmb{b}^{l})\}_{l=1}^{L}$, $\{( W^{\text{emb},m},  \pmb{b}^{\text{emb},m})\}_{m=1}^{M}$, and $\{(W^{\text{hyper},l}, \pmb{b}^{\text{hyper},l})\}_{l=1}^{L}$
   \STATE Initialize $\{\pmb{s}^l\}_{l=1}^{L} = f^{\text{hyper}}(\pmb{\mu}^{\text{target}})$
   \FOR{$epoch=1$ to $ep_2$}
       \STATE Compute forward pass: $u_{\Theta}((\pmb{x},t); \{ \pmb{s}^l (\pmb{\mu}^\text{target}) \}_{l=1}^L)$
       \STATE Compute PINN loss and update model parameters via backpropagation
   \ENDFOR
\end{algorithmic}
\end{algorithm}

\section{Reproducibility}\label{a:rep}

\paragraph{Software and hardware environments} We implement and conduct experiments with \textsc{Python} 3.9.7 and \textsc{Pytorch} 1.13.0, \textsc{CUDA} 11.6.1, \textsc{NVIDIA} Driver 470.74, i9 CPU, \textsc{NVIDIA RTX A5000}, and \textsc{RTX 2080 Ti}. Our source code for the benchmark PDEs is mainly based on \url{https://github.com/a1k12/characterizing-pinns-failure-modes} (MIT License).

\paragraph{Hyperparameters} We collect 256/100/1,000 points from initial/boundary/collocation points, and 1,000 test points for each CDR equation. For the 2D Hemlholtz equation, we collect 400/1,000/10,000 points from boundary/collocation/test points. The train/test sets contain non-overlapping spatial/temporal collocation points. The PINN baselines consist of 6 FC layers with 50 neurons. For our methods, we employ 3 hidden layers ($L\!\!=\!\!M\!\!=\!\!3$). The Adap optimizer is used with learning rate 1e-3 for PINN baselines, and 1e-3 and 2.5e-4 for Phase1 and Phase2 of meta-learning methods. These hyperparameters are common in all our experiments.

\section{Meta-learning baselines: MAML and Reptile}\label{a:maml_reptile}
We consider the two most representative optimization-based meta-learning algorithms: model-agnostic meta learning (MAML) \cite{finn2017model} and Reptile \cite{nichol2018first}.   
In the parameterized PDEs setting, we can define a task, $\tau^{(i)}$, as a specific setting of PDE parameters, $\pmb{\mu}^{(i)}$.  
Both MAML and Reptile seek an initial weights of a PINN, which can serve as a good starting point for gradient-based optimizers when a solution of a new unseen PDE parameters is sought. Both methods consist of an \textit{inner} loop and an \textit{outer} loop. The inner loop takes $k$ optimization gradient descent steps to update model parameters from a current of the meta initial points $\theta_0^l$ given a training task $\tau^{(i)}$. Here, this $k$-step update is denoted as $\theta_k (\theta_0^l, \tau^{(i)})$. Then the outer loop updates the meta-learned initial points using the information obtained from the inner loop such that 
\begin{equation}
  \begin{split}
  \theta_0^{l+1} &= \theta_0^{l} - \beta \nabla_{\theta} L (\theta_k(\theta, \tau^{(l)}))|_{\theta = \theta_0^l}, \qquad (\text{MAML})\\
  \theta_0^{l+1} &= \theta_0^{l} - \beta (\theta_k(\theta_0^l, \tau^{(l)}) - \theta_0^l), \qquad \qquad (\text{Reptile})
  \end{split}
\end{equation}
where Reptile has a simpler update rule, which does not require the second-order gradients. 

\section{Comparisons of baselines and our method}
Table~\ref{tab:baseline_summary} compares the baseline models with our method in three aspects: a target function being approximated, an initialization scheme, and the rank structure. For vanilla PINN and their variants, the function being approximated is $u({\pmb{x},t;\pmb{\mu}})|_{\pmb{\mu} = \pmb{\mu}^i}$, which is the solution of the parameterized PDE that is realized at a certain PDE parameter $\pmb{\mu}^i$. On the other hand, PINN-P and our proposed method models the parameterized solution itself $u(\pmb{x},t;\pmb{\mu})$. The initilization column indicates if meta-learning is used or not. The rank structure column indicates if the rank is adaptively chosen for different values of $\pmb{\mu}$.

\begin{table}[h]
\centering
\small

\caption{Comparisons of baselines and our method}\label{tab:baseline_summary}
\begin{tabular}{|l|c|c|c|c|c|c|}
\hline
& \multicolumn{2}{c|}{Target function} & \multicolumn{2}{c|}{Initialization} & \multicolumn{2}{c|}{Rank structure} \\
\hline
& $u(\pmb{x},t;\pmb{\mu}) | _{\pmb{\mu} = \pmb{\mu}^{(i)}}$ & $u(\pmb{x},t;\pmb{\mu}) $ & Random & Meta-learned & Fixed  & Adaptive\\
\hline
PINN  & \checkmark & &\checkmark&  &\checkmark&\\
PINN-R & \checkmark & &\checkmark & &\checkmark&\\
PINN-S2S & \checkmark & &\checkmark & &\checkmark&\\
Na\"ive \LRPINNplain{} & \checkmark & &\checkmark & &\checkmark&\\
PINN-P & & \checkmark & \checkmark & &\checkmark&\\
MAML & \checkmark & & &\checkmark&\checkmark&\\ 
Reptile & \checkmark & && \checkmark&\checkmark&\\ 
HyperPINN & \checkmark & && \checkmark&\checkmark&\\ 
Hyper-\LRPINNplain{} & & \checkmark &&\checkmark&&\checkmark\\
\hline
\end{tabular}
\end{table}




\section{Train and test datasets generation}
Train/test datasets are collected at collocations points for varying PDE parameters of a specific PDE type; e.g., for convection equation, we only vary $\beta$, resulting in $\{\{(\pmb{x}_k,t_k;\beta^{(i)})\}_{k=1}^{n_{c}^{\ell}}\}_{i=1}^{n_{\beta}}$, where $n_c^{\ell}$ denotes the number of collocation points in train ($\ell=0$) and test ($\ell=1$) sets, $n_{\beta}$ denotes the number of distinct $\beta$ values. 
The test dataset, which also includes the reference solution is constructed by either analytically or numerically solving the CDR equations. 

\section{Ablation Study}\label{a:ablation}
As an ablation study, we check training loss and test MSE with and without the orthogonality penalty Eq.~\eqref{eq:ortho_const} ($w_1=w_2=1$). We observe that when the orthogonality penalty plays an important role in minimizing the train loss and learning the representative basis vectors of the solutions for varying range of parameter values.
\begin{figure}[ht!]
\centering
\subfloat[Training loss]{\includegraphics[width=0.245\columnwidth]{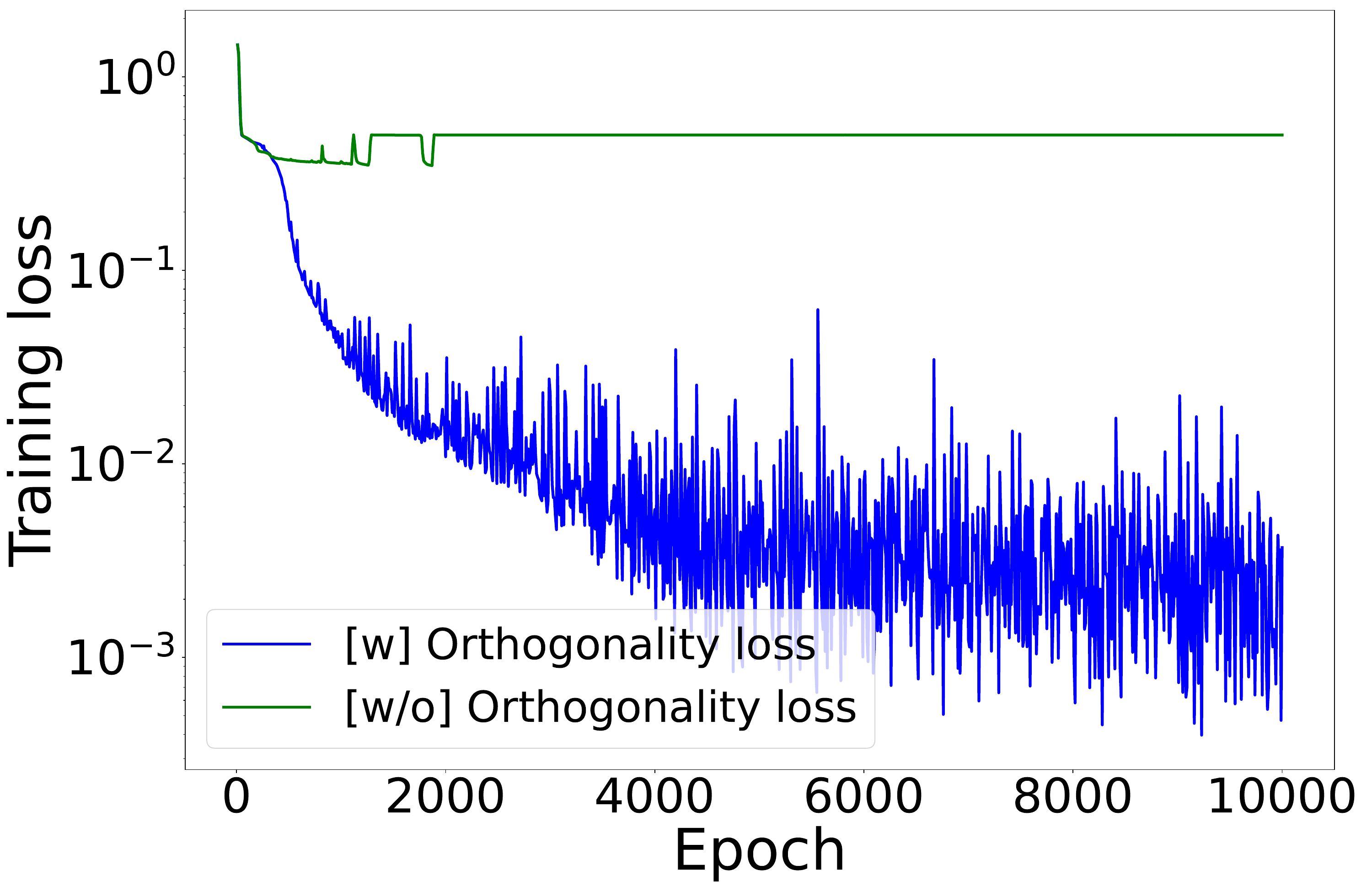}}\hfill
\subfloat[Test MSE]{\includegraphics[width=0.245\columnwidth]{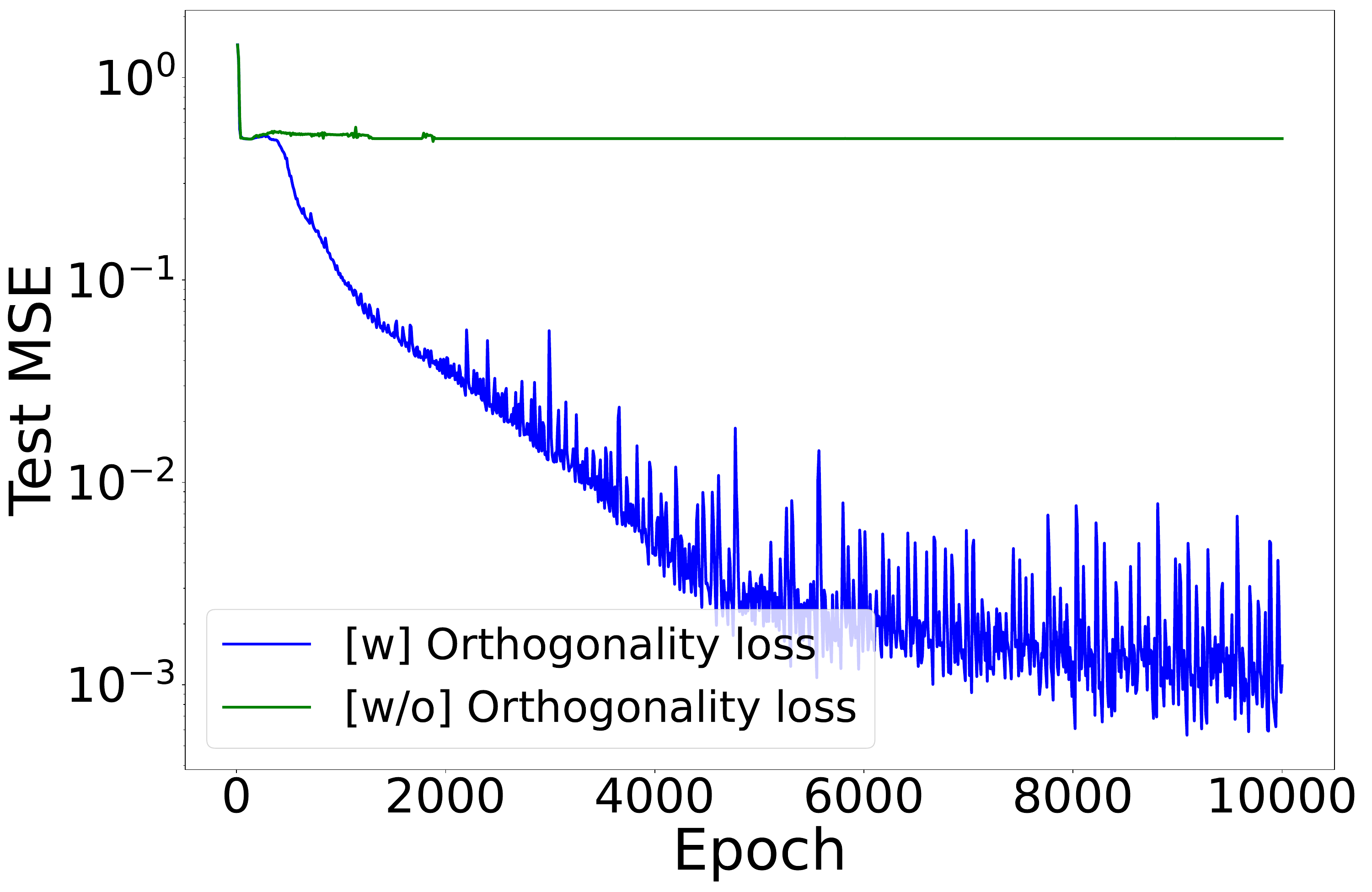}}\hfill
\subfloat[Training loss]
{\includegraphics[width=0.245\columnwidth]{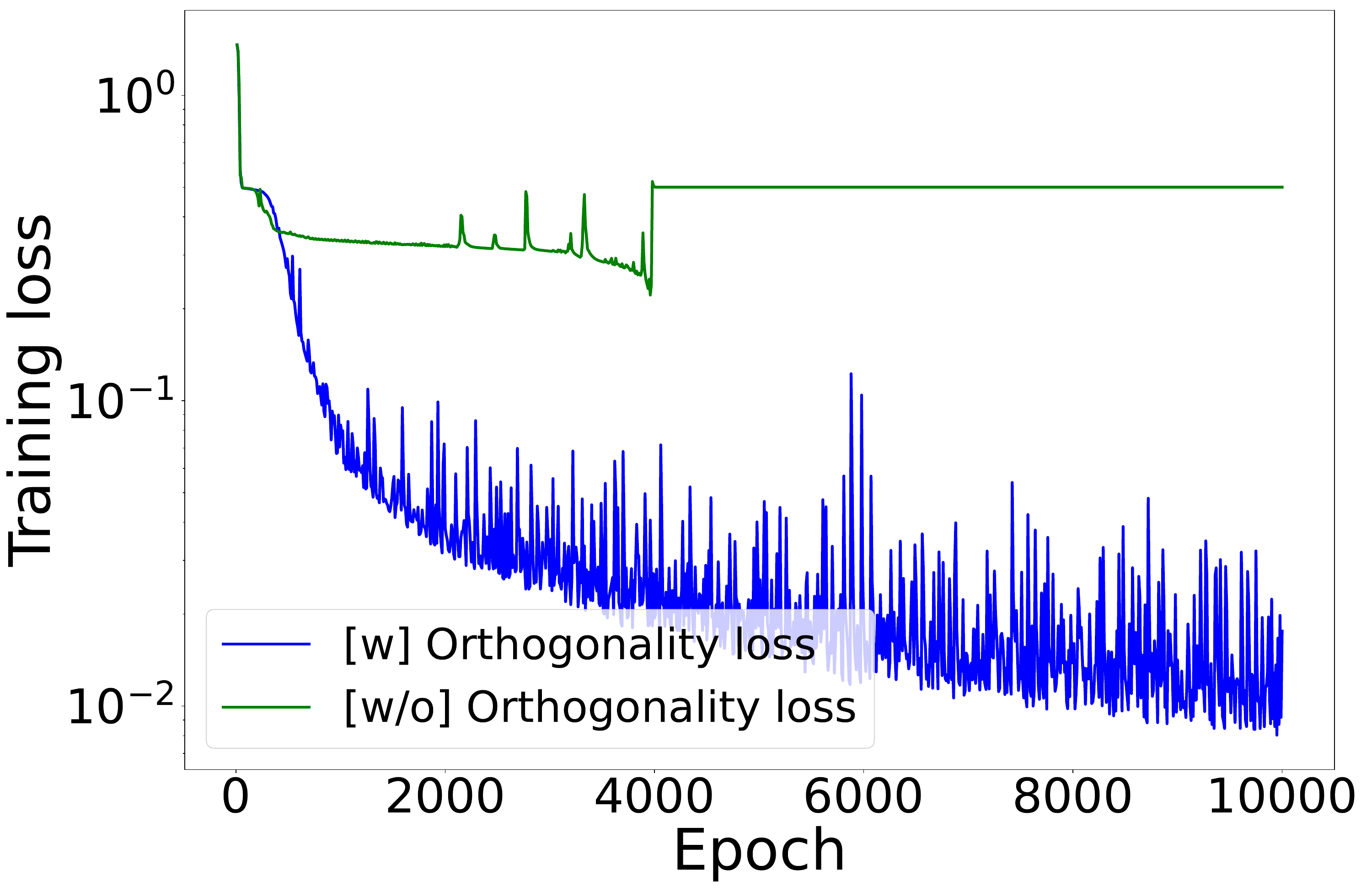}}\hfill
\subfloat[Test MSE]{\includegraphics[width=0.245\columnwidth]{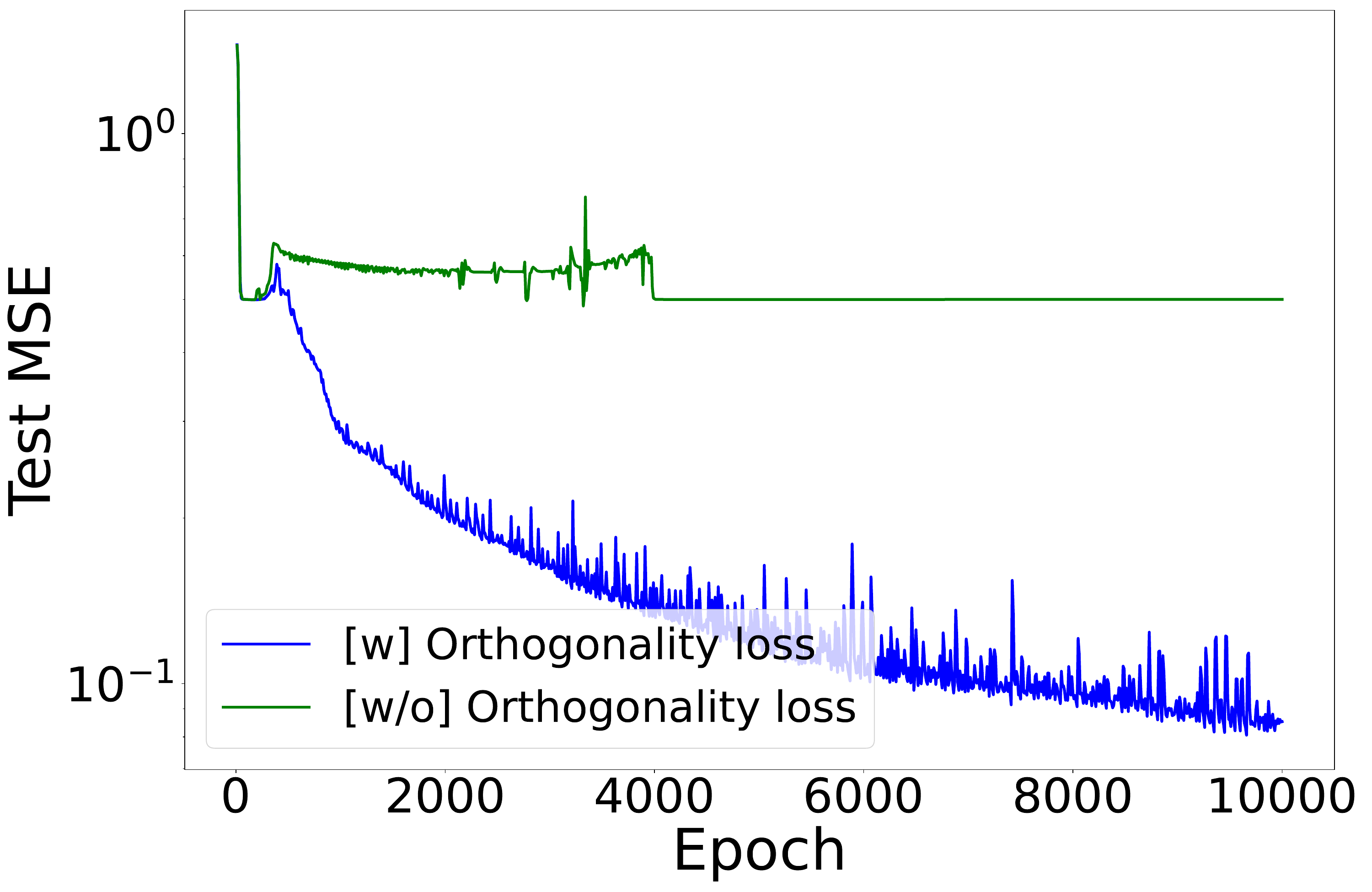}}\\

\caption{[Convection equations] training loss and test MSE with/without the orthogonality penalty (Eq.~\eqref{eq:ortho_const}) for the  $\beta \in [1, 20]$ (a-b), and  $\beta \in [1, 40]$ (c-d).}
\end{figure}

\section{Adaptive rank: learned rank structure of hidden layers}
\begin{figure}[ht!]
    \centering
    \subfloat[$\pmb{s}^1(\pmb{\mu})$]
    {\includegraphics[width=0.32\columnwidth]
    {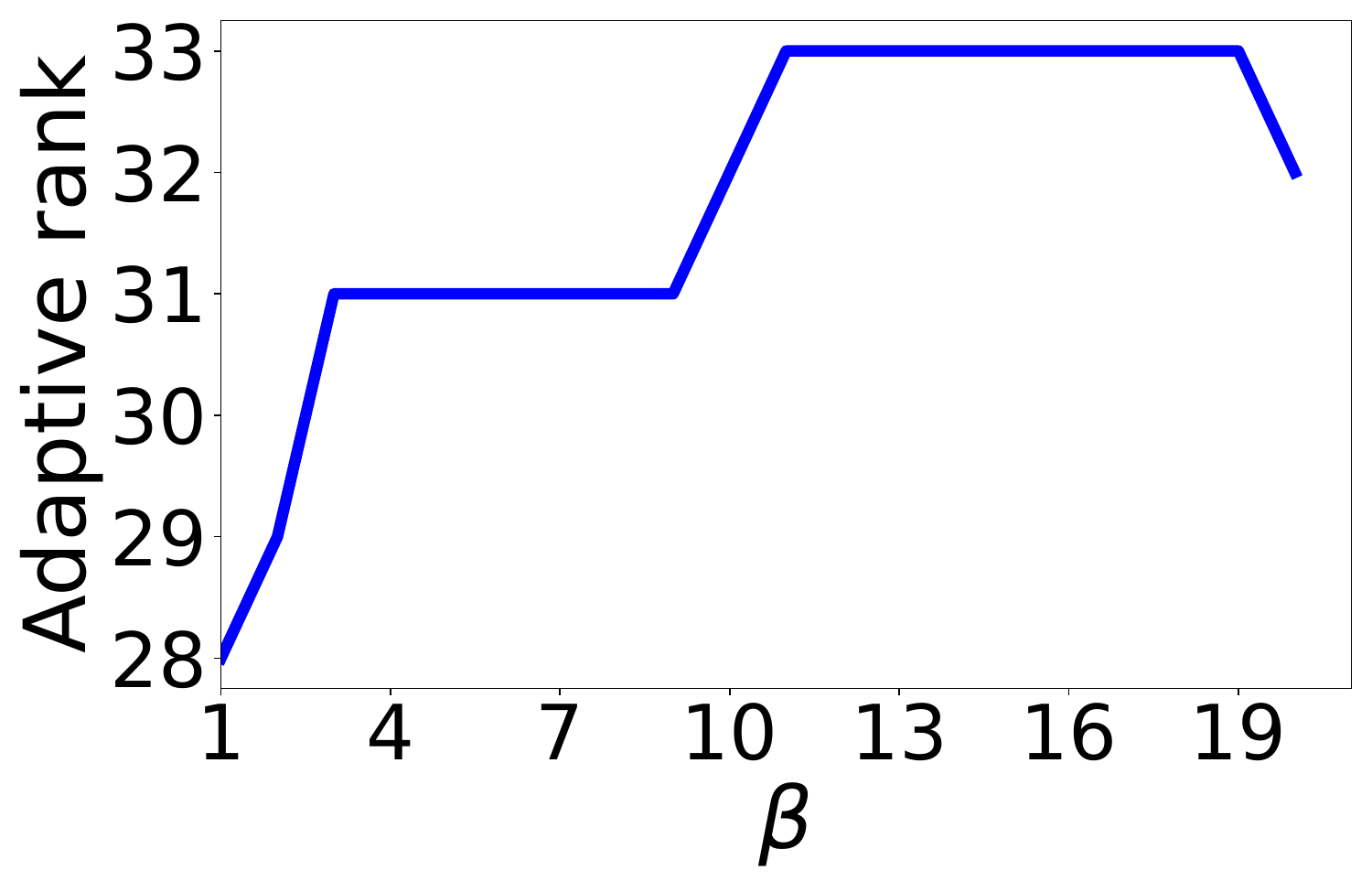}}\hfill
    \subfloat[$\pmb{s}^2(\pmb{\mu})$]
    {\includegraphics[width=0.32\columnwidth]
    {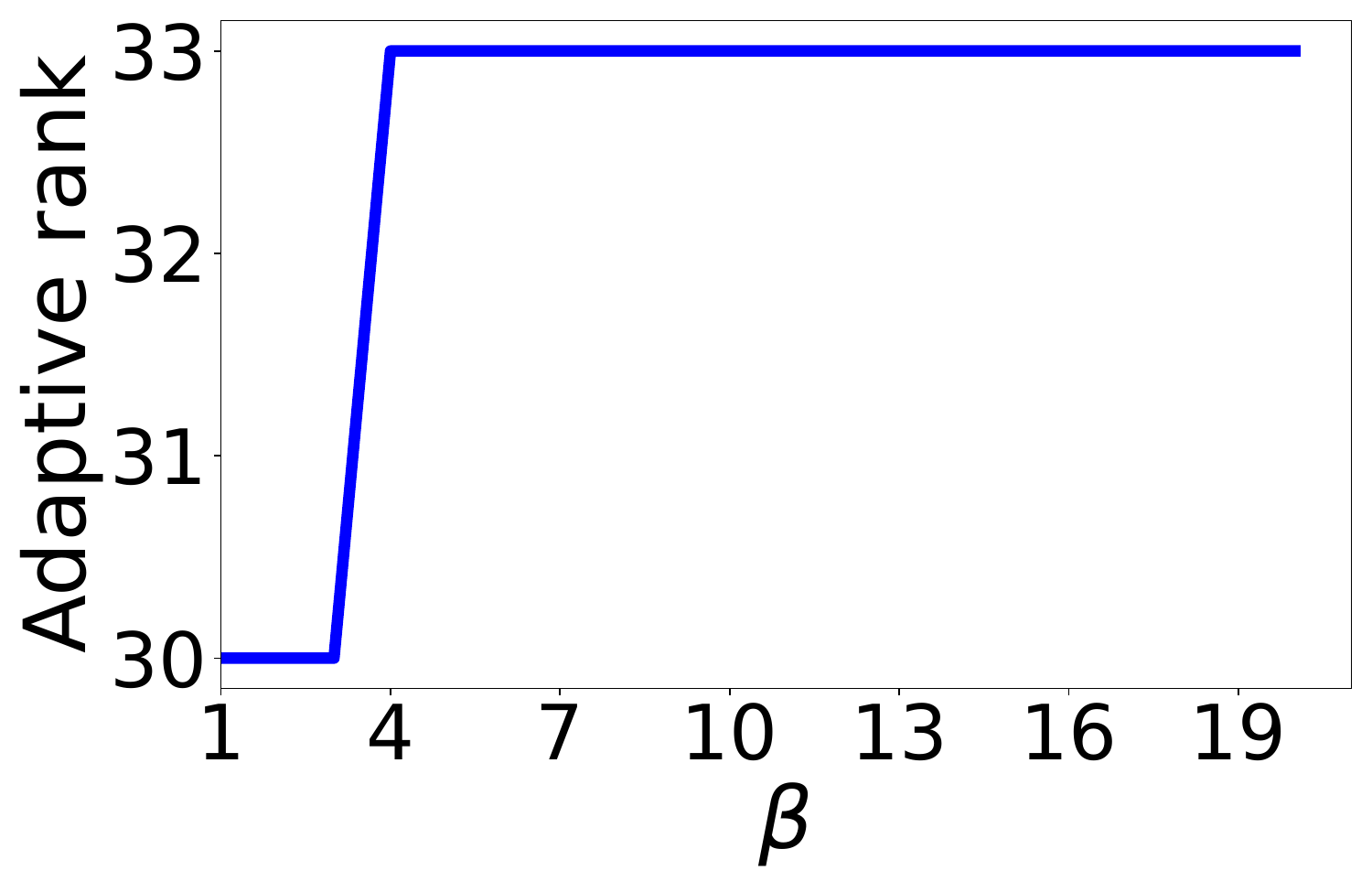}}\hfill
    \subfloat[$\pmb{s}^3(\pmb{\mu})$]
    {\includegraphics[width=0.32\columnwidth]
    {figs/check_low_rank_sin_1_20_3.pdf}} \\
    \caption{[Convection equation] Learned rank  structures of the three hidden layers. $\beta \in [1,20]$}
    \vspace{-1.0em}
\end{figure}

\section{Visualization of learned diagonal elements for varying PDE parameters}\label{a:diag_heat}
\begin{figure}[ht!]
\centering
\subfloat[Hypernetwork output: $\pmb{s}^1(\pmb{\mu})$]{\includegraphics[width=0.33\columnwidth]{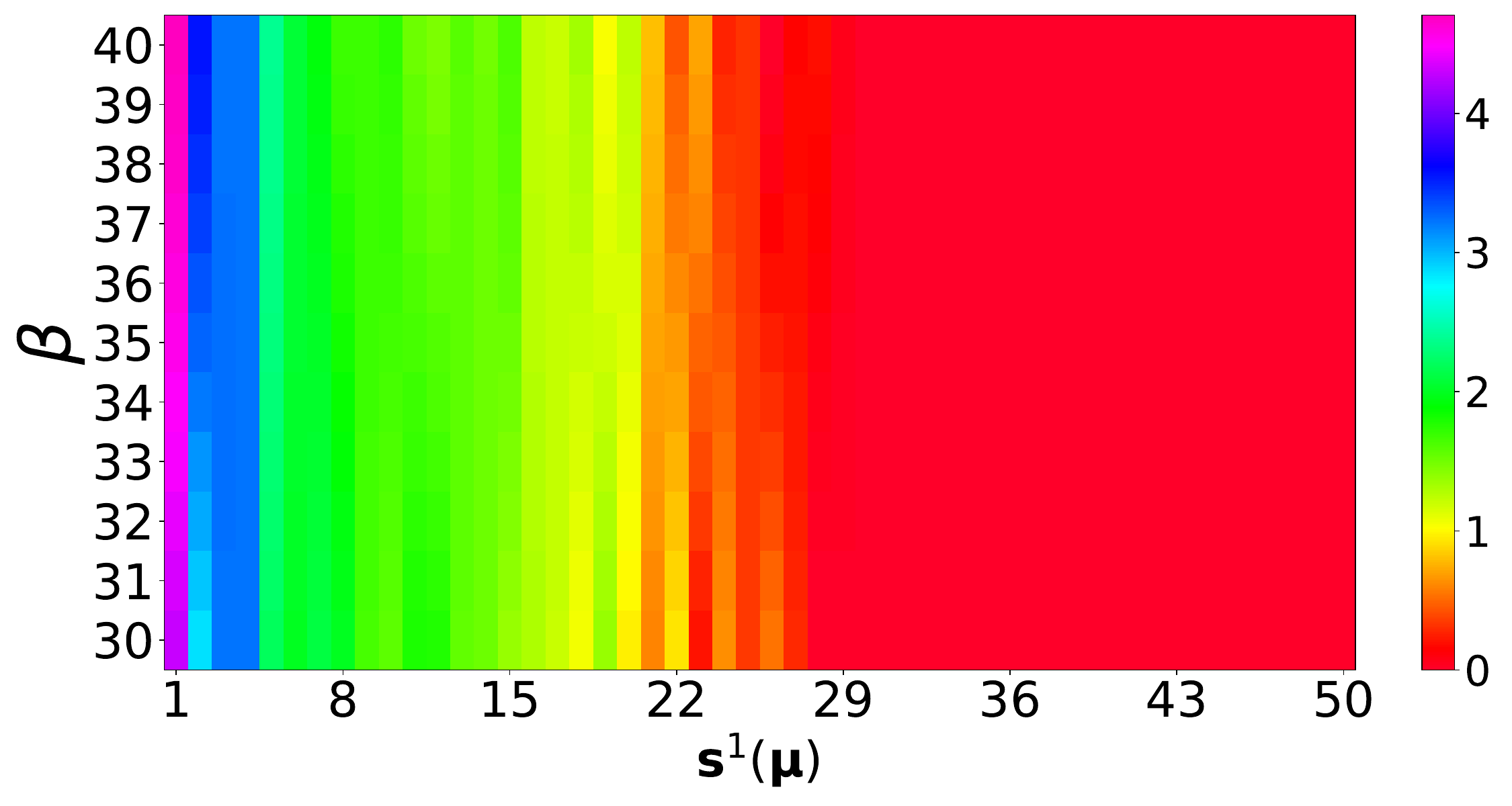}} \hfill
\subfloat[Hypernetwork output: $\pmb{s}^2(\pmb{\mu})$]{\includegraphics[width=0.33\columnwidth]{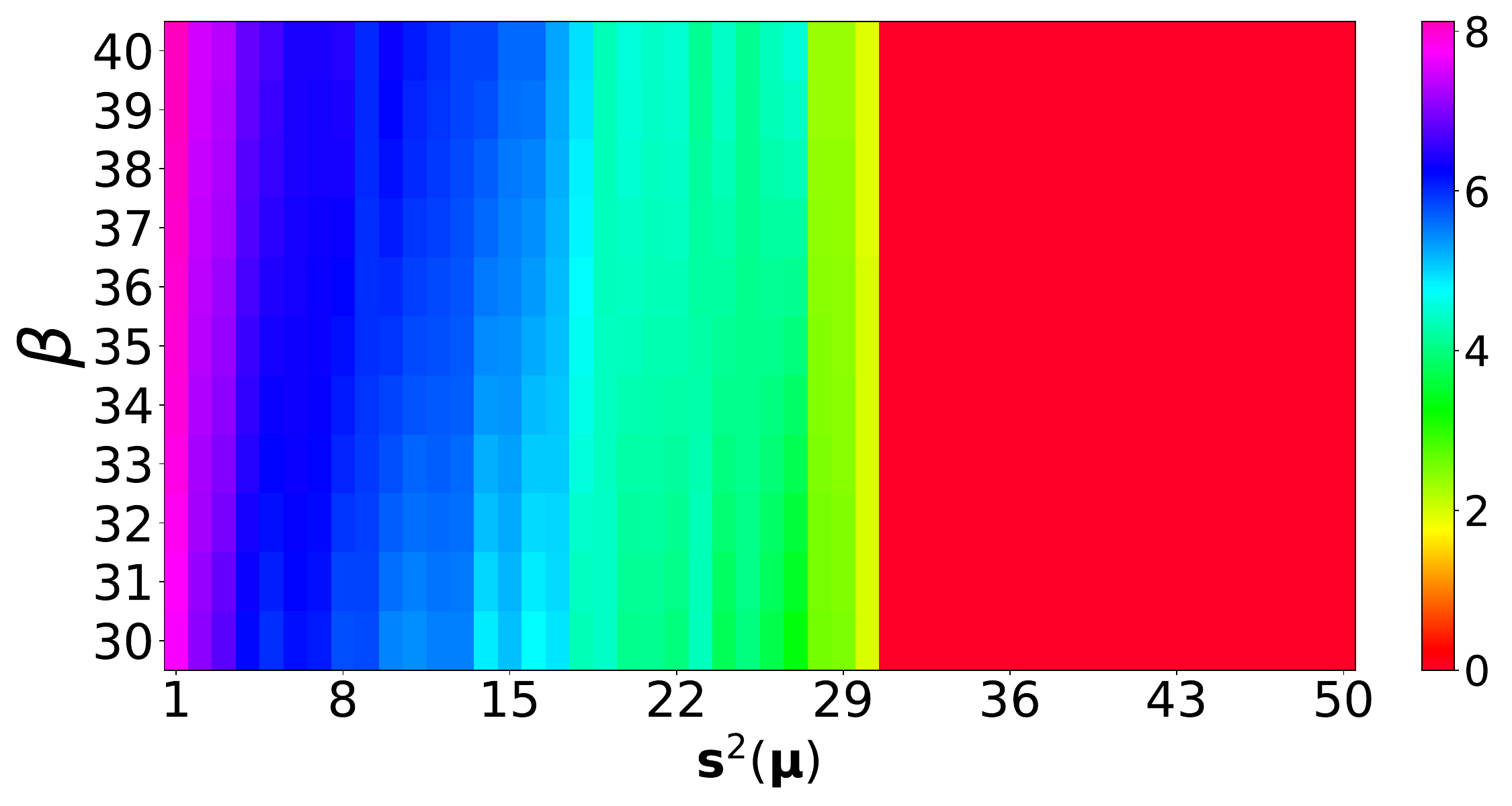}} \hfill
\subfloat[Hypernetwork output: $\pmb{s}^3(\pmb{\mu})$]{\includegraphics[width=0.33\columnwidth]{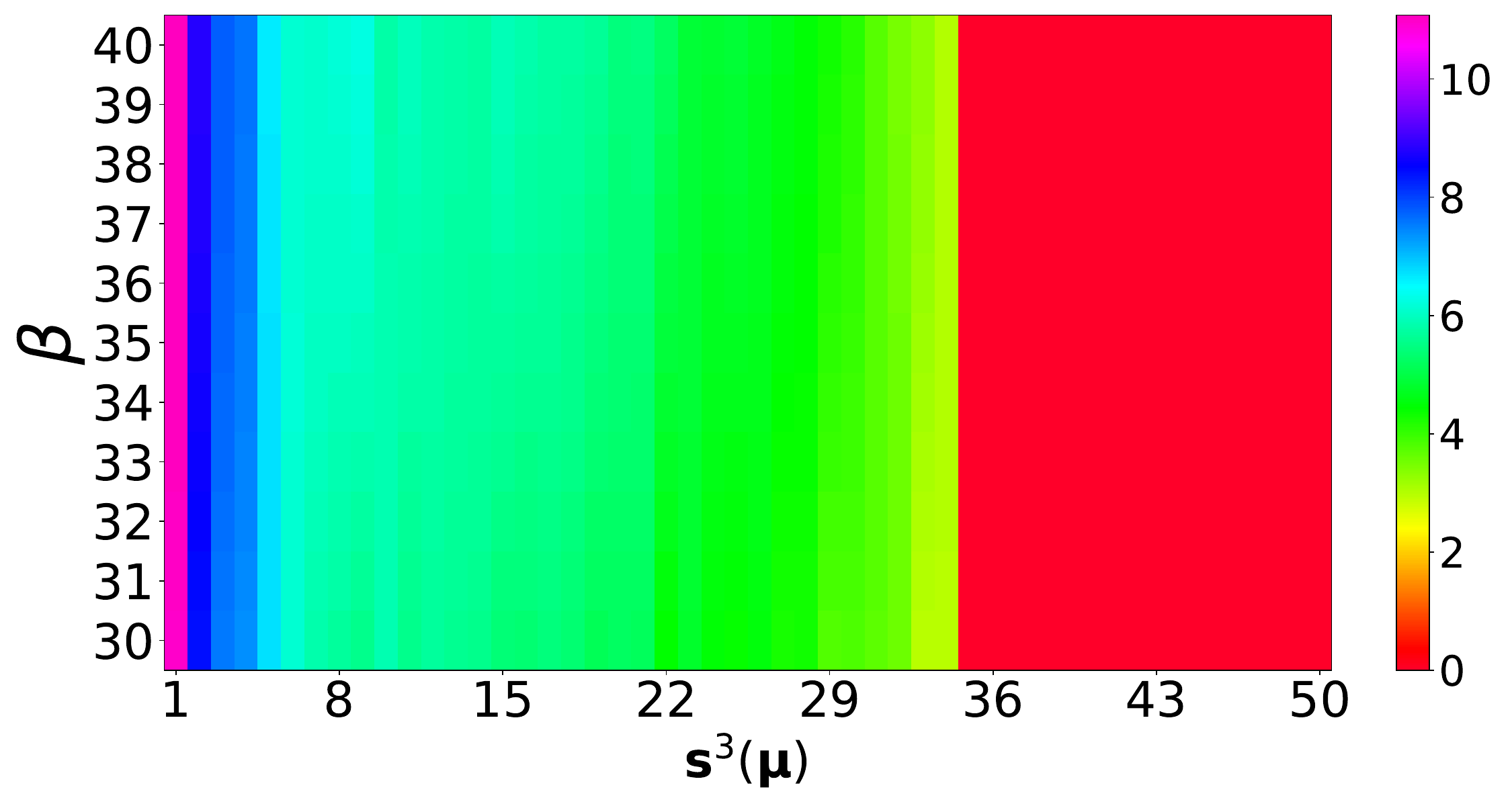}} \\
\caption{Convection equation ($\beta \in [30,40]$)}\label{fig:result_conv_30_40}
\end{figure}

\begin{figure}[ht!]
\centering
\subfloat[Hypernetwork output: $\pmb{s}^1(\pmb{\mu})$]{\includegraphics[width=0.33\columnwidth]{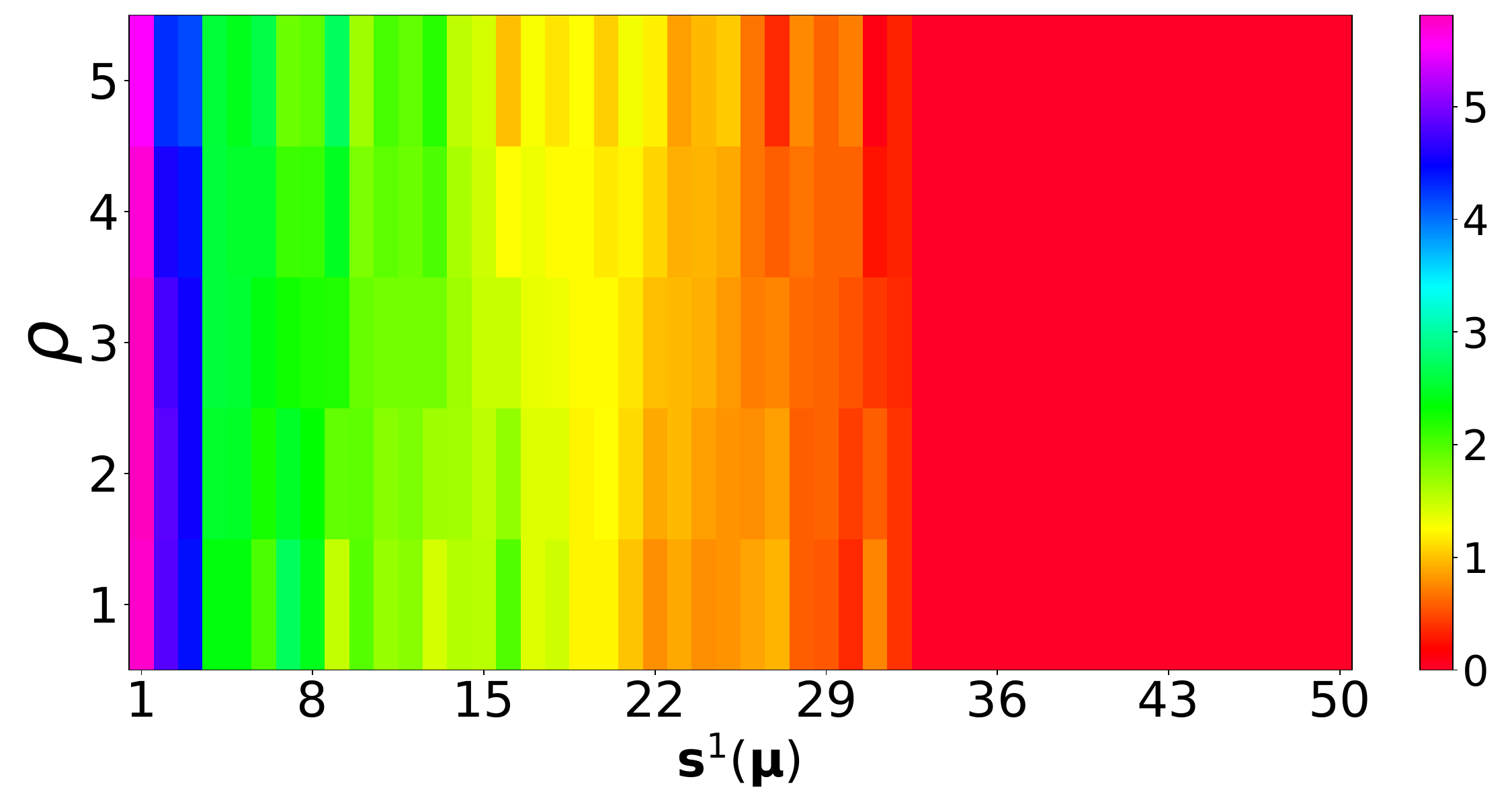}} \hfill
\subfloat[Hypernetwork output: $\pmb{s}^2(\pmb{\mu})$]{\includegraphics[width=0.33\columnwidth]{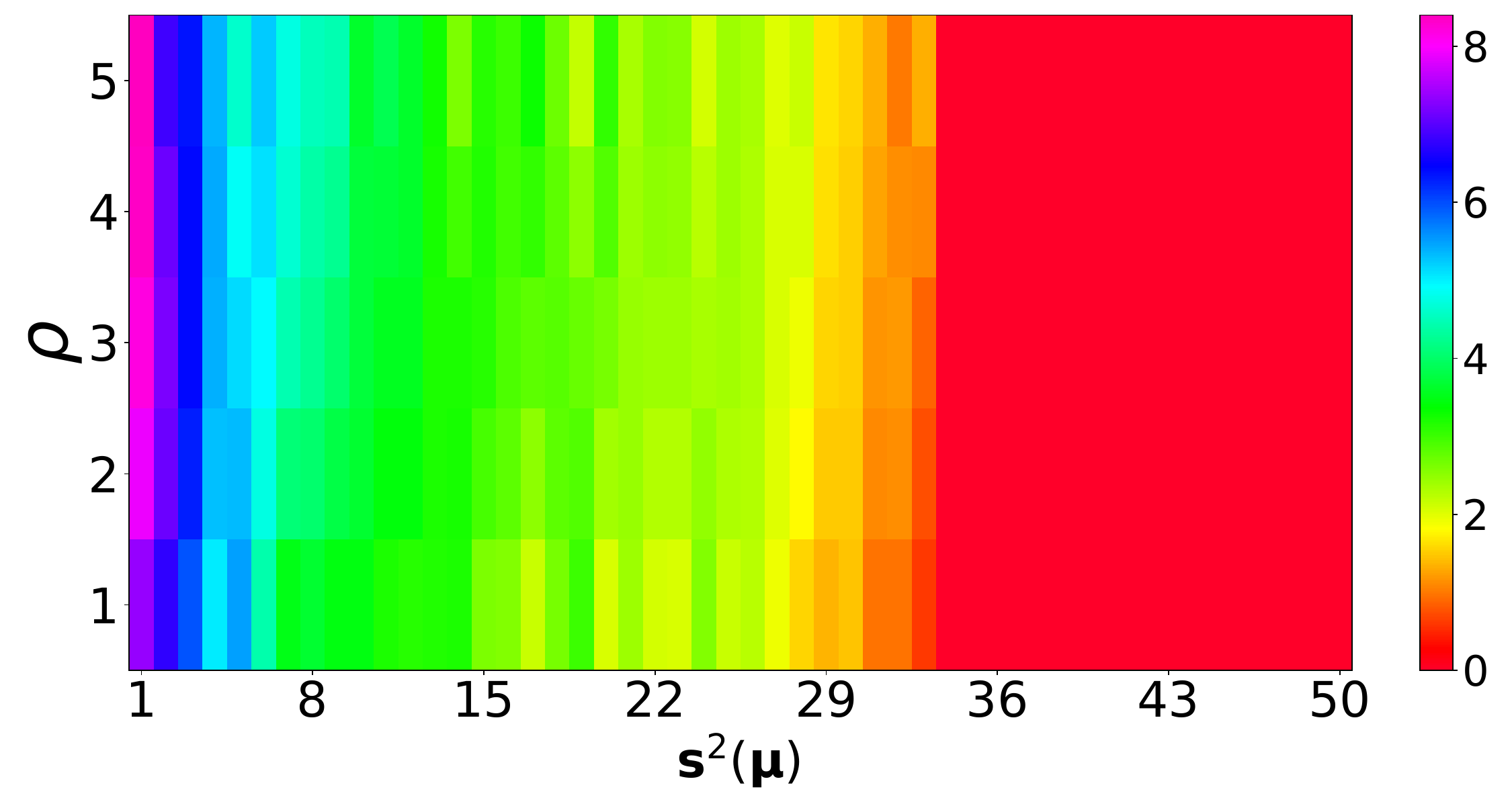}} \hfill
\subfloat[Hypernetwork output: $\pmb{s}^3(\pmb{\mu})$]{\includegraphics[width=0.33\columnwidth]{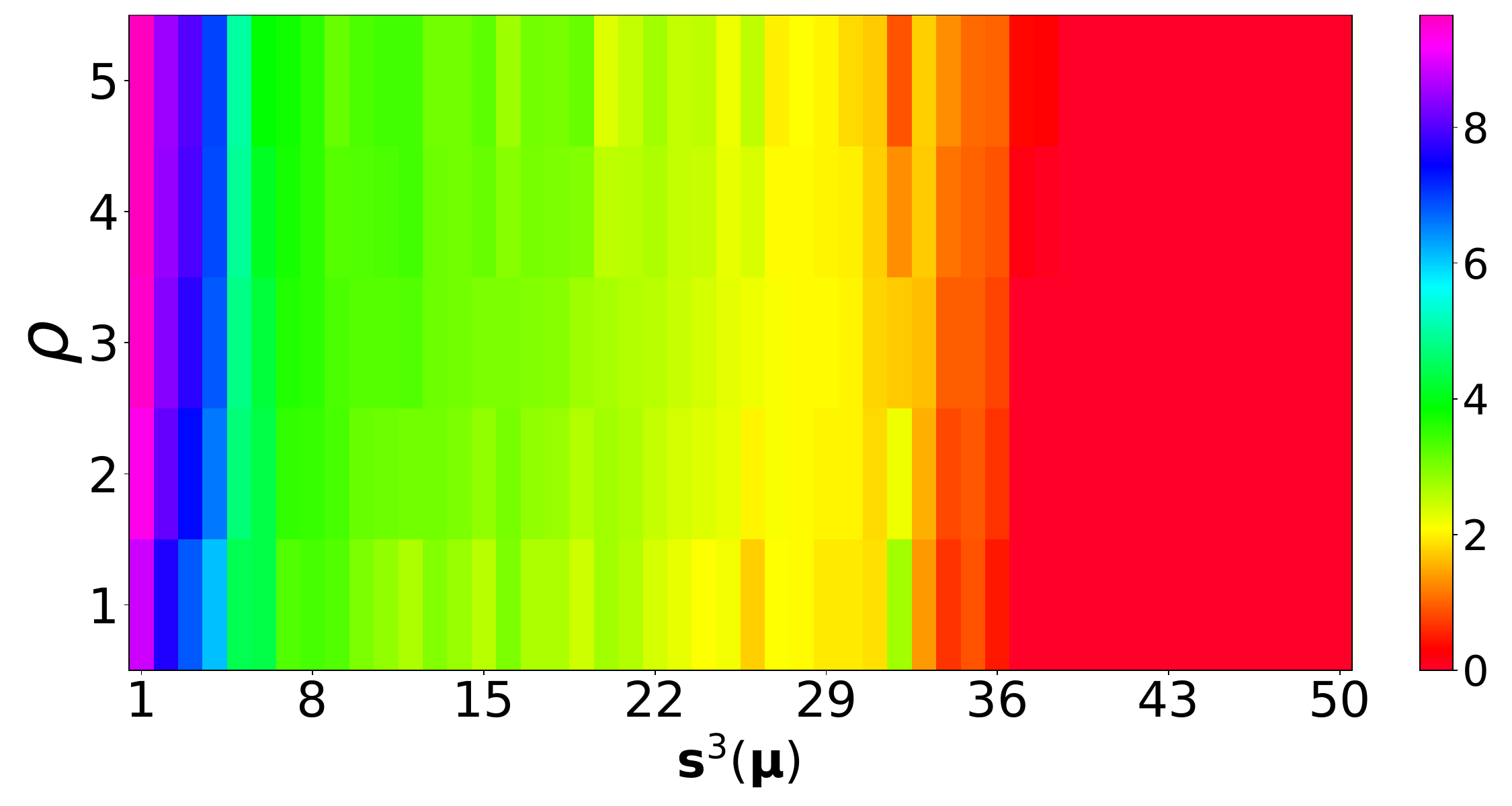}} \\
\caption{Reaction equation ($\rho\in [1,5]$)}\label{fig:result_reac_1_5}
\end{figure}

As we shown in the main manuscript, we visualize the magnitude of learned diagonal coefficients in the heatmap format. The results of the convection equations and reaction equations are plotted. One evident observation is that as we vary the PDE parameter in a wider range (e.g., $\beta\in [1,30]$) (as opposed to a shorter range, e.g., $\beta\in [30,40]$ or $\rho\in [1,5]$), the adaptive way of learning the rank results in more dynamic rank structure and, thus, will be more beneficial in terms of computational/memory efficiency. When the considered model parameters lie in a relative shorter range, the characteristics of the solutions are similar to each other and, thus, it does not require for the network to learn different rank structures (e.g., Figure~\ref{fig:result_conv_30_40}, Figure~\ref{fig:result_reac_1_5}). 
\begin{figure}[ht!]
\centering
\subfloat[Hypernetwork output: $\pmb{s}^1(\pmb{\mu})$]{\includegraphics[width=0.33\columnwidth]{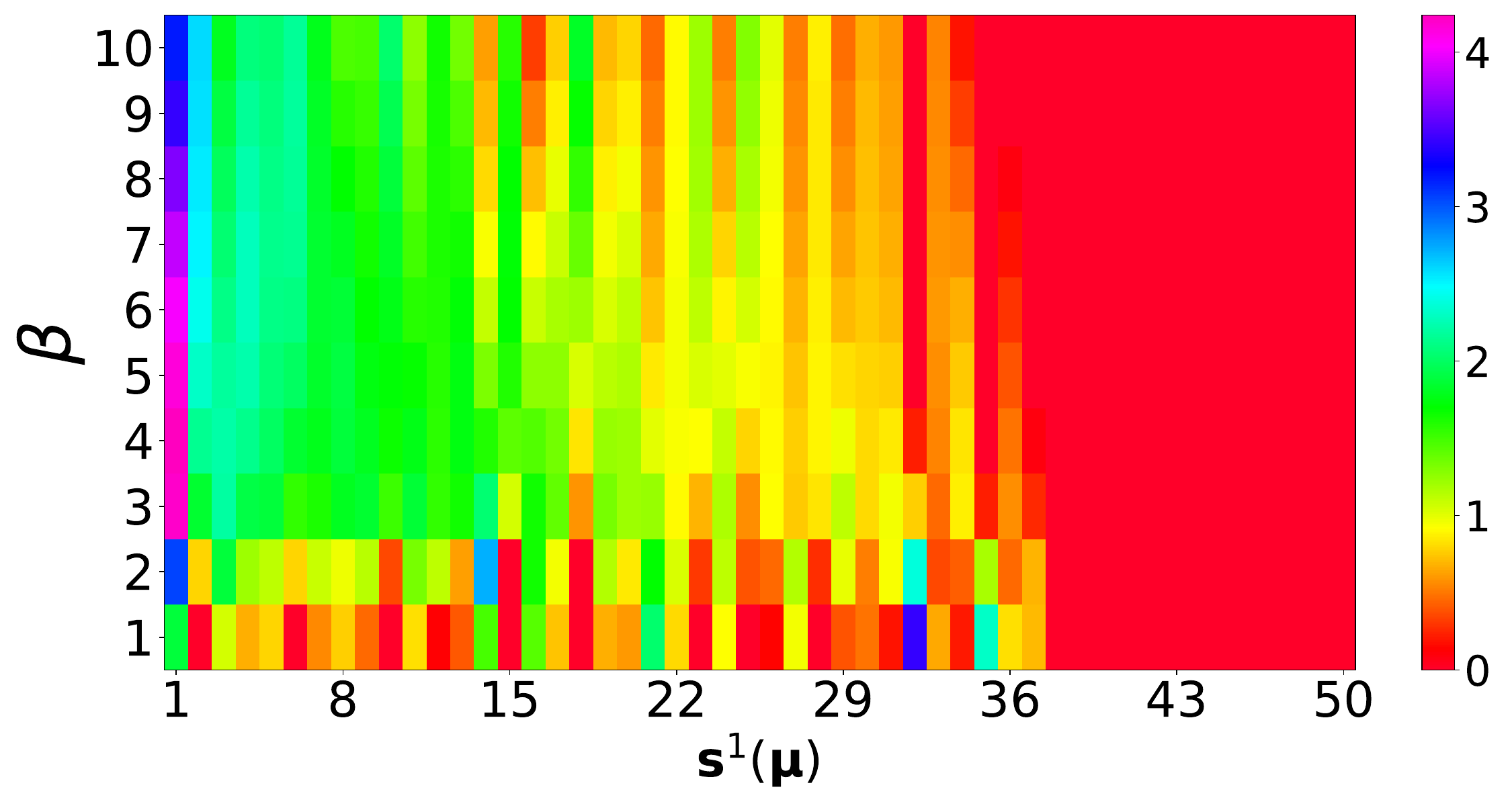}} \hfill
\subfloat[Hypernetwork output: $\pmb{s}^2(\pmb{\mu})$]{\includegraphics[width=0.33\columnwidth]{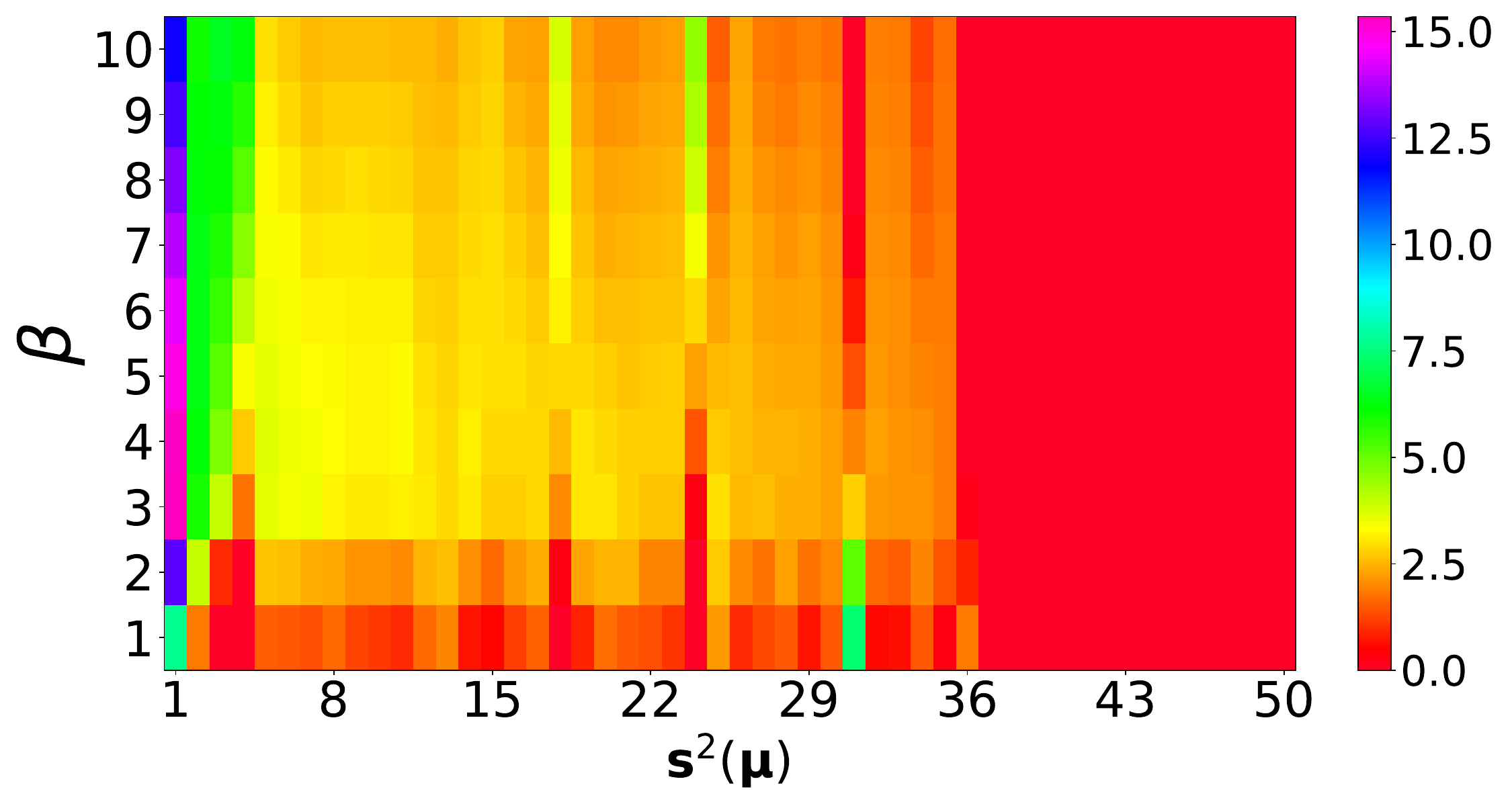}} \hfill
\subfloat[Hypernetwork output: $\pmb{s}^3(\pmb{\mu})$]{\includegraphics[width=0.33\columnwidth]{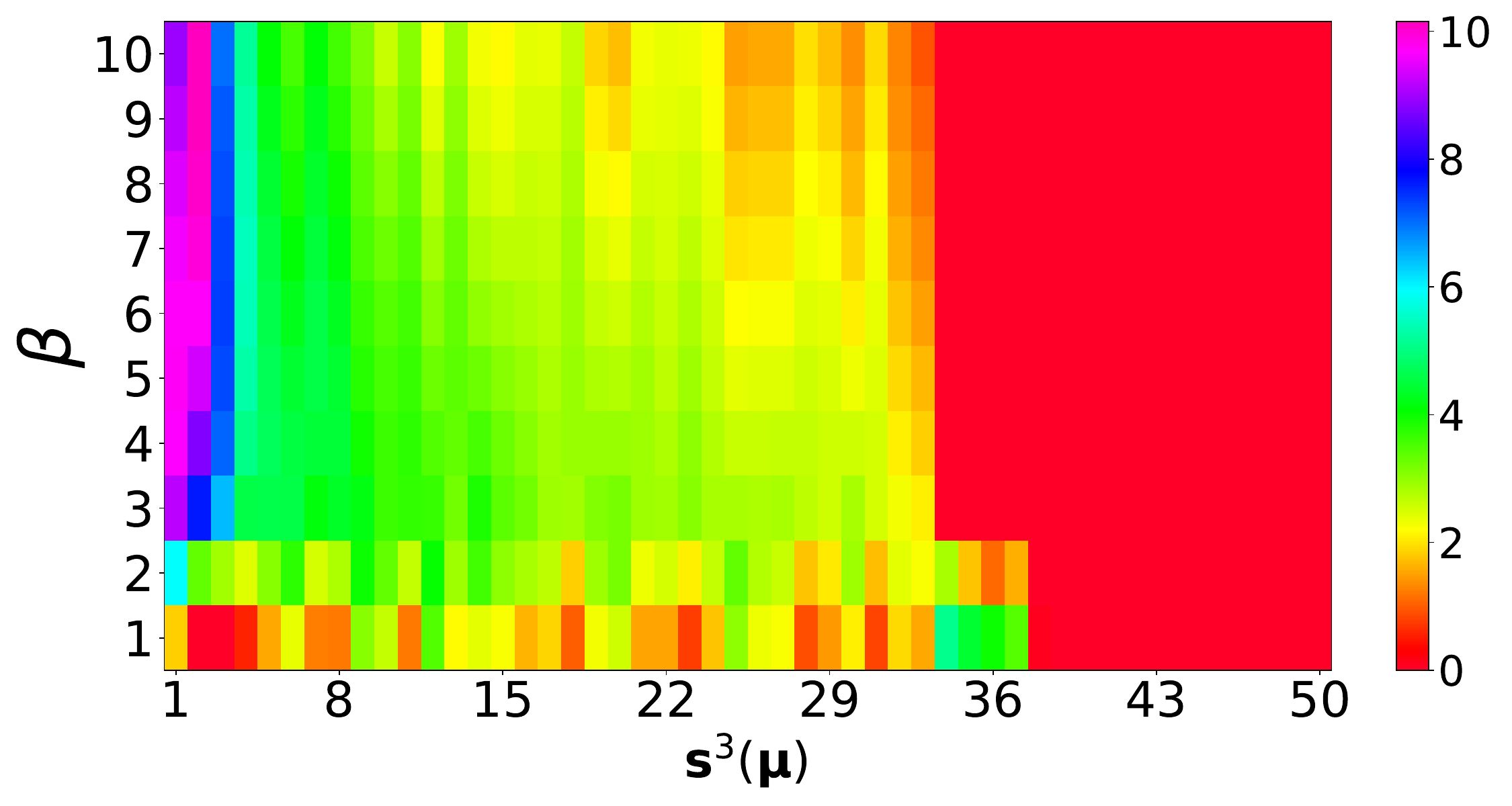}} \\
\caption{Convection equation ($\beta \in [1,10]$)}\label{fig:result_conv_1_10}
\end{figure}
\begin{figure}[ht!]
\centering
\subfloat[Hypernetwork output: $\pmb{s}^1(\pmb{\mu})$]{\includegraphics[width=0.33\columnwidth]{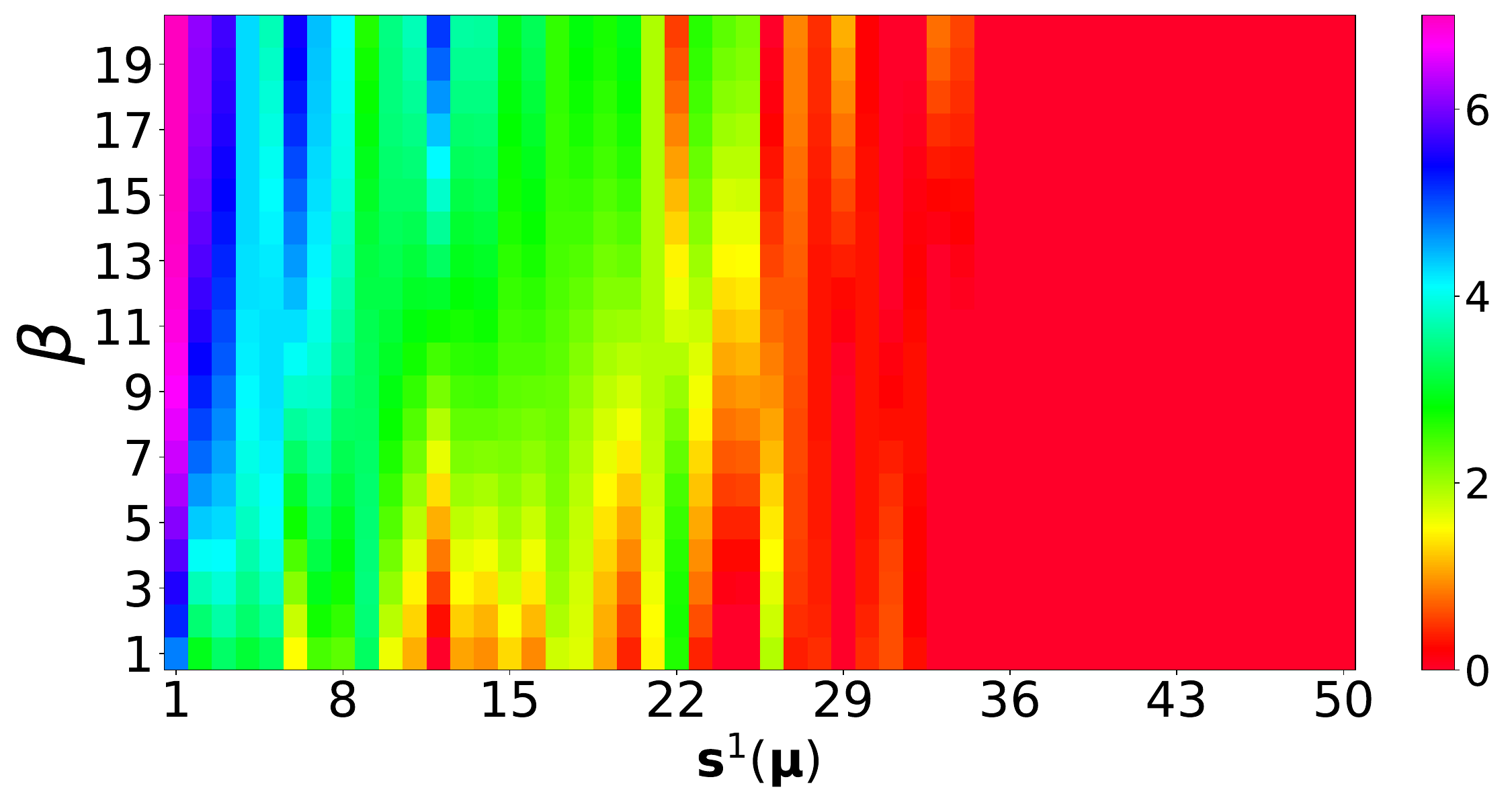}} \hfill
\subfloat[Hypernetwork output: $\pmb{s}^2(\pmb{\mu})$]{\includegraphics[width=0.33\columnwidth]{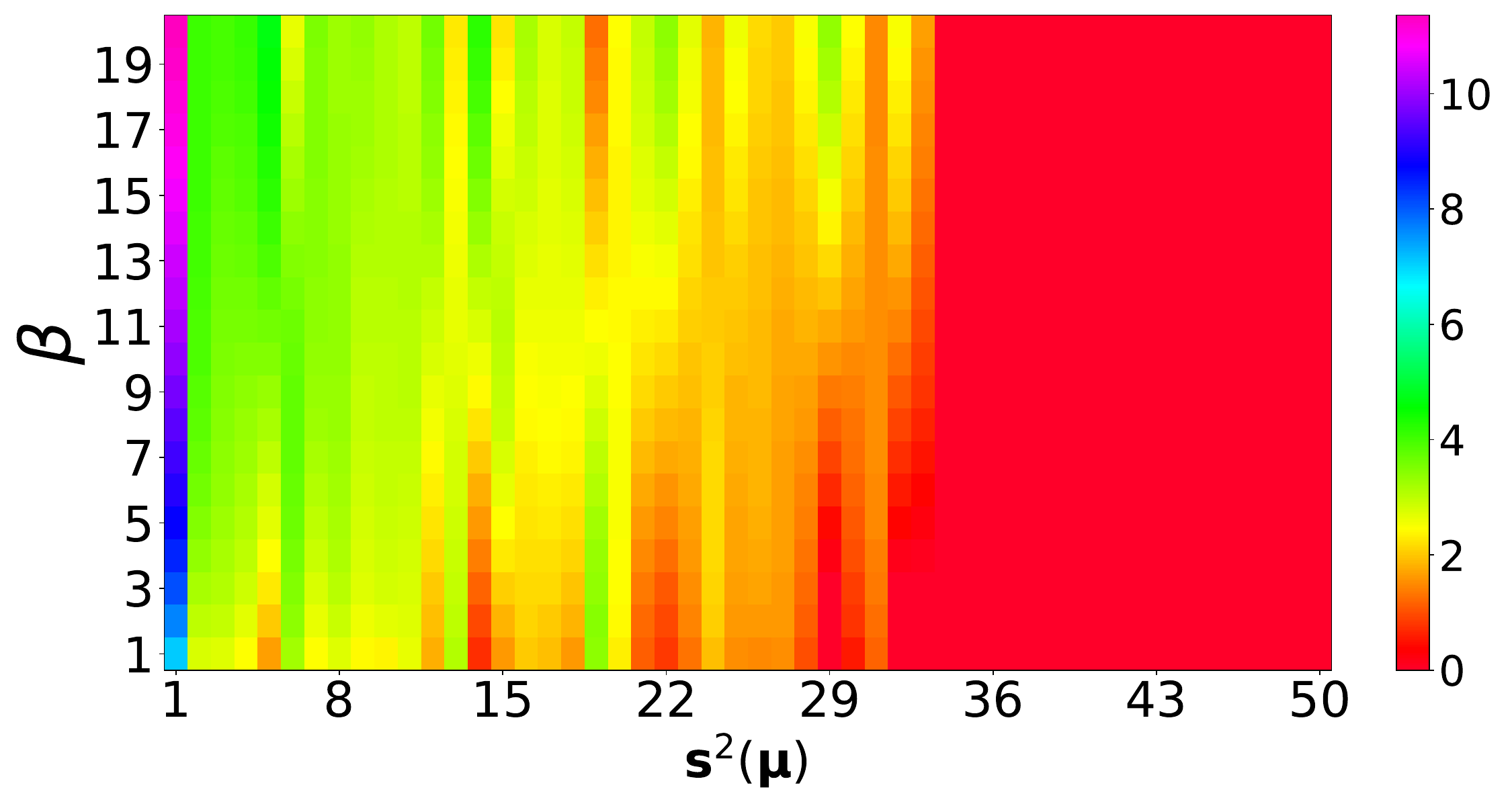}} \hfill
\subfloat[Hypernetwork output: $\pmb{s}^3(\pmb{\mu})$]{\includegraphics[width=0.33\columnwidth]{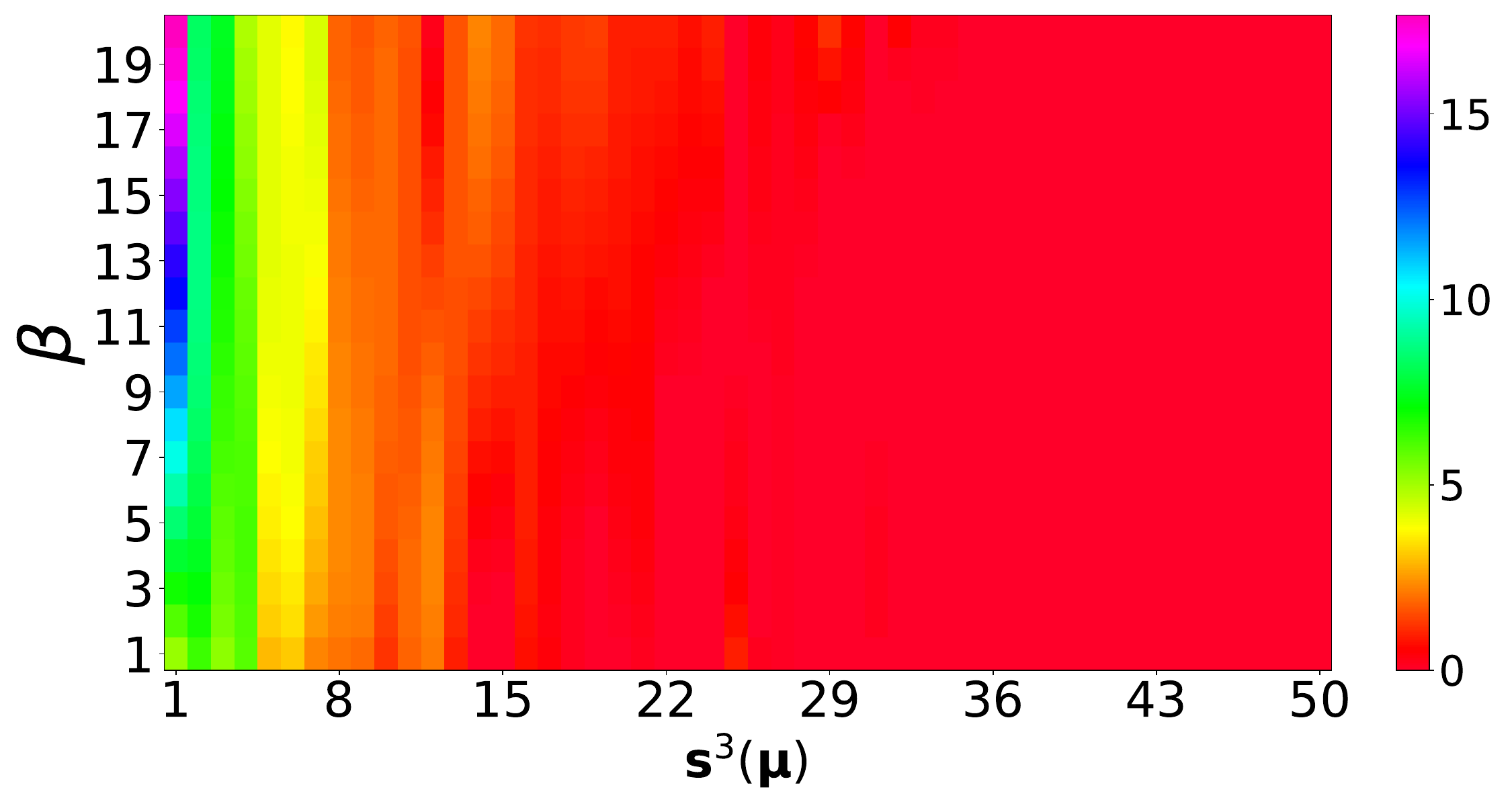}} \\
\caption{Convection equation ($\beta \in [1,20]$)}\label{fig:result_conv_1_20}
\end{figure}
\begin{figure}[ht!]
\centering
\subfloat[Hypernetwork output: $\pmb{s}^1(\pmb{\mu})$]{\includegraphics[width=0.33\columnwidth]{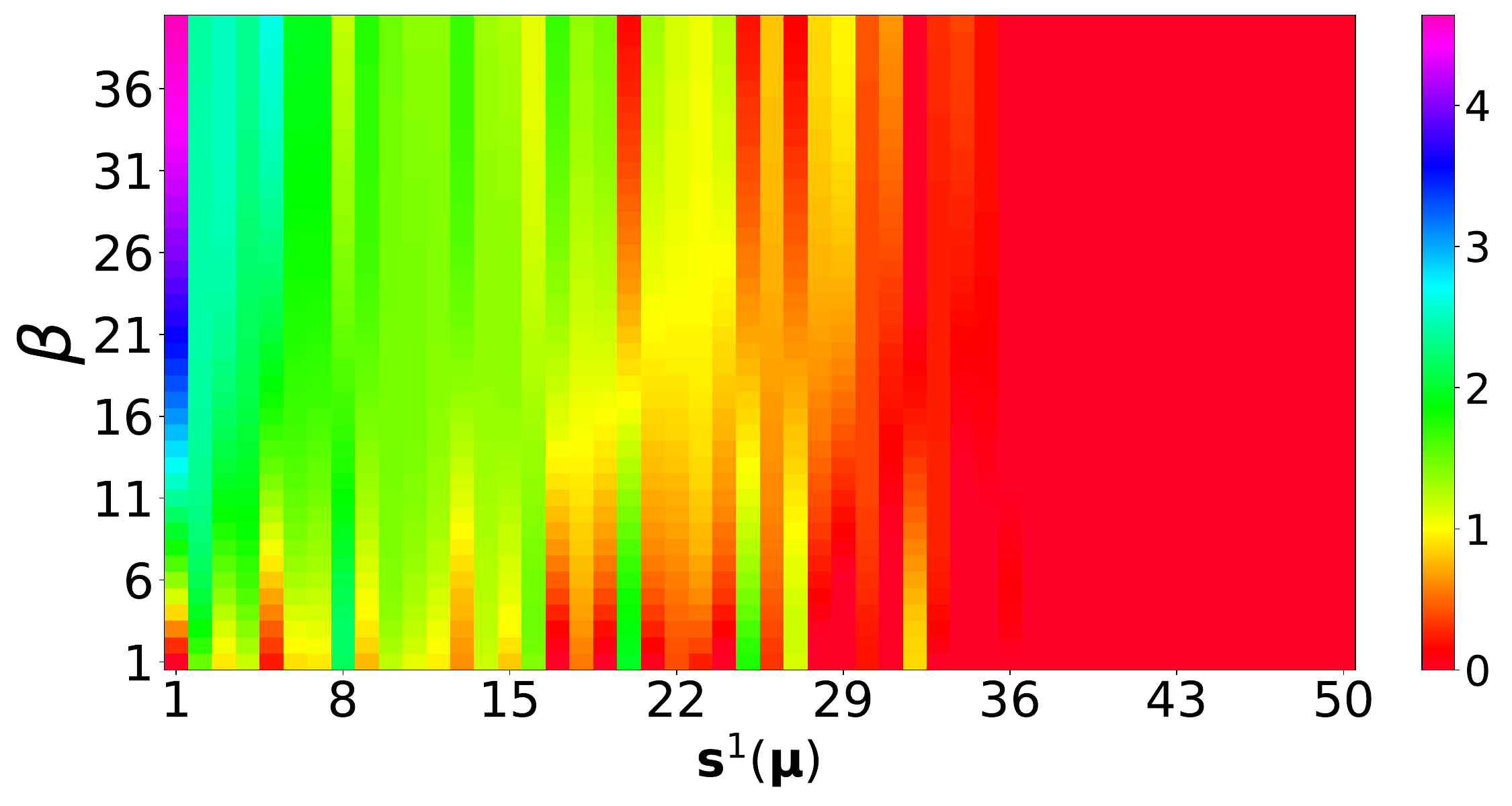}} \hfill
\subfloat[Hypernetwork output: $\pmb{s}^2(\pmb{\mu})$]{\includegraphics[width=0.33\columnwidth]{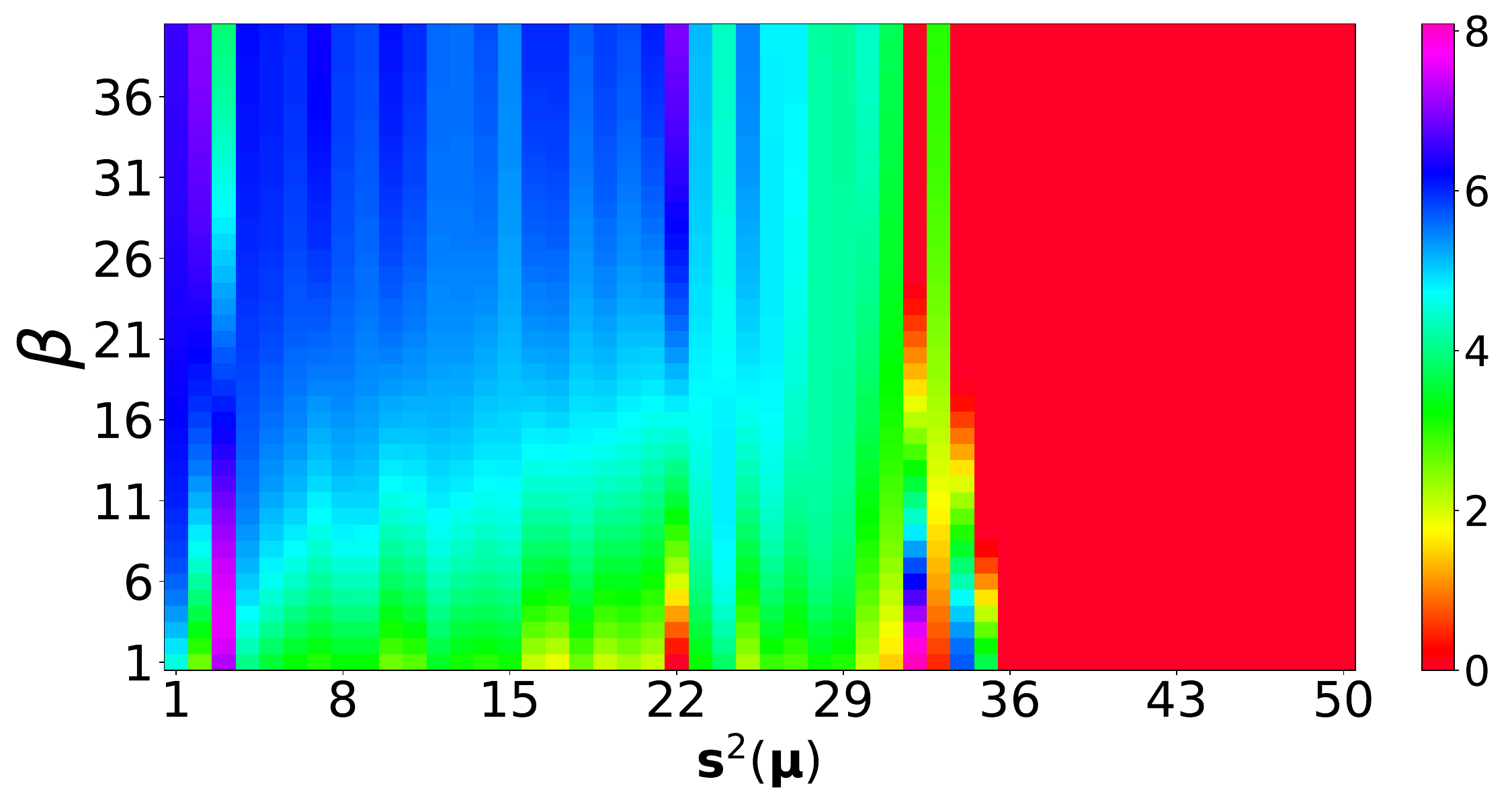}} \hfill
\subfloat[Hypernetwork output: $\pmb{s}^3(\pmb{\mu})$]{\includegraphics[width=0.33\columnwidth]{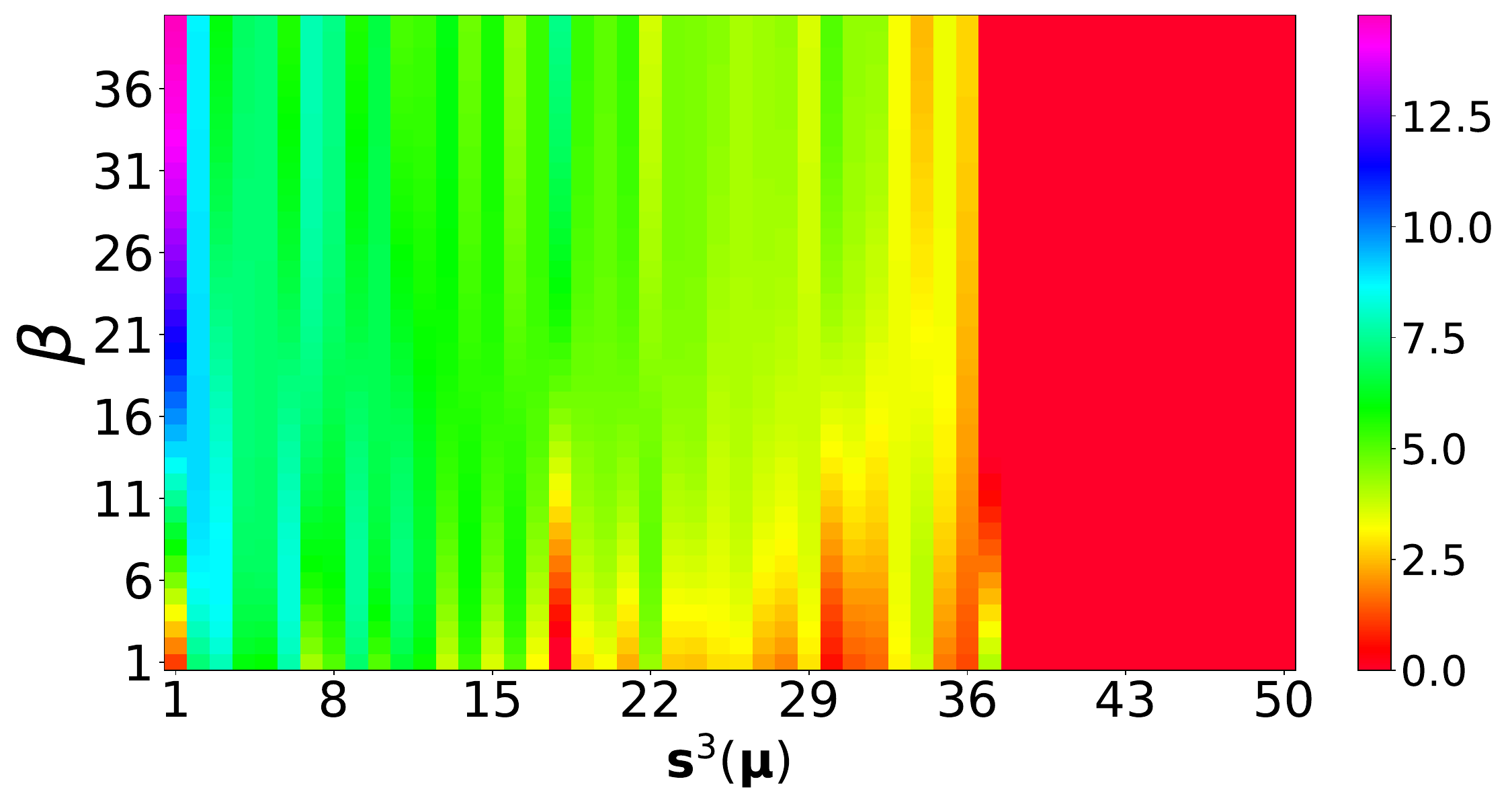}} \\
\caption{Convection equation ($\beta \in [1,40]$)}\label{fig:result_conv_1_40}
\end{figure}
\begin{figure}[ht!]
\centering
\subfloat[Hypernetwork output: $\pmb{s}^1(\pmb{\mu})$]{\includegraphics[width=0.33\columnwidth]{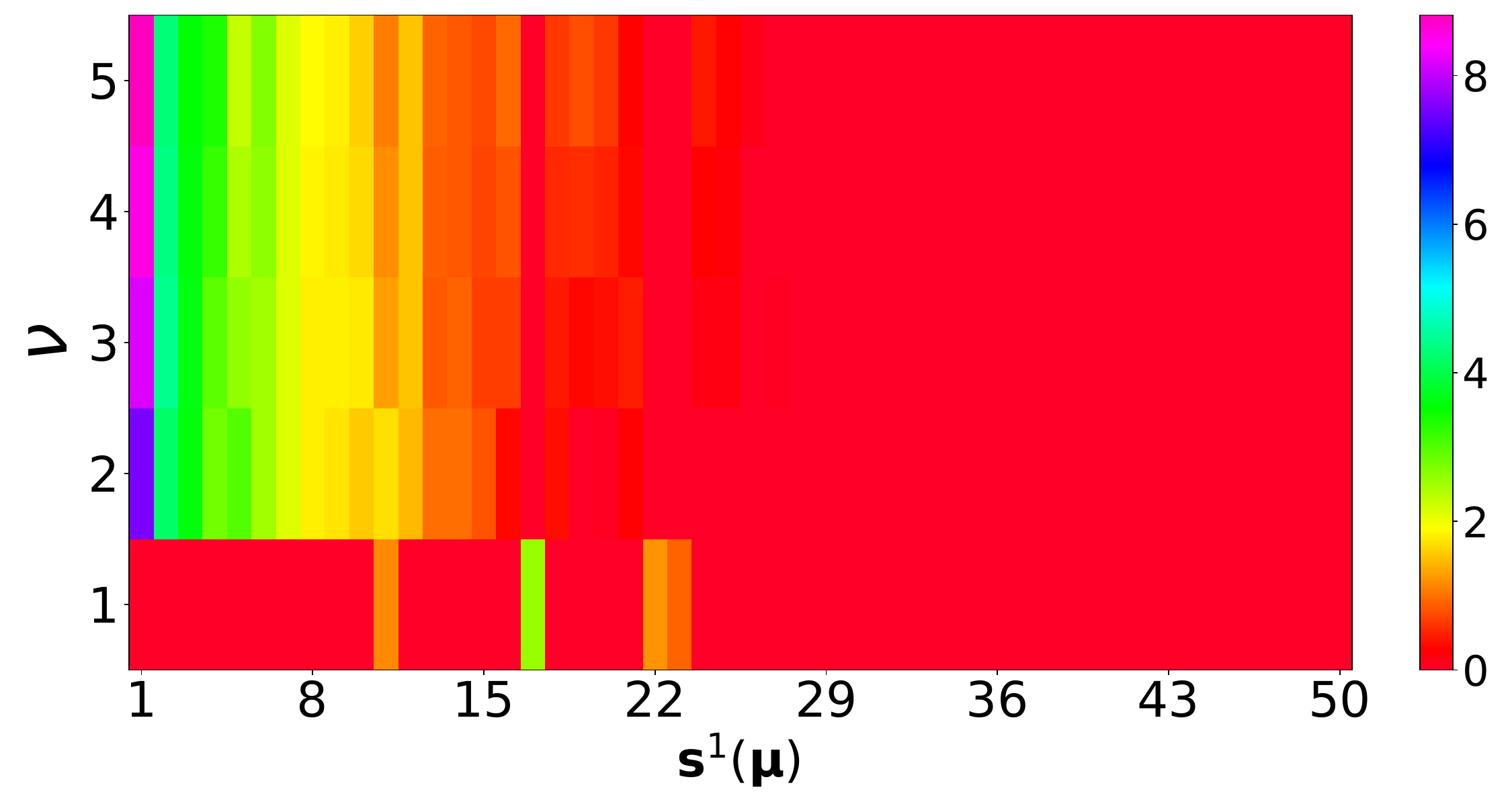}} \hfill
\subfloat[Hypernetwork output: $\pmb{s}^2(\pmb{\mu})$]{\includegraphics[width=0.33\columnwidth]{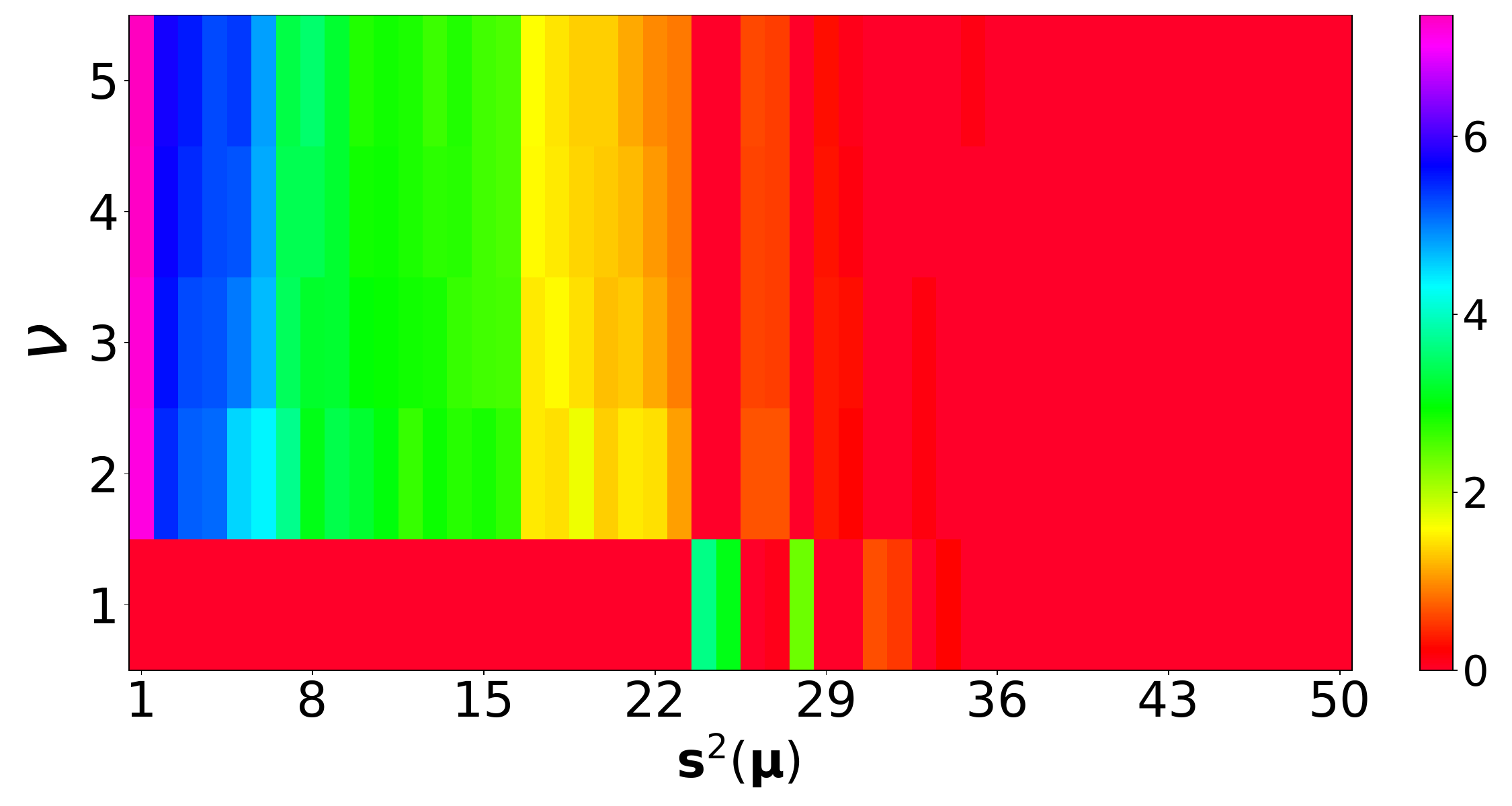}} \hfill
\subfloat[Hypernetwork output: $\pmb{s}^3(\pmb{\mu})$]{\includegraphics[width=0.33\columnwidth]{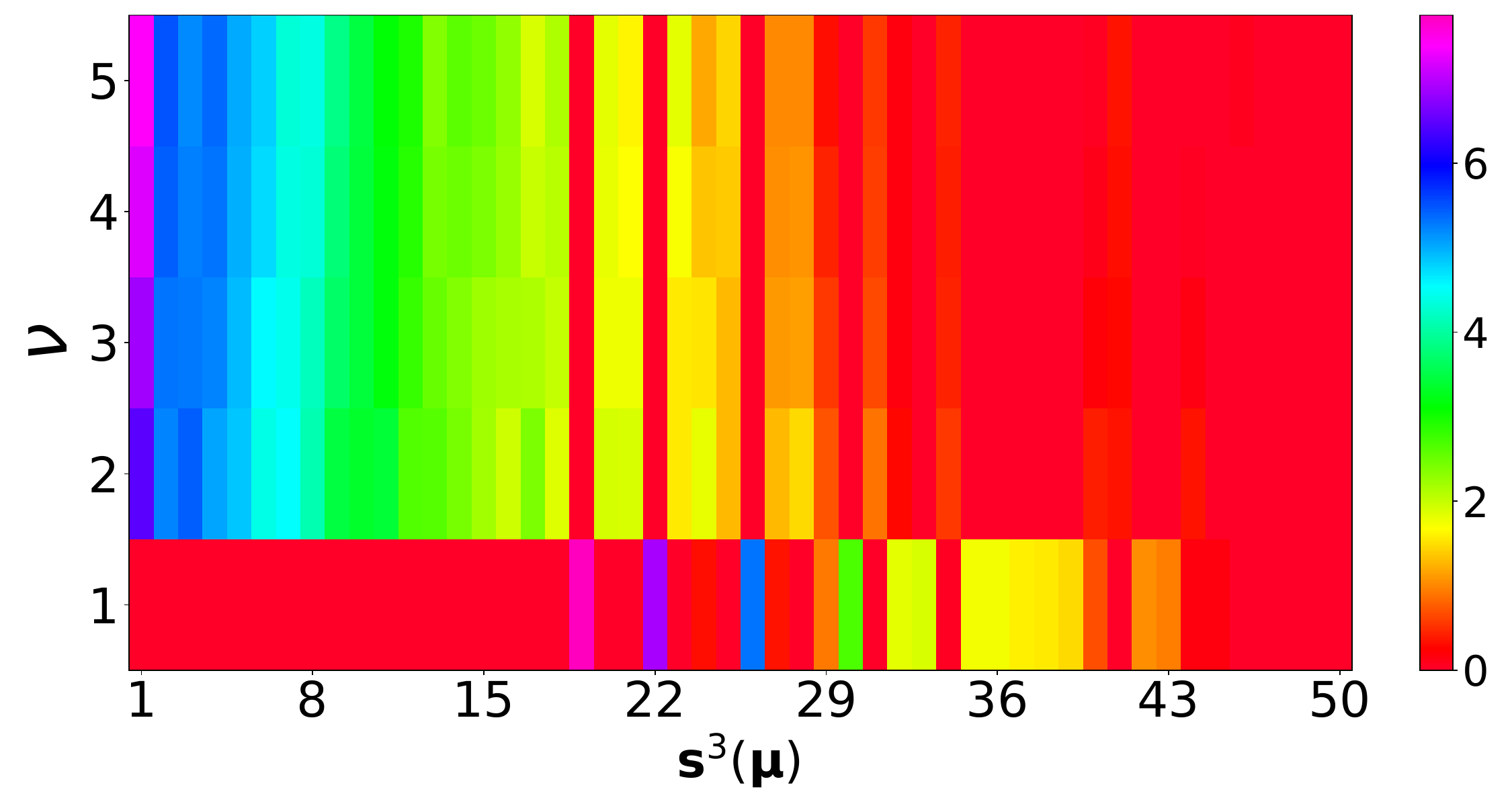}} \\
\caption{Reaction-diffusion equation ($\nu \in [1,5], \rho=5$)}
\end{figure}

Nevertheless, from the experiments of all considered PDEs, we observe the low-rank structures, which our model captures well and leading to the improved training process even compared to the well-known meta-learning algorithms (MAML and Reptile). Also, as we train our model by showing multiple PDEs describing similar physical phenomena, the proposed model overcomes the failure modes without any special learning approaches (i.e., curriculum or sequence-to-sequence).   

\section{More experimental results on general cases}\label{a:general_result}

To verify the performance of Hyper-LR-PINNs in the general cases, we measure the average performance of baselines and our method in various PDEs with $\beta, \nu, \rho\in[1,5]$. The initial condition of each equation is Gaussian distribution $N(\pi, (\pi/2)^2)$.

\begin{table}[h]
\centering
\caption{The average absolute and relative errors on general cases}\label{tab:l2_general}
\resizebox{\textwidth}{!}{
\begin{tabular}{lccccc|ccc}

\specialrule{1pt}{2pt}{2pt}
 \multirow{2}{*}{\textbf{PDE-type}}& \multirow{2}{*}{\textbf{Metric}} & \multicolumn{4}{c|}{\textbf{No pre-training}} & \multicolumn{3}{c}{\textbf{Meta-learning}} \\ \cmidrule{3-6}  \cmidrule{7-9}
 &  & \textbf{PINN} & \textbf{PINN-R} & \textbf{PINN-P} & \textbf{PINN-S2S} & \textbf{MAML} & \textbf{Reptile} & \textbf{Hyper-LR-PINN} \\

\specialrule{1pt}{2pt}{2pt}

\multirow{2}{*}{\textbf{Convection}} & Abs. err. & 0.0183 & 0.0222 & 0.0112 & 0.1281 & 0.0579 & 0.0173 & 0.0038 \\
 & Rel. err. & 0.0327 & 0.0381 & 0.0217 & 0.2160 & 0.1036 & 0.0347 & 0.0085 \\
 \cmidrule{1-9}
\multirow{2}{*}{\textbf{Diffusion}} & Abs. err. & 0.1335 & 0.1665 & 0.1433 & 0.1987 & 0.0803 & 0.0844 & 0.1169 \\
 & Rel. err. & 0.2733 & 0.3462 & 0.2920 & 0.4050 & 0.1673 & 0.1742 & 0.2458 \\
 \cmidrule{1-9}
\multirow{2}{*}{\textbf{Reaction}} & Abs. err. & 0.3341 & 0.3336 & 0.1749 & 0.4714 & 0.0029 & 0.0033 & 0.0025 \\
 & Rel. err. & 0.3907 & 0.3907 & 0.2024 & 0.5907 & 0.0057 & 0.0064 & 0.0045 \\
 \cmidrule{1-9}
\multirow{2}{*}{\textbf{Conv.-Diff.}} & Abs. err. & 0.0610 & 0.0654 & 0.0733 & 0.0979 & 0.0354 & 0.0372 & 0.0331 \\
 & Rel. err. & 0.1175 & 0.1289 & 0.1437 & 0.1950 & 0.0667 & 0.0713 & 0.0664 \\
 \cmidrule{1-9}
\multirow{2}{*}{\textbf{Reac.-Diff.}} & Abs. err. & 0.1900 & 0.1876 & 0.2201 & 0.4201 & 0.0310 & 0.0250 & 0.0684 \\
 & Rel. err. & 0.2702 & 0.2777 & 0.3179 & 0.5346 & 0.0537 & 0.0427 & 0.1168 \\
 \cmidrule{1-9}
\multirow{2}{*}{\textbf{C-D-R}} & Abs. err. & 0.1676 & 0.1629 & 0.1704 & 0.4878 & 0.0090 & 0.0461 & 0.0201 \\
 & Rel. err. & 0.2210 & 0.2149 & 0.2308 & 0.5983 & 0.0144 & 0.0701 & 0.0329 \\
\specialrule{1pt}{2pt}{2pt}
\end{tabular}
}
\end{table}

\section{Experimental results on failure modes} \label{a:fail_result}

In this section, we present the results of additional experiments on the failure mode. We show comparisons with baselines using additional metrics for convection equations, reaction equations, and reaction-diffusion equations, such as maximum error and explained variance. The initial condition of convection equations are $1+\sin(x)$, and the initial conditions of reaction equations and reaction-diffusion equations are Gaussian distribution $N(\pi, (\pi/4)^2)$. As shown in the following Tables, our Hyper-LR-PINNs overwhelmingly outperform other baselines.

\clearpage
\subsection{Convection equation}
For experiments on convection equations, meta-learning methods train convection equations in the range $\beta \in [30,40]$.

\begin{table}[!h]
\centering
\caption{The absolute and relative errors of the solutions of convection equations with $\beta = \{30, 35, 40\}$}\label{tab:result_conv_err_pinn}
\resizebox{\textwidth}{!}{
\begin{tabular}{lcccccccc}
\specialrule{1pt}{2pt}{2pt}
\multirow{2}{*}{$\pmb{\beta}$} & \multicolumn{2}{c}{\textbf{PINN}} & \multicolumn{2}{c}{\textbf{PINN-R}} & \multicolumn{2}{c}{\textbf{PINN-P}} & \multicolumn{2}{c}{\textbf{PINN-S2S}} \\ \cmidrule(lr){2-9}
 & \multicolumn{1}{c}{Abs. err.} & \multicolumn{1}{c}{Rel. err.} & \multicolumn{1}{c}{Abs. err.} & \multicolumn{1}{c}{Rel. err.} & \multicolumn{1}{c}{Abs. err.} & \multicolumn{1}{c}{Rel. err.} & \multicolumn{1}{c}{Abs. err.} & \multicolumn{1}{c}{Rel. err.} \\ 
 \specialrule{1pt}{2pt}{2pt}
\textbf{30} & 0.4015 & 0.4033 & 0.5435 & 0.5219 & 0.3821 & 0.3889 & 0.6342 & 0.5831 \\ \cmidrule(lr){1-9}
\textbf{35} & 0.4785 & 0.4701 & 0.5579 & 0.5309 & 0.1701 & 0.1621 & 0.6396 & 0.5868 \\ \cmidrule(lr){1-9}
\textbf{40} & 0.5490 & 0.5219 & 0.5897 & 0.5558 & 0.4988 & 0.4861 & 0.7319 & 0.7300 \\
\specialrule{1pt}{2pt}{2pt}
\end{tabular}}
\end{table}

\begin{table}[!h]
\centering
\caption{The max error and explained variance score of the solutions of convection equations with $\beta = \{30, 35, 40\}$}\label{tab:result_conv_max_var_pinn}
\resizebox{\textwidth}{!}{
\begin{tabular}{lcccccccc}
\specialrule{1pt}{2pt}{2pt}
\multirow{2}{*}{$\pmb{\beta}$} & \multicolumn{2}{c}{\textbf{PINN}} & \multicolumn{2}{c}{\textbf{PINN-R}} & \multicolumn{2}{c}{\textbf{PINN-P}} & \multicolumn{2}{c}{\textbf{PINN-S2S}} \\ \cmidrule(lr){2-9}
 & Max. err. & Exp. var. & Max. err. & Exp. var. & Max. err. & Exp. var. & Max. err. & Exp. var. \\
\specialrule{1pt}{2pt}{2pt}
\textbf{30} & 1.0471 & 0.5431 & 1.1105 & 0.2453 & 1.0394 & 0.5747 & 1.1719 & 0.0564 \\ \cmidrule(lr){1-9}
\textbf{35} & 1.0685 & 0.3435 & 1.0987 & 0.1707 & 0.4289 & 0.9216 & 1.3323 & 0.0092 \\ \cmidrule(lr){1-9}
\textbf{40} & 1.1018 & 0.2136 & 1.1356 & 0.1175 & 1.0714 & 0.3160 & 1.5677 & -0.0155 \\ 
\specialrule{1pt}{2pt}{2pt}
\end{tabular}}
\end{table}

\begin{table}[!h]
\centering
\small
\caption{
The max error and explained variance score of the solutions of convection equations with $\beta = \{30, 35, 40\}$}\label{tab:result_conv_max_var}
\resizebox{\textwidth}{!}{
\begin{tabular}{lccccccccccccc}
\specialrule{1pt}{2pt}{2pt}
\multirow{5}{*}{$\pmb{\beta}$} & \multirow{5}{*}{\textbf{Rank}} & \multicolumn{2}{c}{\textbf{{[}w/o{]} Pre-training}} & \multicolumn{10}{c}{\textbf{{[}w{]} Pre-training}} \\ \cmidrule(lr){3-4} \cmidrule(lr){5-14}
 &  & \multicolumn{2}{c}{\textbf{Na\"ive-LR-PINN}} & \multicolumn{2}{c}{\begin{tabular}[c]{@{}c@{}}\textbf{Curriculum}\\ \textbf{learning}\end{tabular}} & \multicolumn{2}{c}{\textbf{MAML}} & \multicolumn{2}{c}{\textbf{Reptile}} & \multicolumn{2}{c}{\begin{tabular}[c]{@{}c@{}}\textbf{Hyper-LR-PINN}\\ \textbf{(Full rank)}\end{tabular}} & \multicolumn{2}{c}{\begin{tabular}[c]{@{}c@{}}\textbf{Hyper-LR-PINN}\\ \textbf{(Adaptive rank)}\end{tabular}} \\ \cmidrule(lr){3-14}
 &  & \multicolumn{1}{c}{Max. err.} & \multicolumn{1}{c}{Exp. var.} & \multicolumn{1}{c}{Max. err.} & \multicolumn{1}{c}{Exp. var.} & \multicolumn{1}{c}{Max. err.} & \multicolumn{1}{c}{Exp. var.} & \multicolumn{1}{c}{Max. err.} & \multicolumn{1}{c}{Exp. var.} & Max. err. & Exp. var. & Max. err. & Exp. var. \\
\specialrule{1pt}{2pt}{2pt}
\multirow{7}{*}{\textbf{30}} & 10 & 1.1096 & 0.2067 & 1.0303 & 0.5321 & 1.3550 & 0.0050 & 1.1645 & 0.1445 & \multirow{7}{*}{0.1270} & \multirow{7}{*}{0.9959} & \multirow{7}{*}{0.1229} & \multirow{7}{*}{0.9957} \\ \cmidrule(lr){2-10}
 & 20 & 1.0961 & 0.2322 & 1.0188 & 0.5539 & 1.4006 & 0.0074 & 1.1810 & 0.1008 &  &  &  &  \\ \cmidrule(lr){2-10}
 & 30 & 1.1901 & 0.1314 & 1.0417 & 0.5084 & 1.1445 & 0.1662 & 1.1780 & 0.1214 &  &  &  &  \\ \cmidrule(lr){2-10}
 & 40 & 1.1140 & 0.2850 & 0.9798 & 0.6097 & 1.1580 & 0.1542 & 1.1240 & 0.1712 &  &  &  &  \\ \cmidrule(lr){2-10}
 & 50 & 1.1114 & 0.2709 & 1.0346 & 0.5257 & 1.1670 & 0.1428 & 1.1430 & 0.1592 &  &  &  &  \\
\specialrule{1pt}{2pt}{2pt}
\multirow{7}{*}{\textbf{35}} & 10 & 1.1088 & 0.1592 & 1.1725 & 0.1177 & 1.4062 & 0.0031 & 1.1540 & 0.1260 & \multirow{7}{*}{0.1269} & \multirow{7}{*}{0.9941} & \multirow{7}{*}{0.1390} & \multirow{7}{*}{0.9936} \\ \cmidrule(lr){2-10}
 & 20 & 1.1025 & 0.1546 & 1.1867 & 0.0693 & 1.4412 & -0.0047 & 1.1911 & 0.0966 &  &  &  &  \\ \cmidrule(lr){2-10}
 & 30 & 1.1805 & 0.0784 & 1.1753 & 0.1134 & 1.1683 & 0.1172 & 1.1752 & 0.1125 &  &  &  &  \\ \cmidrule(lr){2-10}
 & 40 & 1.0959 & 0.1949 & 1.1646 & 0.1084 & 1.1603 & 0.1255 & 1.1480 & 0.1276 &  &  &  &  \\ \cmidrule(lr){2-10}
 & 50 & 1.0967 & 0.1999 & 1.1883 & 0.0900 & 1.1759 & 0.1119 & 1.1377 & 0.1358 &  &  &  &  \\
\specialrule{1pt}{2pt}{2pt}
\multirow{7}{*}{\textbf{40}} & 10 & 1.1718 & 0.1008 & 1.1671 & 0.0953 & 1.4451 & 0.0007 & 1.1741 & 0.0908 & \multirow{7}{*}{0.2437} & \multirow{7}{*}{0.9875} & \multirow{7}{*}{0.2672} & \multirow{7}{*}{0.9848} \\
\cmidrule(lr){2-10}
 & 20 & 1.1259 & 0.1225 & 1.1781 & 0.0469 & 1.4981 & -0.0075 & 1.2091 & 0.0715 &  &  &  &  \\ \cmidrule(lr){2-10}
 & 30 & 1.1853 & 0.0757 & 1.1859 & 0.0894 & 1.1878 & 0.0787 & 1.2095 & 0.0817 &  &  &  &  \\ \cmidrule(lr){2-10}
 & 40 & 1.1133 & 0.1953 & 1.1864 & 0.0843 & 1.2176 & 0.0789 & 1.1717 & 0.0885 &  &  &  &  \\ \cmidrule(lr){2-10}
 & 50 & 1.2450 & 0.0947 & 1.1880 & 0.0730 & 1.1987 & 0.0743 & 1.1791 & 0.0886 &  &  &  &  \\ 
\specialrule{1pt}{2pt}{2pt}
\end{tabular}}
\end{table}

\clearpage
\subsection{Reaction equation}
For experiments on reaction equations, meta-learning methods train reaction equations in the range $\rho \in [1,5]$.

\begin{table}[!h]
\centering
\caption{The absolute and relative errors of the solutions of reaction equations with $\rho = \{4, 5, 6, 7\}$}\label{tab:result_reac_err_pinn}
\resizebox{\textwidth}{!}{
\begin{tabular}{lcccccccc}
\specialrule{1pt}{2pt}{2pt}
\multirow{2}{*}{$\pmb{\rho}$} & \multicolumn{2}{c}{\textbf{PINN}} & \multicolumn{2}{c}{\textbf{PINN-R}} & \multicolumn{2}{c}{\textbf{PINN-P}} & \multicolumn{2}{c}{\textbf{PINN-S2S}} \\ \cmidrule(lr){2-9}
 & Abs. err. & Rel. err. & Abs. err. & Rel. err. & Abs. err. & Rel. err. & Abs. err. & Rel. err. \\
\specialrule{1pt}{2pt}{2pt}
\textbf{4} & 0.0242 & 0.0750 & 0.0260 & 0.0720 & 0.0208 & 0.0614 & 0.4085 & 0.8280 \\ \cmidrule(lr){1-9}
\textbf{5} & 0.5501 & 0.9862 & 0.0334 & 0.1017 & 0.4981 & 0.9349 & 0.4497 & 0.8205 \\ \cmidrule(lr){1-9}
\textbf{6} & 0.5987 & 0.9913 & 0.0403 & 0.1200 & 0.5981 & 0.9906 & 0.5434 & 0.9068 \\ \cmidrule(lr){1-9}
\textbf{7} & 0.6431 & 0.9950 & 0.0275 & 0.0848 & 0.6459 & 0.9976 & 0.6285 & 0.9794 \\
\specialrule{1pt}{2pt}{2pt}
\end{tabular}}
\end{table}

\begin{table}[!h]
\caption{The absolute and relative errors of the solutions of reaction equations with $\rho = \{4, 5, 6, 7\}$}\label{tab:result_reac_err}
\renewcommand{\arraystretch}{0.95}
\centering
\small
\resizebox{\textwidth}{!}{
\begin{tabular}{lccccccccccccc}
\specialrule{1pt}{2pt}{2pt}
\multirow{5}{*}{$\pmb{\rho}$} & \multirow{5}{*}{\textbf{Rank}} & \multicolumn{2}{c}{\textbf{{[}w/o{]} Pre-training}} & \multicolumn{10}{c}{\textbf{{[}w{]} Pre-training}} \\ \cmidrule(lr){3-4} \cmidrule(lr){5-14}
 &  & \multicolumn{2}{c}{\textbf{Na\"ive-LR-PINN}} & \multicolumn{2}{c}{\begin{tabular}[c]{@{}c@{}}\textbf{Curriculum}\\ \textbf{learning}\end{tabular}} & \multicolumn{2}{c}{\textbf{MAML}} & \multicolumn{2}{c}{\textbf{Reptile}} & \multicolumn{2}{c}{\begin{tabular}[c]{@{}c@{}}\textbf{Hyper-LR-PINN}\\ \textbf{(Full rank)}\end{tabular}} & \multicolumn{2}{c}{\begin{tabular}[c]{@{}c@{}}\textbf{Hyper-LR-PINN}\\ \textbf{(Adaptive rank)}\end{tabular}} \\ \cmidrule(lr){3-14}
 &  & \multicolumn{1}{c}{Abs. err.} & \multicolumn{1}{c}{Rel. err.} & \multicolumn{1}{c}{Abs. err.} & \multicolumn{1}{c}{Rel. err.} & \multicolumn{1}{c}{Abs. err.} & \multicolumn{1}{c}{Rel. err.} & \multicolumn{1}{c}{Abs. err.} & \multicolumn{1}{c}{Rel. err.} & Abs. err. & Rel. err. & Abs. err. & Rel. err. \\
\specialrule{1pt}{2pt}{2pt}
\multirow{7}{*}{\textbf{4}} & 10 & 0.0448 & 0.1221 & 0.1287 & 0.2897 & 0.1509 & 0.3396 & 0.1426 & 0.3108 & \multirow{7}{*}{0.0126} & \multirow{7}{*}{0.0395} & \multirow{7}{*}{0.0151} & \multirow{7}{*}{0.0460} \\ \cmidrule(lr){2-10}
 & 20 & 0.0498 & 0.1362 & 0.1333 & 0.2977 & 0.1476 & 0.3305 & 0.1243 & 0.3017 &  &  &  &  \\ \cmidrule(lr){2-10}
 & 30 & 0.0511 & 0.1494 & 0.1203 & 0.2683 & 0.0949 & 0.2237 & 0.0570 & 0.1534 &  &  &  &  \\ \cmidrule(lr){2-10}
 & 40 & 0.0425 & 0.1247 & 0.1182 & 0.2629 & 0.1218 & 0.2947 & 0.1076 & 0.2528 &  &  &  &  \\ \cmidrule(lr){2-10}
 & 50 & 0.0551 & 0.1498 & 0.1195 & 0.2644 & 0.0749 & 0.1913 & 0.0741 & 0.1912 &  &  &  &  \\
\specialrule{1pt}{2pt}{2pt}
\multirow{7}{*}{\textbf{5}} & 10 & 0.5381 & 0.9647 & 0.1349 & 0.3031 & 0.5166 & 0.9273 & 0.5472 & 0.9806 & \multirow{7}{*}{0.0157} & \multirow{7}{*}{0.0518} & \multirow{7}{*}{0.0211} & \multirow{7}{*}{0.0655} \\ \cmidrule(lr){2-10}
 & 20 & 0.5471 & 0.9810 & 0.1372 & 0.3049 & 0.5489 & 0.9816 & 0.5483 & 0.9845 &  &  &  &  \\ \cmidrule(lr){2-10}
 & 30 & 0.5385 & 0.9664 & 0.1209 & 0.2724 & 0.5378 & 0.9662 & 0.5454 & 0.9778 &  &  &  &  \\ \cmidrule(lr){2-10}
 & 40 & 0.5413 & 0.9696 & 0.1208 & 0.2701 & 0.5321 & 0.9538 & 0.5450 & 0.9785 &  &  &  &  \\ \cmidrule(lr){2-10}
 & 50 & 0.5475 & 0.9822 & 0.1186 & 0.2648 & 0.5478 & 0.9815 & 0.5536 & 0.9906 &  &  &  &  \\
\specialrule{1pt}{2pt}{2pt}
\multirow{7}{*}{\textbf{6}} & 10 & 0.5987 & 0.9912 & 0.1355 & 0.2957 & 0.5926 & 0.9806 & 0.5969 & 0.9882 & \multirow{7}{*}{0.0167} & \multirow{7}{*}{0.0552} & \multirow{7}{*}{0.0183} & \multirow{7}{*}{0.0572} \\ \cmidrule(lr){2-10}
 & 20 & 0.5981 & 0.9905 & 0.1303 & 0.2822 & 0.5988 & 0.9906 & 0.5989 & 0.9919 &  &  &  &  \\ \cmidrule(lr){2-10}
 & 30 & 0.5959 & 0.9863 & 0.1306 & 0.2854 & 0.5947 & 0.9853 & 0.5992 & 0.9931 &  &  &  &  \\ \cmidrule(lr){2-10}
 & 40 & 0.5988 & 0.9917 & 0.1196 & 0.2613 & 0.5951 & 0.9854 & 0.6001 & 0.9936 &  &  &  &  \\ \cmidrule(lr){2-10}
 & 50 & 0.5989 & 0.9919 & 0.1194 & 0.2586 & 0.6004 & 0.9930 & 0.6022 & 0.9970 &  &  &  &  \\
\specialrule{1pt}{2pt}{2pt}
\multirow{7}{*}{\textbf{7}} & 10 & 0.6431 & 0.9949 & 0.1448 & 0.3011 & 0.6394 & 0.9890 & 0.6413 & 0.9922 & \multirow{7}{*}{0.0180} & \multirow{7}{*}{0.0575} & \multirow{7}{*}{0.0203} & \multirow{7}{*}{0.0456} \\ \cmidrule(lr){2-10}
 & 20 & 0.6425 & 0.9942 & 0.1363 & 0.2837 & 0.6410 & 0.9914 & 0.6423 & 0.9941 &  &  &  &  \\ \cmidrule(lr){2-10}
 & 30 & 0.6424 & 0.9938 & 0.1285 & 0.2764 & 0.6401 & 0.9903 & 0.6431 & 0.9955 &  &  &  &  \\ \cmidrule(lr){2-10}
 & 40 & 0.6430 & 0.9949 & 0.1145 & 0.2445 & 0.6401 & 0.9901 & 0.6436 & 0.9958 &  &  &  &  \\ \cmidrule(lr){2-10}
 & 50 & 0.6428 & 0.9948 & 0.1140 & 0.2418 & 0.6417 & 0.9927 & 0.6437 & 0.9962 &  &  &  &  \\
\specialrule{1pt}{2pt}{2pt}
\end{tabular}}
\end{table}

\clearpage

\begin{table}[!h]
\centering
\caption{The max error and explained variance score of the solutions of reaction equations with $\rho = \{4, 5, 6, 7\}$}\label{tab:result_reac_max_var_pinn}
\resizebox{\textwidth}{!}{
\begin{tabular}{lcccccccc}
\specialrule{1pt}{2pt}{2pt}
\multirow{2}{*}{$\pmb{\rho}$} & \multicolumn{2}{c}{\textbf{PINN}} & \multicolumn{2}{c}{\textbf{PINN-R}} & \multicolumn{2}{c}{\textbf{PINN-P}} & \multicolumn{2}{c}{\textbf{PINN-S2S}} \\ \cmidrule(lr){2-9}
 & Max. err. & Exp. var. & Max. err. & Exp. var. & Max. err. & Exp. var. & Max. err. & Exp. var. \\
\specialrule{1pt}{2pt}{2pt}
\textbf{4} & 0.2282 & 0.9881 & 0.2055 & 0.9881 & 0.1681 & 0.9921 & 0.9895 & 0.0514 \\ \cmidrule(lr){1-9}
\textbf{5} & 0.9961 & 0.0288 & 0.3297 & 0.9745 & 1.0077 & -0.0375 & 0.9746 & -0.0890 \\ \cmidrule(lr){1-9}
\textbf{6} & 0.9982 & 0.0163 & 0.4422 & 0.9596 & 0.9983 & 0.0158 & 1.0877 & -0.1455 \\ \cmidrule(lr){1-9}
\textbf{7} & 0.9996 & 0.0088 & 0.4038 & 0.9750 & 1.0042 & 0.0139 & 1.0267 & -0.0224 \\
\specialrule{1pt}{2pt}{2pt}
\end{tabular}}
\end{table}

\begin{table}[h]
\caption{
The max error and explained variance score of the solutions of reaction equations with $\rho = \{4, 5, 6, 7\}$}\label{tab:result_reac_max_var}
\renewcommand{\arraystretch}{0.95}
\centering
\small
\resizebox{\textwidth}{!}{
\begin{tabular}{lccccccccccccc}
\specialrule{1pt}{2pt}{2pt}
\multirow{5}{*}{$\pmb{\rho}$} & \multirow{5}{*}{\textbf{Rank}} & \multicolumn{2}{c}{\textbf{{[}w/o{]} Pre-training}} & \multicolumn{10}{c}{\textbf{{[}w{]} Pre-training}} \\ \cmidrule(lr){3-4} \cmidrule(lr){5-14}
 &  & \multicolumn{2}{c}{\textbf{Na\"ive-LR-PINN}} & \multicolumn{2}{c}{\begin{tabular}[c]{@{}c@{}}\textbf{Curriculum}\\ \textbf{learning}\end{tabular}} & \multicolumn{2}{c}{\textbf{MAML}} & \multicolumn{2}{c}{\textbf{Reptile}} & \multicolumn{2}{c}{\begin{tabular}[c]{@{}c@{}}\textbf{Hyper-LR-PINN}\\ \textbf{(Full rank)}\end{tabular}} & \multicolumn{2}{c}{\begin{tabular}[c]{@{}c@{}}\textbf{Hyper-LR-PINN}\\ \textbf{(Adaptive rank)}\end{tabular}} \\ \cmidrule(lr){3-14}
 &  & \multicolumn{1}{c}{Max. err.} & \multicolumn{1}{c}{Exp. var.} & \multicolumn{1}{c}{Max. err.} & \multicolumn{1}{c}{Exp. var.} & \multicolumn{1}{c}{Max. err.} & \multicolumn{1}{c}{Exp. var.} & \multicolumn{1}{c}{Max. err.} & \multicolumn{1}{c}{Exp. var.} & Max. err. & Exp. var. & Max. err. & Exp. var. \\
\specialrule{1pt}{2pt}{2pt}
\multirow{7}{*}{\textbf{4}} & 10 & 0.3395 & 0.9699 & 0.6069 & 0.8453 & 0.6653 & 0.8064 & 0.5896 & 0.8443 & \multirow{7}{*}{0.1356} & \multirow{7}{*}{0.9964} & \multirow{7}{*}{0.1524} & \multirow{7}{*}{0.9953} \\ \cmidrule(lr){2-10}
 & 20 & 0.3618 & 0.9649 & 0.5921 & 0.8352 & 0.6820 & 0.8162 & 0.6349 & 0.8420 &  &  &  &  \\ \cmidrule(lr){2-10}
 & 30 & 0.4223 & 0.9556 & 0.5635 & 0.8770 & 0.5249 & 0.9131 & 0.3748 & 0.9559 &  &  &  &  \\ \cmidrule(lr){2-10}
 & 40 & 0.3601 & 0.9680 & 0.5438 & 0.8825 & 0.6538 & 0.8423 & 0.5659 & 0.8874 &  &  &  &  \\ \cmidrule(lr){2-10}
 & 50 & 0.3900 & 0.9576 & 0.5507 & 0.8805 & 0.4316 & 0.9350 & 0.4767 & 0.9320 &  &  &  &  \\
\specialrule{1pt}{2pt}{2pt}
\multirow{7}{*}{\textbf{5}} & 10 & 0.9828 & 0.0690 & 0.6997 & 0.8061 & 0.9639 & 0.1355 & 0.9951 & 0.0418 & \multirow{7}{*}{0.1967} & \multirow{7}{*}{0.9929} & \multirow{7}{*}{0.2287} & \multirow{7}{*}{0.9891} \\ \cmidrule(lr){2-10}
 & 20 & 0.9909 & 0.0371 & 0.6969 & 0.8030 & 0.9910 & 0.0482 & 0.9951 & 0.0261 &  &  &  &  \\ \cmidrule(lr){2-10}
 & 30 & 0.9850 & 0.0626 & 0.6576 & 0.8472 & 0.9883 & 0.0591 & 0.9865 & 0.0452 &  &  &  &  \\ \cmidrule(lr){2-10}
 & 40 & 0.9824 & 0.0640 & 0.6464 & 0.8525 & 0.9776 & 0.0917 & 0.9935 & 0.0377 &  &  &  &  \\ \cmidrule(lr){2-10}
 & 50 & 0.9899 & 0.0333 & 0.6405 & 0.8568 & 0.9928 & 0.0399 & 0.9980 & 0.0279 &  &  &  &  \\
\specialrule{1pt}{2pt}{2pt}
\multirow{7}{*}{\textbf{6}} & 10 & 1.0073 & 0.0163 & 0.7566 & 0.7896 & 0.9942 & 0.0402 & 0.9989 & 0.0223 & \multirow{7}{*}{0.2405} & \multirow{7}{*}{0.9905} & \multirow{7}{*}{0.2418} & \multirow{7}{*}{0.9899} \\ \cmidrule(lr){2-10}
 & 20 & 0.9986 & 0.0165 & 0.7293 & 0.8042 & 0.9968 & 0.0223 & 1.0113 & 0.0136 &  &  &  &  \\ \cmidrule(lr){2-10}
 & 30 & 0.9967 & 0.0272 & 0.7319 & 0.8111 & 0.9985 & 0.0253 & 1.0118 & 0.0081 &  &  &  &  \\ \cmidrule(lr){2-10}
 & 40 & 1.0089 & 0.0135 & 0.6944 & 0.8407 & 0.9965 & 0.0278 & 1.0064 & 0.0118 &  &  &  &  \\ \cmidrule(lr){2-10}
 & 50 & 1.0071 & 0.0131 & 0.6918 & 0.8448 & 1.0089 & 0.0178 & 1.0231 & 0.0049 &  &  &  &  \\
\specialrule{1pt}{2pt}{2pt}
\multirow{7}{*}{\textbf{7}} & 10 & 1.0115 & 0.0084 & 0.7502 & 0.7513 & 0.9991 & 0.0225 & 1.0014 & 0.0142 & \multirow{7}{*}{0.2922} & \multirow{7}{*}{0.9882} & \multirow{7}{*}{0.2150} & \multirow{7}{*}{0.9920}  \\ \cmidrule(lr){2-10}
 & 20 & 1.0048 & 0.0092 & 0.7136 & 0.7820 & 0.9996 & 0.0180 & 1.0064 & 0.0080 &  &  &  &  \\ \cmidrule(lr){2-10}
 & 30 & 1.0057 & 0.0116 & 0.7468 & 0.7961 & 1.0001 & 0.0183 & 1.0126 & 0.0044 &  &  &  &  \\ \cmidrule(lr){2-10}
 & 40 & 1.0110 & 0.0075 & 0.6827 & 0.8394 & 0.9994 & 0.0193 & 1.0061 & 0.0071 &  &  &  &  \\ \cmidrule(lr){2-10}
 & 50 & 1.0090 & 0.0077 & 0.6765 & 0.8443 & 1.0008 & 0.0137 & 1.0142 & 0.0046 &  &  &  &  \\
\specialrule{1pt}{2pt}{2pt}
\end{tabular}}
\end{table}

\clearpage

\subsection{Reaction-diffusion equation}
For experiments on reaction-diffusion equations, meta-learning methods train reaction equations in the range $\nu \in [1,5], \rho=5$. To learn reaction-diffusion equations, PINN baselines employ 2000 epochs, but meta-learning methods train only 10 epochs, after pre-training.

\begin{table}[h]
\centering
\caption{The absolute and relative errors of the solutions of reaction-diffusion equations with $\nu = \{4, 5, 6, 7\}, \rho=5$}\label{tab:result_reac_diff_err_pinn}
\resizebox{\textwidth}{!}{
\begin{tabular}{lcccccccc}
\specialrule{1pt}{2pt}{2pt}
\multirow{2}{*}{$\pmb{(\nu, \rho)}$} & \multicolumn{2}{c}{\textbf{PINN}} & \multicolumn{2}{c}{\textbf{PINN-R}} & \multicolumn{2}{c}{\textbf{PINN-P}} & \multicolumn{2}{c}{\textbf{PINN-S2S}} \\ \cmidrule(lr){2-9}
 & Abs. err. & Rel. err. & Abs. err. & Rel. err. & Abs. err. & Rel. err. & Abs. err. & Rel. err. \\
\specialrule{1pt}{2pt}{2pt}
\textbf{(4, 5)} & 0.2909 & 0.4902 & 0.2909 & 0.4914 & 0.2804 & 0.4744 & 0.7265 & 0.9950 \\ \cmidrule(lr){1-9}
\textbf{(5, 5)} & 0.3186 & 0.5134 & 0.3364 & 0.5400 & 0.3380 & 0.5456 & 0.7383 & 1.0037 \\ \cmidrule(lr){1-9}
\textbf{(6, 5)} & 0.3183 & 0.4891 & 0.3810 & 0.5884 & 0.4255 & 0.6648 & 0.7265 & 0.9881 \\ \cmidrule(lr){1-9}
\textbf{(7, 5)} & 0.4860 & 0.7273 & 0.4126 & 0.6273 & 0.5229 & 0.7925 & 0.6630 & 0.9213 \\
\specialrule{1pt}{2pt}{2pt}
\end{tabular}}
\end{table}

\begin{table}[h]
\caption{The absolute and relative errors of the solutions of reaction-diffusion equations with $\nu = \{4, 5, 6, 7\}, \rho=5$}\label{tab:result_reac_diff_err}
\renewcommand{\arraystretch}{0.95}
\resizebox{\textwidth}{!}{
\begin{tabular}{lccccccccccccc}
\specialrule{1pt}{2pt}{2pt}
\multirow{4}{*}{($\pmb{\nu, \rho}$)} & \multirow{4}{*}{\textbf{Rank}} & \multicolumn{2}{c}{\textbf{{[}w/o{]} Pre-training}} & \multicolumn{10}{c}{\textbf{{[}w{]} Pre-training}} \\ \cmidrule(lr){3-4} \cmidrule(lr){5-14}
 &  & \multicolumn{2}{c}{\multirow{2}{*}{\textbf{Naive-LR-PINN}}} & \multicolumn{2}{c}{\multirow{2}{*}{\begin{tabular}[c]{@{}c@{}}\textbf{Curriculum}\\ \textbf{learning}\end{tabular}}} & \multicolumn{2}{c}{\multirow{2}{*}{\textbf{MAML}}} & \multicolumn{2}{c}{\multirow{2}{*}{\textbf{Reptile}}} & \multicolumn{2}{c}{\multirow{2}{*}{\begin{tabular}[c]{@{}c@{}}\textbf{Hyper-LR-PINN}\\ \textbf{(Full rank)}\end{tabular}}} & \multicolumn{2}{c}{\multirow{2}{*}{\begin{tabular}[c]{@{}c@{}}\textbf{Hyper-LR-PINN}\\ \textbf{(Adaptive rank)}\end{tabular}}} \\ 
 &  & \multicolumn{2}{c}{} & \multicolumn{2}{c}{} & \multicolumn{2}{c}{} & \multicolumn{2}{c}{} & \multicolumn{2}{c}{} & \multicolumn{2}{c}{} \\ \cmidrule(lr){3-14}
 &  & \multicolumn{1}{c}{Abs. err.} & \multicolumn{1}{c}{Rel. err.} & \multicolumn{1}{c}{Abs. err.} & \multicolumn{1}{c}{Rel. err.} & \multicolumn{1}{c}{Abs. err.} & \multicolumn{1}{c}{Rel. err.} & \multicolumn{1}{c}{Abs. err.} & \multicolumn{1}{c}{Rel. err.} & Abs. err. & Rel. err. & Abs. err. & Rel. err. \\
\specialrule{1pt}{2pt}{2pt}
\multirow{7}{*}{\textbf{(4, 5)}} & 10 & 0.0333 & 0.0631 & 0.6862 & 0.9744 & 0.0522 & 0.0873 & 0.7284 & 0.9882 & \multirow{7}{*}{0.0439} & \multirow{7}{*}{0.0850} & \multirow{7}{*}{0.0440} & \multirow{7}{*}{0.0851} \\ \cmidrule(lr){2-10}
 & 20 & 0.0470 & 0.0791 & 0.6633 & 1.0030 & 0.6976 & 0.9408 & 0.7443 & 1.0108 &  &  &  &  \\ \cmidrule(lr){2-10}
 & 30 & 0.0175 & 0.0312 & 0.6628 & 0.9630 & 0.0481 & 0.0817 & 0.7088 & 0.9597 &  &  &  &  \\ \cmidrule(lr){2-10}
 & 40 & 0.0658 & 0.1128 & 0.6667 & 0.9709 & 0.0516 & 0.0874 & 0.7428 & 1.0068 &  &  &  &  \\ \cmidrule(lr){2-10}
 & 50 & 0.0794 & 0.1331 & 0.6693 & 0.9718 & 0.0458 & 0.0767 & 0.7269 & 0.9856 &  &  &  &  \\
\specialrule{1pt}{2pt}{2pt}
\multirow{7}{*}{\textbf{(5, 5)}} & 10 & 0.1036 & 0.1688 & 0.7156 & 0.9985 & 0.0604 & 0.0998 & 0.7407 & 0.9964 & \multirow{7}{*}{0.0518} & \multirow{7}{*}{0.0934} & \multirow{7}{*}{0.0519} & \multirow{7}{*}{0.0936} \\ \cmidrule(lr){2-10}
 & 20 & 0.1001 & 0.1633 & 0.7084 & 1.0030 & 0.7106 & 0.9511 & 0.7567 & 1.0190 &  &  &  &  \\ \cmidrule(lr){2-10}
 & 30 & 0.0731 & 0.1161 & 0.7079 & 0.9928 & 0.0824 & 0.1258 & 0.7198 & 0.9667 &  &  &  &  \\ \cmidrule(lr){2-10}
 & 40 & 0.0946 & 0.1569 & 0.7073 & 0.9968 & 0.0678 & 0.1116 & 0.7563 & 1.0164 &  &  &  &  \\ \cmidrule(lr){2-10}
 & 50 & 0.0875 & 0.1416 & 0.7126 & 1.0036 & 0.0733 & 0.1136 & 0.7390 & 0.9935 &  &  &  &  \\
\specialrule{1pt}{2pt}{2pt}
\multirow{7}{*}{\textbf{(6, 5)}} & 10 & 0.1009 & 0.1593 & 0.6936 & 0.9542 & 0.1100 & 0.1626 & 0.7512 & 1.0045 & \multirow{7}{*}{0.0665} & \multirow{7}{*}{0.1106} & \multirow{7}{*}{0.0665} & \multirow{7}{*}{0.1104} \\ \cmidrule(lr){2-10}
 & 20 & 0.1024 & 0.1616 & 0.6370 & 0.8897 & 0.7125 & 0.9499 & 0.7669 & 1.0265 &  &  &  &  \\ \cmidrule(lr){2-10}
 & 30 & 0.0938 & 0.1454 & 0.6962 & 0.9589 & 0.1354 & 0.1903 & 0.7291 & 0.9735 &  &  &  &  \\ \cmidrule(lr){2-10}
 & 40 & 0.0983 & 0.1561 & 0.6470 & 0.8983 & 0.1145 & 0.1707 & 0.7672 & 1.0250 &  &  &  &  \\ \cmidrule(lr){2-10}
 & 50 & 0.1125 & 0.1804 & 0.7022 & 0.9694 & 0.1298 & 0.1822 & 0.7492 & 1.0014 &  &  &  &  \\
\specialrule{1pt}{2pt}{2pt}
\multirow{7}{*}{\textbf{(7, 5)}} & 10 & 0.0926 & 0.1430 & 0.2754 & 0.4140 & 0.1676 & 0.2312 & 0.7607 & 1.0127 & \multirow{7}{*}{0.0796} & \multirow{7}{*}{0.1262} & \multirow{7}{*}{0.0800} & \multirow{7}{*}{0.1269} \\ \cmidrule(lr){2-10}
 & 20 & 0.1158 & 0.1785 & 0.2935 & 0.4473 & 0.7135 & 0.9485 & 0.7760 & 1.0341 &  &  &  &  \\ \cmidrule(lr){2-10}
 & 30 & 0.0907 & 0.1381 & 0.2463 & 0.3696 & 0.1679 & 0.2306 & 0.7379 & 0.9811 &  &  &  &  \\ \cmidrule(lr){2-10}
 & 40 & 0.1065 & 0.1666 & 0.2583 & 0.3926 & 0.1571 & 0.2218 & 0.7763 & 1.0327 &  &  &  &  \\ \cmidrule(lr){2-10}
 & 50 & 0.1417 & 0.2231 & 0.2560 & 0.3837 & 0.1674 & 0.2287 & 0.7583 & 1.0090 &  &  &  &  \\ 
 \specialrule{1pt}{2pt}{2pt}
\end{tabular}}
\end{table}

\clearpage

\begin{table}[h]
\caption{The max error and explained variance score of the solutions of reaction-diffusion equations with $\nu = \{4, 5, 6, 7\}, \rho=5$}\label{tab:result_reac_diff_max_var_pinn}
\centering
\renewcommand{\arraystretch}{0.95}
\resizebox{\textwidth}{!}{
\begin{tabular}{lcccccccc}
\specialrule{1pt}{2pt}{2pt}
\multirow{2}{*}{$\pmb{(\nu, \rho)}$} & \multicolumn{2}{c}{\textbf{PINN}} & \multicolumn{2}{c}{\textbf{PINN-R}} & \multicolumn{2}{c}{\textbf{PINN-P}} & \multicolumn{2}{c}{\textbf{PINN-S2S}} \\ \cmidrule(lr){2-9}
 & Max. err. & Exp. var. & Max. err. & Exp. var. & Max. err. & Exp. var. & Max. err. & Exp. var. \\
\specialrule{1pt}{2pt}{2pt}
\textbf{(4, 5)} & 0.9624 & 0.0038 & 0.9697 & -0.0073 & 0.9400 & 0.0542 & 1.0706 & -0.0136 \\ \cmidrule(lr){1-9}
\textbf{(5, 5)} & 0.9452 & -0.0278 & 0.9782 & -0.1213 & 1.0000 & -0.1643 & 1.1295 & -0.0634 \\ \cmidrule(lr){1-9}
\textbf{(6, 5)} & 0.8441 & 0.8441 & 0.9931 & -0.2086 & 1.1412 & -0.6138 & 1.1584 & -0.1617 \\ \cmidrule(lr){1-9}
\textbf{(7, 5)} & 1.1444 & -0.6415 & 1.0150 & -0.348 & -0.6138 & -0.6138 & 1.2363 & -0.4552 \\
\specialrule{1pt}{2pt}{2pt}
\end{tabular}
}
\end{table}

\begin{table}[h]
\caption{The max error and explained variance score of the solutions of reaction-diffusion equations with $\nu = \{4, 5, 6, 7\}, \rho=5$}\label{tab:result_reac_diff_max_var}
\renewcommand{\arraystretch}{0.95}
\resizebox{\textwidth}{!}{
\begin{tabular}{lccccccccccccc}
\specialrule{1pt}{2pt}{2pt}
\multirow{4}{*}{{($\pmb{\nu, \rho}$)}} & \multirow{4}{*}{\textbf{Rank}} & \multicolumn{2}{c}{\textbf{{[}w/o{]} Pre-training}} & \multicolumn{10}{c}{\textbf{{[}w{]} Pre-training}} \\ \cmidrule(lr){3-4} \cmidrule(lr){5-14}
 &  & \multicolumn{2}{c}{\multirow{2}{*}{\textbf{Na\"ive-LR-PINN}}} & \multicolumn{2}{c}{\multirow{2}{*}{\begin{tabular}[c]{@{}c@{}}\textbf{Curriculum}\\ \textbf{learning}\end{tabular}}} & \multicolumn{2}{c}{\multirow{2}{*}{\textbf{MAML}}} & \multicolumn{2}{c}{\multirow{2}{*}{\textbf{Reptile}}} & \multicolumn{2}{c}{\multirow{2}{*}{\begin{tabular}[c]{@{}c@{}}\textbf{Hyper-LR-PINN}\\ (\textbf{Full rank})\end{tabular}}} & \multicolumn{2}{c}{\multirow{2}{*}{\begin{tabular}[c]{@{}c@{}}\textbf{Hyper-LR-PINN}\\ (\textbf{Adaptive rank})\end{tabular}}} \\
 &  & \multicolumn{2}{c}{} & \multicolumn{2}{c}{} & \multicolumn{2}{c}{} & \multicolumn{2}{c}{} & \multicolumn{2}{c}{} & \multicolumn{2}{c}{} \\ \cmidrule(lr){3-14}
 &  & \multicolumn{1}{c}{Max. err.} & \multicolumn{1}{c}{Exp. var.} & \multicolumn{1}{c}{Max. err.} & \multicolumn{1}{c}{Exp. var.} & \multicolumn{1}{c}{Max. err.} & \multicolumn{1}{c}{Exp. var.} & \multicolumn{1}{c}{Max. err.} & \multicolumn{1}{c}{Exp. var.} & Max. err. & Exp. var. & Max. err. & Exp. var. \\
 \specialrule{1pt}{2pt}{2pt}
\multirow{7}{*}{\textbf{(4, 5)}} & 10 & 0.1859 & 0.9745 & 1.3388 & -0.5457 & 0.2907 & 0.9450 & 1.0065 & 0.1604 & \multirow{7}{*}{0.2189} & \multirow{7}{*}{0.9588} & \multirow{7}{*}{0.2189} & \multirow{7}{*}{0.9588} \\ \cmidrule(lr){2-10} 
 & 20 & 0.1967 & 0.9719 & 1.4119 & -0.9933 & 1.0161 & 0.3333 & 1.0145 & 0.1023 &  &  &  &  \\ \cmidrule(lr){2-10}
 & 30 & 0.1668 & 0.9934 & 1.3659 & -0.8454 & 0.3037 & 0.9530 & 1.0199 & 0.2398 &  &  &  &  \\ \cmidrule(lr){2-10}
 & 40 & 0.2753 & 0.9439 & 1.3921 & -0.9100 & 0.2747 & 0.9402 & 1.0225 & 0.1462 &  &  &  &  \\ \cmidrule(lr){2-10}
 & 50 & 0.3133 & 0.9265 & 1.3897 & -0.8704 & 0.3118 & 0.9613 & 1.0200 & 0.1754 &  &  &  &  \\ 
\specialrule{1pt}{2pt}{2pt}
\multirow{7}{*}{\textbf{(5, 5)}} & 10 & 0.3998 & 0.8848 & 1.2749 & -0.5216 & 0.2794 & 0.9345 & 1.0176 & 0.1469 & \multirow{7}{*}{0.2317} & \multirow{7}{*}{0.9523} & \multirow{7}{*}{0.2317} & \multirow{7}{*}{0.9523} \\ \cmidrule(lr){2-10}
 & 20 & 0.3778 & 0.8922 & 1.3262 & -0.7929 & 1.0325 & 0.3102 & 1.0250 & 0.0883 &  &  &  &  \\ \cmidrule(lr){2-10}
 & 30 & 0.2498 & 0.9470 & 1.2717 & -0.5948 & 0.3300 & 0.9435 & 1.0310 & 0.2275 &  &  &  &  \\ \cmidrule(lr){2-10}
 & 40 & 0.3739 & 0.8953 & 1.2902 & -0.6908 & 0.2792 & 0.9250 & 1.0352 & 0.1322 &  &  &  &  \\ \cmidrule(lr){2-10}
 & 50 & 0.3216 & 0.9213 & 1.3184 & -0.7047 & 0.3214 & 0.9503 & 1.0314 & 0.1628 &  &  &  &  \\ 
\specialrule{1pt}{2pt}{2pt}
\multirow{7}{*}{\textbf{(6, 5)}} & 10 & 0.3527 & 0.9007 & 1.1203 & -0.2824 & 0.3264 & 0.9240 & 1.0264 & 0.1345 & \multirow{7}{*}{0.2570} & \multirow{7}{*}{0.9430} & \multirow{7}{*}{0.2578} & \multirow{7}{*}{0.9429} \\ \cmidrule(lr){2-10}
 & 20 & 0.3580 & 0.8992 & 1.1396 & -0.3386 & 1.0354 & 0.2814 & 1.0345 & 0.0753 &  &  &  &  \\ \cmidrule(lr){2-10}
 & 30 & 0.3095 & 0.9235 & 1.1204 & -0.3159 & 0.3742 & 0.9327 & 1.0397 & 0.2153 &  &  &  &  \\ \cmidrule(lr){2-10}
 & 40 & 0.3493 & 0.9037 & 1.1130 & -0.2780 & 0.3305 & 0.9126 & 1.0457 & 0.1182 &  &  &  &  \\ \cmidrule(lr){2-10}
 & 50 & 0.4116 & 0.8689 & 1.1388 & -0.3871 & 0.3679 & 0.9397 & 1.0407 & 0.1501 &  &  &  &  \\ 
\specialrule{1pt}{2pt}{2pt}
\multirow{7}{*}{\textbf{(7, 5)}} & 10 & 0.3048 & 0.9225 & 0.7000 & 0.4155 & 0.3745 & 0.9149 & 1.0334 & 0.1238 & \multirow{7}{*}{0.2827} & \multirow{7}{*}{0.9284} & \multirow{7}{*}{0.2833} & \multirow{7}{*}{0.9286} \\ \cmidrule(lr){2-10}
 & 20 & 0.3802 & 0.8820 & 0.7686 & 0.2740 & 1.0365 & 0.2589 & 1.0418 & 0.0635 &  &  &  &  \\ \cmidrule(lr){2-10}
 & 30 & 0.2823 & 0.9318 & 0.6453 & 0.5257 & 0.3956 & 0.9202 & 1.0467 & 0.2049 &  &  &  &  \\ \cmidrule(lr){2-10}
 & 40 & 0.3676 & 0.8915 & 0.6842 & 0.4392 & 0.3789 & 0.8998 & 1.0537 & 0.1066 &  &  &  &  \\ \cmidrule(lr){2-10}
 & 50 & 0.5088 & 0.8037 & 0.6560 & 0.4977 & 0.3944 & 0.9276 & 1.0482 & 0.1398 &  &  &  &  \\
 \specialrule{1pt}{2pt}{2pt}

\end{tabular}}
\end{table}

\clearpage

\section{Loss curves of meta-learning methods in Phase 2}\label{a:loss_curve_appendix}
\begin{figure}[ht!]
    \centering
    \subfloat[$\beta = 30$]
    {\includegraphics[width=0.32\columnwidth]
    {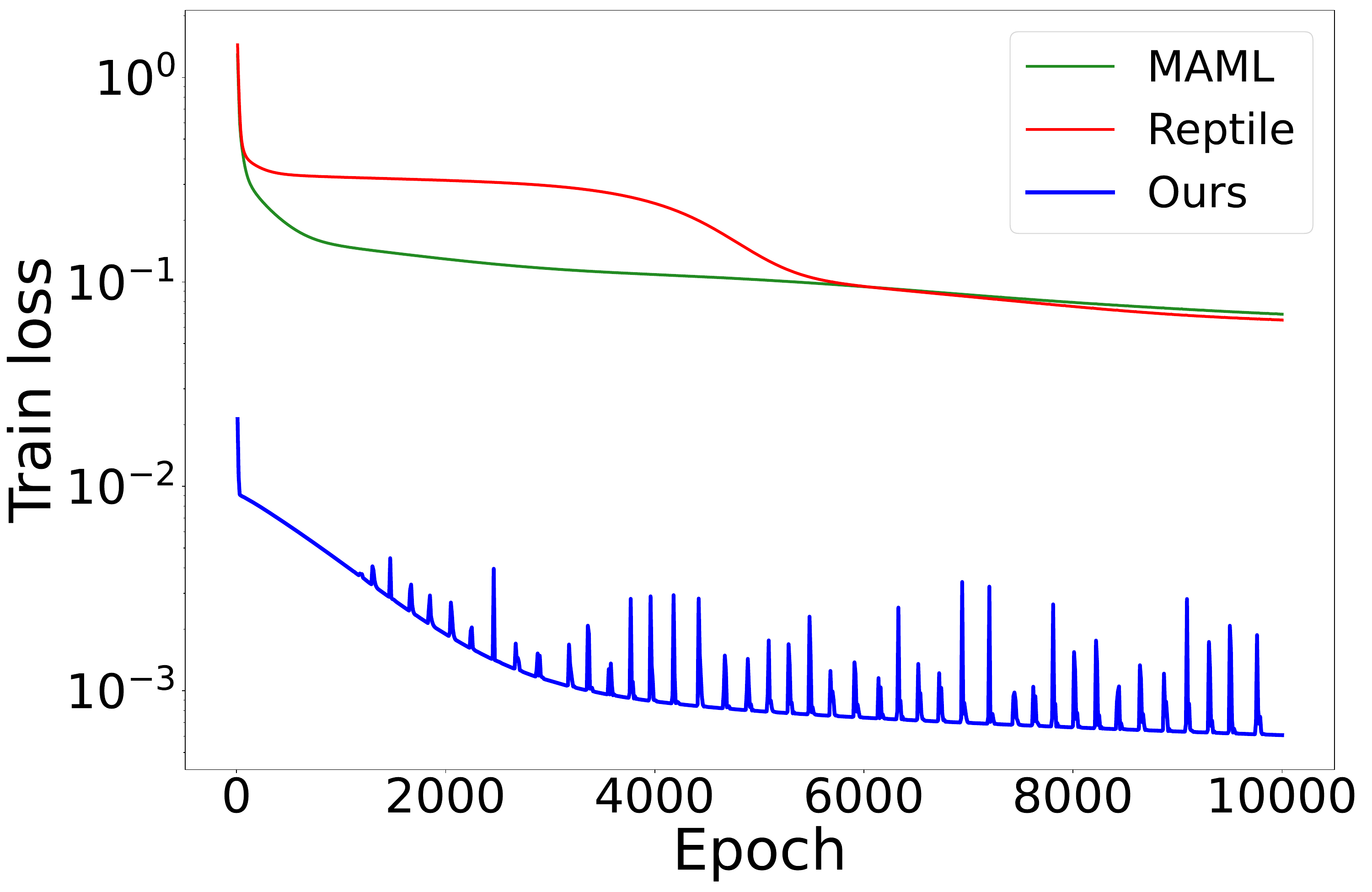}}\hfill
    \subfloat[$\beta = 35$]
    {\includegraphics[width=0.32\columnwidth]
    {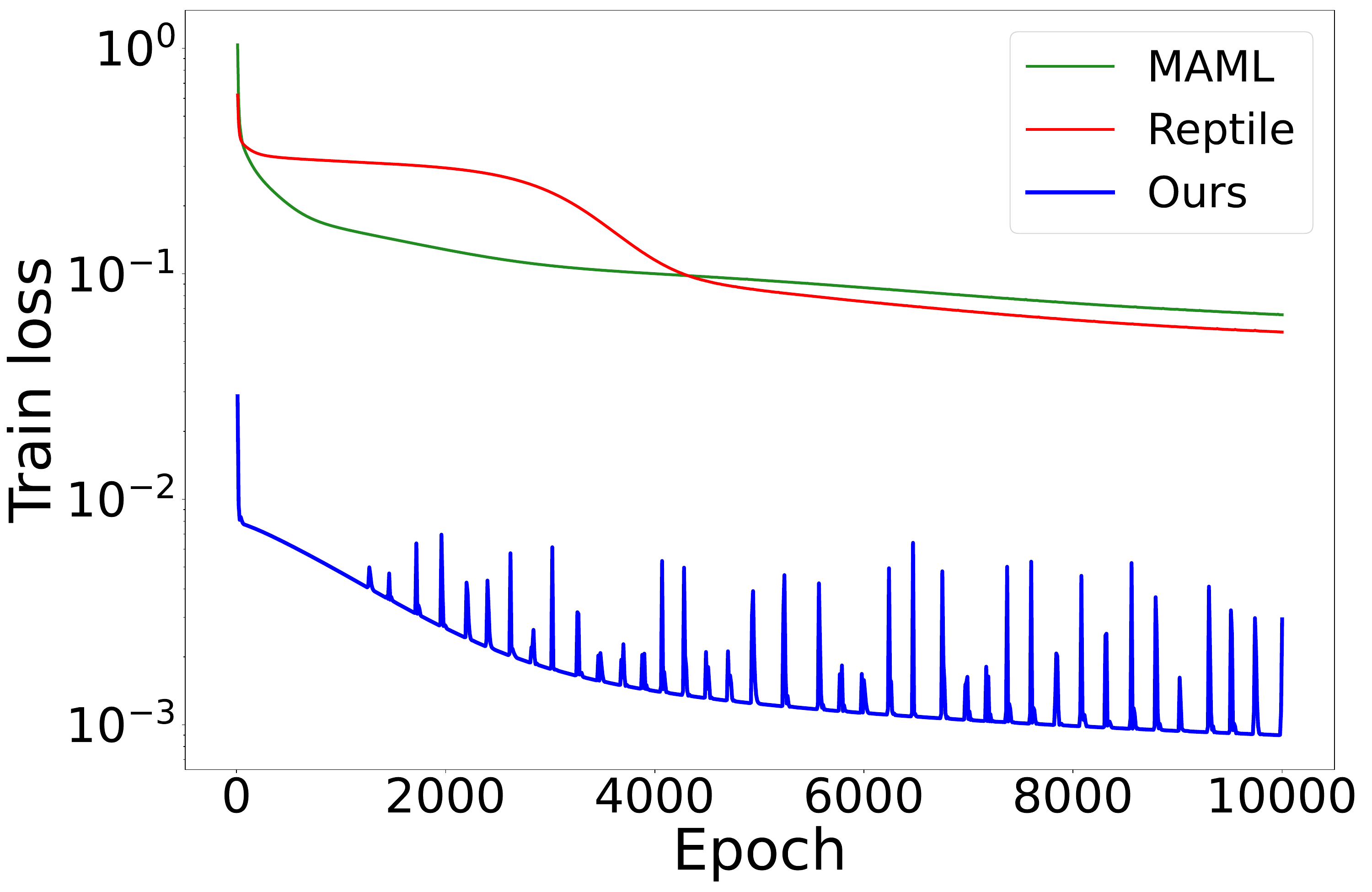}}\hfill
    \subfloat[$\beta = 40$]
    {\includegraphics[width=0.32\columnwidth]
    {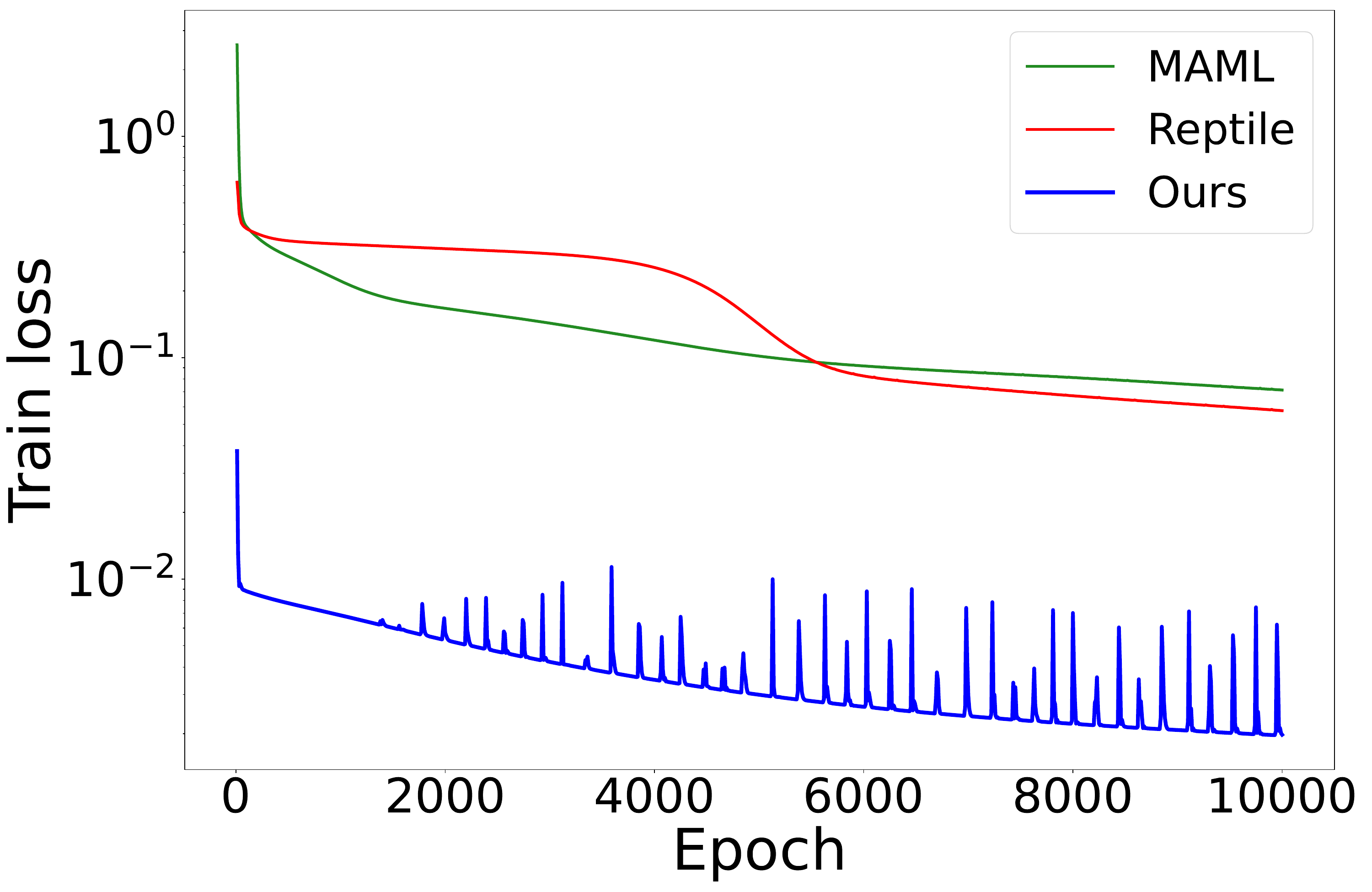}} \\
    \caption{Training loss curve in phase 2. We follow the experimental setting of Figure 5. $\beta = \{30, 35, 40\}$}
\end{figure}

\begin{figure}[ht!]
    \centering
    \subfloat[$\beta = 30$ (Abs. err.)]
    {\includegraphics[width=0.32\columnwidth]
    {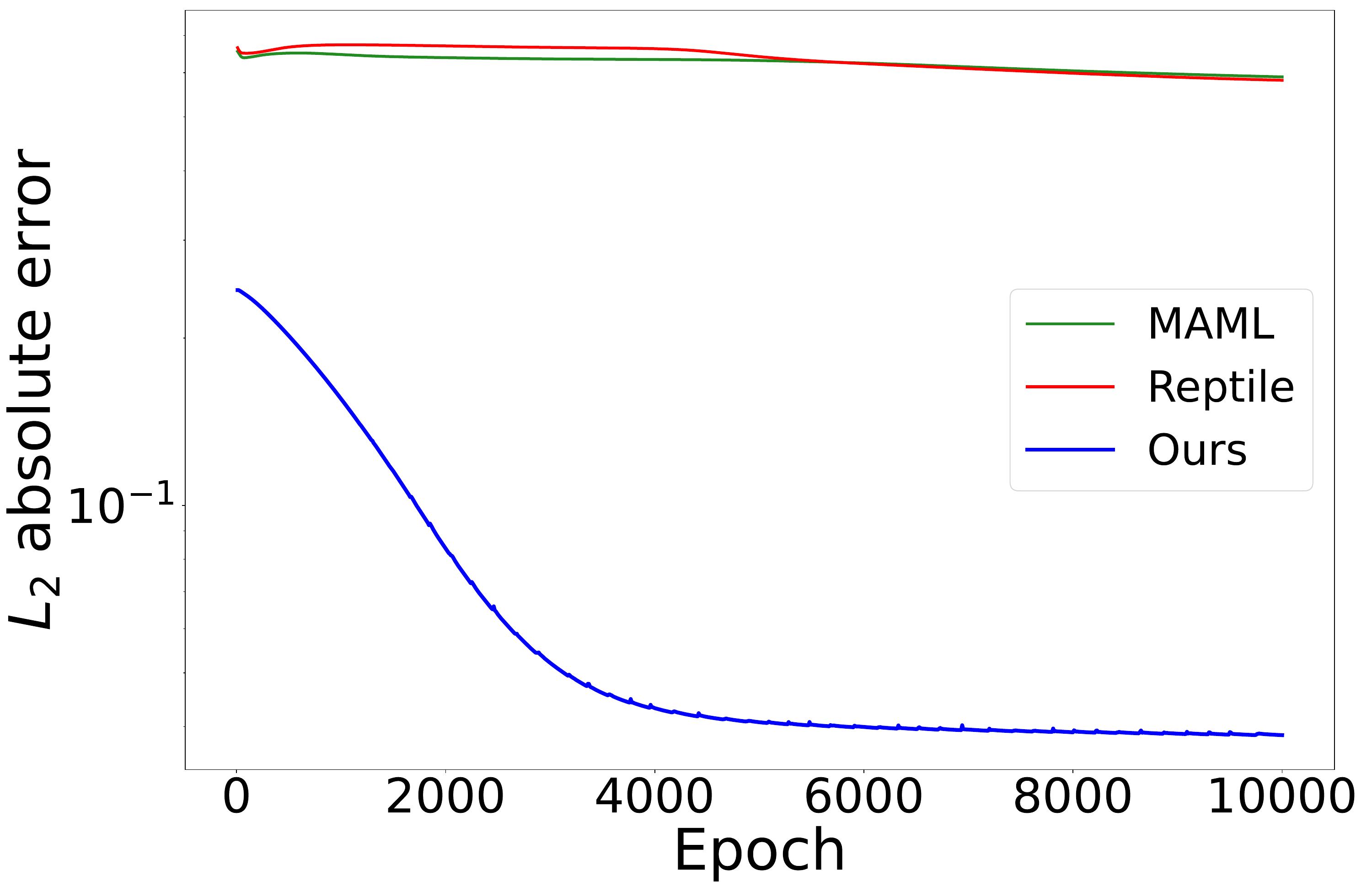}}\hfill
    \subfloat[$\beta = 35$ (Abs. err.)]
    {\includegraphics[width=0.32\columnwidth]
    {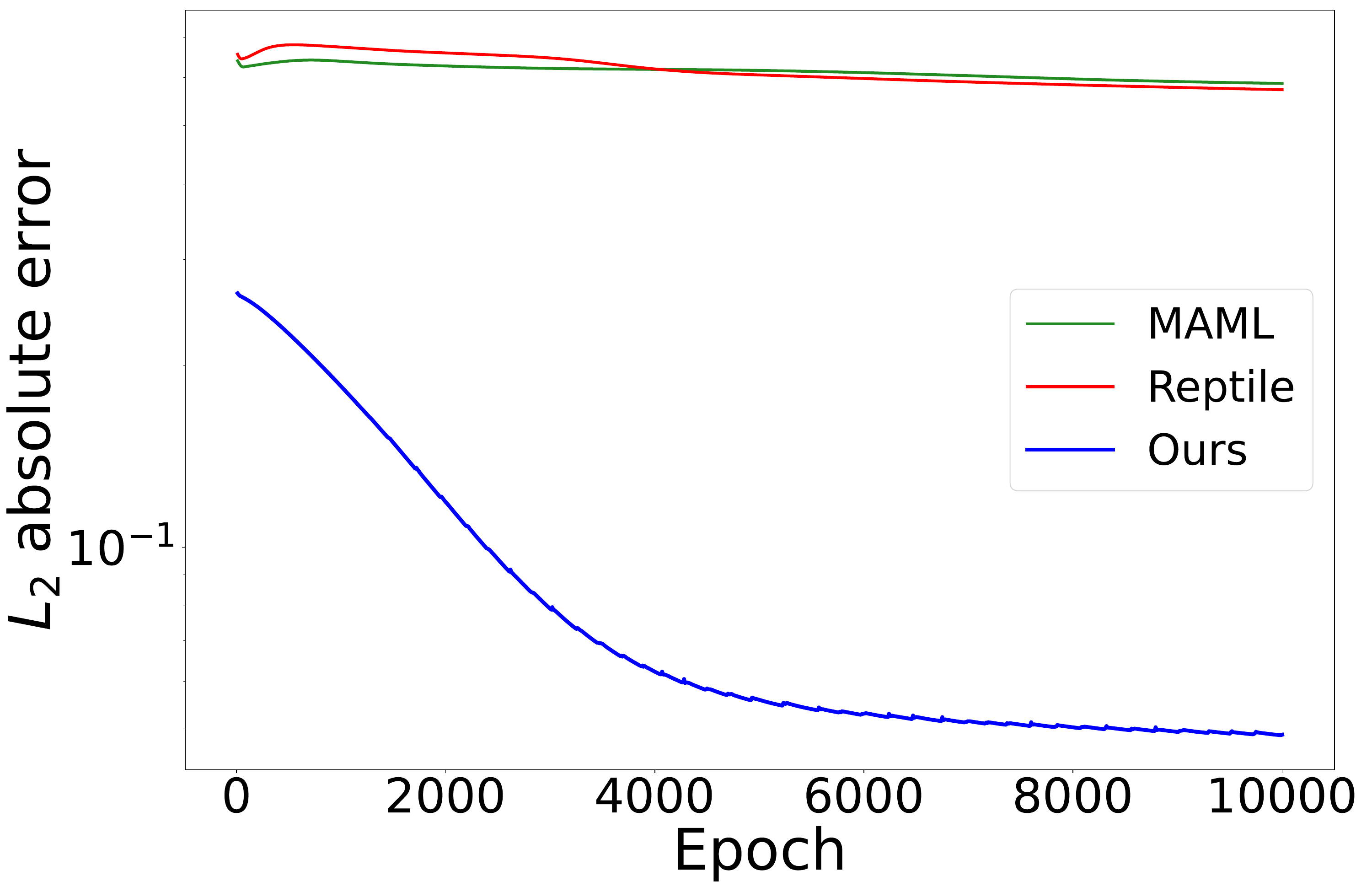}}\hfill
    \subfloat[$\beta = 40$ (Abs. err.)]
    {\includegraphics[width=0.32\columnwidth]
    {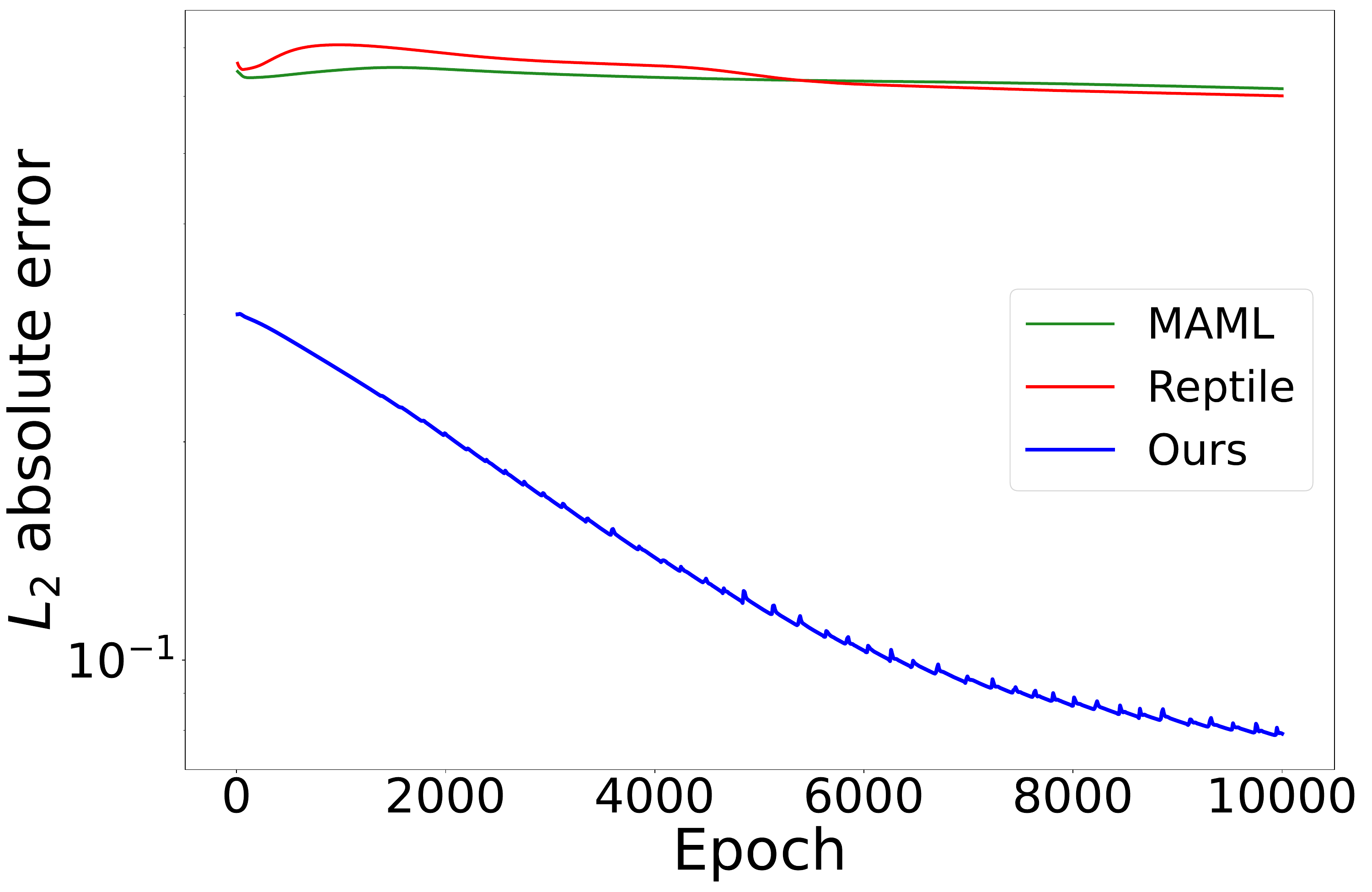}} \\

    \subfloat[$\beta = 30$ (Rel. err.)]
    {\includegraphics[width=0.32\columnwidth]
    {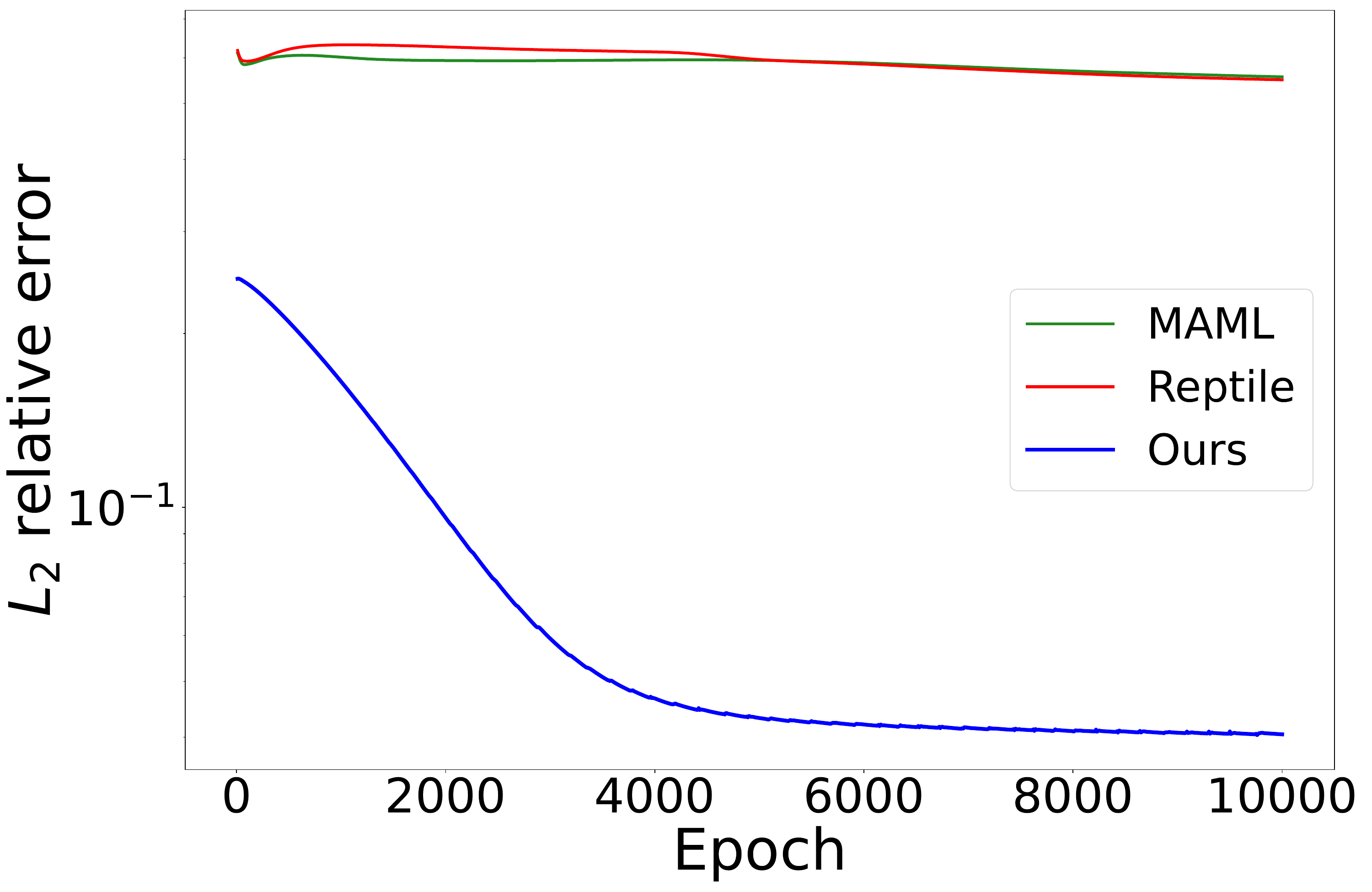}}\hfill
    \subfloat[$\beta = 35$ (Rel. err.)]
    {\includegraphics[width=0.32\columnwidth]
    {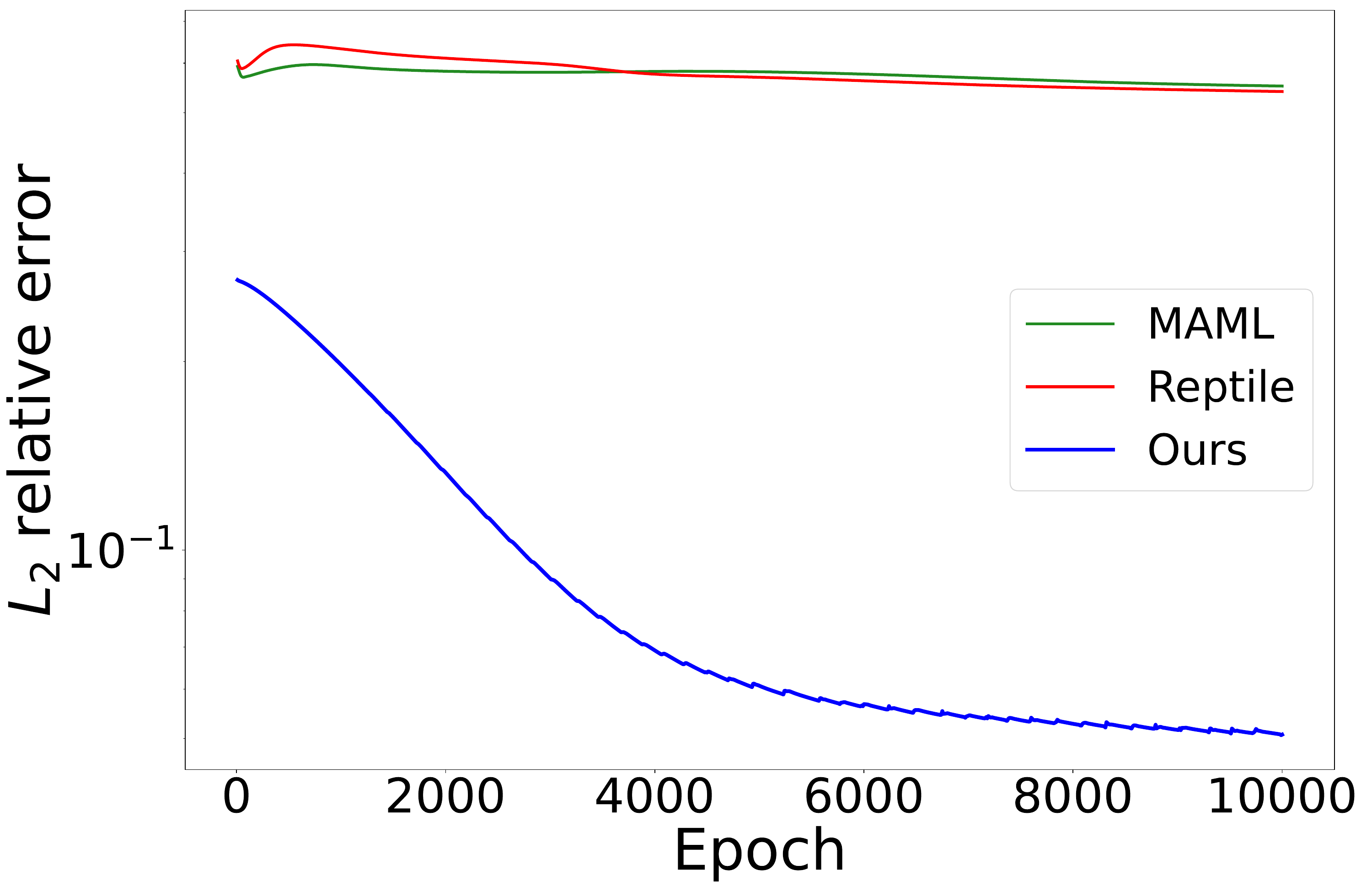}}\hfill
    \subfloat[$\beta = 40$ (Rel. err.)]
    {\includegraphics[width=0.32\columnwidth]
    {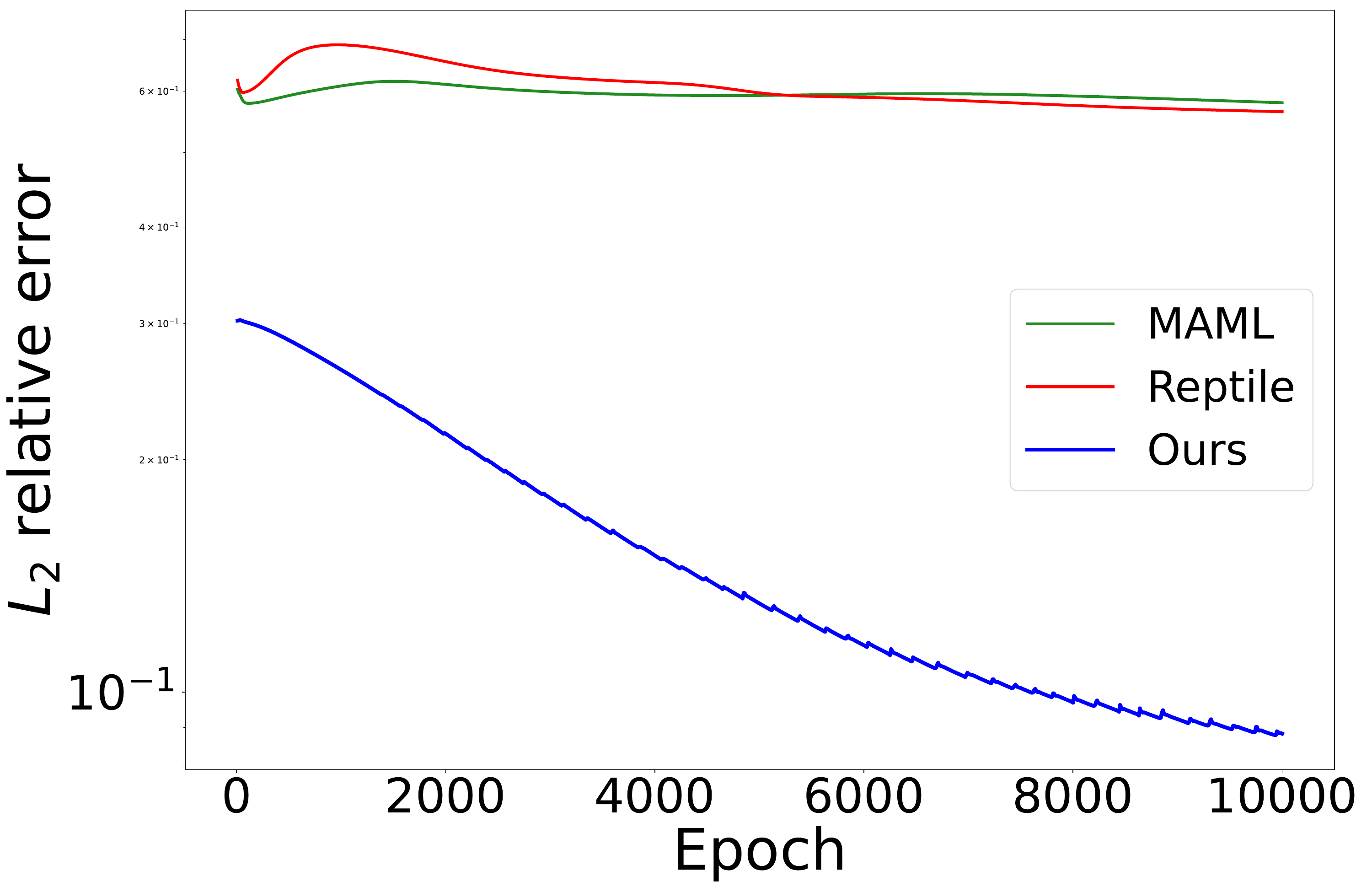}} \\
    \caption{Curves of $L_2$ absolute error (top row) and $L_2$ relative error (bottom row) in phase2. We follow the experimental setting of Figure 5. $\beta = \{30, 35, 40\}$}
\end{figure}

\clearpage

\section{Adaptive rank: $\beta$ outside of the range in training phase 1}
\begin{figure}[ht!]
    \centering
    \subfloat[$\pmb{s}^1(\pmb{\mu})$]
    {\includegraphics[width=0.32\columnwidth]
    {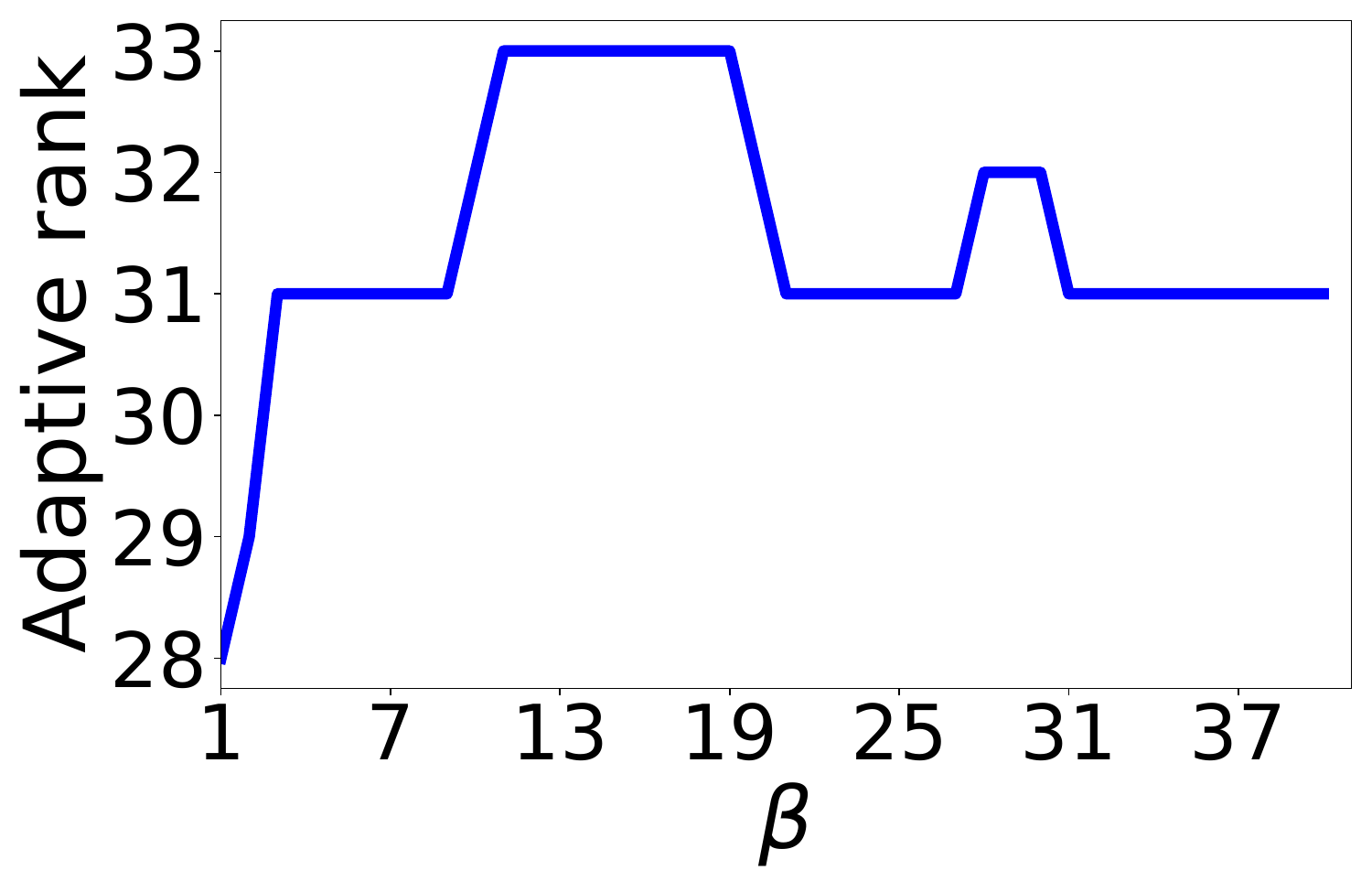}}\hfill
    \subfloat[$\pmb{s}^2(\pmb{\mu})$]
    {\includegraphics[width=0.32\columnwidth]
    {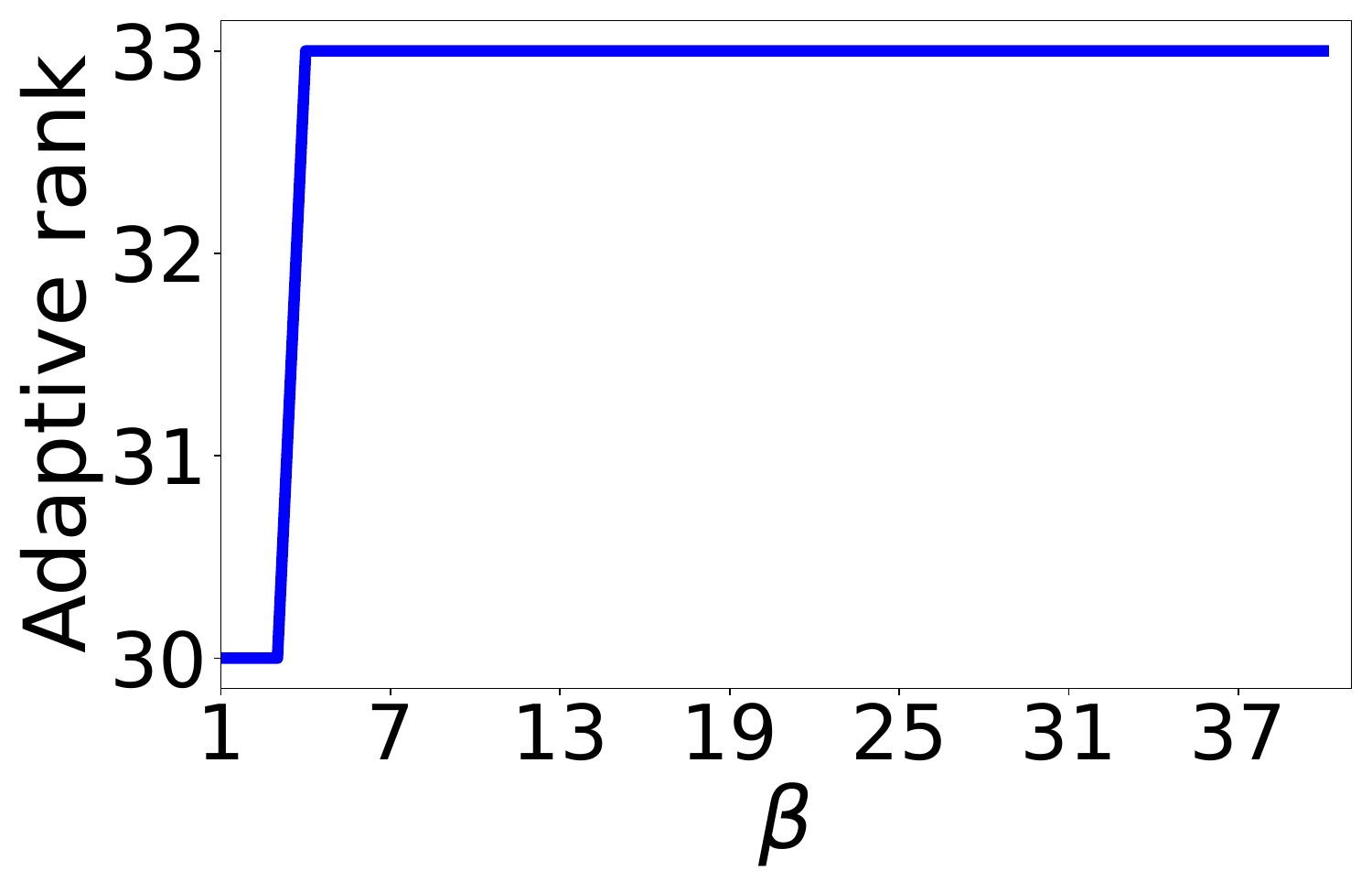}}\hfill
    \subfloat[$\pmb{s}^3(\pmb{\mu})$]
    {\includegraphics[width=0.32\columnwidth]
    {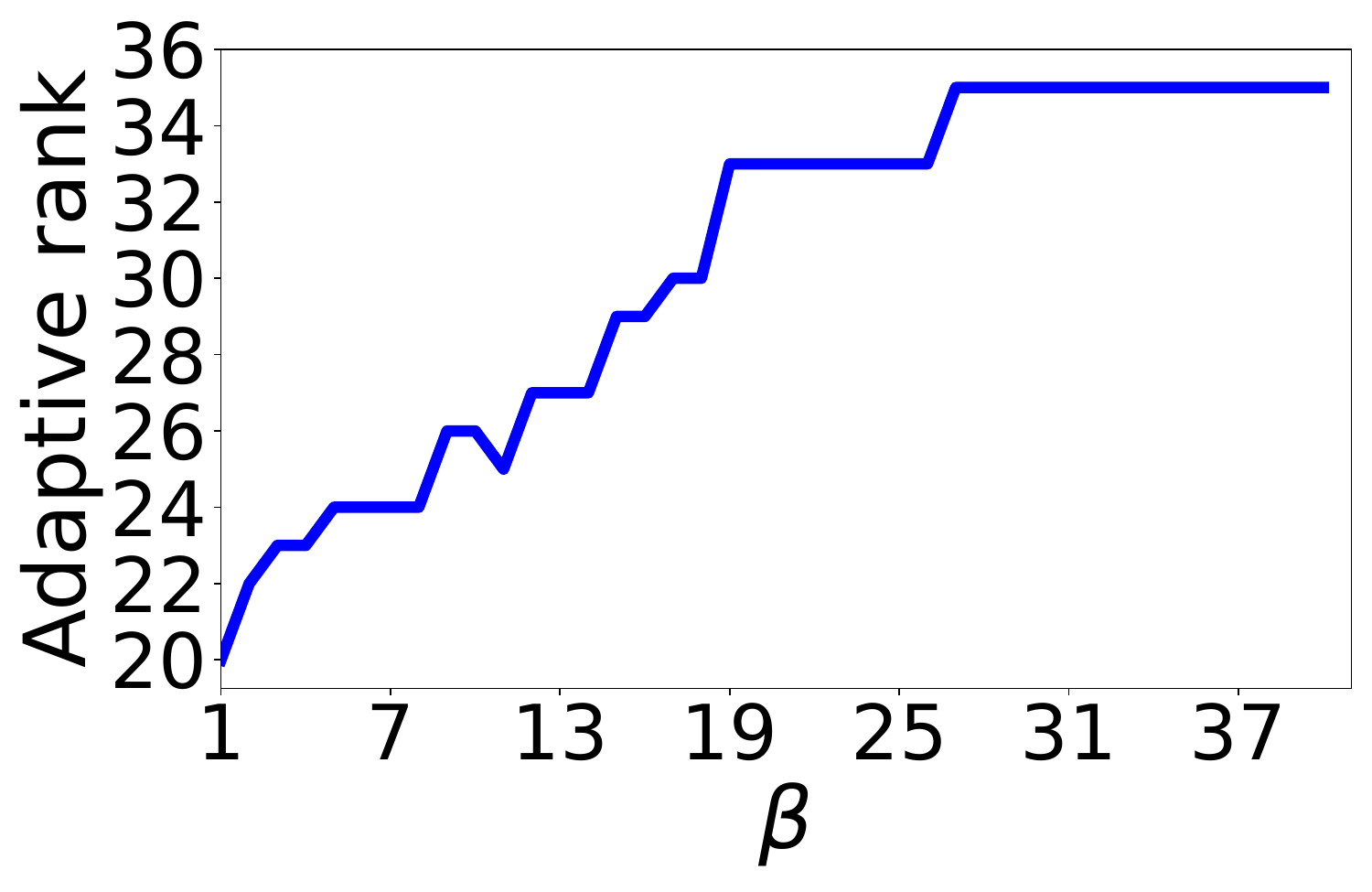}} \\
    \caption{[Convection equation] Learned rank  structures of the three hidden layers. $\beta \in [1,20]$ are used in phase 1.}
    \vspace{-1.0em}
\end{figure}

\section{Comparision of learnable basis and fixed basis on Hyper-LR-PINN}\label{a:learn_fixed_basis}

\begin{figure}[ht!]
    \centering
    \subfloat[$\beta = 30$]
    {\includegraphics[width=0.32\columnwidth]
    {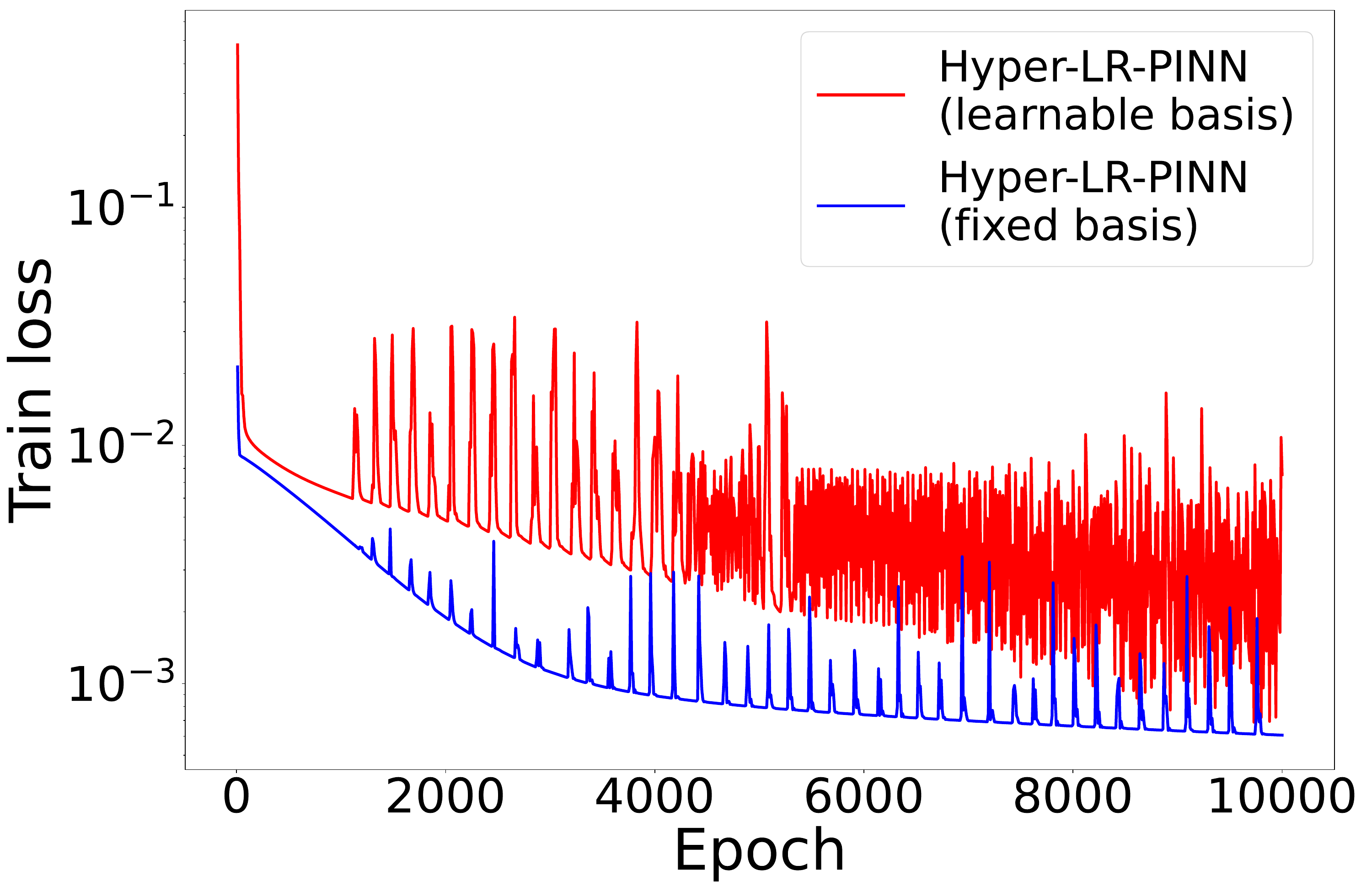}}\hfill
    \subfloat[$\beta = 35$]
    {\includegraphics[width=0.32\columnwidth]
    {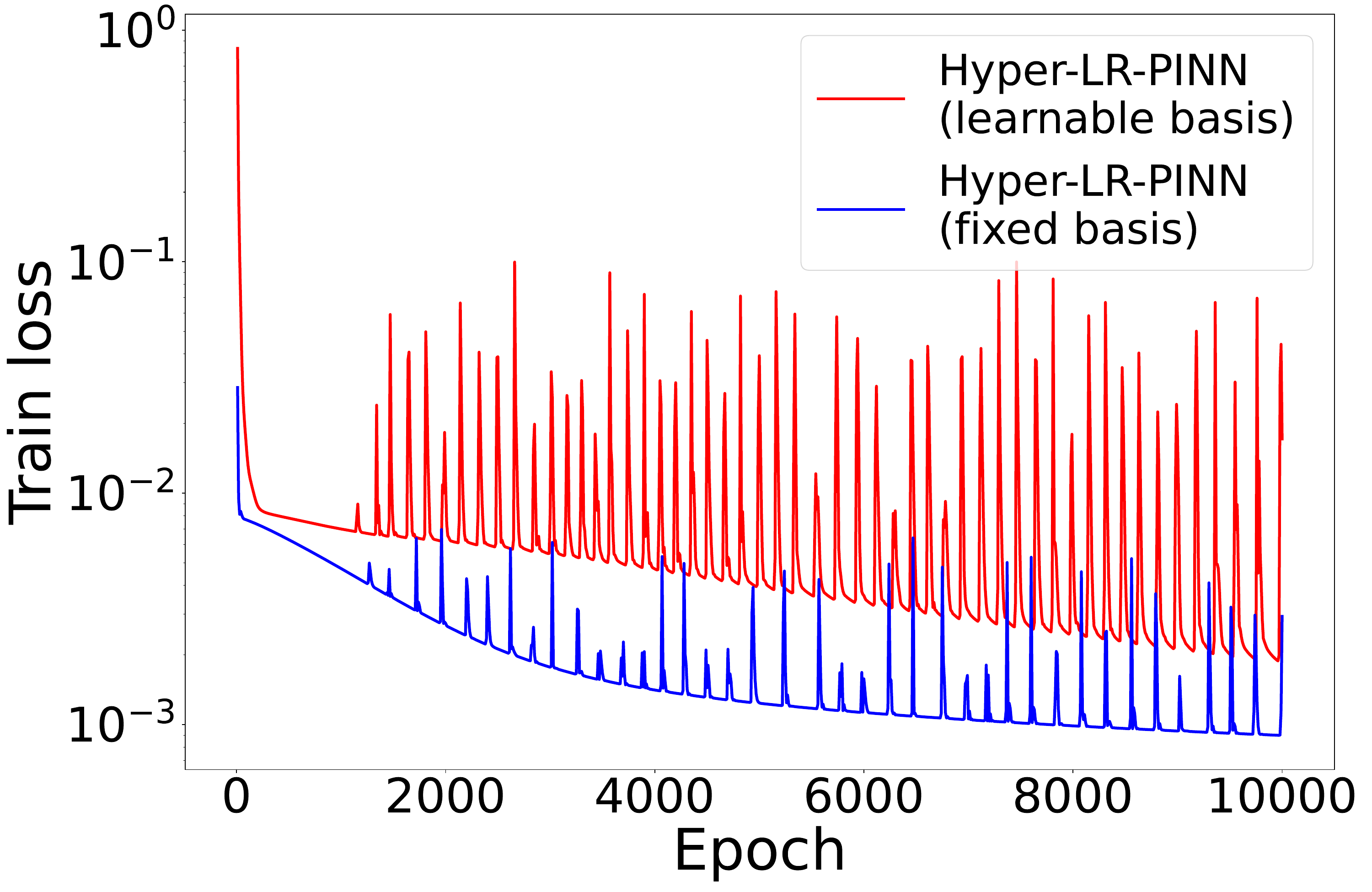}}\hfill
    \subfloat[$\beta = 40$]
    {\includegraphics[width=0.32\columnwidth]
    {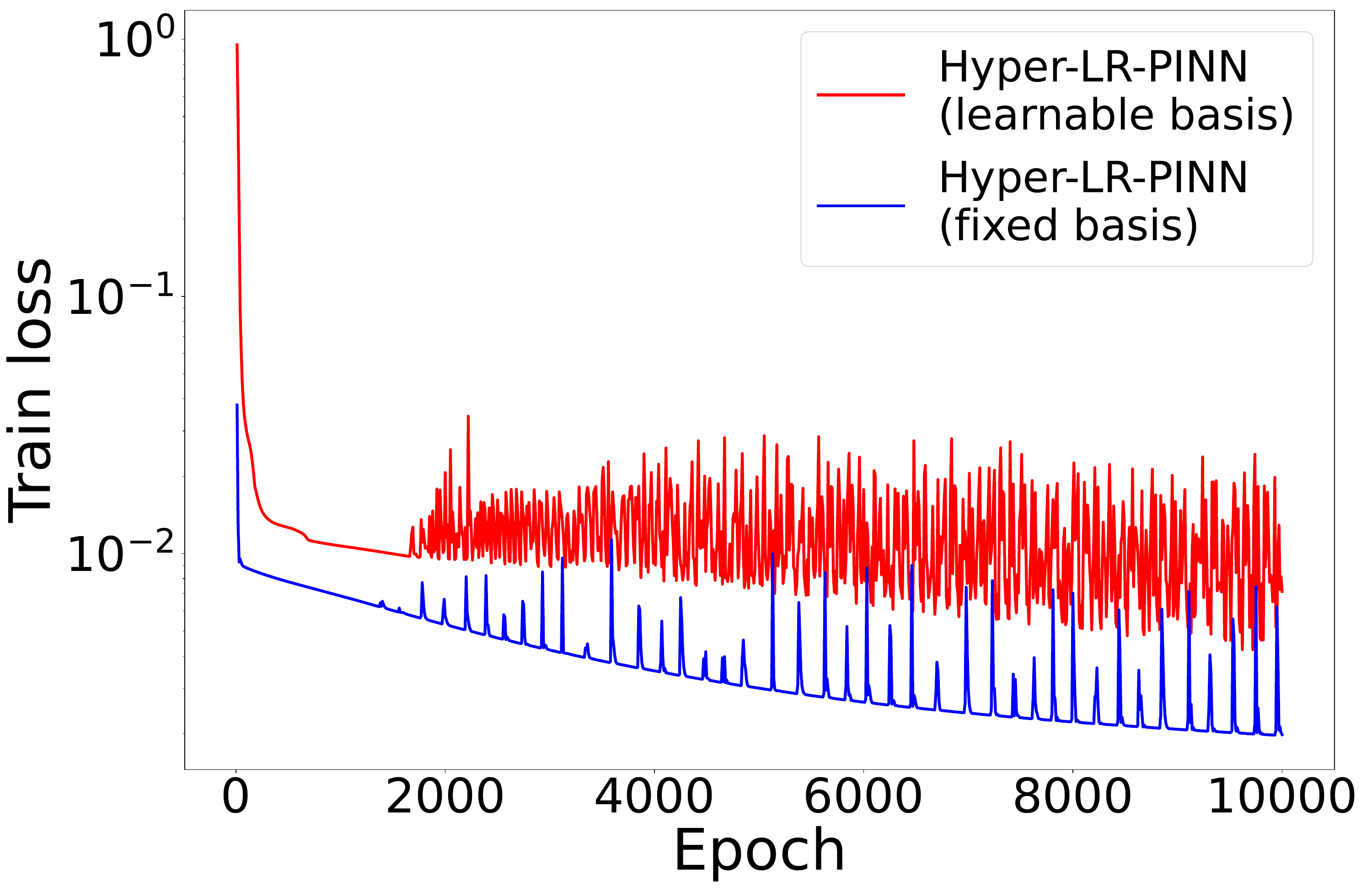}} \\
    \caption{Training loss curve on convection equations in phase 2. $\beta = \{30, 35, 40\}$}\label{fig:learn_fix_train}
\end{figure}

\begin{figure}[ht!]
    \centering
    \subfloat[$\beta = 30$ (Abs. err.)]
    {\includegraphics[width=0.32\columnwidth]
    {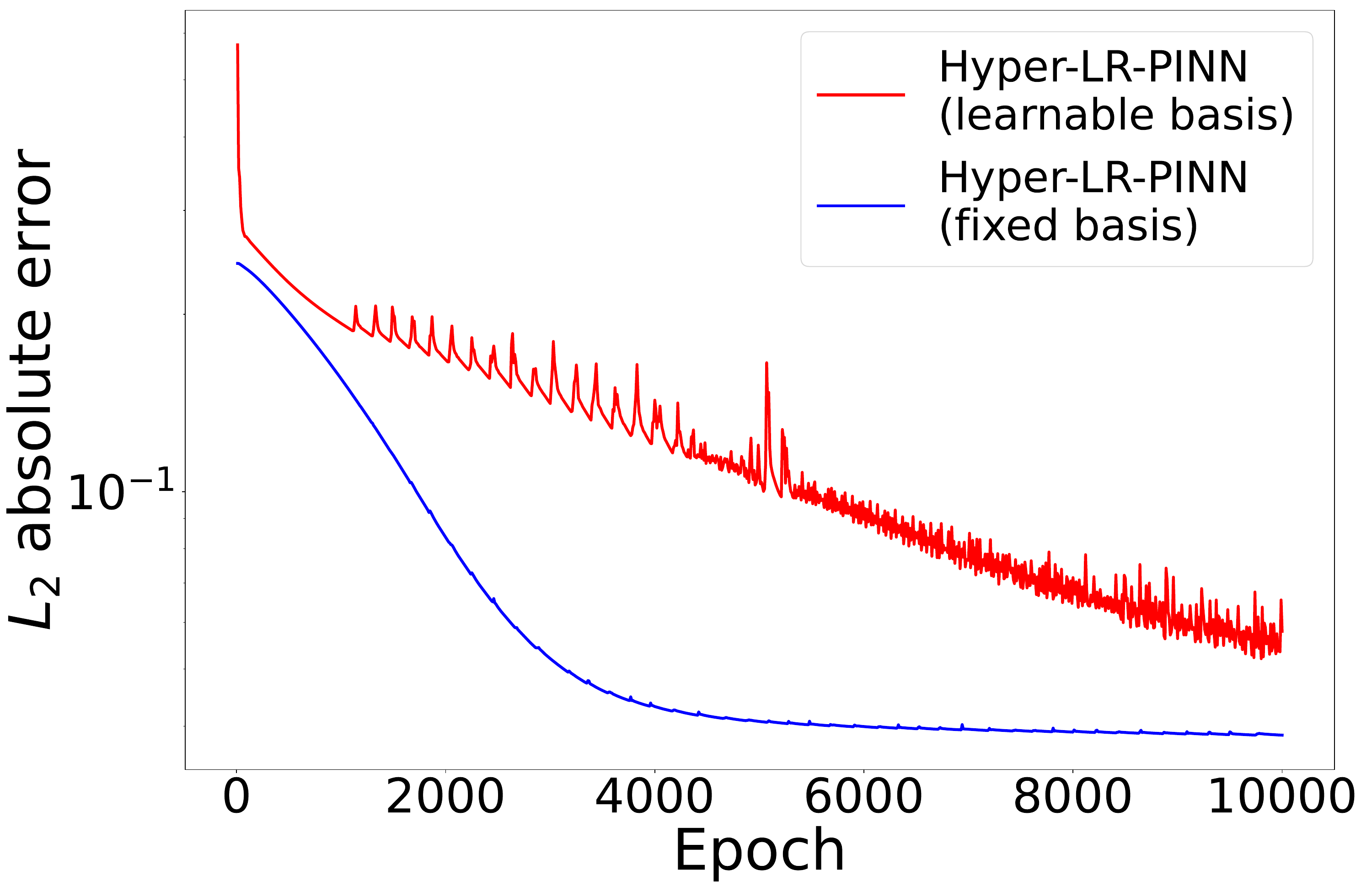}}\hfill
    \subfloat[$\beta = 35$ (Abs. err.)]
    {\includegraphics[width=0.32\columnwidth]
    {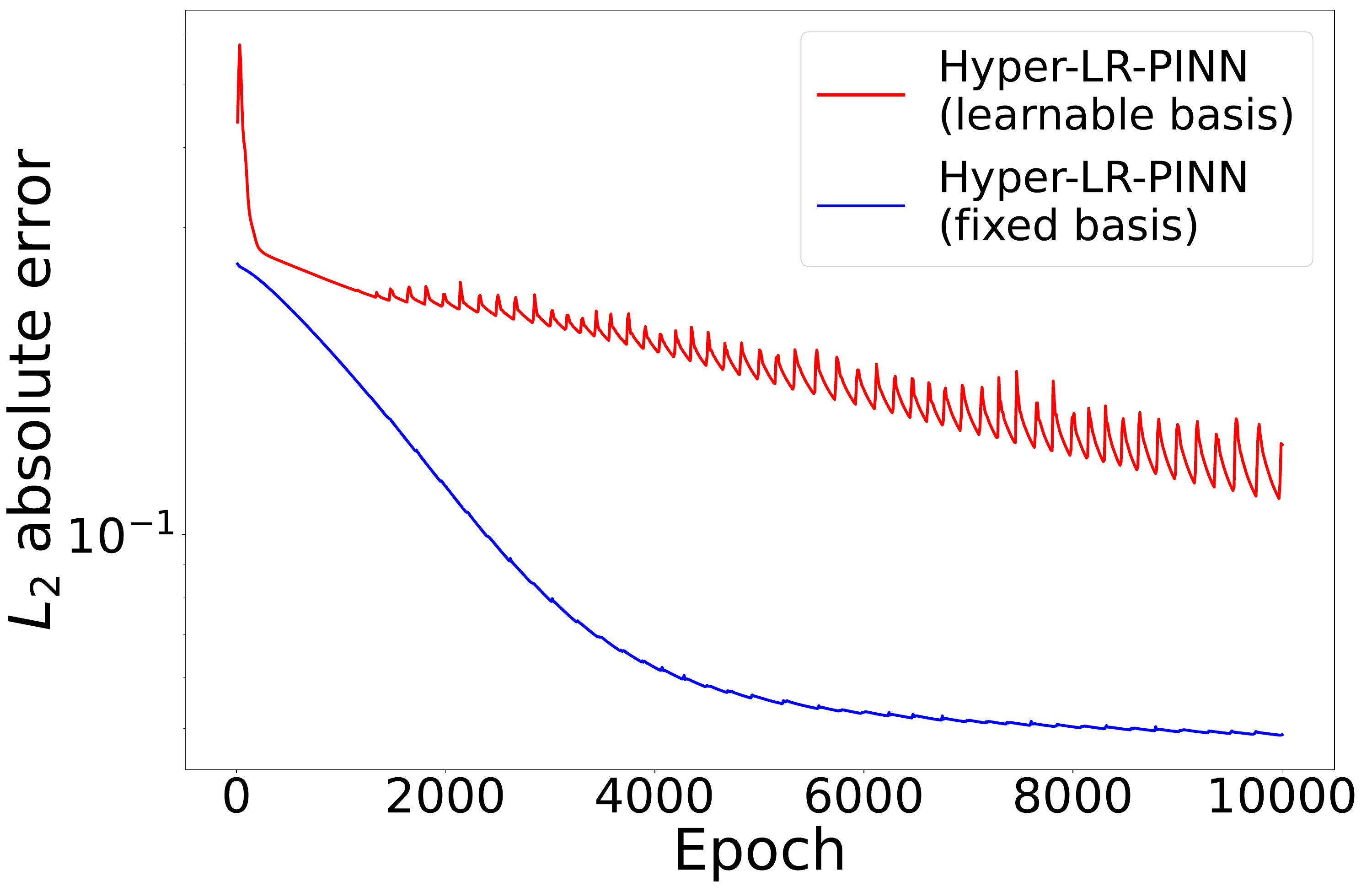}}\hfill
    \subfloat[$\beta = 40$ (Abs. err.)]
    {\includegraphics[width=0.32\columnwidth]
    {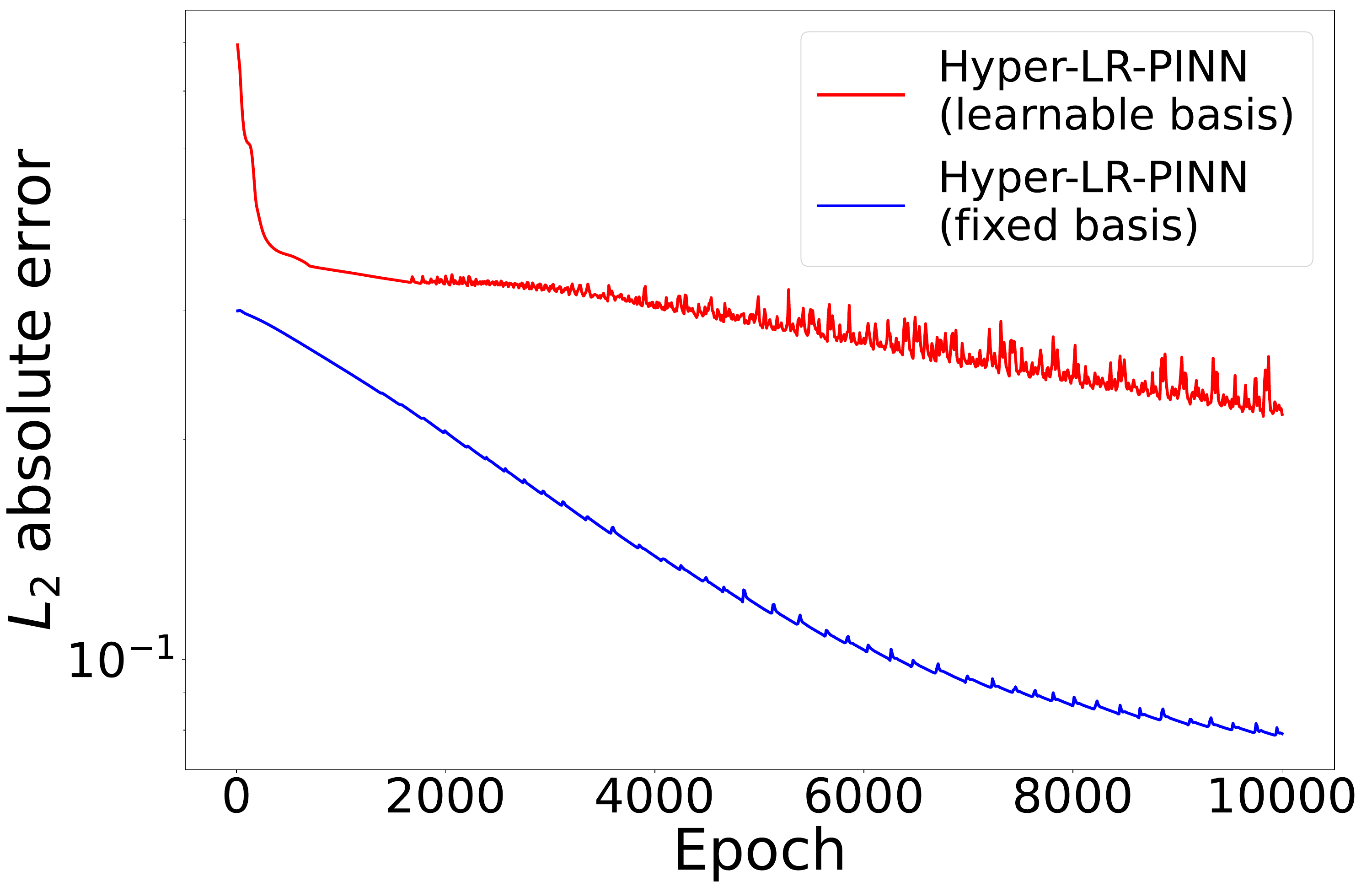}} \\

    \subfloat[$\beta = 30$ (Rel. err.)]
    {\includegraphics[width=0.32\columnwidth]
    {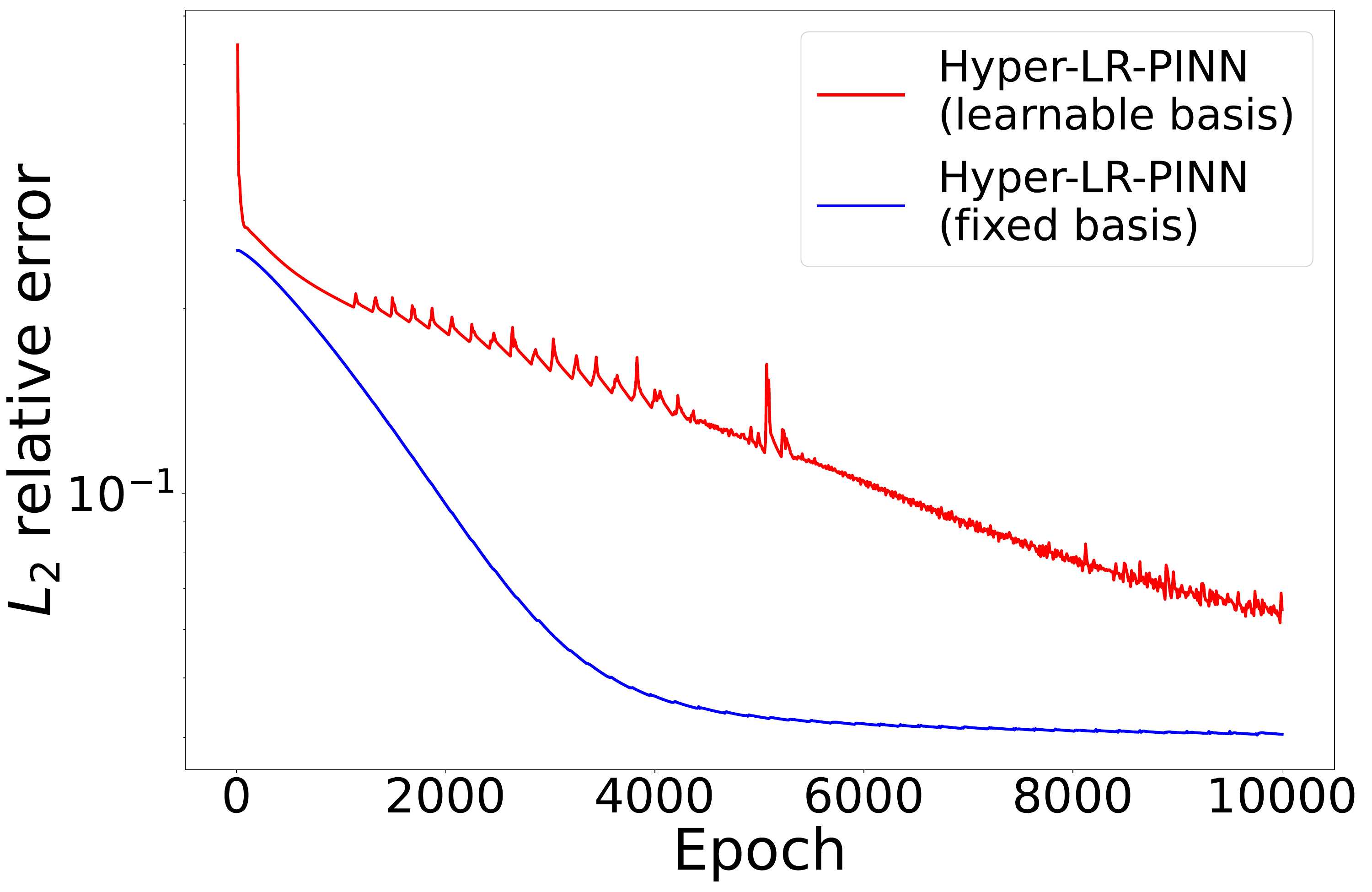}}\hfill
    \subfloat[$\beta = 35$ (Rel. err.)]
    {\includegraphics[width=0.32\columnwidth]
    {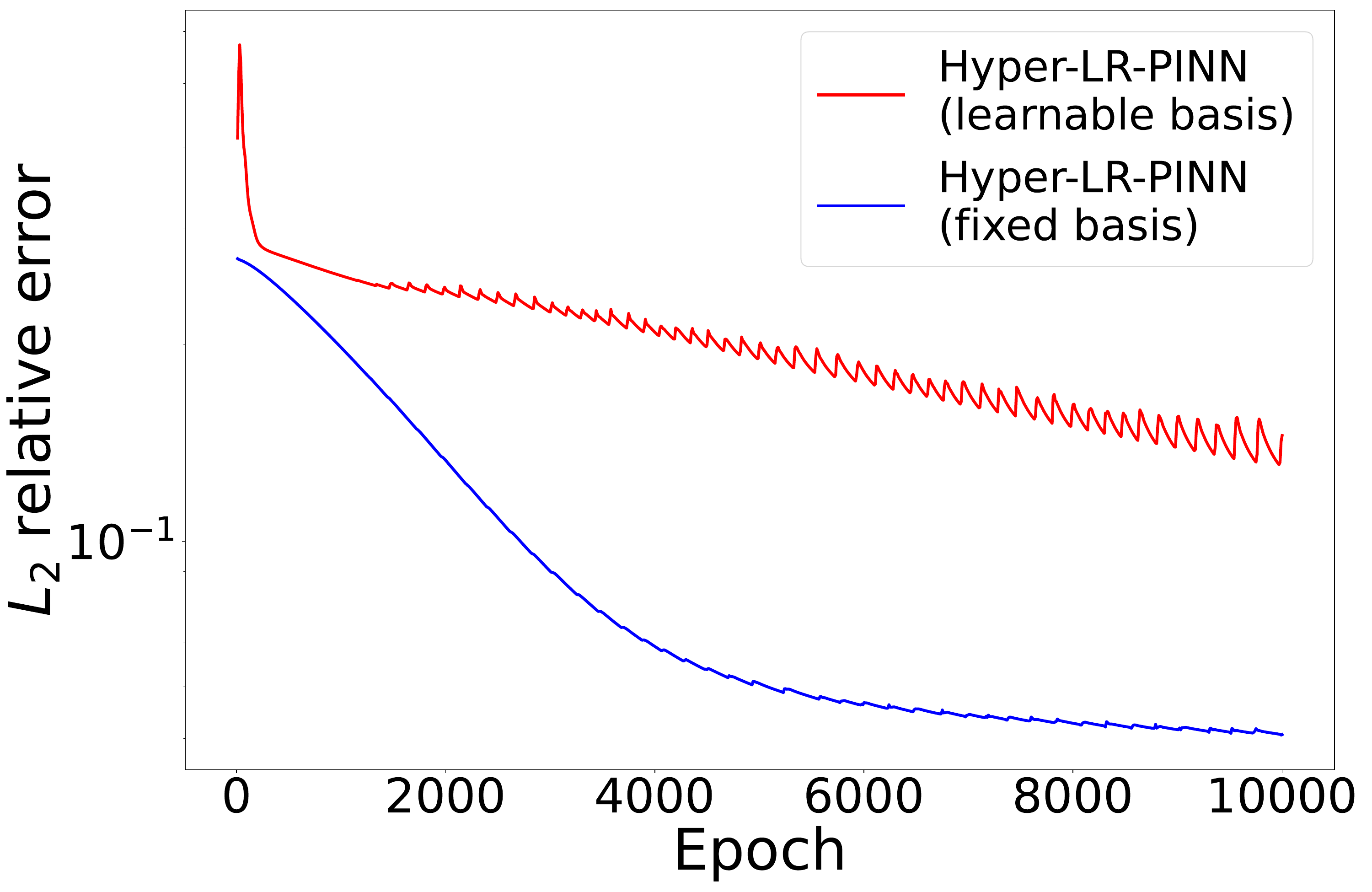}}\hfill
    \subfloat[$\beta = 40$ (Rel. err.)]
    {\includegraphics[width=0.32\columnwidth]
    {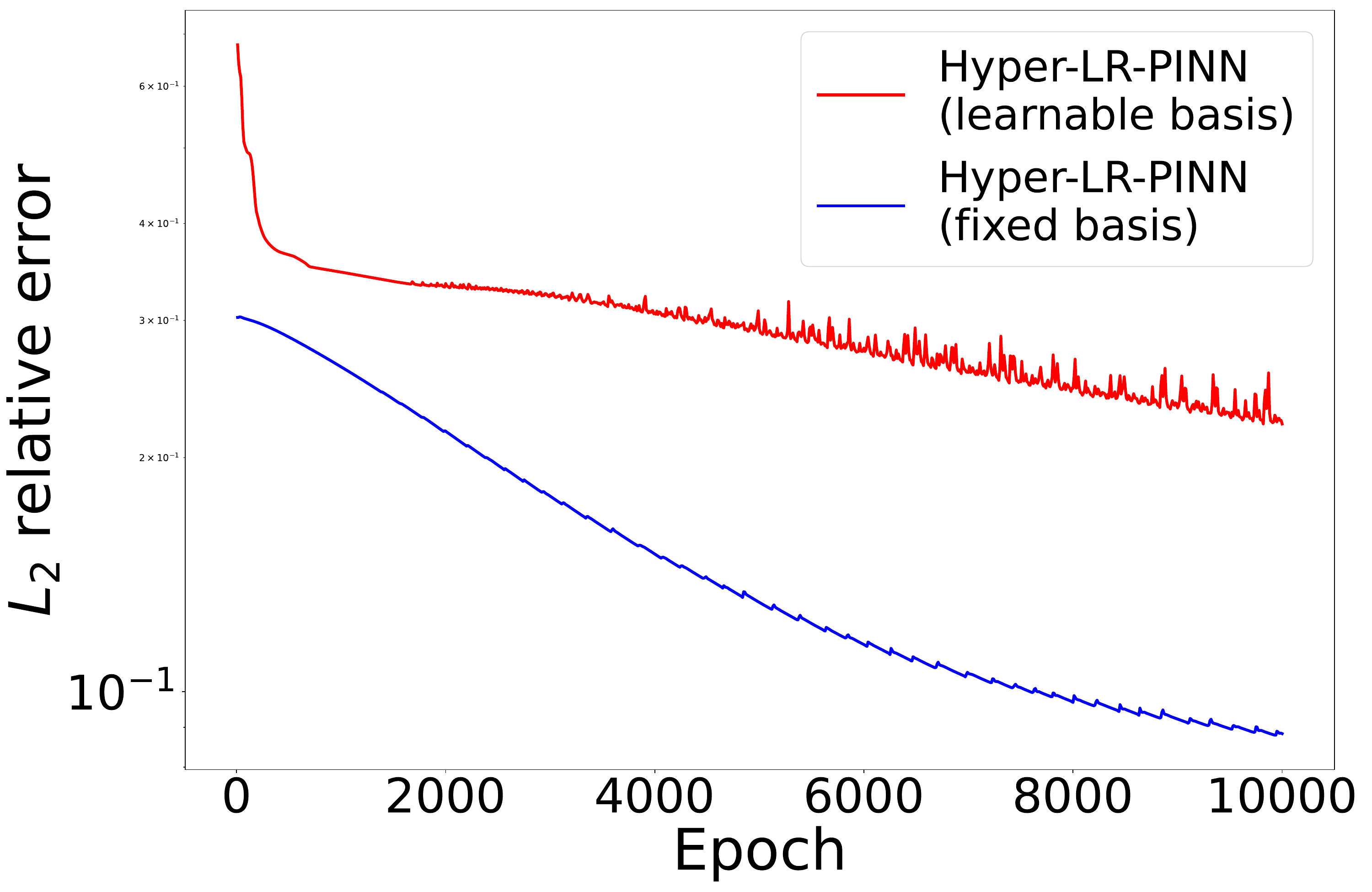}} \\
    \caption{Curves of $L_2$ absolute error (top row) and $L_2$ relative error (bottom row) on convection equations in phase 2. $\beta = \{30, 35, 40\}$}\label{fig:learn_fix_test}
\end{figure}

\clearpage

\begin{table}[ht!]
\centering
\caption{Experimental results of learnable basis and fixed basis. $\beta = \{30, 35, 40\}$}\label{tab:learn_fix_err}
\begin{tabular}{lcccc}
\specialrule{1pt}{2pt}{2pt}
\multirow{2}{*}{$\pmb{\beta}$} & \multicolumn{2}{c}{\textbf{Learnable basis}} & \multicolumn{2}{c}{\textbf{Fixed basis}} \\ \cmidrule(lr){2-5}
   & Abs. err. & Rel. err. & Abs. err. & Rel. err. \\ 
\specialrule{1pt}{2pt}{2pt}
\textbf{30} & 0.0567    & 0.0656    & 0.0386    & 0.0405    \\
\textbf{35} & 0.1344    & 0.1454    & 0.0490    & 0.0509    \\
\textbf{40} & 0.2169    & 0.2208    & 0.0790    & 0.0883    \\
\specialrule{1pt}{2pt}{2pt}
\end{tabular}
\end{table}

\begin{table}[ht!]
\centering
\caption{Number of learnable parameters in Phase2}\label{tab:learn_fix_model_size}
\begin{tabular}{lrrr}
\specialrule{1pt}{2pt}{2pt}
                & $\beta=30$    & $\beta=35$    & $\beta=40$    \\
\specialrule{1pt}{2pt}{2pt}
Learnable basis & 9,392 & 9,594 & 9,594 \\
Fixed basis     & 292   & 294   & 293  \\
\specialrule{1pt}{2pt}{2pt}
\end{tabular}
\end{table}

To compare fixed basis and learnable basis, we provide training loss curves in Figure~\ref{fig:learn_fix_train} and curves of absolute error (Abs.err.) and relative error (Rel.err.) in Figure~\ref{fig:learn_fix_test}. In all cases, fixed basis exhibits more stable learning and superior performance compared to learnable basis. Additionally, as can be seen in Table~\ref{tab:learn_fix_model_size}, fixed basis is also significantly more efficient with a model size over 30 times smaller.

\section{Visualization of the results in phase 1 and phase 2}

\begin{figure}[hbt!]
    \centering
    \subfloat[$\beta = 30$ (phase 1)]
    {\includegraphics[width=0.32\columnwidth]
    {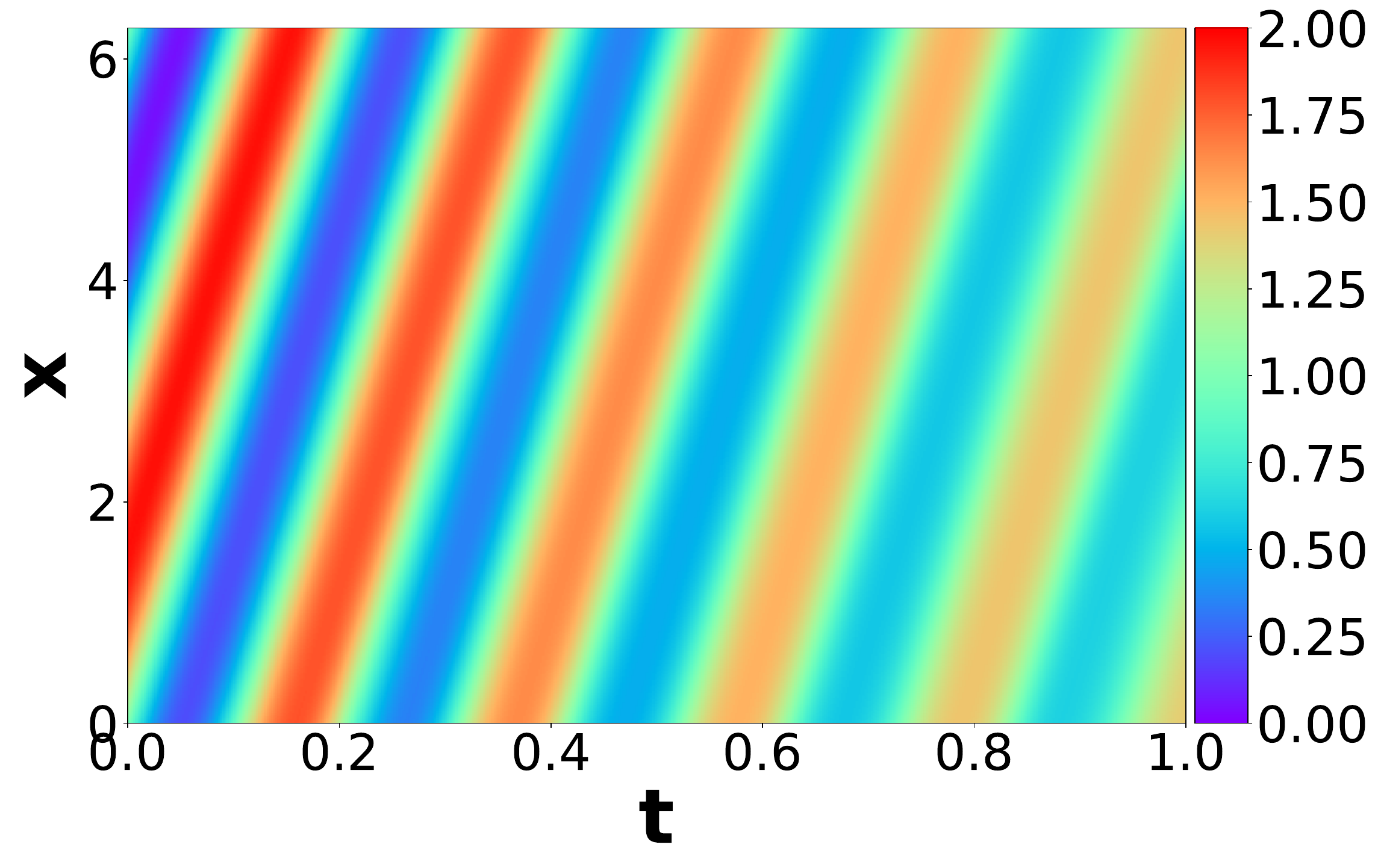}}\hfill
    \subfloat[$\beta = 35$ (phase 1)]
    {\includegraphics[width=0.32\columnwidth]
    {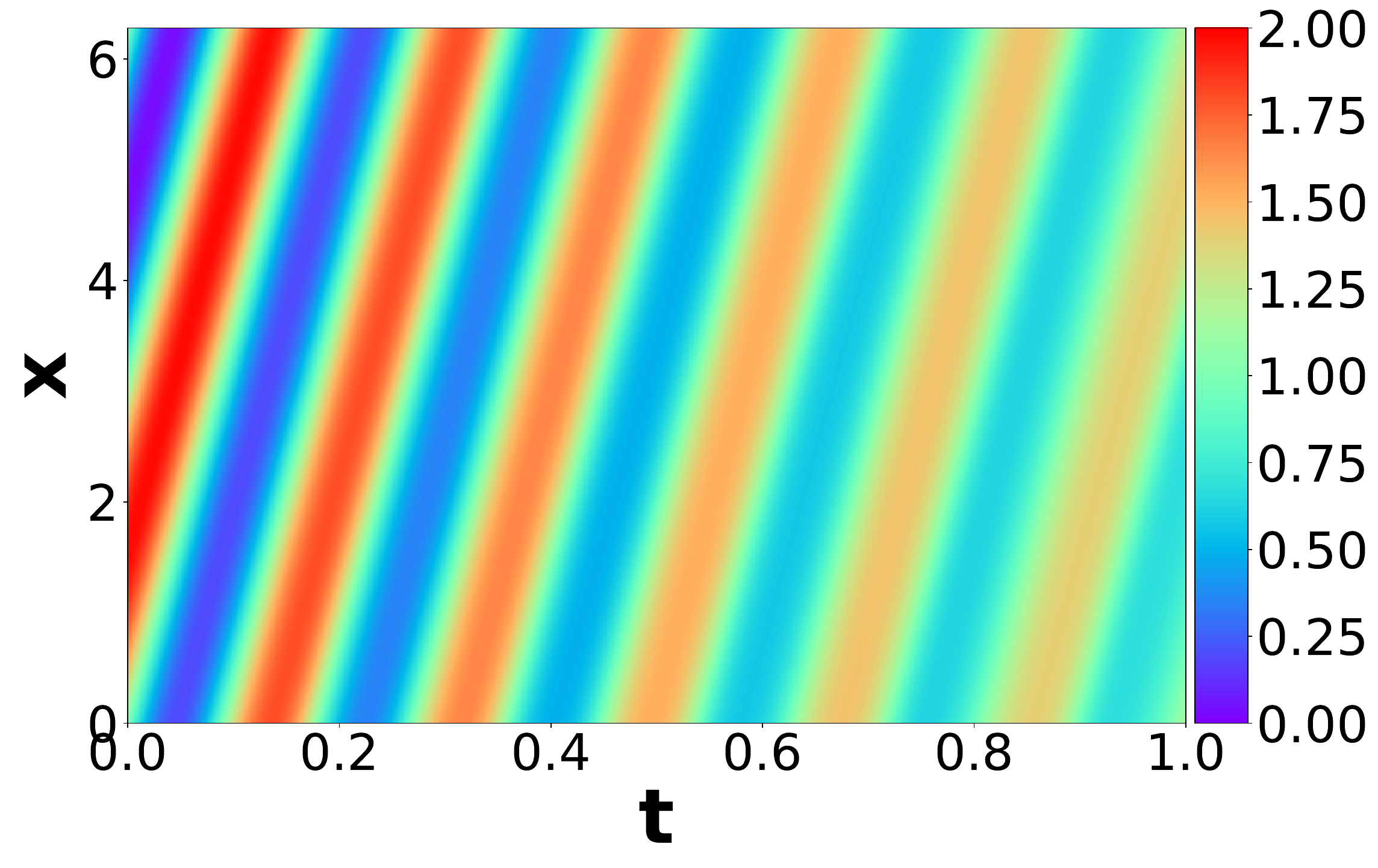}}\hfill
    \subfloat[$\beta = 40$ (phase 1)]
    {\includegraphics[width=0.32\columnwidth]
    {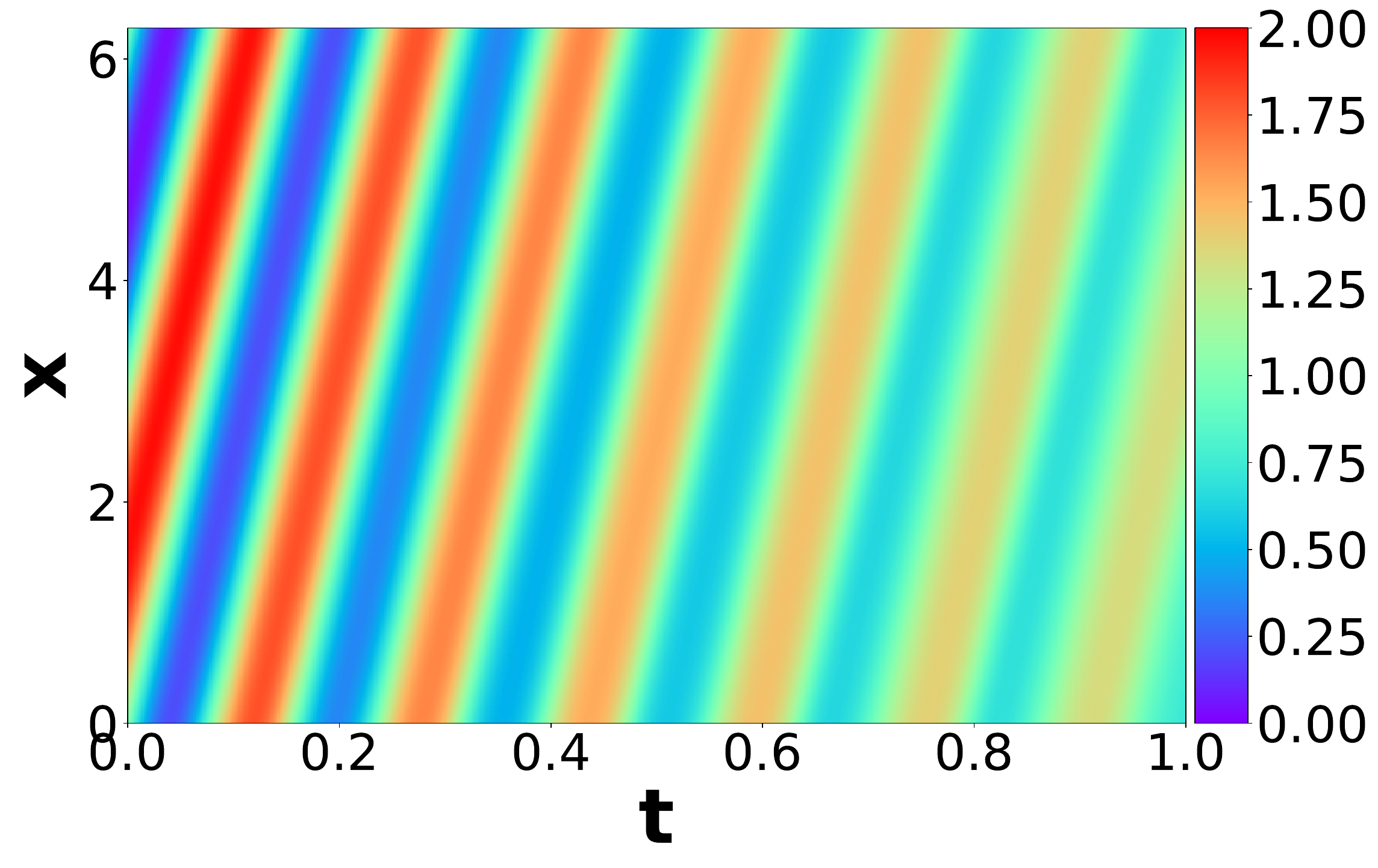}} \\

    \subfloat[$\beta = 30$ (phase 2)]
    {\includegraphics[width=0.32\columnwidth]
    {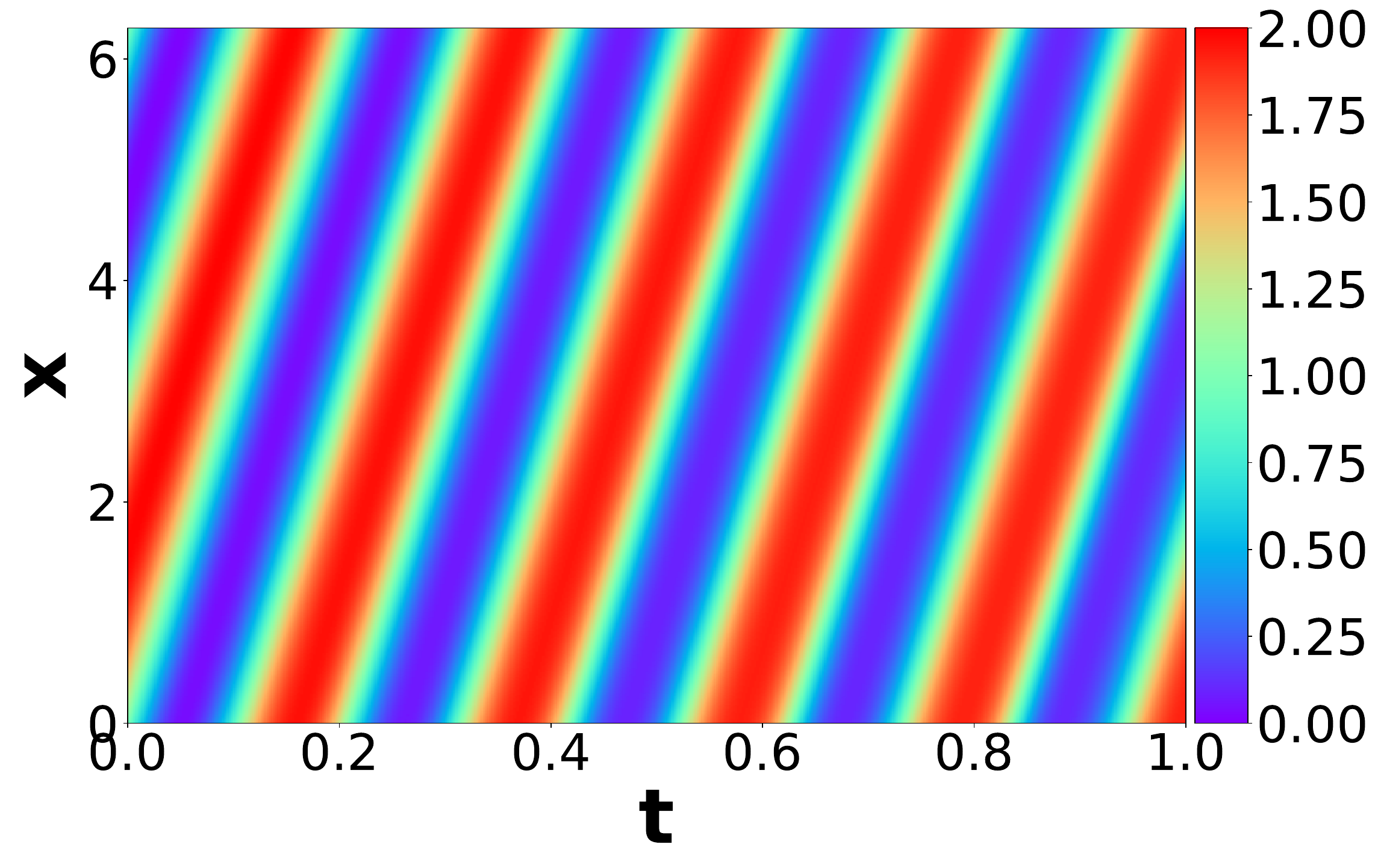}}\hfill
    \subfloat[$\beta = 35$ (phase 2)]
    {\includegraphics[width=0.32\columnwidth]
    {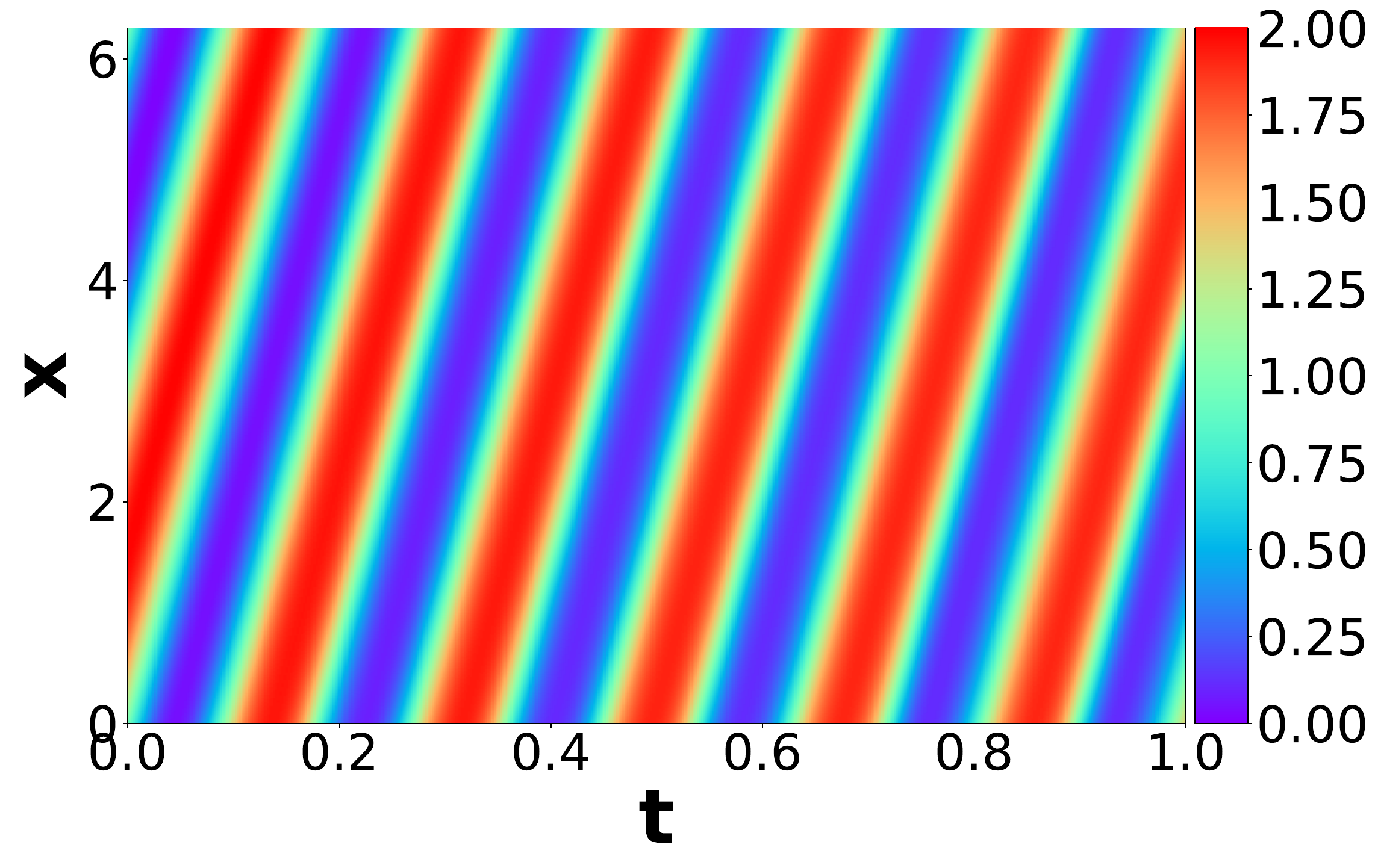}}\hfill
    \subfloat[$\beta = 40$ (phase 2)]
    {\includegraphics[width=0.32\columnwidth]
    {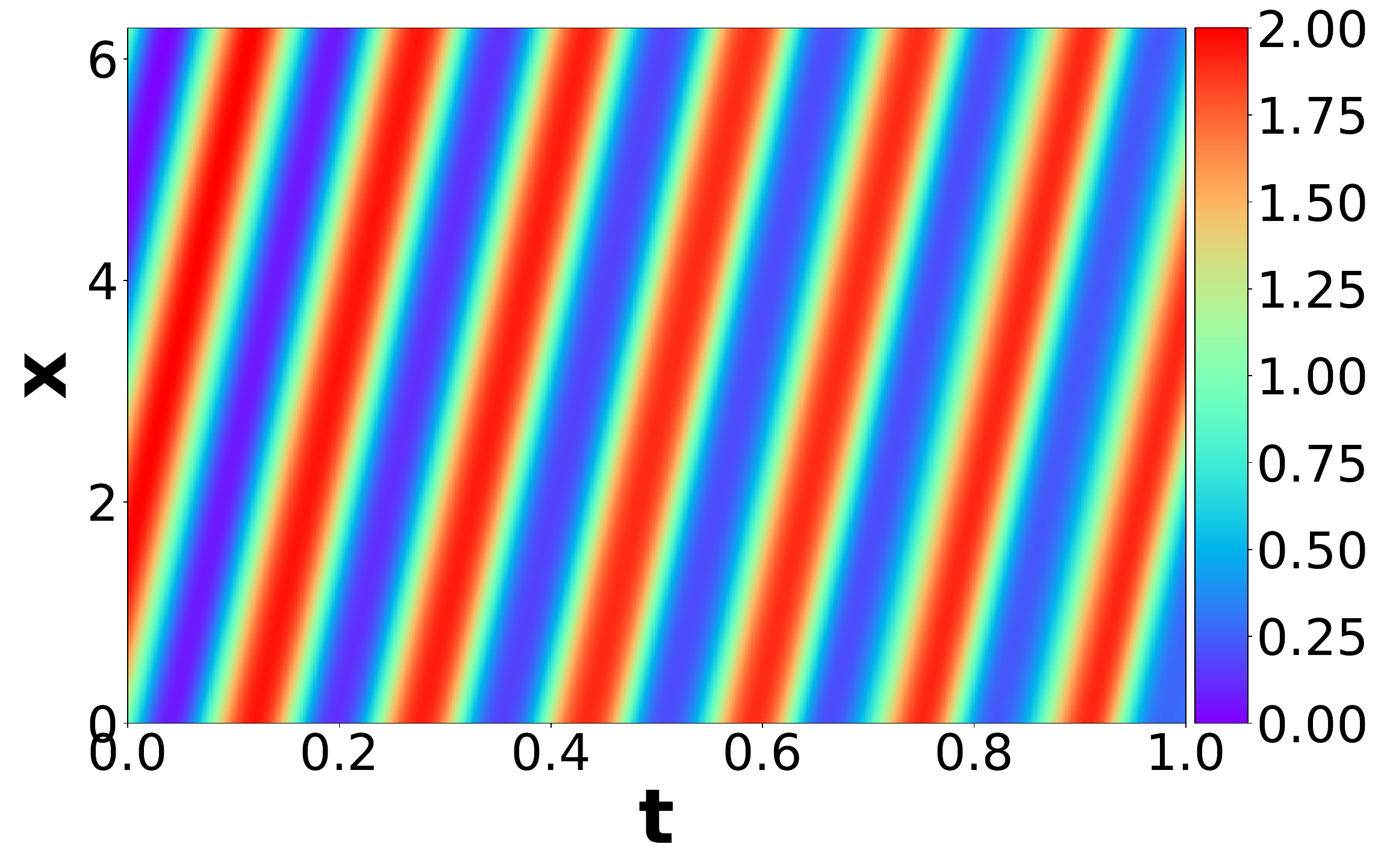}} \\
    \caption{The figures in the top row are the results of phase 1, and the figures in the bottom row are the results of phase 2. We follow the experimental setting of Table 2. $\beta = \{30, 35, 40\}$}
\end{figure}

\section{Other experiments}\label{sec:other_exps}
\subsection{Model size study for PINNs}
In Table~\ref{tab:PINN_model_size}, we report the performance by the model size, i.e., the number of model parameters.

\begin{table}[ht!]
    \centering
    \caption{PINNs performance by varying the model size. The numbers inside parentheses are model parameter numbers.}\label{tab:PINN_model_size}
    \resizebox{\textwidth}{!}{
    \begin{tabular}{lrrrrrrrrrrrr}
    \specialrule{1pt}{2pt}{2pt}
    \multirow{2}{*}{\pmb{$\beta$}} &
      \multicolumn{2}{r}{dim=50 (10,401)} &
      \multicolumn{2}{r}{dim=40 (6,721)} &
      \multicolumn{2}{r}{dim=30 (3,841)} &
      \multicolumn{2}{r}{dim=20 (1,761)} &
      \multicolumn{2}{r}{dim=10 (481)} &
      \multicolumn{2}{r}{dim=8 (321)} \\ \cmidrule(lr){2-13}

       & Abs.err. & Rel.err. & Abs.err. & Rel.err. & Abs.err. & Rel.err. & Abs.err. & Rel.err. & Abs.err. & Rel.err. & Abs.err. & Rel.err. \\ \cmidrule(lr){1-13}
    \textbf{30} & 0.4015   & 0.4033   & 0.4077   & 0.4114   & 0.4576   & 0.4524   & 0.4365   & 0.4369   & 0.4398   & 0.4409   & 0.4479   & 0.4478   \\
    \textbf{35} & 0.4785   & 0.4701   & 0.4815   & 0.4736   & 0.5132   & 0.4971   & 0.4491   & 0.4464   & 0.4823   & 0.4763   & 0.5023   & 0.4903   \\
    \textbf{40} & 0.5490   & 0.5219   & 0.5599   & 0.5333   & 0.5792   & 0.5467   & 0.5302   & 0.5097   & 0.5088   & 0.4966   & 0.5410   & 0.5207   \\
    \specialrule{1pt}{2pt}{2pt}
    \end{tabular}}
\end{table}
\subsection{Performance comparisons on full-rank and adaptive-rank in Hyper-LR-PINNs}

\begin{table}[ht!]
\centering
\caption{Comparison of full rank and adaptive rank in Phase2 }\label{tab:comparison_full_adap}
\begin{tabular}{lcccccc}
\specialrule{1pt}{2pt}{2pt}
\multirow{2}{*}{$\pmb{\beta}$} &
  \multicolumn{2}{c}{\textbf{Phase1}} &
  \multicolumn{2}{c}{\begin{tabular}[c]{@{}c@{}}\textbf{Phase2}\\ \textbf{(Full rank)}\end{tabular}} &
  \multicolumn{2}{c}{\begin{tabular}[c]{@{}c@{}}\textbf{Phase2}\\ \textbf{(Adaptive rank)}\end{tabular}} \\ \cmidrule(lr){2-7}
   & Abs.err. & Rel.err. & Abs.err. & Rel.err. & Abs.err. & Rel.err. \\ \cmidrule(lr){1-7}
\textbf{30} & 0.2294   & 0.2346   & 0.0360   & 0.0379   & 0.0375   & 0.0389   \\
\textbf{35} & 0.2443   & 0.2534   & 0.0428   & 0.0443   & 0.0448   & 0.0461   \\
\textbf{40} & 0.2443   & 0.2534   & 0.0603   & 0.0655   & 0.0656   & 0.0722   \\
\specialrule{1pt}{2pt}{2pt}
\end{tabular}
\end{table}

In Table~\ref{tab:comparison_full_adap}, we compare the cases where we fix weights to be of full-rank and where we make them adaptive (i.e., some singular values are truncated by ReLU). The adaptive rank approach still achieve the comparable performance.

\subsection{Performance comparisons against Hyper-PINNs}

\begin{table}[ht!]
\centering
\caption{Comparison of HyperPINN and Hyper-LR-PINN}\label{tab:comparison_hyper_PINN}
\begin{tabular}{lcccc}
\specialrule{1pt}{2pt}{2pt}
\multirow{2}{*}{$\pmb{\beta}$} & \multicolumn{2}{c}{\textbf{HyperPINN}} & \multicolumn{2}{c}{\textbf{Hyper-LR-PINN}} \\ \cmidrule(lr){2-5}
                      & Abs.err.      & Rel.err.      & Abs.err.        & Rel.err.        \\ \cmidrule(lr){1-5}
\textbf{30}                    & 0.9491        & 0.9856        & 0.0375          & 0.0389          \\
\textbf{35}                    & 0.9639        & 0.9831        & 0.0448          & 0.0461          \\
\textbf{40}                    & 0.9773        & 0.9989        & 0.0656          & 0.0722          \\
\specialrule{1pt}{2pt}{2pt}
\end{tabular}
\end{table}

We provide the results of the comparison between HyperPINN and Hyper-LR-PINN (Adaptive rank) in Table~\ref{tab:comparison_hyper_PINN}. In all cases, Hyper-LR-PINN demonstrates overwhelmingly superior performance. All experimental settings are the same as in Table~\ref{tab:result_conv_err}.

\subsection{Performance on extrapolation in the PDE parameter domain}
\begin{table}[ht!]
\centering
\caption{Detailed results of the experiment in Figure~\ref{fig:interpolation_task}}\label{tab:detailed_results}
\begin{tabular}{lllllll}
\specialrule{1pt}{2pt}{2pt}
 &
  $\pmb{\beta=27}$ &
  $\pmb{\beta=28}$ &
  $\pmb{\beta=29}$ &
  $\pmb{\beta=41}$ &
  $\pmb{\beta=42}$ &
  $\pmb{\beta=43}$ \\ \cmidrule(lr){1-7}
Abs. err. &
  0.0311 &
  0.0351 &
  0.0358 &
  0.1002 &
  0.1089 &
  0.1490 \\
Rel. err. &
  0.0318 &
  0.0366 &
  0.0377 &
  0.1174 &
  0.1238 &
  0.1624 \\
\specialrule{1pt}{2pt}{2pt}
\end{tabular}
\end{table}

\clearpage
\section{2D-Helmholtz equation}\label{a:helmholtz_appendix}
We present the results of applying Hyper-LR-PINN to compute solutions of parameterized 2D Helmholtz equations:
\begin{align}
    \begin{split}
    \frac{\partial^2 u(x,y)}{\partial x^2} + 
    \frac{\partial^2 u(x,y)}{\partial y^2} + k^2 u(x,y) -q(x,y) = 0, \\
    q(x,y) = (-(a_1 \pi)^2-(a_2 \pi)^2+k^2)\sin(a_1 \pi x) \sin(a_2 \pi y), \\
    \end{split}
\label{eq:helmholtz_eq}
\end{align} 
where the forcing term $q$ is chosen such that the analytical expression for the solutions are available: 
\begin{align}
    u(x,y) = k^2 \sin(a_1 \pi x) \sin(a_2 \pi y).
\label{eq:helmholtz_eq_sol}
\end{align} 

\begin{figure}[ht!]
\vspace{-1.0em}

\subfloat[PINN ($a=2.0$)]{\includegraphics[width=0.32\columnwidth]{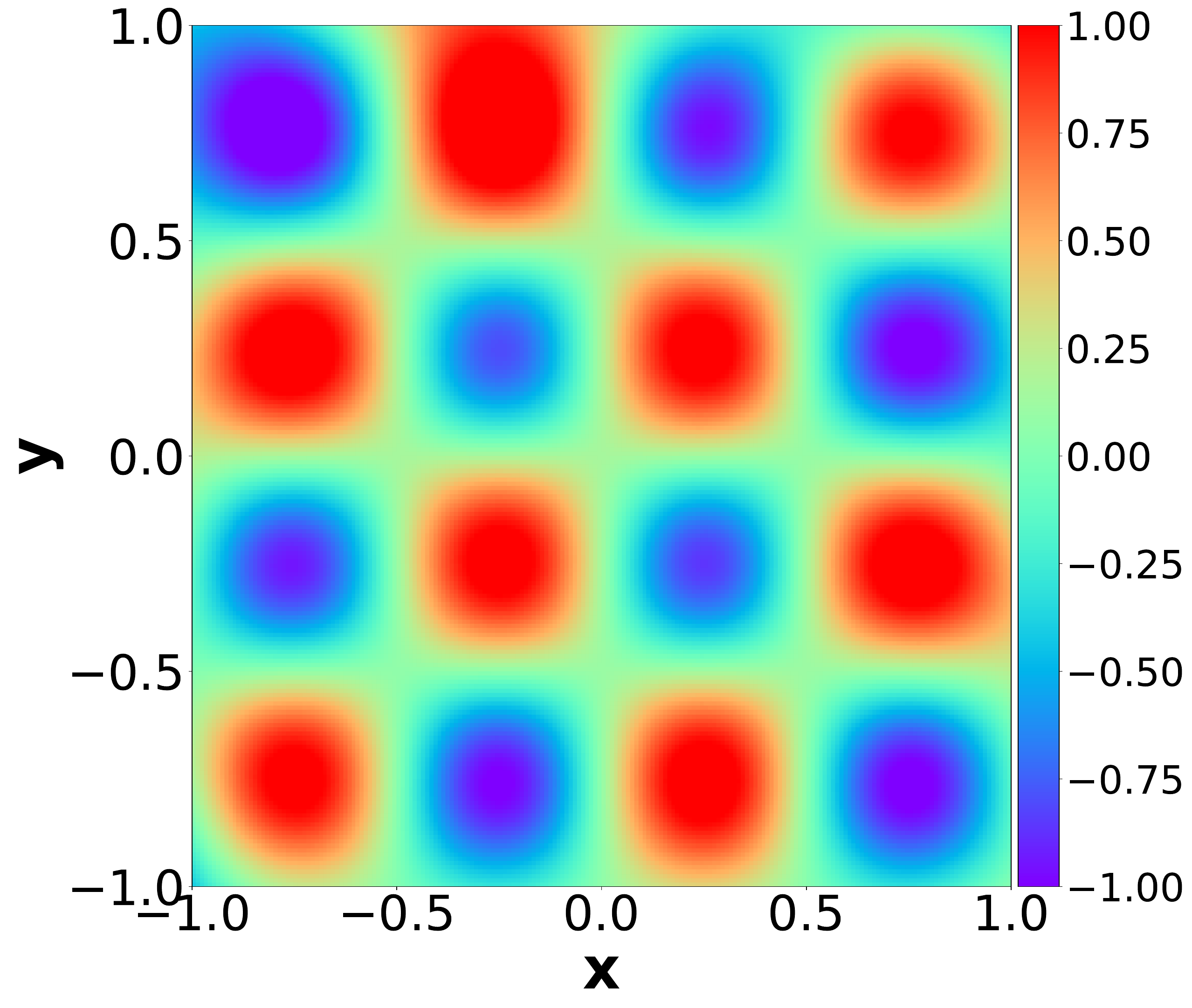}}\hfill
\subfloat[Ours ($a=2.0$)]{\includegraphics[width=0.32\columnwidth]{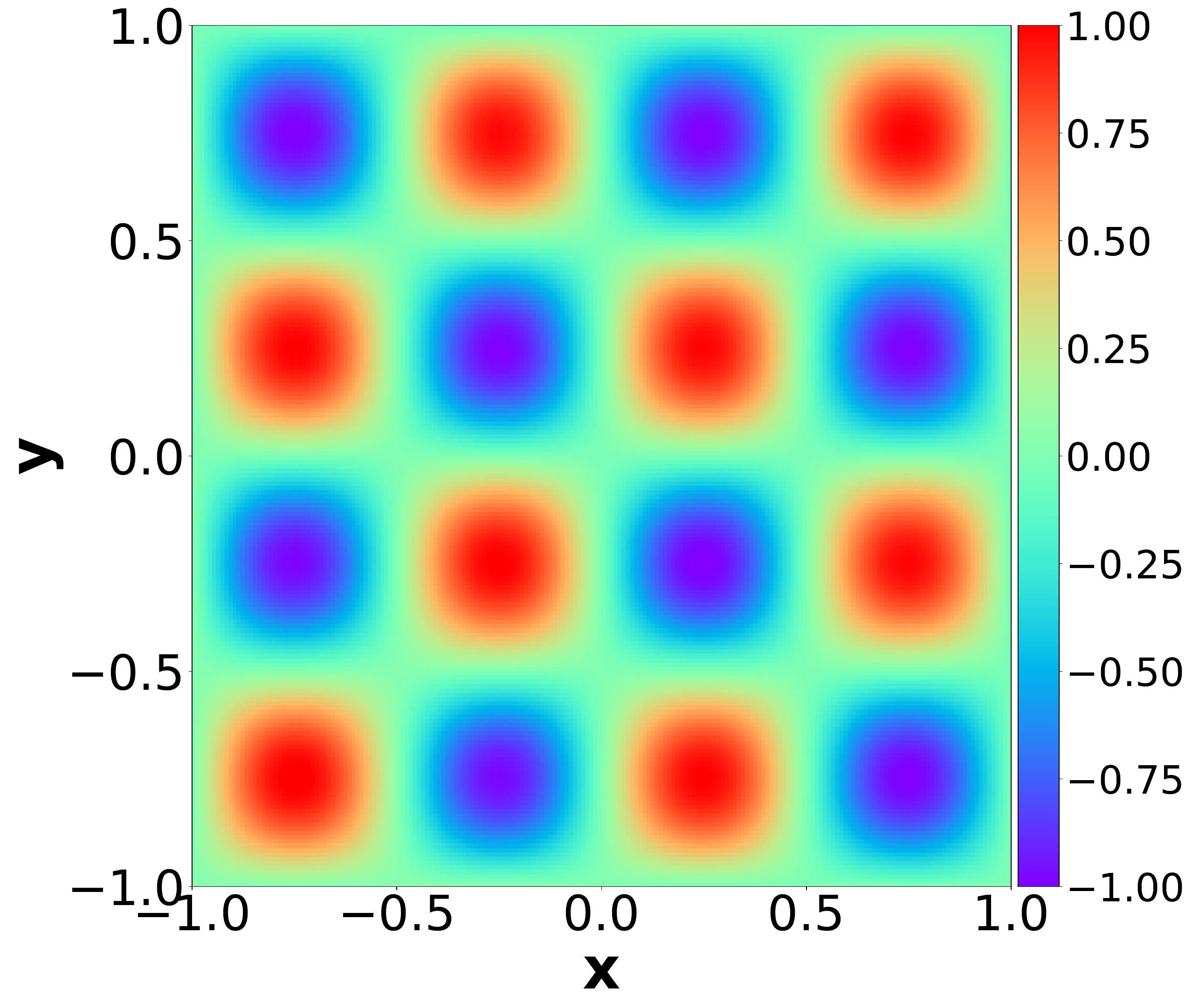}}\hfill
\subfloat[Exact ($a=2.0$)]{\includegraphics[width=0.32\columnwidth]{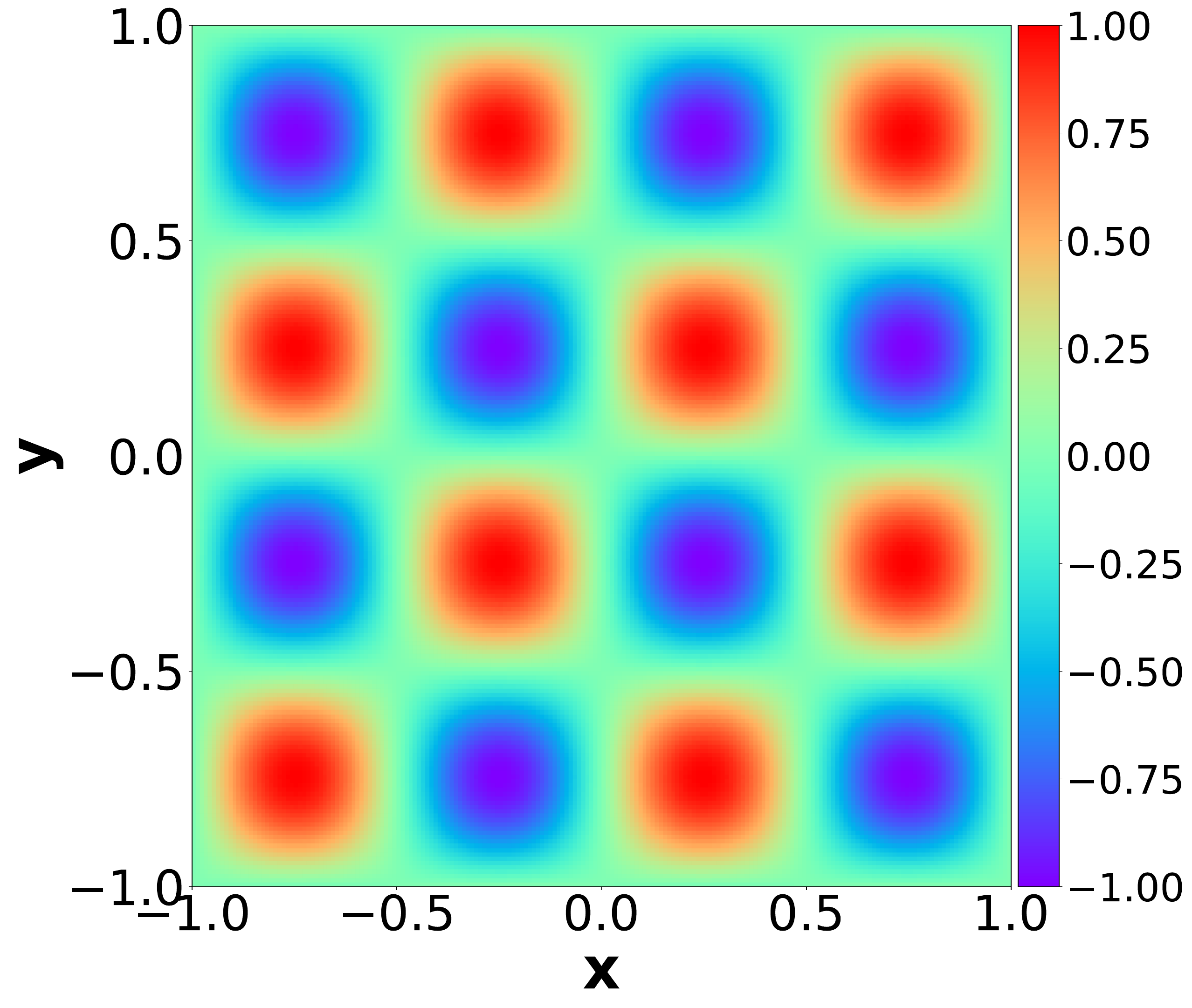}}\\
\vspace{-0.5em}

\subfloat[PINN ($a=2.5$)]{\includegraphics[width=0.32\columnwidth]{figs/pinn_helm_2.5.pdf}}\hfill
\subfloat[Ours ($a=2.5$)]{\includegraphics[width=0.32\columnwidth]{figs/lr_pinn_helm_2.5.pdf}}\hfill
\subfloat[Exact ($a=2.5$)]{\includegraphics[width=0.32\columnwidth]{figs/helm_2.5_2.5.pdf}}\\
\vspace{-0.5em}

\subfloat[PINN ($a=3.0$)]{\includegraphics[width=0.32\columnwidth]{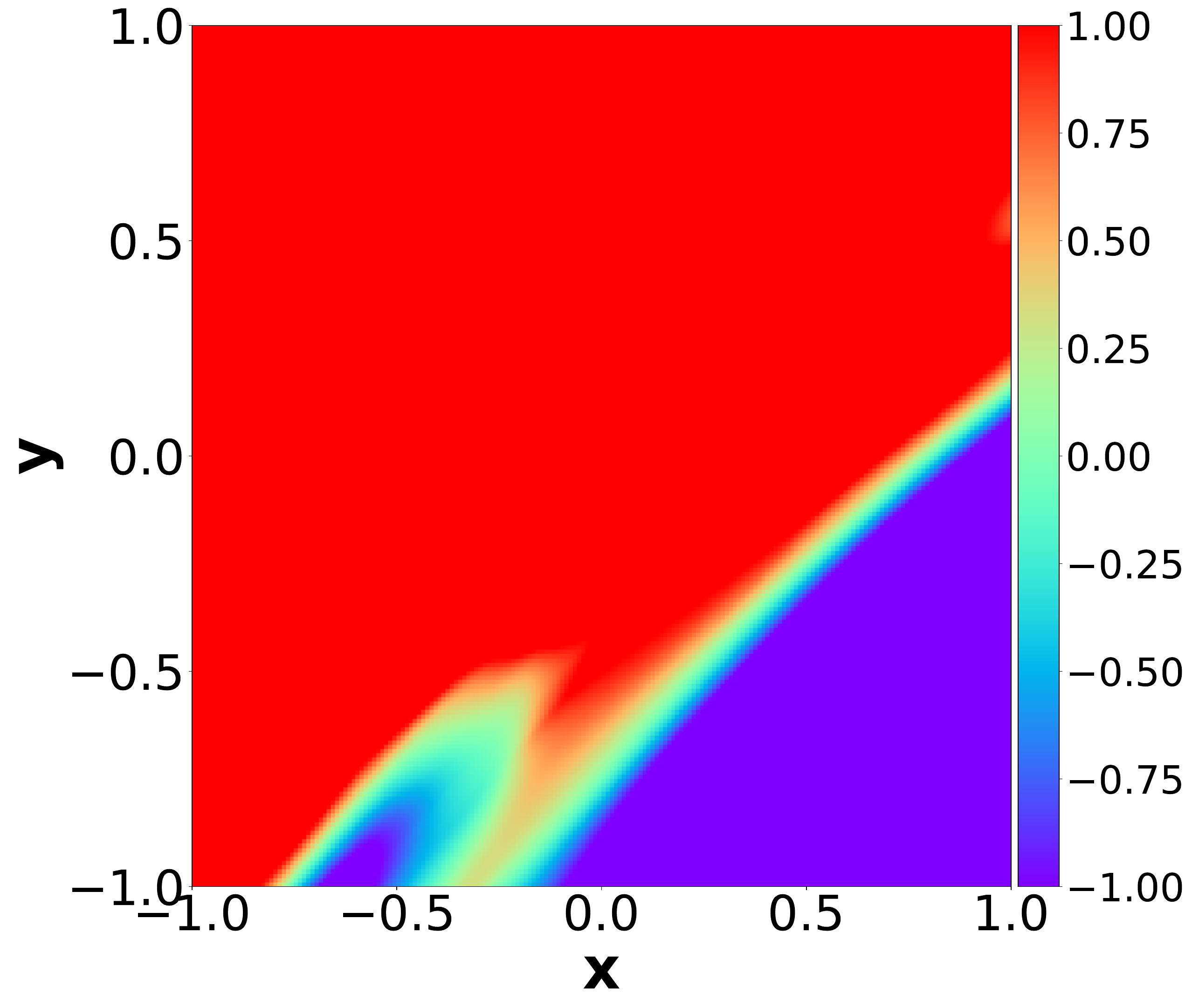}}\hfill
\subfloat[Ours ($a=3.0$)]{\includegraphics[width=0.32\columnwidth]{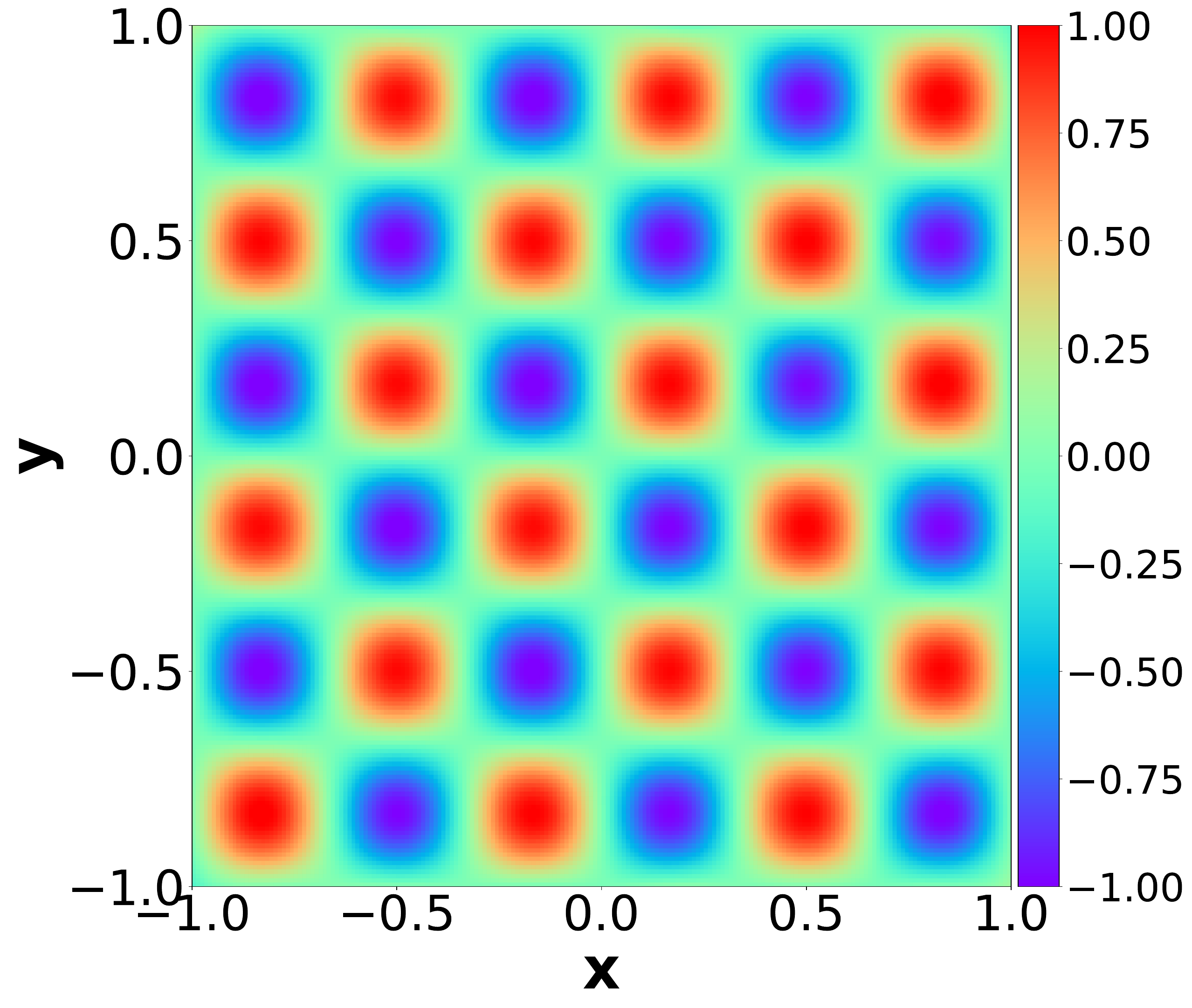}}\hfill
\subfloat[Exact ($a=3.0$)]{\includegraphics[width=0.32\columnwidth]{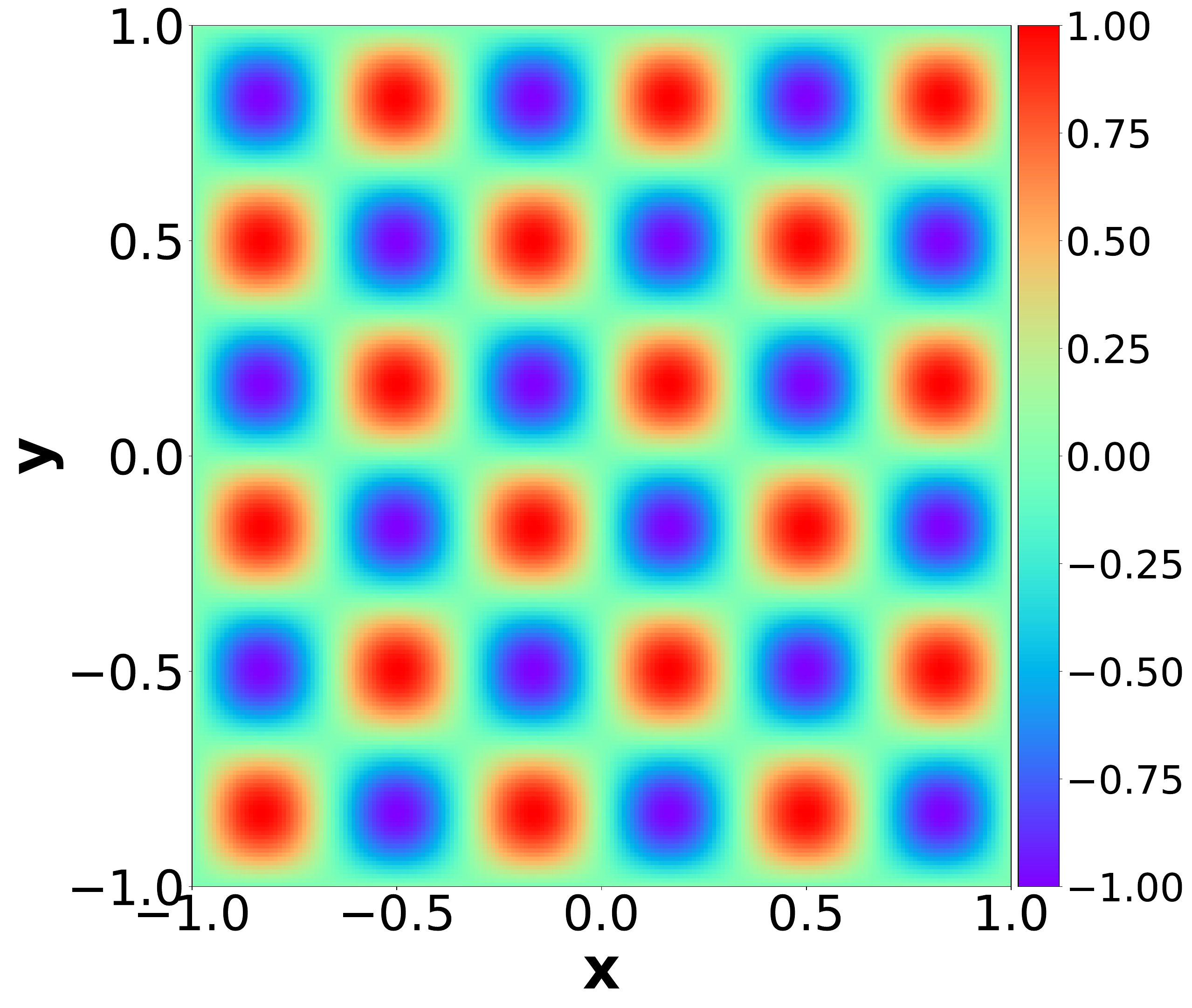}}\\

\caption{[2D-Helmholtz equation] Solution snapshots}\label{fig:helmholtz}
\vspace{-.5em}
\end{figure}

Figure~\ref{fig:helmholtz} shows the solution snapshots of 2D-Helmholtz equations for varying parameters and demonstrates that the proposed approach outperforms the vanilla PINNs. 

\clearpage

The comparison results of the absolute and relative errors between PINN and Hyper-LR-PINN can be observed in Table~\ref{tab:helmholtz}. In all cases, Hyper-LR-PINN consistently exhibits significantly smaller errors compared to PINN.

\begin{table}[ht!]
    \centering
    \small
    \caption{[2D-Helmholtz equation] The absolute and relative errors of the solutions of 2D-Helmholtz equations with $a = \{2.0,2.5,3.0\}$}\label{tab:helmholtz}
    \resizebox{\textwidth}{!}{
    \begin{tabular}{cccccccc}
    \specialrule{1pt}{2pt}{2pt}
    \multirow{2}{*}{} &
      \multicolumn{2}{c}{$a=2.0$} &
      \multicolumn{2}{c}{$a=2.5$} &
      \multicolumn{2}{c}{$a=3.0$}  \\ \cmidrule(lr){2-7}

       & Abs.err. & Rel.err. & Abs.err. & Rel.err. & Abs.err. & Rel.err. \\ \cmidrule(lr){1-7}
    \textbf{PINN}          & 0.1698\scriptsize{$\pm$0.0553} & 0.4258\scriptsize{$\pm$0.0939} & 0.7403\scriptsize{$\pm$0.1050} & 1.8232\scriptsize{$\pm$0.2464} & 2.0537\scriptsize{$\pm$0.4904} & 4.8401\scriptsize{$\pm$0.7366} \\
    \textbf{Hyper-LR-PINN} & 0.0137\scriptsize{$\pm$0.0021} & 0.0330\scriptsize{$\pm$0.0032} & 0.0285\scriptsize{$\pm$0.0029} & 0.0611\scriptsize{$\pm$0.0071} & 0.0153\scriptsize{$\pm$0.0028} & 0.0389\scriptsize{$\pm$0.0060} \\
    \specialrule{1pt}{2pt}{2pt}
    \end{tabular}}
\end{table}


\section{Standard deviation}\label{a:std}

Table~\ref{tab:result_conv_err} presents the mean error of the experimental results and additionally, we provide the standard deviation for the evaluation metrics in Tables~\ref{tbl:std_abs_rel} and~\ref{tbl:std_max_exp}.

\begin{table}[!h]
\renewcommand{\arraystretch}{0.85}
\centering
    \caption{Standard deviation of the evaluation metrics (Abs. err. and Rel. err.) $\beta = \{30, 35, 40\}$}\label{tbl:std_abs_rel}
\resizebox{\textwidth}{!}{
\begin{tabular}{lccccccccccccc}
\specialrule{1pt}{2pt}{2pt}
\multirow{5}{*}{$\pmb{\beta}$} & \multirow{5}{*}{\textbf{Rank}} & \multicolumn{2}{c}{\textbf{{[}w/o{]} Pre-training}} & \multicolumn{10}{c}{\textbf{{[}w{]} Pre-training}} \\ \cmidrule(lr){3-4} \cmidrule(lr){5-14}
 &  & \multicolumn{2}{c}{\textbf{Na\"ive-LR-PINN}} & \multicolumn{2}{c}{\begin{tabular}[c]{@{}c@{}}\textbf{Curriculum}\\ \textbf{learning}\end{tabular}} & \multicolumn{2}{c}{\textbf{MAML}} & \multicolumn{2}{c}{\textbf{Reptile}} & \multicolumn{2}{c}{\begin{tabular}[c]{@{}c@{}}\textbf{Hyper-LR-PINN}\\ \textbf{(Full rank)}\end{tabular}} & \multicolumn{2}{c}{\begin{tabular}[c]{@{}c@{}}\textbf{Hyper-LR-PINN}\\ \textbf{(Adaptive rank)}\end{tabular}} \\ \cmidrule(lr){3-14}
 &  & \multicolumn{1}{c}{Abs. err.} & \multicolumn{1}{c}{Rel. err.} & \multicolumn{1}{c}{Abs. err.} & \multicolumn{1}{c}{Rel. err.} & \multicolumn{1}{c}{Abs. err.} & \multicolumn{1}{c}{Rel. err.} & \multicolumn{1}{c}{Abs. err.} & \multicolumn{1}{c}{Rel. err.} & Abs. err. & Rel. err. & Abs. err. & Rel. err. \\
\specialrule{1pt}{2pt}{2pt}
\multirow{7}{*}{\textbf{30}}  
 & 10 & $\pm$0.0526 & $\pm$0.0397 & $\pm$0.0000	& $\pm$0.0000 & $\pm$0.0172 & $\pm$0.0247 & $\pm$0.0521 & $\pm$0.0474 & \multirow{7}{*}{$\pm$0.0023} & \multirow{7}{*}{$\pm$0.0023} & \multirow{7}{*}{$\pm$0.0025}& \multirow{7}{*}{$\pm$0.0014} \\ \cmidrule(lr){2-10}
 & 20 & $\pm$0.0374 & $\pm$0.0300 & $\pm$0.0000	& $\pm$0.0000 & $\pm$0.0164 & $\pm$0.0243 & $\pm$0.0260 & $\pm$0.0217 &  &  &  &  \\ \cmidrule(lr){2-10}
 & 30 & $\pm$0.0522 & $\pm$0.0423 & $\pm$0.0000	& $\pm$0.0000 & $\pm$0.0508 & $\pm$0.0446 & $\pm$0.0135 & $\pm$0.0125 &  &  &  &  \\ \cmidrule(lr){2-10}
 & 40 & $\pm$0.0446 & $\pm$0.0339 & $\pm$0.0015	& $\pm$0.0005 & $\pm$0.0780 & $\pm$0.0870 & $\pm$0.0024 & $\pm$0.0021 &  &  &  &  \\ \cmidrule(lr){2-10}
 & 50 & $\pm$0.0385 & $\pm$0.0298 & $\pm$0.0001	& $\pm$0.0003 & $\pm$0.0610 & $\pm$0.0647 & $\pm$0.0055 & $\pm$0.0047 &  &  &  &  \\

\specialrule{1pt}{2pt}{2pt}
\multirow{7}{*}{\textbf{35}}
 & 10 & $\pm$0.0484 & $\pm$0.0382 & $\pm$0.0016	& $\pm$0.0006 & $\pm$0.0243 & $\pm$0.0353 & $\pm$0.0545	& $\pm$0.0498 & \multirow{7}{*}{$\pm$0.0029} & \multirow{7}{*}{$\pm$0.0028} & \multirow{7}{*}{$\pm$0.0025}& \multirow{7}{*}{$\pm$0.0027} \\ \cmidrule(lr){2-10}
 & 20 & $\pm$0.0254 & $\pm$0.0201 & $\pm$0.0000	& $\pm$0.0000 & $\pm$0.0267 & $\pm$0.0414 & $\pm$0.0218	& $\pm$0.0163 &  &  &  &  \\ \cmidrule(lr){2-10}
 & 30 & $\pm$0.0143 & $\pm$0.0151 & $\pm$0.0000	& $\pm$0.0000 & $\pm$0.0483 & $\pm$0.0437 & $\pm$0.0069	& $\pm$0.0051 &  &  &  &  \\ \cmidrule(lr){2-10}
 & 40 & $\pm$0.0303 & $\pm$0.0224 & $\pm$0.0001	& $\pm$0.0002 & $\pm$0.0788 & $\pm$0.0942 & $\pm$0.0016	& $\pm$0.0012 &  &  &  &  \\ \cmidrule(lr){2-10}
 & 50 & $\pm$0.0307 & $\pm$0.0228 & $\pm$0.0001	& $\pm$0.0000 & $\pm$0.0734 & $\pm$0.0848 & $\pm$0.0041	& $\pm$0.0028 &  &  &  &  \\

\specialrule{1pt}{2pt}{2pt}
\multirow{7}{*}{\textbf{40}}
 & 10 & $\pm$0.0331 & $\pm$0.0273 & $\pm$0.0014	& $\pm$0.0005 & $\pm$0.0418 & $\pm$0.0648 & $\pm$0.0506	& $\pm$0.0537 &\multirow{7}{*}{$\pm$0.0038} & \multirow{7}{*}{$\pm$0.0043} & \multirow{7}{*}{$\pm$0.0043}& \multirow{7}{*}{$\pm$0.0042} \\ \cmidrule(lr){2-10}
 & 20 & $\pm$0.0215 & $\pm$0.0188 & $\pm$0.0000	& $\pm$0.0000 & $\pm$0.0410 & $\pm$0.0589 & $\pm$0.0233	& $\pm$0.0214 &  &  &  &  \\ \cmidrule(lr){2-10}
 & 30 & $\pm$0.0107 & $\pm$0.0116 & $\pm$0.0000	& $\pm$0.0000 & $\pm$0.0468 & $\pm$0.0510 & $\pm$0.0091	& $\pm$0.0109 &  &  &  &  \\ \cmidrule(lr){2-10}
 & 40 & $\pm$0.0363 & $\pm$0.0305 & $\pm$0.0001	& $\pm$0.0001 & $\pm$0.0703 & $\pm$0.0899 & $\pm$0.0005	& $\pm$0.0009 &  &  &  &  \\ \cmidrule(lr){2-10}
 & 50 & $\pm$0.0065 & $\pm$0.0080 & $\pm$0.0004	& $\pm$0.0005 & $\pm$0.0800 & $\pm$0.1027 & $\pm$0.0086	& $\pm$0.0071 &  &  &  &  \\

\specialrule{1pt}{2pt}{2pt}
\end{tabular}}
\end{table}

\begin{table}[!h]
\centering
\small
\caption{Standard deviation of the evaluation metrics (Max. err. and Exp. var.) $\beta = \{30, 35, 40\}$}\label{tbl:std_max_exp}
\resizebox{\textwidth}{!}{
\begin{tabular}{lccccccccccccc}
\specialrule{1pt}{2pt}{2pt}
\multirow{5}{*}{$\pmb{\beta}$} & \multirow{5}{*}{\textbf{Rank}} & \multicolumn{2}{c}{\textbf{{[}w/o{]} Pre-training}} & \multicolumn{10}{c}{\textbf{{[}w{]} Pre-training}} \\ \cmidrule(lr){3-4} \cmidrule(lr){5-14}
 &  & \multicolumn{2}{c}{\textbf{Na\"ive-LR-PINN}} & \multicolumn{2}{c}{\begin{tabular}[c]{@{}c@{}}\textbf{Curriculum}\\ \textbf{learning}\end{tabular}} & \multicolumn{2}{c}{\textbf{MAML}} & \multicolumn{2}{c}{\textbf{Reptile}} & \multicolumn{2}{c}{\begin{tabular}[c]{@{}c@{}}\textbf{Hyper-LR-PINN}\\ \textbf{(Full rank)}\end{tabular}} & \multicolumn{2}{c}{\begin{tabular}[c]{@{}c@{}}\textbf{Hyper-LR-PINN}\\ \textbf{(Adaptive rank)}\end{tabular}} \\ \cmidrule(lr){3-14}
 &  & \multicolumn{1}{c}{Max. err.} & \multicolumn{1}{c}{Exp. var.} & \multicolumn{1}{c}{Max. err.} & \multicolumn{1}{c}{Exp. var.} & \multicolumn{1}{c}{Max. err.} & \multicolumn{1}{c}{Exp. var.} & \multicolumn{1}{c}{Max. err.} & \multicolumn{1}{c}{Exp. var.} & Max. err. & Exp. var. & Max. err. & Exp. var. \\
\specialrule{1pt}{2pt}{2pt}
\multirow{7}{*}{\textbf{30}}
 & 10 &  $\pm$0.0629	& $\pm$0.1001	& $\pm$0.0000	& $\pm$0.0000	& $\pm$0.0802	& $\pm$0.0126   & $\pm$0.1296 & $\pm$0.0803 &\multirow{7}{*}{$\pm$0.0025} & \multirow{7}{*}{$\pm$0.0015} & \multirow{7}{*}{$\pm$0.0009}& \multirow{7}{*}{$\pm$0.0015} \\ \cmidrule(lr){2-10}
 & 20 &  $\pm$0.0525	& $\pm$0.0760	& $\pm$0.0000	& $\pm$0.0000	& $\pm$0.0572	& $\pm$0.0089   & $\pm$0.0349 & $\pm$0.0512 &  &  &  &  \\ \cmidrule(lr){2-10}
 & 30 &  $\pm$0.0198	& $\pm$0.0332	& $\pm$0.0000	& $\pm$0.0000	& $\pm$0.1220	& $\pm$0.0842   & $\pm$0.0180 & $\pm$0.0222 &  &  &  &  \\ \cmidrule(lr){2-10}
 & 40 &  $\pm$0.0362	& $\pm$0.0943	& $\pm$0.0004	& $\pm$0.0026	& $\pm$0.1935	& $\pm$0.0999   & $\pm$0.0078 & $\pm$0.0056 &  &  &  &  \\ \cmidrule(lr){2-10}
 & 50 &  $\pm$0.0402	& $\pm$0.0809	& $\pm$0.0017	& $\pm$0.0004	& $\pm$0.1697	& $\pm$0.0815   & $\pm$0.0223 & $\pm$0.0137 &  &  &  &  \\

\specialrule{1pt}{2pt}{2pt}
\multirow{7}{*}{\textbf{35}}
 & 10 & $\pm$0.1273	& $\pm$0.0817	& $\pm$0.0023	& $\pm$0.0019	& $\pm$0.0571	& $\pm$0.0082  & $\pm$0.1590 & $\pm$0.0742 &\multirow{7}{*}{$\pm$0.0024} & \multirow{7}{*}{$\pm$0.0019} & \multirow{7}{*}{$\pm$0.0057}& \multirow{7}{*}{$\pm$0.0018} \\ \cmidrule(lr){2-10}
 & 20 & $\pm$0.0640	& $\pm$0.0463	& $\pm$0.0000	& $\pm$0.0000	& $\pm$0.0879	& $\pm$0.0135  & $\pm$0.0349 & $\pm$0.0435 &  &  &  &  \\ \cmidrule(lr){2-10}
 & 30 & $\pm$0.0771	& $\pm$0.0219	& $\pm$0.0000	& $\pm$0.0000	& $\pm$0.1478	& $\pm$0.0686  & $\pm$0.0192 & $\pm$0.0115 &  &  &  &  \\ \cmidrule(lr){2-10}
 & 40 & $\pm$0.0511	& $\pm$0.0603	& $\pm$0.0014	& $\pm$0.0002	& $\pm$0.2414	& $\pm$0.0882  & $\pm$0.0036 & $\pm$0.0026 &  &  &  &  \\ \cmidrule(lr){2-10}
 & 50 & $\pm$0.0556	& $\pm$0.0602	& $\pm$0.0001	& $\pm$0.0001	& $\pm$0.2254	& $\pm$0.0763  & $\pm$0.0152 & $\pm$0.0063 &  &  &  &  \\

\specialrule{1pt}{2pt}{2pt}
\multirow{7}{*}{\textbf{40}}
 & 10 & $\pm$0.0672	& $\pm$0.0446	& $\pm$0.0010	& $\pm$0.0019	& $\pm$0.1190	& $\pm$0.0165  & $\pm$0.1741 & $\pm$0.0574 &\multirow{7}{*}{$\pm$0.0106} & \multirow{7}{*}{$\pm$0.0011} & \multirow{7}{*}{$\pm$0.0105}& \multirow{7}{*}{$\pm$0.0004} \\ \cmidrule(lr){2-10}
 & 20 & $\pm$0.0645	& $\pm$0.0381	& $\pm$0.0000	& $\pm$0.0000	& $\pm$0.1143	& $\pm$0.0205  & $\pm$0.0456 & $\pm$0.0578 &  &  &  &  \\ \cmidrule(lr){2-10}
 & 30 & $\pm$0.0365	& $\pm$0.0143	& $\pm$0.0000	& $\pm$0.0000	& $\pm$0.1748	& $\pm$0.0564  & $\pm$0.0269 & $\pm$0.0093 &  &  &  &  \\ \cmidrule(lr){2-10}
 & 40 & $\pm$0.0539	& $\pm$0.0744	& $\pm$0.0016	& $\pm$0.0001	& $\pm$0.2441	& $\pm$0.0668  & $\pm$0.0047 & $\pm$0.0006 &  &  &  &  \\ \cmidrule(lr){2-10}
 & 50 & $\pm$0.0249	& $\pm$0.0156	& $\pm$0.0025	& $\pm$0.0001	& $\pm$0.2586	& $\pm$0.0626  & $\pm$0.0211 & $\pm$0.0133 &  &  &  &  \\

\specialrule{1pt}{2pt}{2pt}
\end{tabular}}
\end{table}




\clearpage

\end{document}